\documentclass[twoside,11pt]{article}
\usepackage{pgm}

\usepackage{bm}
\usepackage{amsmath}
\usepackage{subcaption}
\usepackage[linesnumbered,lined,boxed,commentsnumbered,ruled,vlined]{algorithm2e}
\usepackage{multicol}
\usepackage{booktabs}
\usepackage{longtable}

\usepackage[utf8]{inputenc}
\inputencoding{utf8}

\usepackage{hyperref,color,soul,booktabs}
\setulcolor{blue}
\newcommand{\BibTeX}{\textsc{B\kern-0.1emi\kern-0.017emb}\kern-0.15em\TeX}

\usepackage[dvipsnames]{xcolor}

\usepackage{newtxmath,newtxtext}
\DeclareSymbolFont{boldlargesymbols}{LMX}{ntxexx}{b}{n}
\DeclareMathAccent{\bwidetilde}{\mathord}{boldlargesymbols}{"65}

\newcommand{\X}{\ensuremath{\mathbf{X}}}

\newcommand{\Y}{\ensuremath{\mathbf{Y}}}

\newcommand{\x}{\ensuremath{\boldsymbol{x}}}

\newcommand{\y}{\ensuremath{\boldsymbol{y}}}
\newcommand{\PC}{\ensuremath{\mathcal{C}}}
\newcommand{\CLT}{\ensuremath{\mathcal{T}}}
\newcommand{\mixture}{\mathcal{M}}
\newcommand{\vtree}{\mathcal{V}}
\newcommand{\p}{\mathsf{p}}

\newcommand{\RAND}{\textsc{eRAND}}
\newcommand{\VRAND}{\textsc{vRAND}}
\newcommand{\FLOW}{\textsc{eFLOW}}
\newcommand{\vMI}{\textsc{vMI}}
\newcommand{\Pa}{\ensuremath{\mathrm{Pa}}}
\newcommand{\paras}{\ensuremath{\boldsymbol{\theta}}}
\newcommand{\Paras}{\ensuremath{\boldsymbol{\Theta}}}
\newcommand{\LL}{\mathcal{LL}}
\newcommand{\Flow}{\ensuremath{\mathbf{F}}}
\newcommand{\AFlow}{{a}}

\newcommand{\id}[1]{\llbracket{#1}\rrbracket}

\newcommand{\ch}{\ensuremath{\mathsf{ch}}}

\newcommand{\ourlearner}{\textsc{Strudel}}

\newcommand{\learnpsdd}{\textsc{LearnPsdd}}

\newcommand{\data}{\mathcal{D}}

\DeclareMathOperator*{\argmax}{argmax}

\definecolor{petroil2} {RGB} {36, 165, 175}
\definecolor{petroil4} {RGB} {30, 132, 149}
\definecolor{petroil6} {RGB} {23, 101, 115}

\definecolor{gold1} {RGB} {255, 180, 0}
\definecolor{gold2} {RGB} {255, 130, 0}
\definecolor{gold4} {RGB} {250, 100, 0}
\definecolor{gold6} {RGB} {245, 90, 0}

\usepackage{tikz}
\usetikzlibrary{circuits.logic.US}
\usetikzlibrary{shapes.misc}
\usetikzlibrary{shapes.arrows}
\usetikzlibrary{arrows,shapes.geometric}
\usetikzlibrary{patterns,calc,backgrounds}
\usetikzlibrary{trees}

\tikzstyle{nnf}=[
  >=latex, thick, >=stealth, font=\small,auto,scale=0.9,every node/.style={scale=0.9}
]
\tikzstyle{nnfnode}=[
  line width=1.0pt, draw
]
\tikzstyle{nnfand}=[
  nnfnode, and gate, rotate=90
]
\tikzstyle{nnfor}=[
  nnfnode, or gate, rotate=90
]
\tikzstyle{nnf2or}=[
  nnfor, inputs=nn
]
\tikzstyle{nnf2and}=[
  nnfand, inputs=nn
]
\tikzstyle{nnf3or}=[
  nnfor, inputs=nnn, scale=0.75
]
\tikzstyle{nnf3orfull}=[
  nnfor, inputs=nnn, scale=0.85
]
\tikzstyle{nnf4and}=[
  nnfand, inputs=nnnn, scale=0.65
]
\tikzstyle{nnfedge}=[
    line width=0.9
]
\tikzstyle{nnfterm}=[
  draw,fill=gray!10,inner sep=2.5pt, font=\small
]
\definecolor{hotcolor}{rgb}{0.85,0.0,0.0}
\tikzstyle{hot}=[
  draw=hotcolor, line width=1.1pt
]
\tikzstyle{hotparam}=[
  text=hotcolor
]

\ShortHeadings{Strudel: Learning Structured-Decomposable Probabilistic Circuits}{Dang \& Vergari \& Van den Broeck}

\begin{document}

\title{\begin{sc}{Strudel}\end{sc}: Learning Structured-Decomposable Probabilistic Circuits}

\author{\Name{Meihua Dang} \Email{mhdang@cs.ucla.edu}\and
\Name{Antonio Vergari} \Email{aver@cs.ucla.edu}\and
\Name{Guy Van den Broeck} \Email{guyvdb@cs.ucla.edu}\and
\addr Computer Science Department, University of California, Los Angeles
}

\maketitle

\begin{abstract}%
Probabilistic circuits (PCs) represent a probability distribution as a computational graph. 
Enforcing structural properties on these graphs guarantees that several inference scenarios become tractable.
Among these properties, structured decomposability is a particularly appealing one: it enables the efficient and exact computations of the probability of complex logical formulas, and can be used to reason about the expected output of certain predictive models under missing data.
This paper proposes \ourlearner, a simple, fast and accurate learning algorithm for structured-decomposable PCs.
Compared to prior work for learning structured-decomposable PCs, \ourlearner\ delivers more accurate single PC models in fewer iterations, and dramatically scales learning when building ensembles of PCs. It achieves this scalability by exploiting another structural property of PCs, called determinism, and by sharing the same computational graph across mixture components.
We show these advantages on standard density estimation benchmarks and challenging inference~scenarios. 
\end{abstract}


\section{Introduction}
\label{sec:intro}

In several real-world scenarios, decision making requires \textit{advanced} probabilistic reasoning, 
i.e., the ability to answer \textit{complex probabilistic queries}~\citep{tutorial-pc}.
Consider, for instance, querying a generative model for the probability of events described as logical constraints~\citep{Bekker2015}, e.g., rankings of user preferences~\citep{choi2015tractable,Shen2017}; 
or the probability of Bayesian classifiers agreeing on their prediction~\citep{ChoiIJCAI18}; or to conform to human expectations~\citep{KhosraviIJCAI19,KhosraviNeurips19}.
Answering these queries goes beyond the capabilities of intractable probabilistic models like classical Bayesian networks (BNs) and more recent neural estimators such as variational autoencoders and normalizing flows~\citep{papamakarios2019normalizing}.

Moreover, in many sensitive domains like healthcare and finance, the result of these queries is required (i) to be \textit{exact}, as approximations without guarantees would make the decision making process brittle, and (ii) to be provided in a \textit{limited amount of time}.
This explains the recently growing interest around tractable probabilistic models (TPMs) which guarantee both (i) and (ii) by design.

Probabilistic circuits (PCs)~\citep{choi2020pc,tutorial-pc} propose a unifying framework to abstract from the myriad of different TPM representations.
Among these, arithmetic circuits~\citep{darwiche2003differential}, probabilistic sentential decision diagrams~\citep{KisaKR14}, sum-product networks~\citep{Poon2011}, and cutset networks~\citep{rahman2014cutset} naturally fit under the umbrella of PCs. Classical bounded-treewidth graphical models~\citep{koller2009probabilistic} and their mixtures~\citep{Meila2000} are easily cast into a PC.
Within the framework of PCs, one can reason about the tractable inference capabilities of a model via the structural properties of its computational graph.
In turn, this enables learning routines that, by enforcing such specific structural properties, deliver PCs guaranteeing tractable inference for the desired classes of queries.

The \textit{structured} \textit{decomposability}~\citep{KisaKR14} property of PCs enables the largest class of tractable inference scenarios.
Indeed, all the advanced probabilistic queries we mentioned in the introductory paragraph can be exactly and efficiently answered using structured-decomposable PCs.
In a nutshell, a structured-decomposable PC encodes a probability distribution in a computational graph by recursively decomposing it into smaller distributions according to a hierarchical partitioning of the random variables, also called \textit{vtree}. 
However, while inference on structured-decomposable PCs has been extensively studied~\citep{Bekker2015,choi2015tractable,Shen2016,LiangXAI17}, relatively little attention has gone to \textit{learning} these circuits from data.
The only learning algorithm fully tailored towards structured-decomposable PCs is \learnpsdd~\citep{LiangUAI17}. 
\learnpsdd\ had the merit of introducing the task of learning a vtree.
However, this vtree learning step is detached from the actual PC structure learning step as local search, which starts from a fully-factorized circuit that ignores the discovered dependencies in the previous step.  
Furthermore, \learnpsdd\ performs a costly local search: it evaluates many candidate PC structures and computes for each of them their penalized likelihood scores, which makes it hard to scale to larger real-world datasets.

In this paper we introduce \ourlearner, a simpler and faster way to learn structured-decomposable PCs.
Specifically, \ourlearner\ drastically simplifies the search by not computing a likelihood score for each candidate structure but greedily growing the circuit.
Moreover, we do not perform vtree learning and instead initialize the PC structure in \ourlearner\ using the best TPM that can be learned with guarantees. %
It considerably speeds up learning while still delivering accurate PCs.
This is even more relevant when learning large mixtures of PCs; here we propose to scale even further by mixing components that share the same structure.
We demonstrate these performance gains on 20 standard benchmarks and on the  more challenging task of computing the expected predictions of regression models in the context of missing data~\citep{KhosraviNeurips19}. 
The rest of paper is organized as follows.
Sections~\ref{sec:back} and~\ref{sec:circuit-flows} introduce the framework of PCs for tractable probabilistic inference.
Sections~\ref{sec:strudel} and~\ref{sec:mix-strudel} describe  \ourlearner\ for learning single PCs and mixtures thereof.
Lastly, Section~\ref{sec:exp} discusses our experimental results.

\section{Probabilistic Circuits}
\label{sec:back}

Recently, the great interest in tractable probabilistic modeling propelled the introduction of a multitude of representations.
Many of these representations can be understood under a unifying computational framework, which we refer to as \textit{probabilistic circuits} (PCs)~\citep{choi2020pc,tutorial-pc}.
PCs reconcile and abstract from the different graphical and syntactic representations of recently introduced formalisms such as arithmetic circuits~\citep{darwiche2003differential}, probabilistic sentential decision diagrams (PSDDs)~\citep{KisaKR14}, sum-product networks (SPNs)~\citep{Poon2011}, cutset networks~\citep{rahman2014cutset}, and bounded-treewidth graphical models~\citep{koller2009probabilistic,Meila2000}.
Moreover, PCs enable reasoning about the tractable inference scenarios they support in terms of a unified set of structural properties.

\textbf{Notation.}
We use upper-case letters for random variables (RVs), e.g.,~$X,Y$, and lowercase ones for their assignments e.g.,~$x, y$. 
Analogously, sets of RVs are denoted by upper-case bold letters, e.g.,~$\X$, $\Y$, and their joint values by the corresponding lower-case ones, e.g.,~$\x$, $\y$.
Here we consider discrete RVs, specifically represented as Boolean variables, i.e., having values in $\{0, 1\}$.

\begin{figure}[!t]
    \centering
\begin{subfigure}[b]{0.3\textwidth}
    \footnotesize
    \setlength{\tabcolsep}{2pt}
    \begin{tabular}{l| r r}
    \toprule
    $\p(X_i=1|X_{\Pa_i})$ & $X_{\Pa_i}=0$ & $X_{\Pa_i}=1$\\\hline
    $\p(X_4=1)$   &   0.4 &   0.4\\
    $\p(X_3=1|X_4)$   &   0.8 &   0.3\\
    $\p(X_2=1|X_3)$   &   0.5 &   0.9\\
    $\p(X_1=1|X_3)$   &   0.7 &   0.4\\
    \bottomrule
    \end{tabular}
    \caption{\label{fig:cpt}}
\end{subfigure}
\begin{subfigure}[b]{0.1\textwidth}
    \centering
    \includegraphics[width=.9\textwidth]{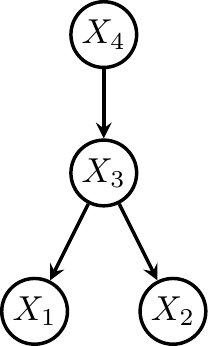}
    \caption{\label{fig:clt}}
    \end{subfigure}\begin{subfigure}[b]{0.1\textwidth}
    \includegraphics[width=.9\textwidth]{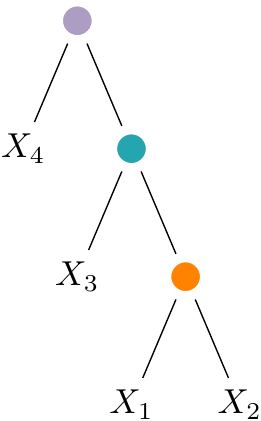}
    \caption{\label{fig:vtree}}
    \end{subfigure}
    \begin{subfigure}[b]{0.45\textwidth}
    \includegraphics[width=.99\textwidth]{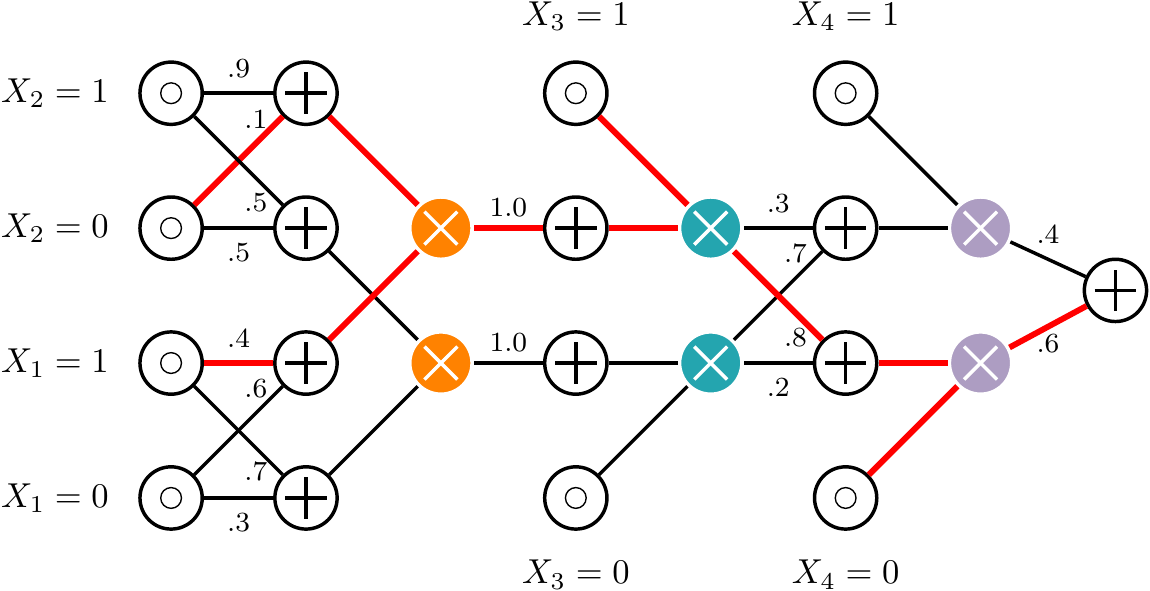}
    \caption{\label{fig:pc}}
    \end{subfigure}
    \caption{\textbf{\textit{A structural decomposable PC and its corresponding vtree and equivalent BN.}} The CLT over RVs $\X=\{X_{1},X_{2},X_{3},X_{4}\}$ in~(\ref{fig:clt}) with CPTs in~(\ref{fig:cpt}) is turned into the structured-decomposable PC in~(\ref{fig:pc}) whose extracted vtree is shown in~(\ref{fig:vtree}).
    In red, all the edges that are ``active'' in the circuit flow for the input configuration $\{X_1=1, X_2=0, X_3=1, X_4=0\}$. Root node is on the right.}
    \label{fig:pc-summary}
\end{figure}{}

\textbf{Representation.} 
A probabilistic circuit $\PC$ over RVs $\X$ is a directed acyclic graph (DAG) representing a computational graph that encodes a probability distribution $\p_{\PC}(\X)$ in a recursive manner.
This DAG has three kinds of nodes: \textit{input distributions} (leaves), \textit{product} nodes and \textit{sum} nodes. 
Figure~\ref{fig:pc} shows an example of a PC.
Each node $n$ encodes a distribution $\p_{n}$, defined as follows.
Let $\PC_{n}$ denote the sub-circuit rooted at node $n$ and $\ch(n)$ its child nodes. 
An input distribution $n$ encodes a tractable probability distribution $\p_{n}$ over some RVs $\phi(n) \subseteq \X$, also called its \textit{scope}.
A product node $n$ defines the factorized distribution $\p_{n}(\X)=\prod_{c\in\ch(n)}\p_{c}(\X)$. W.l.o.g. we will consider product nodes to have only two children.
A sum node $n$ defines a mixture model $\p_{n}(\X)=\sum_{c\in\ch(n)}\theta_{n,c} \p_{c}(\X)$ parameterized by edge weights $\theta_{n,c}$.
The scope of a product or sum node is the union of the scopes of its children:  $\phi(n)=\cup_{c \in \ch(n)}\phi(c)$.
In this work targeting Boolean RVs, we consider univariate input distributions, specifically leafs for RV $X_{i}\in\X$ will be indicator functions of the form $\p(X_{i}=1)=\id{X_{i}=1}$.
Let $\paras$ denote all the parameters in a circuit $\PC$ (only attached to sum node edges, since input distributions are indicators).

Even if they are \textit{probabilistic} models expressed via a \textit{graphical} formalism, probabilistic circuits are \textit{not} classical PGMs.
The clear \textit{operational} semantics we described above makes PCs peculiar neural networks~\citep{vergari2019visualizing,peharz2020einsum} whose inner units are either products (acting as non-linearities) or convex combinations of their inputs.
Overall, a PC $\PC$ defines a multilinear polynomial~\citep{darwiche2003differential} whose indeterminates are the distributions equipping the leaves of $\PC$.

\textbf{Inference.}
Any PC $\PC$ over RVs $\X$ that represents a normalized distribution supports computing the likelihood $\p_{\PC}(\x)$ given a \textit{complete configuration} $\x$ (a complete evidence query) by evaluating the circuit bottom up: starting from the input distributions and computing the output of children before parents.
Additional structural properties of the PC, such as \textit{decomposability}, \textit{smoothness} and \textit{determinism}, can extend the set of probabilistic queries that are guaranteed to be answered exactly and in time linear in the \textit{size} of the PC, that is, its number of edges.
Then, query answering reduces to traversing the PC bottom-up (forward) and top-down (backward) given values for the leaf~nodes.

A PC is \textit{decomposable} if for every product node, the children have disjoint scopes.
That is, product nodes encode well-defined factorized probability distributions.
A PC is \textit{smooth} if for every sum node, the children have the same scope. 
That is, sum nodes encode mixtures of distributions that are well-defined over identical sets of RVs.
Smooth and decomposable PCs enable the tractable computation of any marginal query~\citep{darwiche2003differential}.
SPNs~\citep{Poon2011} are examples of smooth and decomposable PCs.
A circuits is deterministic if for every sum node $n$ and assignment $\x$, at most one of the children of $n$ have a non-zero output.
That is, a deterministic sum defines a mixture model whose components have disjoint support, enabling tractable MAP queries \citep{chan2006robustness}.
Examples of smooth, decomposable and deterministic PCs are cutset networks~\citep{rahman2014cutset,DiMauro2015a} and selective SPNs~\citep{Peharz2014b}.

\textbf{Structured-Decomposable PCs.} 
More recently, the stronger property of \textit{structured decomposability} has been introduced to enable a larger class of tractable inference scenarios~\citep{pipatsrisawat2008new,KisaKR14}.
Briefly, the product nodes in a structured-decomposable PC cannot decompose in arbitrary ways, but must agree on a ``contract.''
A PC is structured-decomposable if it is \textit{normalized} for a \textit{vtree},
a binary tree encoding a hierarchical decomposition of RVs.
Each leaf in a vtree denotes a RV, while an internal node
indicates how to decompose a set of RVs in two subsets mapping to its left and right branch. 
 A PC is normalized for a vtree if the scope of every product node decomposes over its children as its corresponding node in the vtree.
An example of a vtree and a structured-decomposable PC normalized for it are shown in Figure~\ref{fig:vtree} and Figure~\ref{fig:pc}.

By enforcing structured decomposability, several classes of advanced probabilistic queries 
become computable exactly and efficiently.
For instance, structured-decomposable PCs allow to compute symmetric and group queries~\citep{Bekker2015}
and, given certain constrained vtrees, same-decision probabilities~\citep{OztokD15}, their expected version~\citep{ChoiIJCAI17} and classifier agreement~\citep{ChoiIJCAI18}.
Moreover, if two PCs conform to the same vtree, it is possible to efficiently compute the KL divergence between them~\citep{LiangXAI17};
to multiply them~\citep{Shen2016}; or to take the expectation of one according to the distribution encoded by the other~\citep{choi2015tractable,KhosraviNeurips19}.
AND-OR graphs~\citep{dechter2007and} and PSDDs~\citep{KisaKR14} are examples of smooth, deterministic and structured-decomposable PCs.\footnote{In their original formulation~\citep{KisaKR14}, PSDDs required a stronger notion of determinism, which has no practical implication for tractable probabilistic inference.}
Despite all these advanced inference scenarios that structured-decomposable PCs enable, relatively little attention has been dedicated to learning these circuits from data, and the only attempt so far is difficult to scale~\citep{LiangUAI17}.

\section{Circuit Flows: Fast Inference and Parameter Learning}
\label{sec:circuit-flows}

Before explaining \ourlearner, we briefly introduce \textit{circuit flows} -- a computational tool that allows us to scale up our learner.
Determinism not only makes MAP inference tractable, but also enables closed-form parameter estimation in PCs~\citep{KisaKR14} and dramatically speeds up inference by leveraging circuit flows.
We refer to \citet{LiangAAAI19} for a formal definition.
Figure \ref{fig:pc} shows an example of a circuit flow. Intuitively, a circuit flow in a deterministic PC $\PC$ encodes which parameters are activated by different input configurations. 

\textbf{Fast inference.}
For a given sample $\x$, a flow acts as a transformation $f_{\PC}:\X\mapsto\{0,1\}^{|\paras|}$
mapping $\x$ to a binary encoding, called \textit{flow embedding},  
with as many entries as there are parameters in $\PC$.
The $k$-th entry in $f_{\PC}(\x)$ is 1 if sample $\x$ \textit{flows} through the edge associated with the $k$-th parameter in~$\paras$, reaching the output, and 0 otherwise.
As such, the log-likelihood $\mathcal{LL}_{\PC}(\paras; \x)$ of a deterministic circuit $\PC$ parameterized by $\paras$, given a single input configuration $\x$, is efficiently computed~as
\begin{equation*}
    \LL_{\PC}(\paras; \x) = \log(\p_{\PC}(\x)) = {f}_{\PC}(\x)^{T} \cdot \log(\paras).
\end{equation*}
Similarly, the circuit flow of a batch of samples $\data$ can be represented as a binary matrix $\Flow_{\PC}(\data)\in\{0, 1\}^{|\data|\times |\paras|}$.
Then, the log-likelihood given the entire batch at once is efficiently vectorized as
\begin{equation*}
    \LL_{\PC}(\paras; \data) = \Flow_{\PC}(\data)\cdot\log(\paras).
\end{equation*}

This formulation of the likelihood has the clear benefit that a vectorized computation can yield significant speedups. 
Furthermore, the matrix $\Flow_{\PC}(\data)$ can be computed by propagating bit-vectors up and down the circuit $\PC$. Recall that flow embeddings are binary vectors by virtue of the determinism property. Bit-vector arithmetic is extremely efficient, much more so than using floating-point vectors to compute the same likelihoods.
Lastly, circuit flows will greatly benefit inference in our large ensembles sharing the same structure (cf.~Section~\ref{sec:mix-strudel}). 
Since a flow only depends on the circuit structure, for a mixture $\mixture=\{\PC_{i}\}_{i=1}^{k}$ of $k$ PCs sharing the same structure, we need to evaluate a single flow $f_{\mixture}$ once.
As such, the log-likelihood of $\mixture$ given $\x$ can be efficiently computed as:
\begin{equation}
    \LL_{\mixture}(\Paras; \x) = %
    \mathsf{logsumexp}(f_{\mixture}(\x)^{T} \cdot \log(\Paras) + \log(\boldsymbol{w})),
    \label{eq:matrix-mix}
\end{equation}
where $\boldsymbol{w} =\{w_{i}\}_{i=1}^{k}$ are the mixture weights, and $\Paras$ is the matrix whose columns are the parameters $\paras_{i}$  of the $i$th PC in the mixture. 
We empirically show these speedups in Section~\ref{sec:exp} and Appendix \ref{sec:app-fastflows}. 

\textbf{Parameter learning.}
More instrumental to our purpose, flow embeddings can be used for parameter learning of a PC $\PC$.
We define the \textit{aggregate flow} $\AFlow_{\PC}(i, j;\data)$ of one edge $e_{i,j}$ associated with the $k$th weight $\theta_{i,j}$, as the total number of configurations in the dataset $\mathcal{D}$ that flow through edge $e_{i,j}$, and whose likelihood therefore contains the $k$th weight $\theta_{i,j}$ as a factor. That is,  $\AFlow_{\PC}(i, j;\data)={\sum}_{h=1}^{|\mathcal{D}|} \Flow_{\PC}(\data)[h, k]$.
Now, the maximum likelihood estimator (MLE) of weight $\theta_{i, j}$ %
can be computed in closed form as the ratio
\begin{equation}
\theta_{i,j}^{\mathsf{MLE}} = \frac
{\AFlow_{\PC}(i, j;\data)}
{{\sum}_* \AFlow_{\PC}(i, *;\data)}.
    \label{eq:theta-mle}
\end{equation}
In other words, $\theta_{i,j}^{\mathsf{MLE}}$ can be computed as the ratio of the number of samples in $\data$ flowing through edge $e_{i,j}$ over the total number of samples flowing through node $i$.

In summary, the circuit flow formulation has the benefit of (1) vectorizing the computation, (2) allowing for a single computation of the flow embeddings to be reused across PCs with the same structure but different parameters, for example in large ensembles, and (3) yielding simple closed-form parameter estimates. 
Moreover, we will show in Section \ref{sec:strudel} that flows can act as one of the building blocks in structure learning. Thus, both our learner for expressive single models in Section~\ref{sec:strudel} and for scalable ensembles in Section~\ref{sec:mix-strudel} will make use of this efficient formulation.

\section{\textsc{Strudel}: Learning Structured-Decomposable Probabilistic Circuits}
\label{sec:strudel}

The objective of structure learning for PCs is to find a circuit structure and parameters that approximate well the data distribution. 
If the learned PC has to guarantee tractable inference for certain classes of queries,
its structure has to enforce the corresponding properties discussed in Section \ref{sec:back}.
For the advanced inference scenarios we are interested in, and to retain efficient parameter learning (cf.~Section~\ref{sec:circuit-flows}), we require structured decomposability and determinism.
So far, the only alternative learner to deliver such PCs is \learnpsdd~\citep{LiangUAI17}, which however performs a search procedure that can be computationally inefficient.
We review \learnpsdd~next, before proposing \ourlearner\ and explaining how it addresses some deficiencies of the prior work. 

\textbf{\learnpsdd\ and its limitations.}
\learnpsdd\ performs a local search over the space of possible structured-decomposable PCs, given a vtree as input.
To learn a vtree from data, a hierarchical clustering step is performed over the RVs discovering some independence relationships:
they are recursively grouped bottom up (or split top down) so as to maximize their pairwise mutual information.
Next, local search starts from a fully-factorized PC, that is, one where all RVs are considered to be independent, reshaped to conform to the learned vtree.
Each search iteration locally changes the circuit while preserving its semantics and structural properties of smoothness, determinism and structured decomposability.
To propose candidates, \learnpsdd\ consistently applies two structural transformations -- \textit{split} and \textit{clone} -- to all possible nodes in the circuits.
These candidates are then ranked by their log-likelihood score, penalized by their circuit size.

We highlight the following shortcomings of \learnpsdd:
(i) vtree learning as a separate pre-processing step has a limited effect on structure learning, which starts from a fully-factorized distribution, discarding the dependencies discovered in vtree learning.
Moreover, while circuit flows speed up likelihood computation in deterministic circuits, (ii) using likelihood to score candidate structures drastically slows down learning, especially in large data regimes.
As a result, when employed in mixture models \learnpsdd\ has not been able to scale beyond tens of~components.

To overcome these shortcomings, we propose to:
(i) extract a vtree structure from the best graphical model that can be learned in tractable time, and then compile it into a structured-decomposable PC, which provides a more informative starting point, 
(ii) dramatically reduce learning time by employing a greedier local search employing a single transformation, split, and
(iii) effectively use circuit flows to speed up parameter learning and likelihood computation.
The resulting algorithm is 
a \textit{simpler}, \textit{faster} structure learning scheme, yet yielding competitively \textit{accurate} PCs and enabling fast learning of large mixtures.
We name it \ourlearner: a STRUctured-DEcomposable Learner.

\subsection{From Chow-Liu Trees to Structured-Decomposable PCs}
\label{sec:clt-comp}
Instead of coming up with a fixed vtree and distilling an initial PC structure out of it, we propose to do the inverse: \textit{learn the best initial PC from data, and distill a vtree from it}.

 \textbf{Chow-Liu trees.} Here we use \textit{Chow-Liu trees} (CLTs) as the ``best'' initial PC structure.
 CLTs are tree-shaped BNs that satisfy our desiderata: they guarantee to encode the best tree model in terms of KL divergence with the data distribution; they support linear time marginals and MAP inference; and can be learned in time $O(|\X|^{2}|\data|)$.
Furthermore, as we will show next, we can quickly compile CLTs into smooth, deterministic and structured-decomposable PCs and extract a vtree from them.

More formally, a CLT $\CLT$ over RVs $\X$ is a tree-shaped BN equipped with parameters $\theta_{i|\Pa_{i}}$ defining the conditional probability table (CPT) of node $i$ associated to RV $X_{i}$ with parent node $\Pa_{i}$.
An example of a CLT is shown in Figure~\ref{fig:clt}.
The classic Chow-Liu~(\citeyear{ChowLiu}) algorithm learns a CLT~$\CLT$ from  data $\data$ by running a maximum spanning tree algorithm over a complete graph induced by the pairwise mutual information (MI) matrix over variables $\X$ as estimated from data $\data$.
These MI estimates are used to compute the $\theta_{i|\Pa_{i}}$ parameters, and can be smoothed by adding a Laplace correction factor $\alpha$. 
See Appendix~\ref{sec:app-clt} for a detailed algorithm.
Computation-wise, the cost of learning a CLT is the same as learning a vtree in \learnpsdd, i.e., that of computing the pairwise MI.
Learning-wise, we are not discarding this information but embed it in the initial PC structure.

\textbf{Compiling CLTs.} We now turn our attention to compiling a CLT into a structured-decomposable PC.
Compiling generic BNs into smooth, deterministic and decomposable PCs,\footnote{Specifically to Arithmetic Circuits~\citep{darwiche2003differential} represented as DAGs having parameters attached to leaf nodes.} has been extensively researched in the literature~\citep{darwiche2003differential}, and compilation of a BN into a structure decomposable circuit has also been explored~\citep{choi2013compiling,Shen2016}.
However, compiling a BN (even a CLT) for an arbitrary vtree can lead to an exponentially larger PC.
Therefore, we adopt a simple strategy tailored for CLTs that extracts a vtree guaranteeing a linear-size PC in the number of RVs.

We start from the observation that a \textit{rooted} CLT provides a natural variable decomposition.
While rooting the CLT can be done arbitrarily,
we root it at its \textit{Jordan center} as a heuristics to minimize the resulting vtree depth and thus yielding smaller PCs.
Then we traverse the CLT top-down to build the vtree, for each node $X_i \in \CLT$, if $X_i$ is a leaf node, compile it to a vtree leaf node $v_i$ containing variable $X_i$; otherwise build an inner node with $v_i$ as its one branch, and the vtree for its children $\ch(X_i)$ as its other branch.
After a vtree is fully grown, we proceed compiling the CLT bottom-up, caching the previously compiled sub-circuits, which guarantees that we obtain a PC of linear size~\citep{darwiche2003differential}.
For every node $X_{i}\in\CLT$ that we visit, and for every parent configuration, we introduce sum nodes selecting a value of $X_{i}$ with edge weights $\p(X_{i}|X_{\Pa_{i}}=x_{\Pa_{i}})$, i.e., its distribution conditioned on the parent configuration. 
A corresponding indicator leaf, following the vtree structure, is introduced in the product.
This yields a smooth deterministic sum node branching over the possible values for the considered RV. 
Figure~\ref{fig:pc} illustrates the compiled PC for the example CLT in~Figure~\ref{fig:clt}.

\subsection{How And What to Split}
\label{sec:split-heu}
From this initial circuit, \ourlearner\ applies the \textit{split} operation at each step, performing a \textit{greedy search}.
It waives the necessity for computing the penalized log-likelihoods of possible candidate structures.

Our {split} operation is based on the one proposed in~\citet{LiangUAI17}, which builds on \citet{LowdD08}.
Given a PC structure $\PC^{t}$ at iteration $t$, we create $\PC^{t+1}$ in a \textit{two-step procedure}.
First, we select one \textit{edge} $e_{n,c}$ to split, from one sum node $n$ to one of its child product nodes $c$, 
and one \textit{variable} $X_{i}$ to split in the scope of $c$.
Second, we make two \textit{partial copies} of sub-circuits rooted at $c$ conditioned on $\id{X_{i}=0}$ and $\id{X_{i}=1}$ respectively. 
These partial copies are carried out by copying nodes up to a certain depth bound.
We thus create $\PC^{t+1}$ by removing child $c$ from node $n$ in $\PC^{t}$, while adding both of the new copies.
This splitting operation preserves smoothness and determinism since sums are conditioned on the RV $X_{i}$, while copying preserves structured decomposability.

\textbf{Selecting edges to split.} To select an {edge} and {variable} to split, the simplest but uninformed strategy would be to pick one edge and one variable randomly.
We name these two strategies \RAND\ and \VRAND.
Clearly, a more informed heuristics, but less expensive than the computation of the likelihood,  would benefit search.
We introduce two novel heuristics, \FLOW\ and \vMI\ for selecting an edge and a variable respectively. 
Specifically, \FLOW\  selects edge $e_{i, j}$ which maximizes the aggregate circuit flow (cf.~Section~\ref{sec:circuit-flows}):
\begin{equation}
\mathsf{score}_{\mathsf{eFLOW}}(e_{i,j}; \PC^{t},\data) := %
\AFlow_{\PC^t}(i,j;\data)
    \label{eq:edge-flow}
\end{equation}{}
That is, the \FLOW\ picks the edges where more samples in $\data$ flow through, indicating that introducing a new sub-circuit there could  potentially better model the distribution over those samples.

\textbf{Selecting RVs to split.}
Once we pick edge $e_{i, j}$,
we then select the RV among those in the scope of node $j$. 
Specifically our \vMI\ heuristics selects the RV $X_{k}$ sharing more dependencies with the others in the scope.
That is, we maximize the score:
\begin{equation}
\mathsf{score}_{\mathsf{vMI}}(X_{k}; \PC^{t}, \data) := 
{\sum}_{X_{h} \neq X_{k}} \overline{\mathsf{MI}}(X_{h}; X_{k})
    \label{eq:var-mi}
\end{equation}{}
where $\overline{\mathsf{MI}}$ is the pairwise mutual information estimated on the samples of $\data$ ``flowing'' through edge $e_{i,j}$.
By introducing new parameters for highly dependent RVs we can learn more accurate PCs.
The entire local search loop performed by \ourlearner\ is summarized in Algorithm~\ref{alg:learnstrude} in Appendix~\ref{sec:app-learn}.

\section{Fast Mixtures with \ourlearner}
\label{sec:mix-strudel}
Learning mixtures of PCs greatly improves their performance as density estimators~\citep{Vergari2015,DiMauro2015b,DiMauro2017b,rahman2016learning}.

\textbf{Issues in learning structured-decomposable mixtures.} 
Building a mixture of several PCs results in a joint non-deterministic PC, as it introduces a sum node marginalizing a latent variable.
While such a PC would not allow exact MAP inference, it could still be used for queries requiring structured decomposability.
However, to answer complex queries like the expectation of %
predictive models~\citep{KhosraviNeurips19}, one would require a \textit{structured-decomposable mixture of PCs}, i.e., an ensemble whose components are structured-decomposable \textit{and} share the same vtree.
\textit{To force such a constraint, while preserving the mixture expressiveness or containing its circuit size, is a non-trivial research question.}
Consider learning several PCs with \ourlearner\ while requiring them to share the same vtree.
If we learn each component from a different CLT, compiling them to PCs while enforcing a unique vtree might lead to an exponential blow-up in the size of some PCs.
Alternatively, enforcing Algorithm~\ref{algo:learnclt} to output a CLT that can be compactly compiled according to a vtree, would result in losing the algorithm's optimality guarantee.

\textbf{Shared-structure mixtures.} We propose a simpler and faster ensembling strategy which proves to be very effective in practice.
\textit{We build ensembles of PCs sharing the same structure}, concretely the structure learned by \ourlearner\ for single models on that data, while letting each mixture component have different parameters.
This has a number of advantages: we (i) need to perform structure learning only once, (ii) materialize a single flow $f_{\mixture}$ once (as identical structures will generate the same flow), and 
(ii) can evaluate the likelihood of the whole mixture efficiently, as shown in Eq.~\ref{eq:matrix-mix}.
This strategy is compatible with classical ensembling schemes such as EM, bagging and boosting.
All these scenarios, e.g., each M step in EM, reduce to learning each mixture component parameters on a weighted version of the original data.
Determinism allows to do this in closed form (cf.~Eq.~\ref{eq:theta-mle}).

\section{Experiments}
\label{sec:exp}
In this section, we rigorously evaluate \ourlearner\ empirically.\footnote{Code and experiments are available at \url{https://github.com/UCLA-StarAI/Strudel}.}
The natural competitor for \ourlearner\ is \learnpsdd, as they both aim to learn PCs with the same structural properties (cf.~Section~\ref{sec:strudel}).
We evaluate both learners as density estimators on a series of 20 standard benchmark datasets.
Specifically, we aim to answer the following research questions:
\textbf{(Q1)} What is the effect of initializing structure learning with a CLT? 
\textbf{(Q2)} How is the splitting heuristic in \ourlearner\ affecting structure learning?
\textbf{(Q3)} How do single PCs learned by \ourlearner\ compare to those learned by \learnpsdd? %
\textbf{(Q4)} Are ensembles of PCs learned by \ourlearner\ competitive with \learnpsdd?
\textbf{(Q5)} Is our inference approach based on circuit flows speeding up likelihood computations on ensembles of PCs? 
\textbf{(Q6)} Are ensembles of PCs learned by \ourlearner\ helpful for advanced probabilistic queries?

\begin{figure}[!t]
    \includegraphics[width=0.24\textwidth]{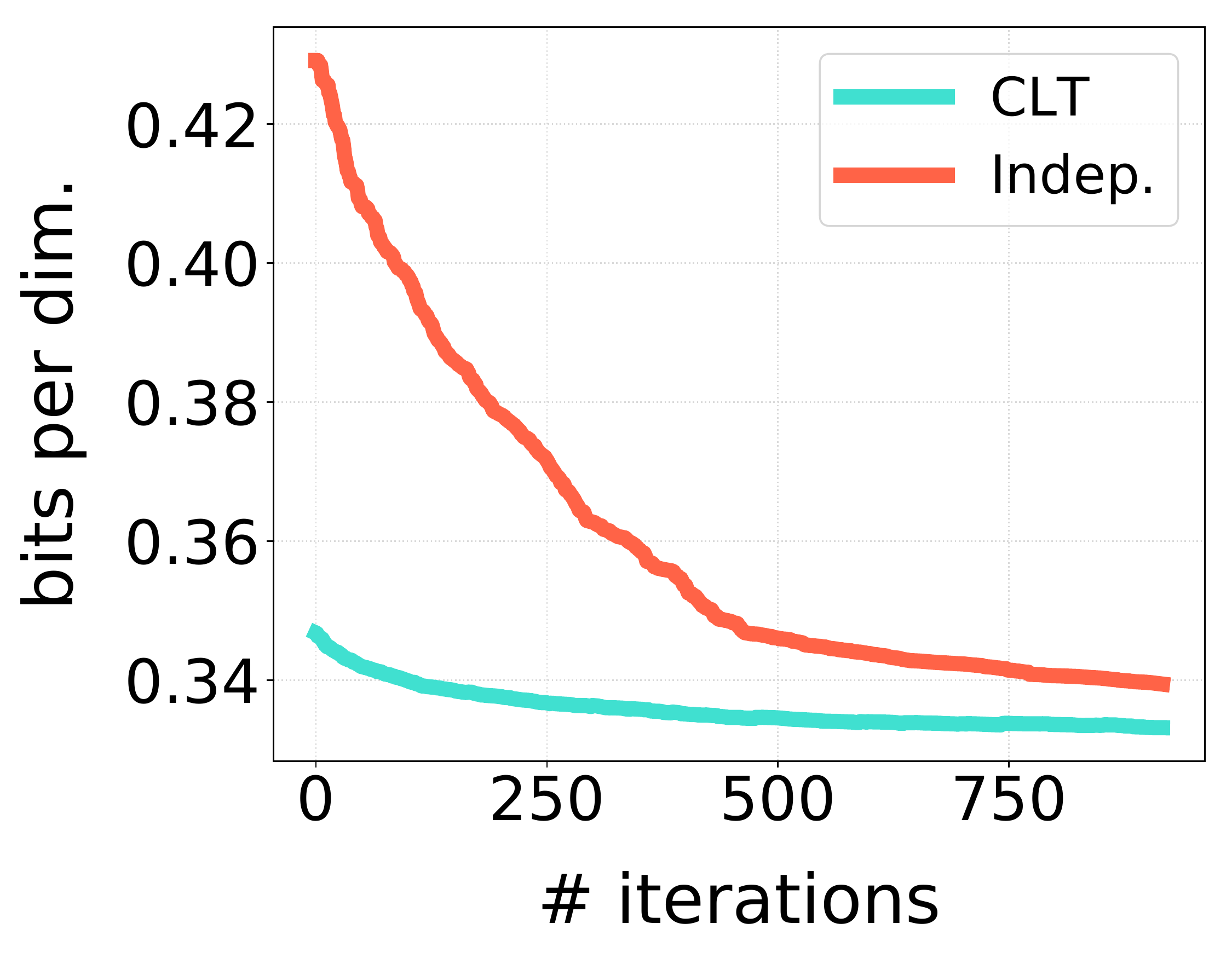}
    \hspace{10pt}
    \includegraphics[width=0.26\textwidth]{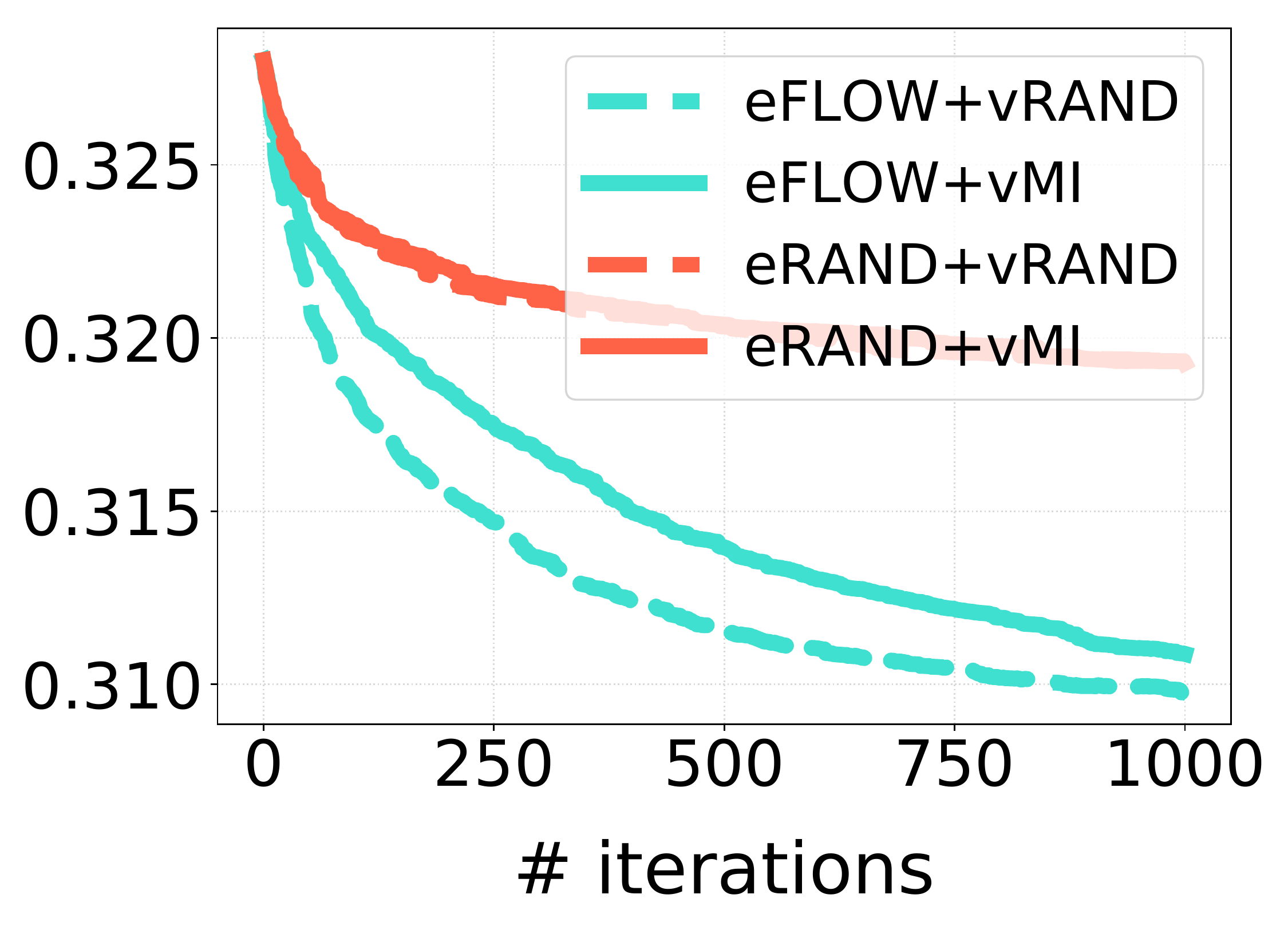}
    \hspace{10pt}
    \includegraphics[width=0.28\textwidth]{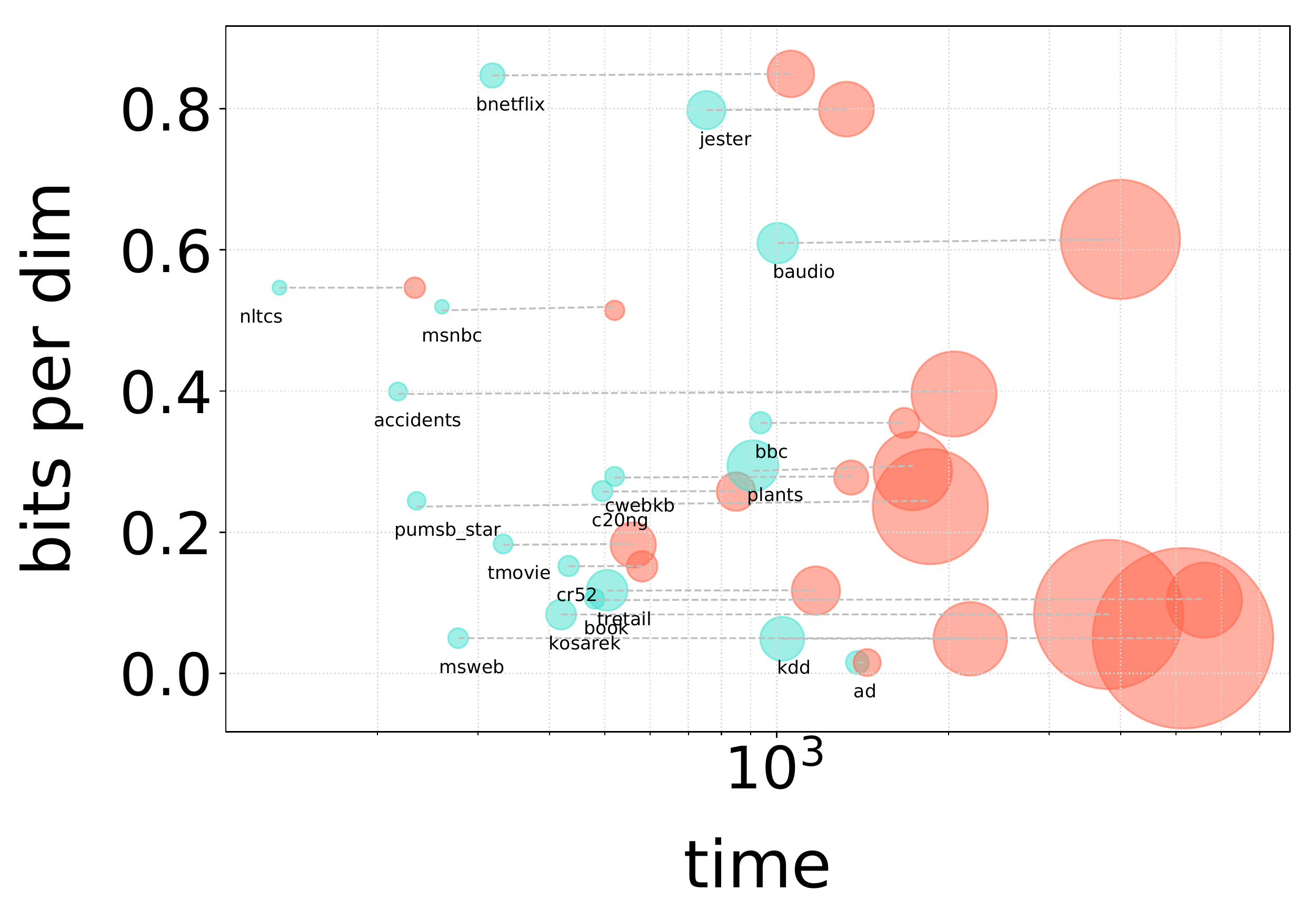}
    \includegraphics[width=0.075\textwidth]{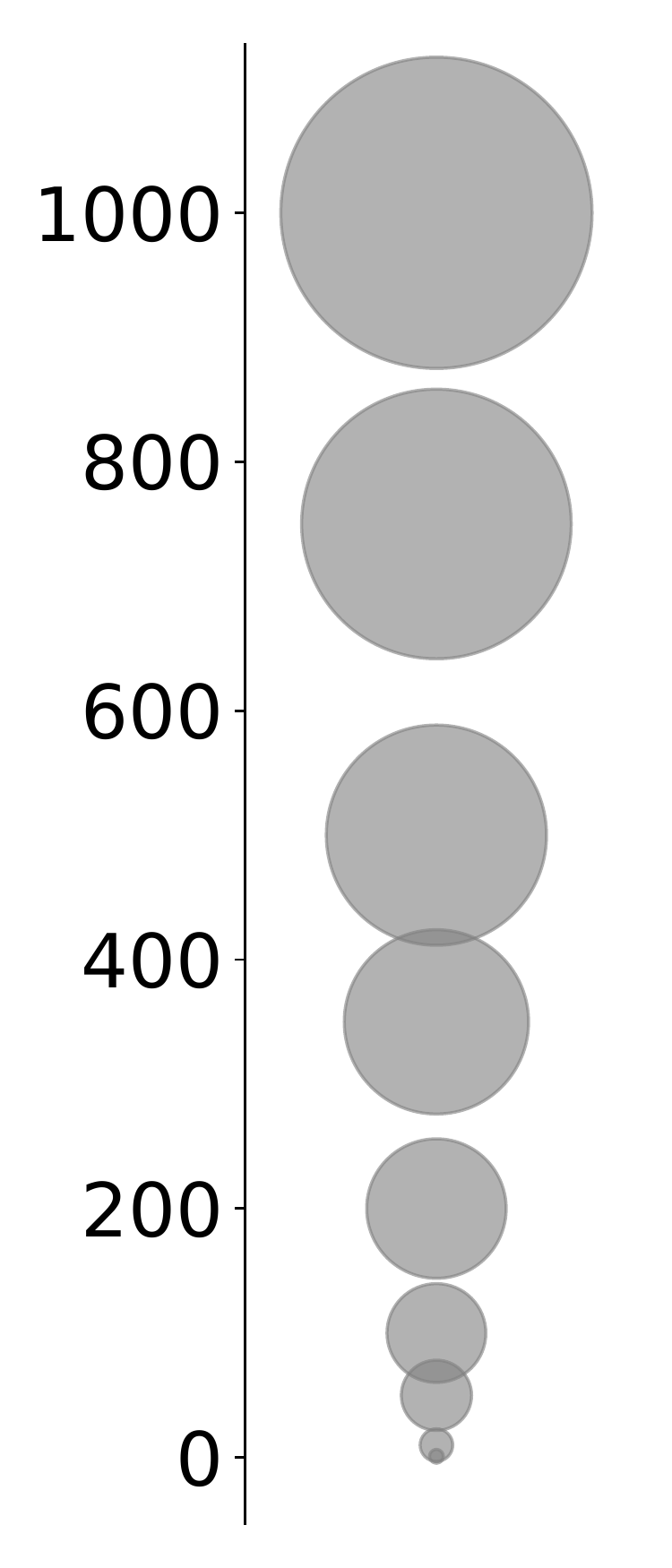}
    \caption{\textit{\textbf{Effect of different initializations and heuristics in \ourlearner.}} We report the mean test bits per dimensions (bpd) averaged across all datasets (y-axis) for each iteration (x-axis) as scored by the different initialization schemes (left) and with four possible splitting heuristics (cf.~Section~\ref{sec:strudel}) in \ourlearner\  when using the CLT initialization (center). On the right, the per-dataset bpd scores (y-axis) versus total learning time (in secs, x-axis) and learned PC sizes (proportional to circle radiuses times 1000, cf.~legend) of the heuristics \FLOW+\vMI (blue) and \FLOW+\VRAND (red).
        \label{fig:heuristics}
    }

\end{figure}

\textbf{(Q1) Effect of using CLTs.}
To evaluate the impact of different initializations in \ourlearner, we compare the CLT initialization scheme we proposed in Section~\ref{sec:clt-comp} against the scheme of \learnpsdd, where the initial PC is a fully-factorized distribution, normalized for a vtree learned in advance.
For both, we learn PCs with up to 1000 splits on every dataset, then report the mean test bits-per-dimension (bpd)\footnote{$\mathsf{bpd}(\paras;\data)=-{\sum_{i=1}^{|\data|}\mathcal{LL}(\paras;\x_{i})}/(\log(2) \cdot |\data| \cdot m)$ where $m$ is the number of features in dataset $\data$.}
averaged across all datasets as a function of iterations in Figure~\ref{fig:heuristics}~(left).
On average, employing CLTs in \ourlearner\ not only delivers more accurate initial PCs as expected, but better bpds in the long run.
Detailed per-dataset curves can be found in Appendix~\ref{sec:app-init}.

\textbf{(Q2) Effect of splitting heuristics.}
We adopt the same setting of \textbf{Q1} and we mix and match all possible combinations of splitting heuristics: 1) \FLOW-\vMI, 2) \FLOW-\VRAND, 3) \RAND-\vMI, 4) \RAND-\VRAND\ (cf.~Section~\ref{sec:split-heu}). 
Figure \ref{fig:heuristics}~(center) reports the mean test bpd per iteration averaged across all datasets. 
Detailed per-datasets plots can be found in Appendix~\ref{sec:app-heuristics}.
It is apparent how selecting edges at random with \RAND\ delivers suboptimal PCs when compared to \FLOW, regardless of the heuristic to select the RV.
On the other hand, \FLOW-\VRAND\ delivers slightly more accurate PCs, on average, than \FLOW-\vMI.
However, this comes at a high price which is highlighted in
Figure \ref{fig:heuristics}~(right): circuits learned by \FLOW-\VRAND\ are one or two orders of magnitude larger and each splitting iteration on them is much slower than for \FLOW-\vMI.
This can be explained as follows: \VRAND\ can greatly increase the PC size by arbitrarily picking a RV far from the root of the sub-circuit selected by \FLOW\ hence duplicating a larger PC, while \vMI\ picks more informative RVs  which generally are closer to the root of such sub-circuit.
Given all the above, we employ \FLOW-\vMI\ and the CLT initialization scheme in our remaining experiments.

\begin{figure}[!t]
\centering
    \includegraphics[width=0.28\textwidth]{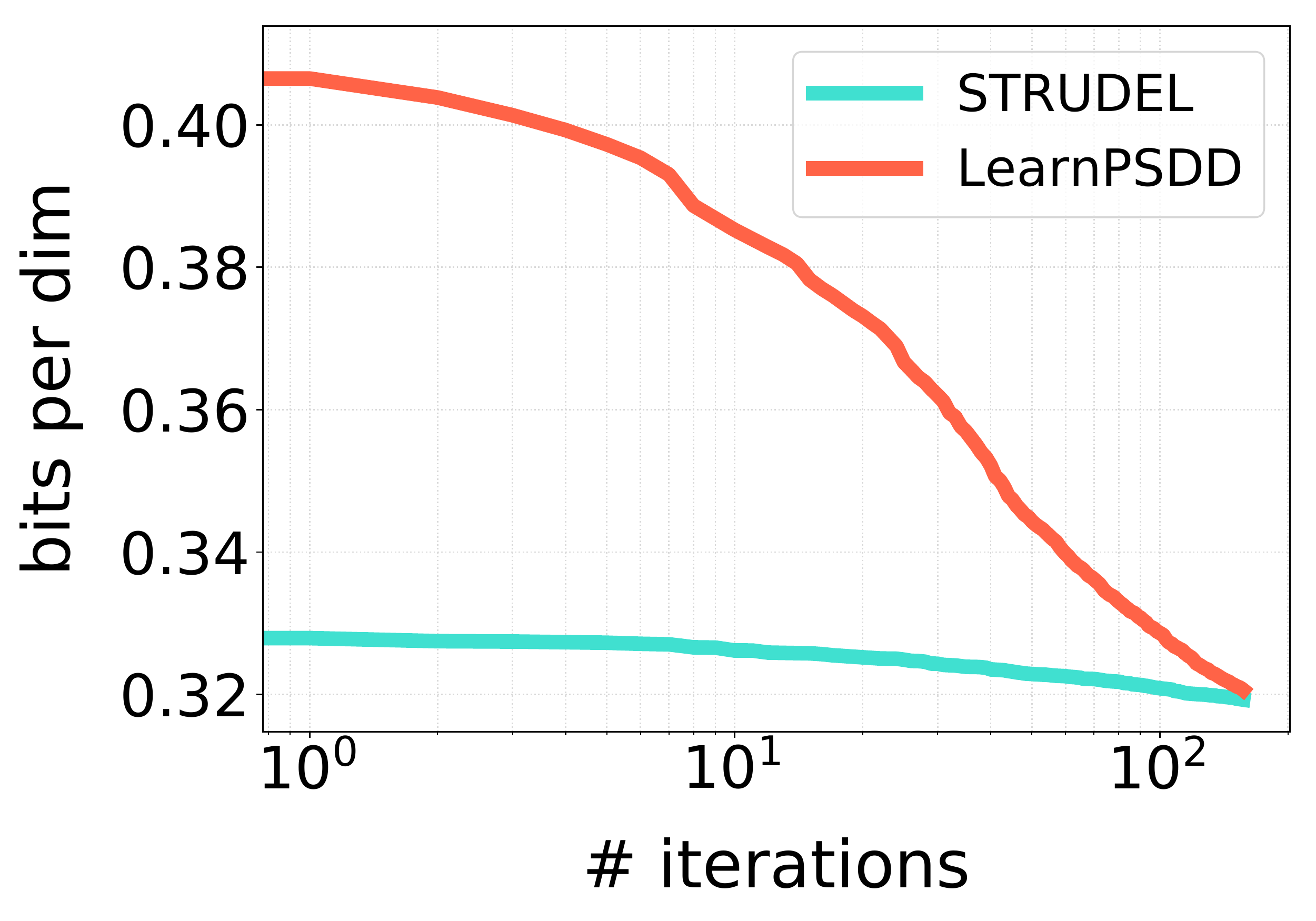}\hspace{10pt}
    \includegraphics[width=0.28\textwidth]{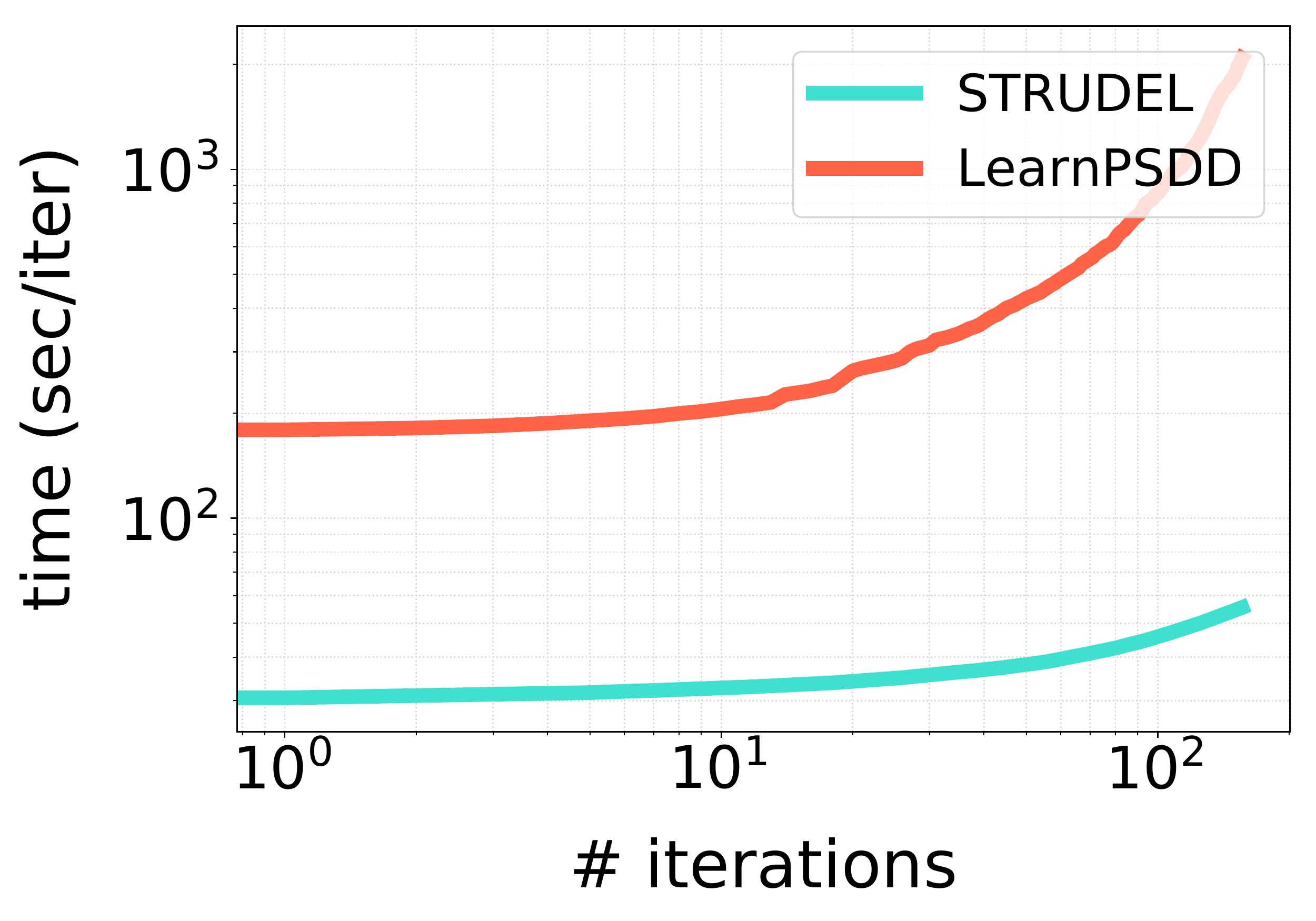}\hspace{10pt}
    \includegraphics[width=0.29\textwidth]{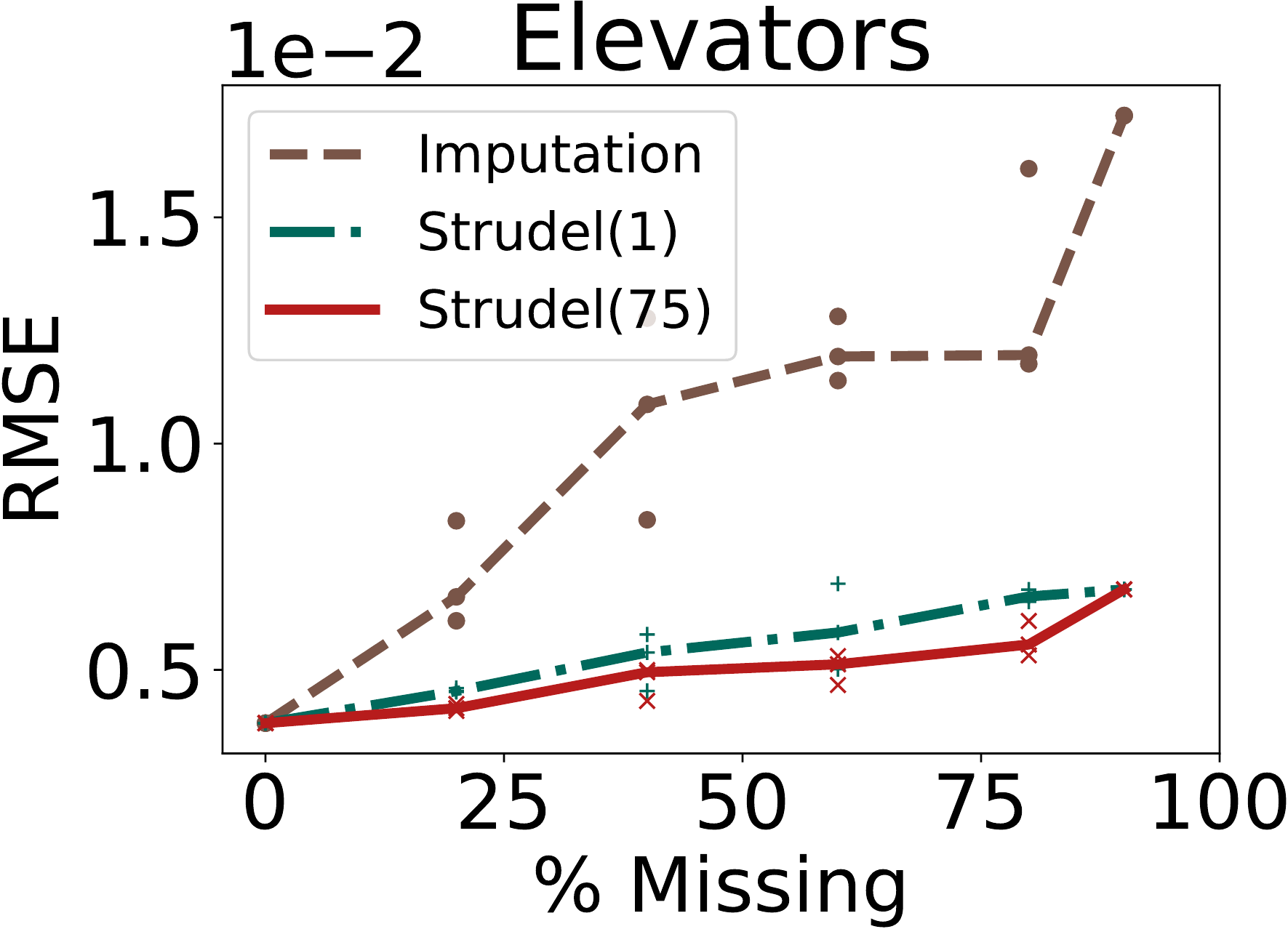}
    \caption{\textit{\textbf{Comparing \ourlearner\ and \learnpsdd\ accuracy and learning times}}. We report the mean bpd averaged across all datasets (y-axis) (left), and seconds per iteration during learning (y-axis) (center) for each iteration (x-axis) as scored by single models learned by \ourlearner\ (blue) and \learnpsdd\ (red). \textit{\textbf{Expected prediction benchmarks}}. On the right, the Root Mean Square Error (RMSE) for median imputation, \ourlearner\ and \ourlearner-Ensemble models (number in parenthesis shows the number of components in ensemble models).}
    \label{fig:missing-body}
\end{figure}

\textbf{(Q3) Single models.}
We perform early stopping after 100 iterations with no improvement for \ourlearner, 
and \learnpsdd\ which we re-run on all datasets by using the hyperparameters reported in~\citet{LiangUAI17}.
Table \ref{tab:single-model} in Appendix~\ref{sec:app-lls} reports the mean test log-likelihood for each dataset.
For single models, \ourlearner\ consistently learns more (or equally) accurate PCs than \learnpsdd\ on 10 datasets out of 20.\footnote{The likelihoods we are reporting for \learnpsdd\ are significantly better than those originally reported in~\citet{LiangUAI17}. Compared to those original results, \ourlearner\ is more accurate than \learnpsdd\ 16 times out of 20.}
More strikingly, \textit{\ourlearner\ delivers equivalently accurate PCs sooner than \learnpsdd}.
This is clearly shown in Figure~\ref{fig:missing-body}~(left \& center) for all datasets: it takes fewer iterations for \ourlearner\ to achieve comparable bpds and each of its iterations takes at least one order of magnitude less time than \learnpsdd.
Lastly, PCs learned with \ourlearner\  are still comparable on many datasets to other PC learners performing local search like selective SPNs (\textsc{selSPN})~\citep{Peharz2014b} and normal SPNs (\textsc{seaSPN})~\citep{Dennis2015}.
Note however, that these competitors learn PCs with less structural requirements (\textsc{selSPNs} are not structured-decomposable and \textsc{seaSPNs} are not deterministic), and hence support less tractable inference scenarios (cf.~Section~\ref{sec:back}). 
Detailed learning times, circuit sizes and statistical tests are reported in Appendix \ref{sec:app-single}.

\textbf{(Q4) Mixtures with \ourlearner.}
We evaluate the shared-structure mixtures of PCs learned by \ourlearner\ by reusing for each dataset the best single PC learned for setting \textbf{Q1}, and performing parameter learning with EM (\ourlearner-EM) or a combination of EM and bagging ({\ourlearner-BEM}) to alleviate overfitting as in~\citet{LiangUAI17}. 
For \ourlearner-EM, 
we choose the number of mixture components $k_{EM}$ by a grid search on \{2, 5, 10, 15, 20, 25, 30\}. 
\ourlearner-BEM instead trains mixtures of PCs using EM on 10 different bagged datasets. 
Table \ref{tab:ensembles} in Appendix \ref{sec:app-ensemble} reports the mean test log-likelihoods of our mixtures and the corresponding ones learned by \learnpsdd.
On 15 datasets \ourlearner-EM outperforms \learnpsdd-EM, while \ourlearner-BEM is more accurate than its counterpart 11 times.
This is remarkable if one notes that PCs in \learnpsdd-(B)EM are allowed to take arbitrary structures and update the structures for each component during learning.
As expected, \ourlearner\ drastically reduces the learning times of large mixtures, as single PCs can be learned much faster, and mixtures can be learned in a fraction of the time by virtue of shared flows (cf.\ Section~\ref{sec:circuit-flows}).

\textbf{(Q5) Effectiveness of flows.} Appendix~\ref{sec:app-fastflows} reports detailed timings comparing our circuit flow approach to classical circuit evaluation for computing mixture likelihoods. The circuit flow approach is orders of magnitude faster (around $10^2$--$10^3$), and up to 4591 times faster on the `msnbc' dataset.

\textbf{(Q6) Advanced probabilistic queries.}
Finally, we evaluate how PCs learned with \ourlearner\ can be exploited for advanced inference scenarios requiring structured decomposability.
We adopt the experimental setting of~\citet{KhosraviNeurips19} aiming to compute the expected predictions of a regressor $r$ w.r.t.\ a generative model represented as a structured-decomposable PC sharing the same vtree of $r$.
We compute the expected predictions of regressors learned on 4 different benchmarks and employ a single PC or a mixture learned by \ourlearner-BEM with 5 bags and a number of EM components cross-validated in \{5, 10, 15, 20\}.
Figure~\ref{fig:missing-body}~(right) and Figure~\ref{fig:missingvalue} in Appendix~\ref{sec:app-exp} show the RMSE of our models for different percentages of missing values, when compared to common imputation schemes like median imputation.
Not only do single PC perform better than the baseline, but the cheap mixtures of PCs with shared structure help further reduce the error.

\section{Conclusions}
\label{sec:conclusions}

In this paper we introduced \ourlearner\ to learn structured-decomposable PCs in a fast and simple way.
\ourlearner\ delivers  accurate single PCs in a fraction of the time of its competitor and effectively scales up to learning mixtures of PCs sharing the same structure.
We consider \ourlearner\  as an initial stepping stone to learn PCs for several application scenarios
where advanced probabilistic inference is required and out-of-the-scope of the current landscape of tractable probabilistic models.

\scalebox{0.01}{quam dulcis sit structura breviter proloquar}
\vspace{-10pt}
\paragraph{Acknowledgments}
This work is partially supported by NSF grants \#IIS-1943641, \#IIS-1633857,
\#CCF-1837129, DARPA grant \#N66001-17-2-4032, a Sloan Fellowship, Intel, and Facebook.

\bibliography{parle-juice}

\clearpage
\appendix

\clearpage
\appendix
\section{Learning a Chow-Liu Tree}
\label{sec:app-clt}
We list the algorithm for learning Chow-Liu Trees~\citep{ChowLiu} here.
\begin{algorithm}[!h]
\SetAlgoLined
  \caption{\textsf{LearnCLT}($\data, \X, \alpha$)}
  \label{algo:learnclt}
    \SetKwInOut{Input}{Input}
    \SetKwInOut{Output}{Output}
    \ResetInOut{Output}
    \Input{a dataset $\data$ over RVs, $\X=\{X_{i}\}_{i=1}^{n}$, Laplace
    smoothing factor $\alpha$}
    \Output{a Chow-Liu tree model $\langle \CLT,\paras=\{\theta_{i|\Pa_{i}}\}_{i=1}^{n}\rangle$ estimating $\p(\X)$}
    $\mathsf{MI}\leftarrow\mathbf{0}_{n\times n}$\\
    \For {\textbf{each} $X_{i},X_{j}\in\mathbf{X}$}{
        $\mathsf{MI}_{ij}\leftarrow
        \mathsf{estimateMI}(\data, X_{i}, X_{j}, \alpha)$
    }
    $T \leftarrow$ \textsf{maximumSpanningTree}($\mathsf{MI}$)\\
    $\CLT \leftarrow \mathsf{traverseTree}(T)$\\
    $\paras \leftarrow\{\theta_{i,\Pa_{i}}\leftarrow\mathsf{estimateCPT}(\data, X_{i},X_{\Pa_{i}},\alpha)\}$\\
    \Return $\langle \CLT, \paras \rangle$
\end{algorithm}

\section{\ourlearner\ Algorithm}
\label{sec:app-learn}
The complete pseudocode for \ourlearner\ is listed in Algorithm \ref{alg:learnstrude}.
\begin{algorithm}[!ht]  
    \caption{\textsf{\ourlearner}$(\data, \X)$}
    \label{alg:learnstrude}
    \SetKwInOut{Input}{Input}
    \SetKwInOut{Output}{Output}
    \ResetInOut{Output}
    \Input{a dataset $\data$ over RVs $\X$}
    \Output{a structured-decomposable PC $\PC$}
    $\CLT \leftarrow$ \textsf{LearnCLT}$(\data, \mathbf{X})$\\
    $\vtree \leftarrow\mathsf{getVTree}(\CLT)$\\
    $\PC\leftarrow\mathsf{compile}(\CLT, \mathcal{V})$\\
    \While{\PC \ is not overfitting}{
        $e_{i,j}^*$ $\leftarrow$ $\argmax_{e_{i,j} \in \mathsf{edges}(\PC)} \mathsf{score}_{\mathsf{eFLOW}}(e_{i,j};\PC, \data)$\\
        $X^{*}$ $\leftarrow$ $\argmax_{X_{k} \in \phi(\PC_{i})} \mathsf{score}_{\mathsf{vMI}}(X_{k}; \PC, \data)$\\
        $\PC \leftarrow  \textsf{SplitOperation}(\PC, e^{*}_{i,j}, X^{*};\data)$\\
        
    }
    \Return $\PC$ 
\end{algorithm}

\section{Circuit Flows for Fast Inference}
\label{sec:app-fastflows}

To empirically show that our shared circuit flows implementation (Section~\ref{sec:circuit-flows}) benefits the efficient evaluation of ensembles, compared a vectorized version of the classical algorithm that evaluates the circuit bottom-up \citep{darwiche2003differential}. We report the time taken to compute likelihoods for an ensemble as a function of the number of components in Table \ref{tab:fast-flows}.

\begin{longtable}[!ht]{cccc}
    \centering
    \includegraphics[width=0.23\textwidth]{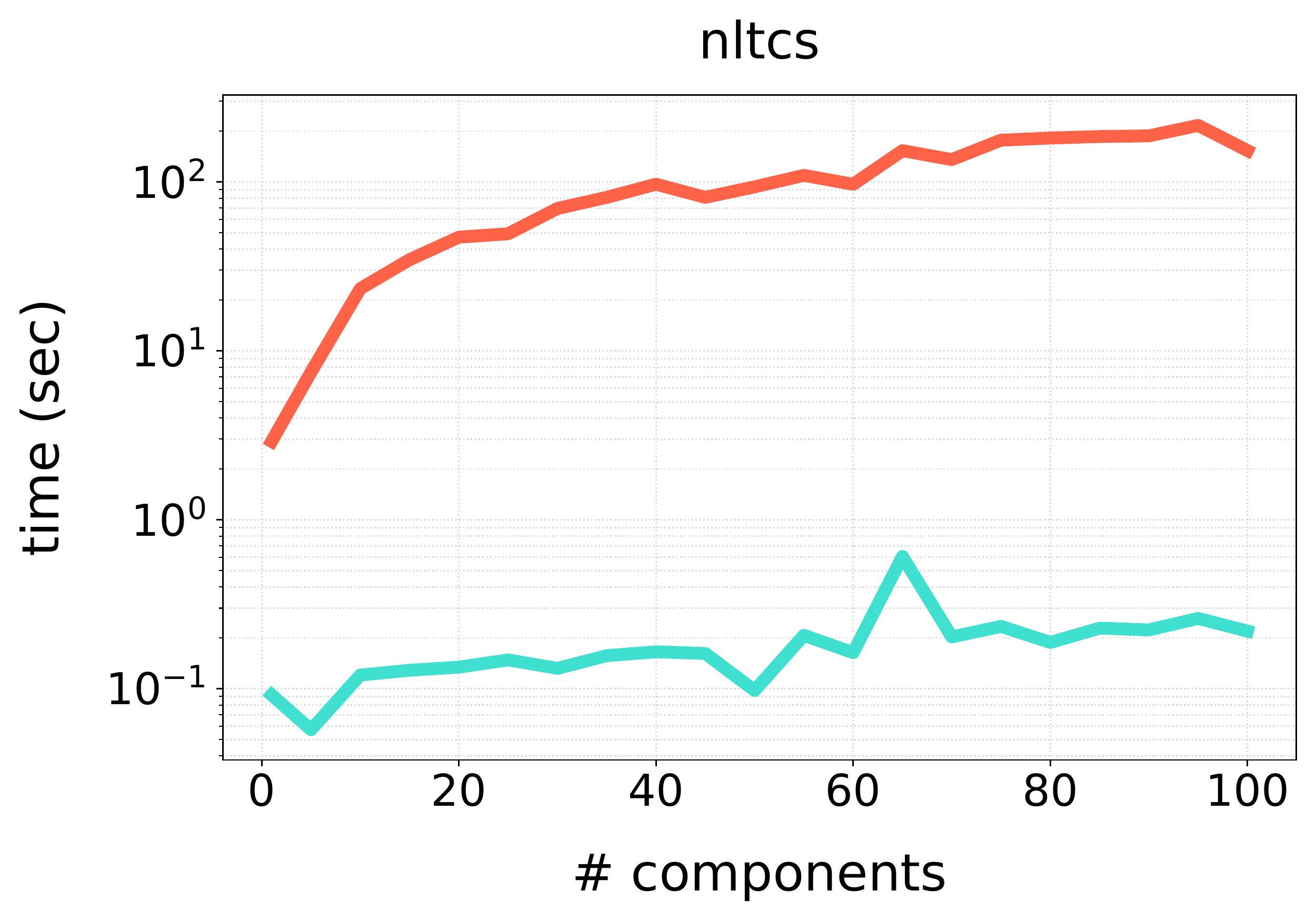}&
    \includegraphics[width=0.23\textwidth]{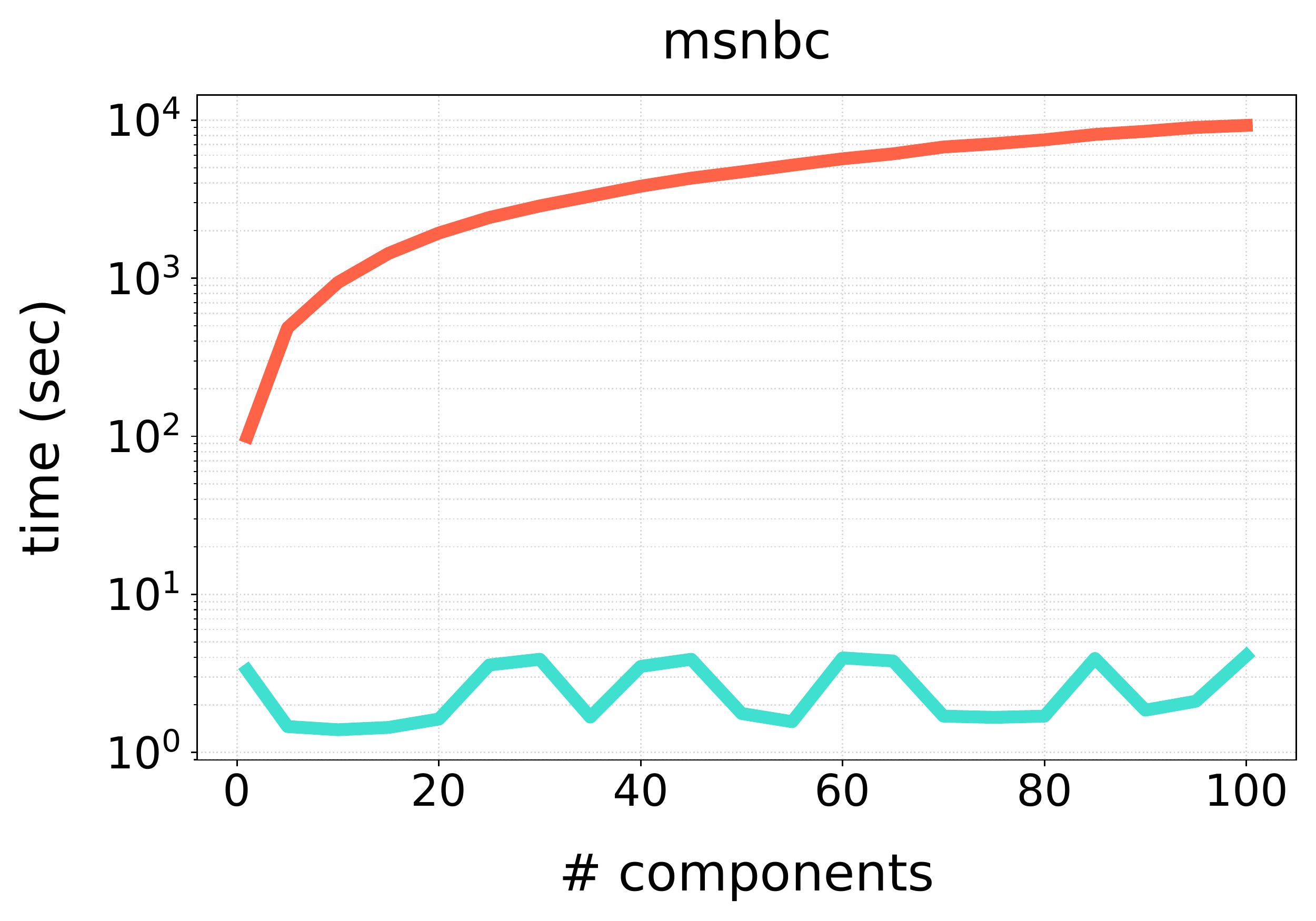}&
    \includegraphics[width=0.23\textwidth]{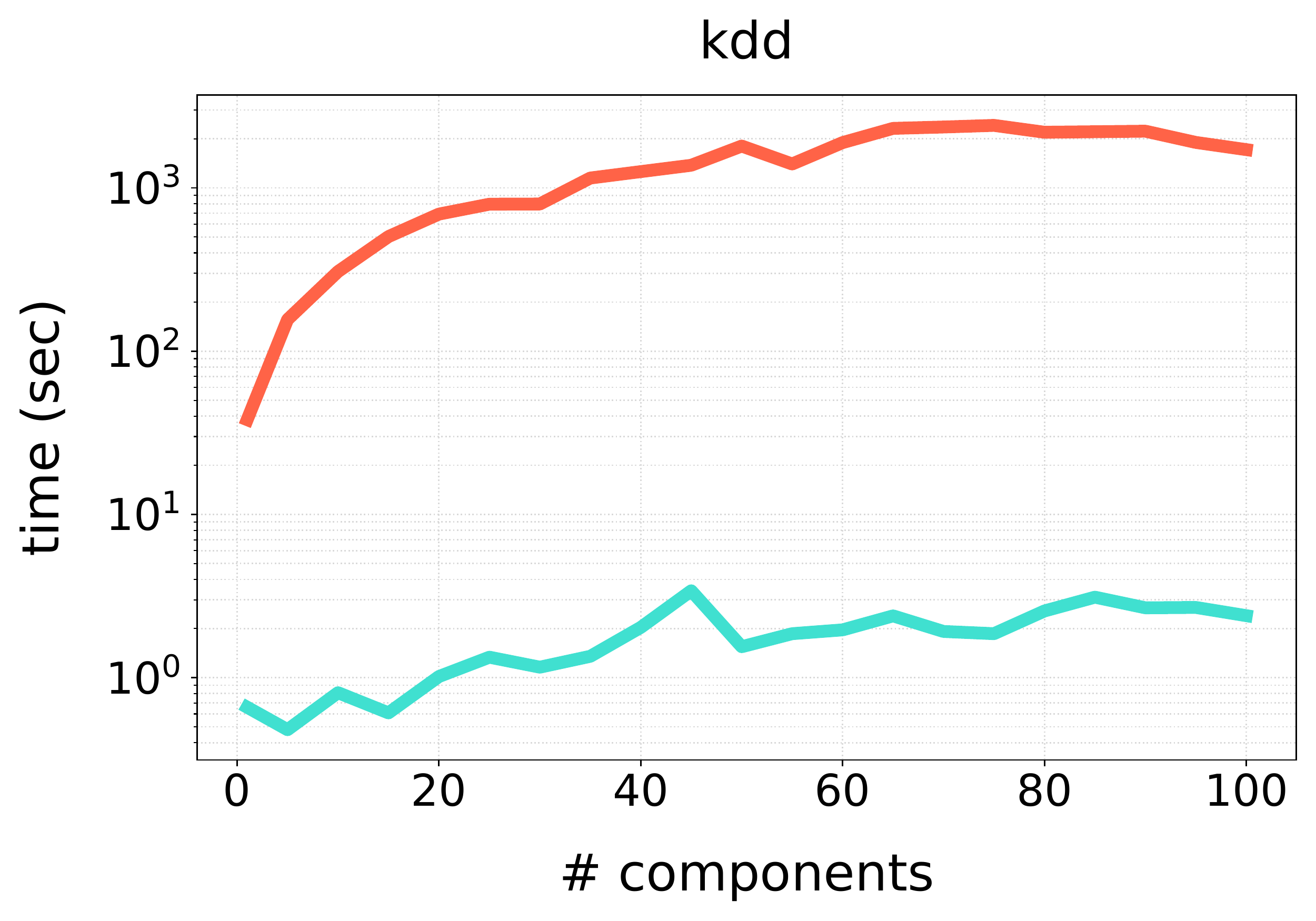}&
    \includegraphics[width=0.23\textwidth]{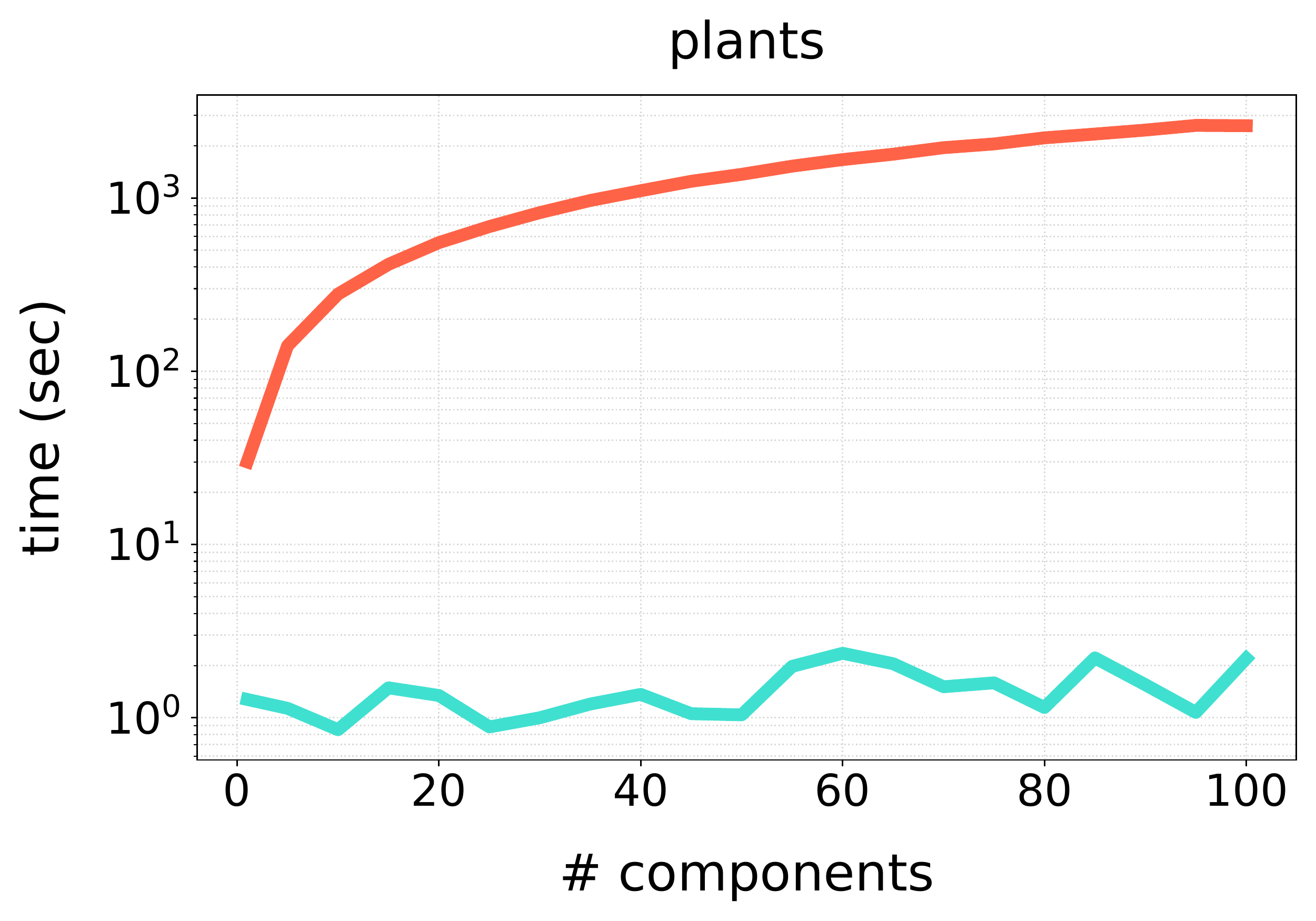}\\
    \includegraphics[width=0.23\textwidth]{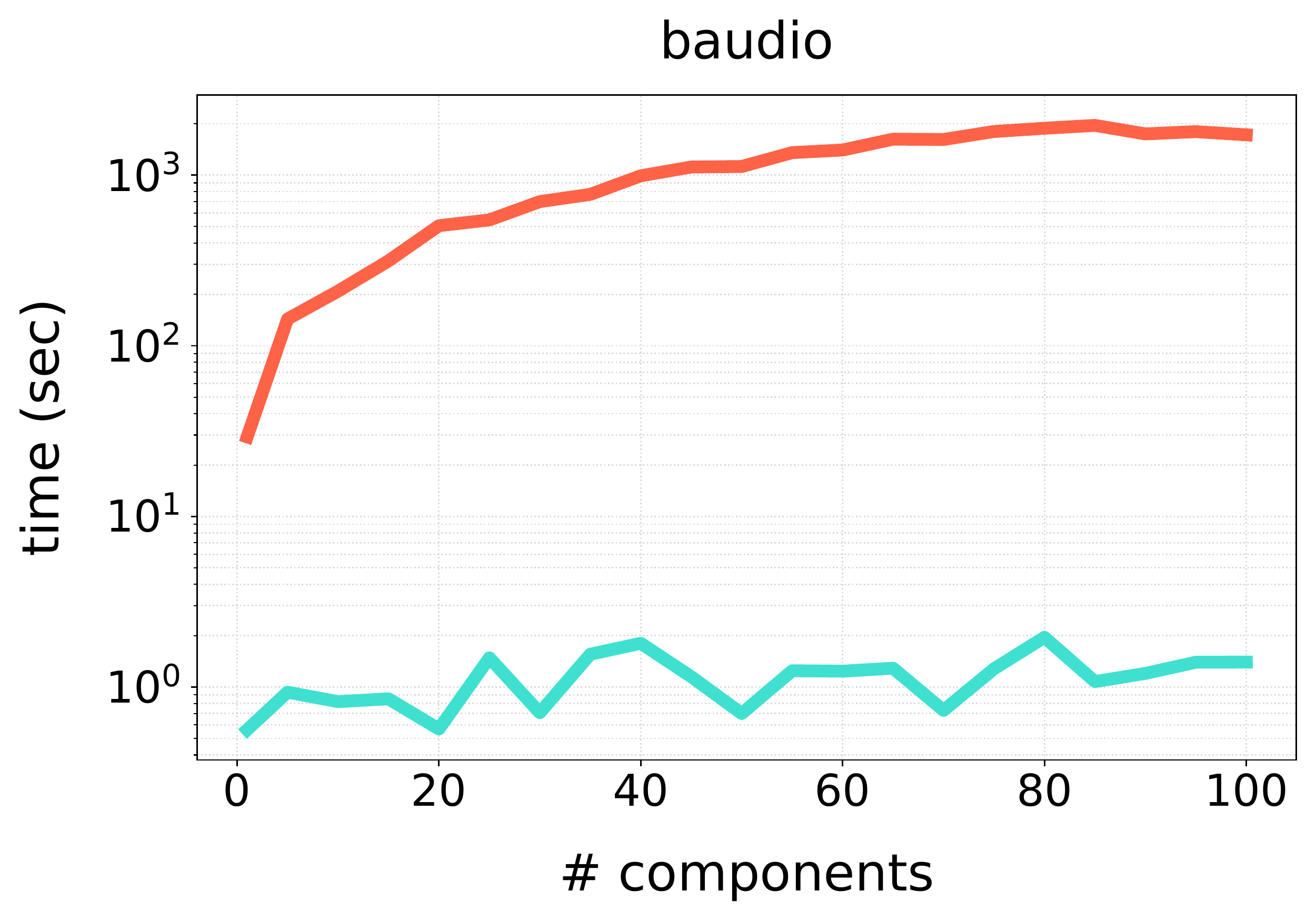}&
    \includegraphics[width=0.23\textwidth]{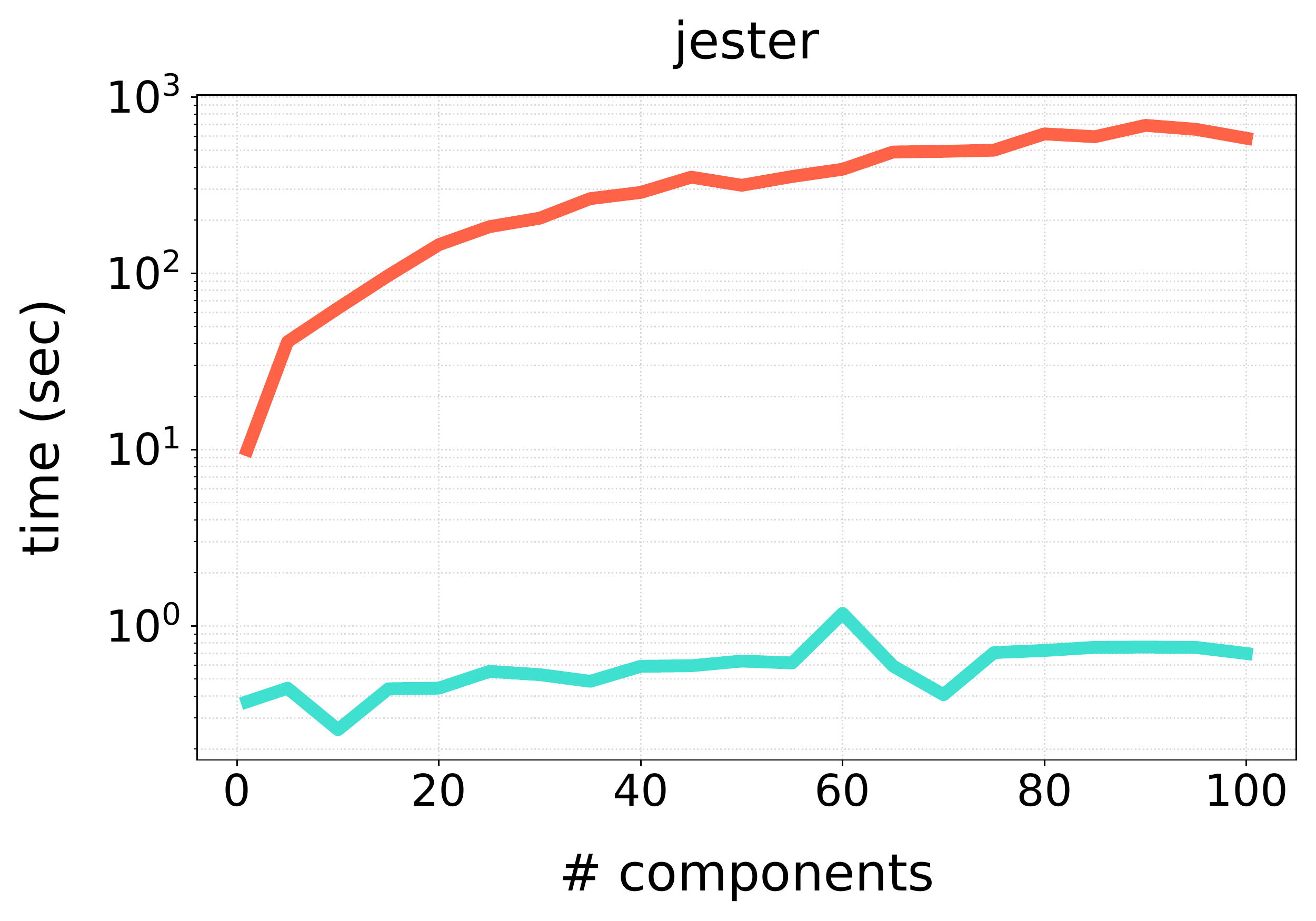}&
    \includegraphics[width=0.23\textwidth]{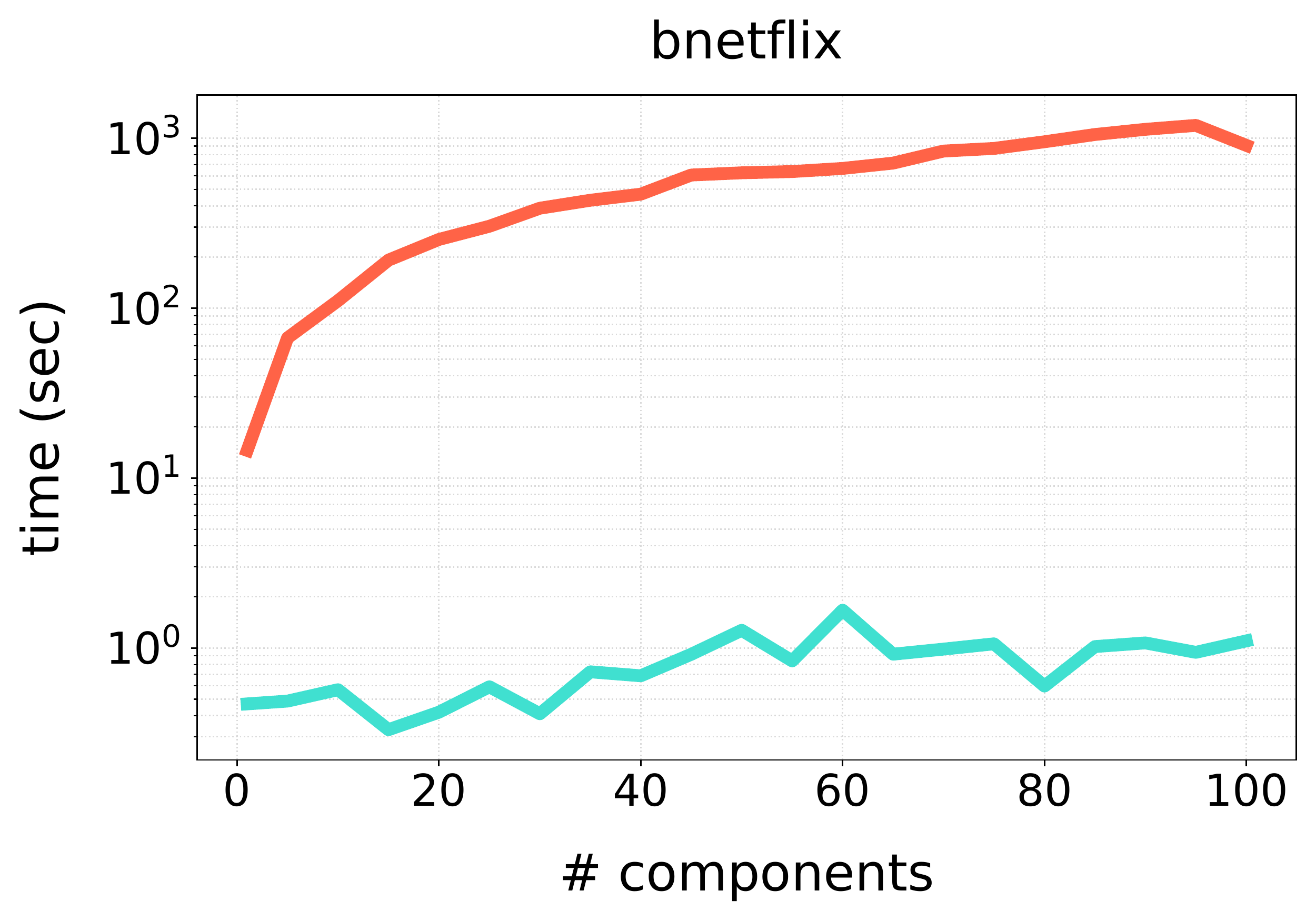}&
    \includegraphics[width=0.23\textwidth]{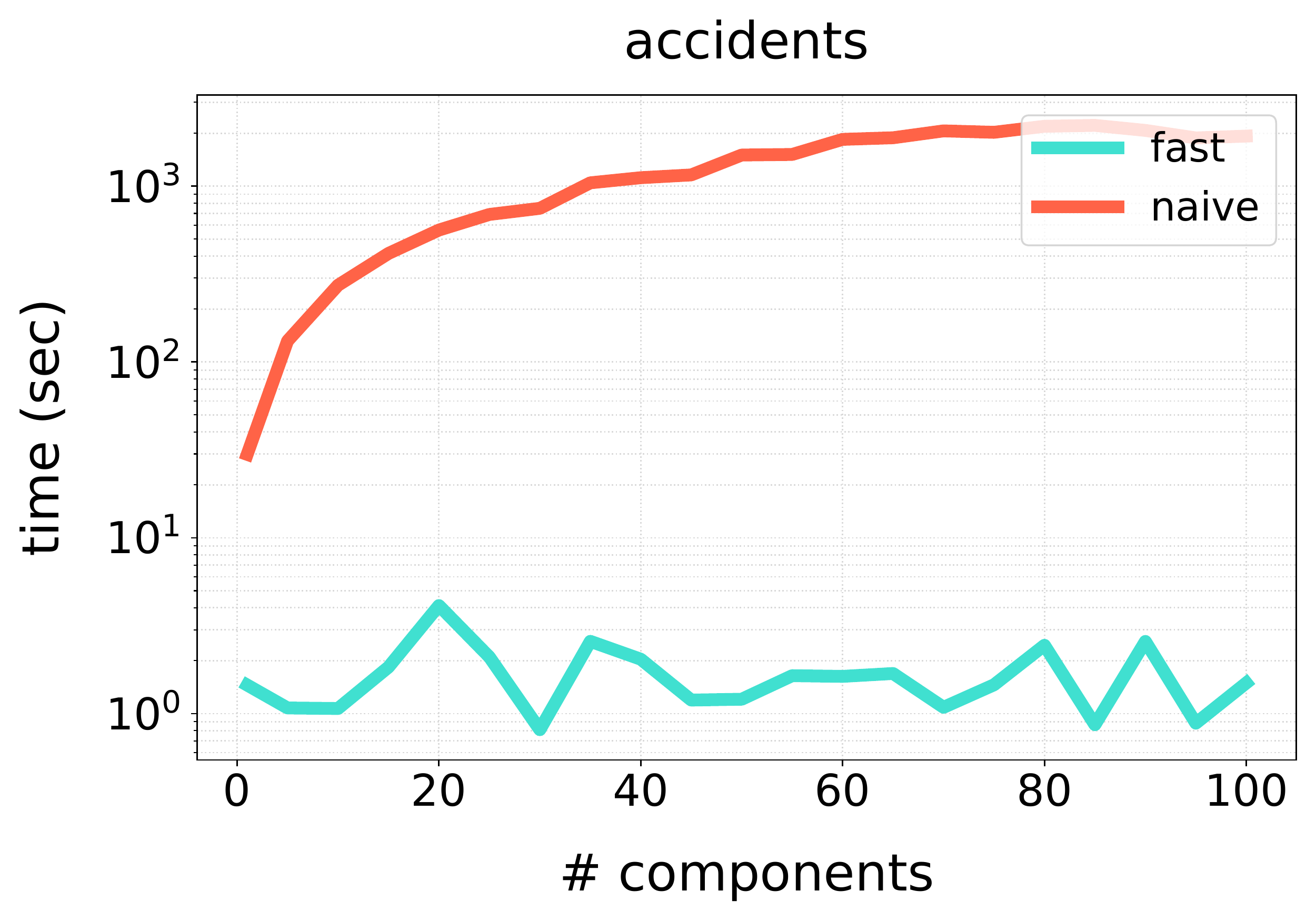}\\
    \includegraphics[width=0.23\textwidth]{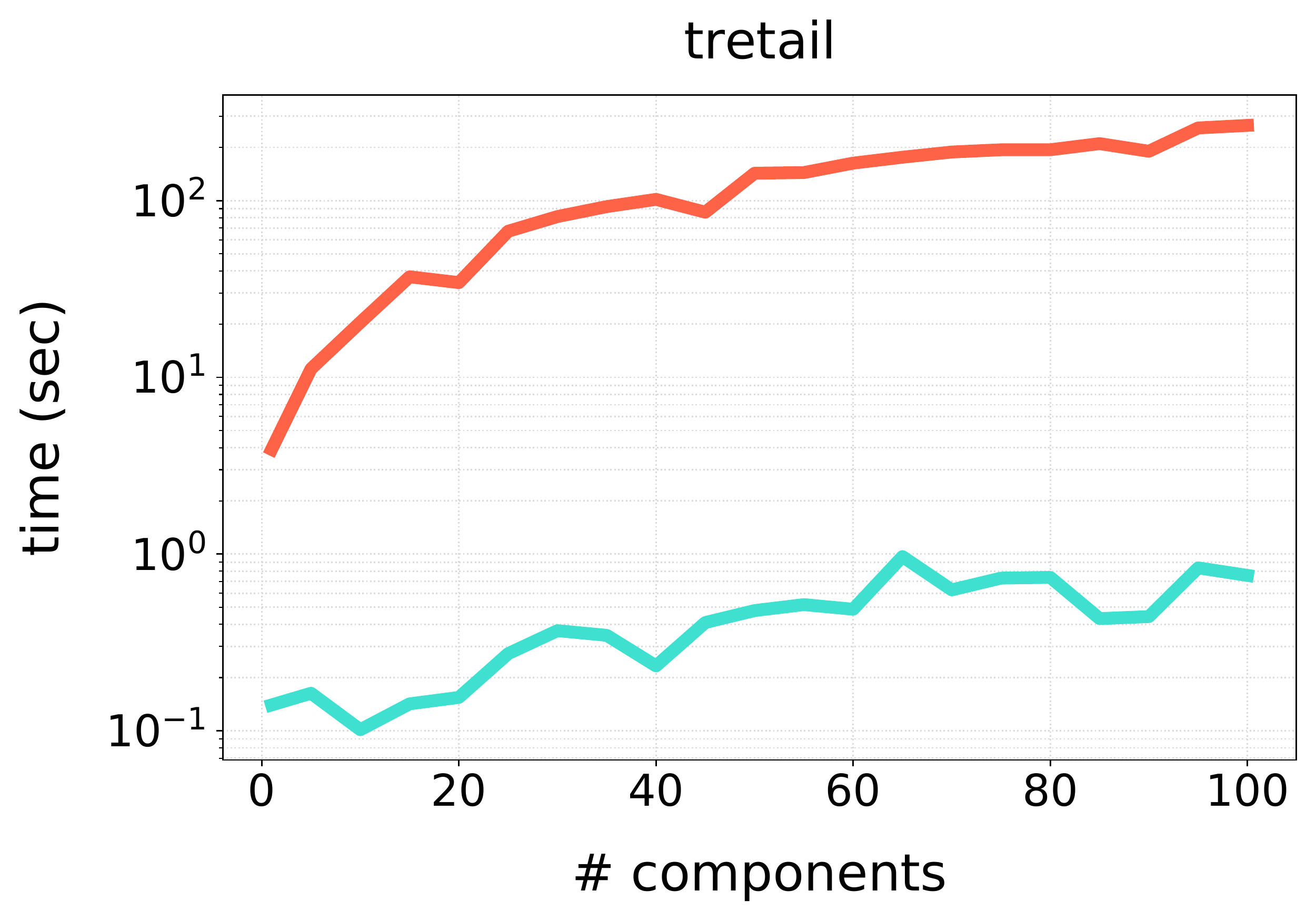}&
    \includegraphics[width=0.23\textwidth]{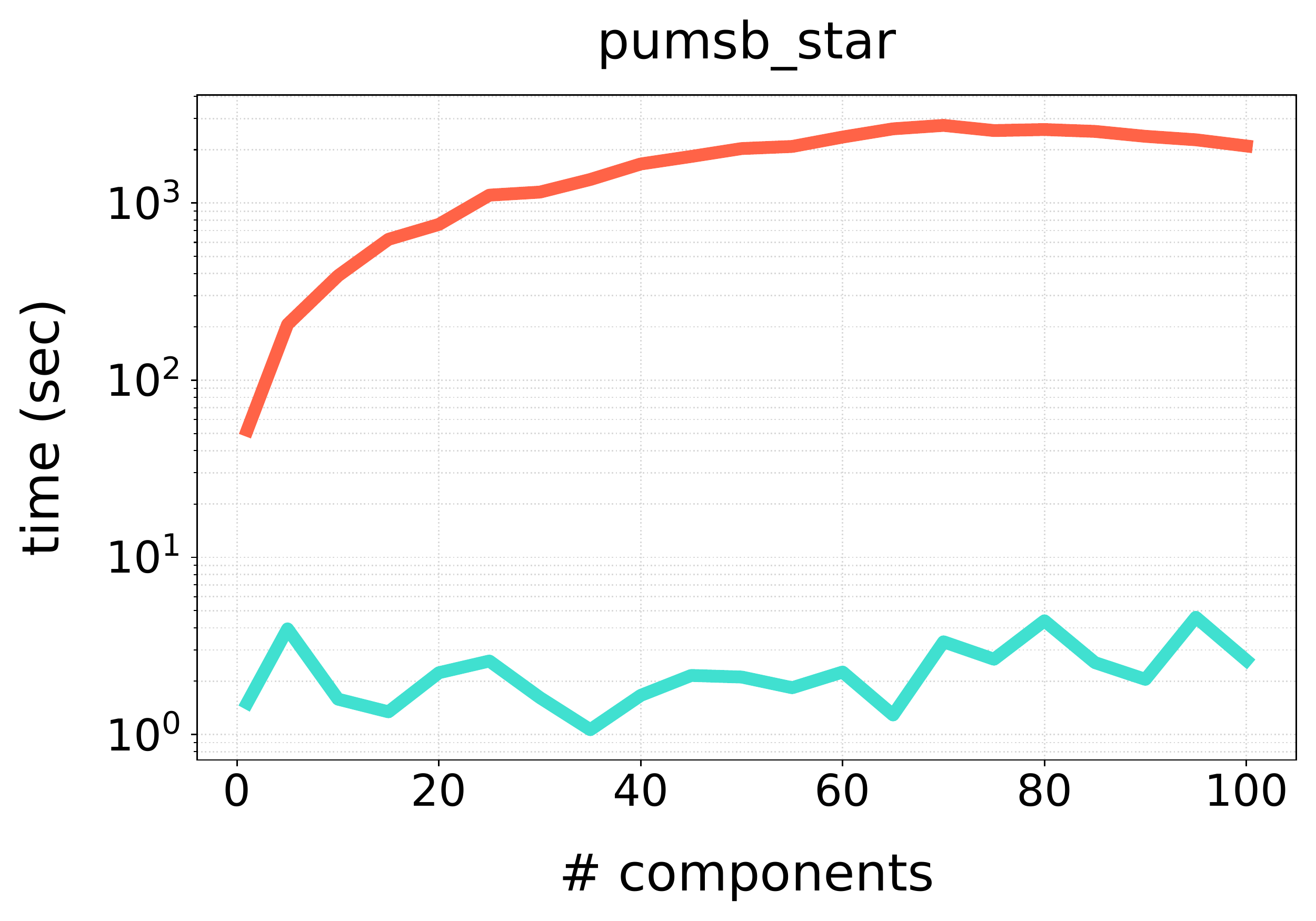}&
    \includegraphics[width=0.23\textwidth]{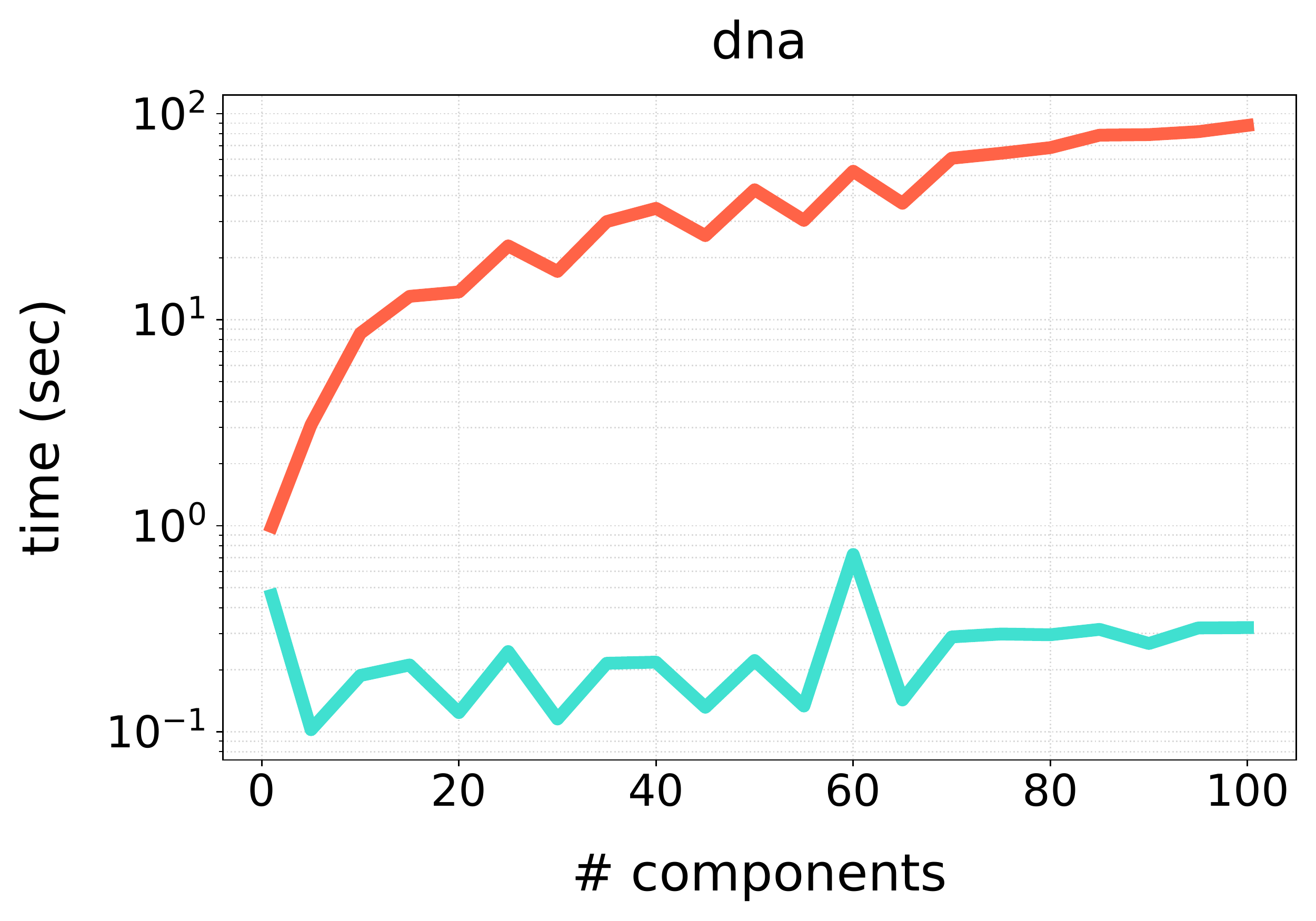}&
    \includegraphics[width=0.23\textwidth]{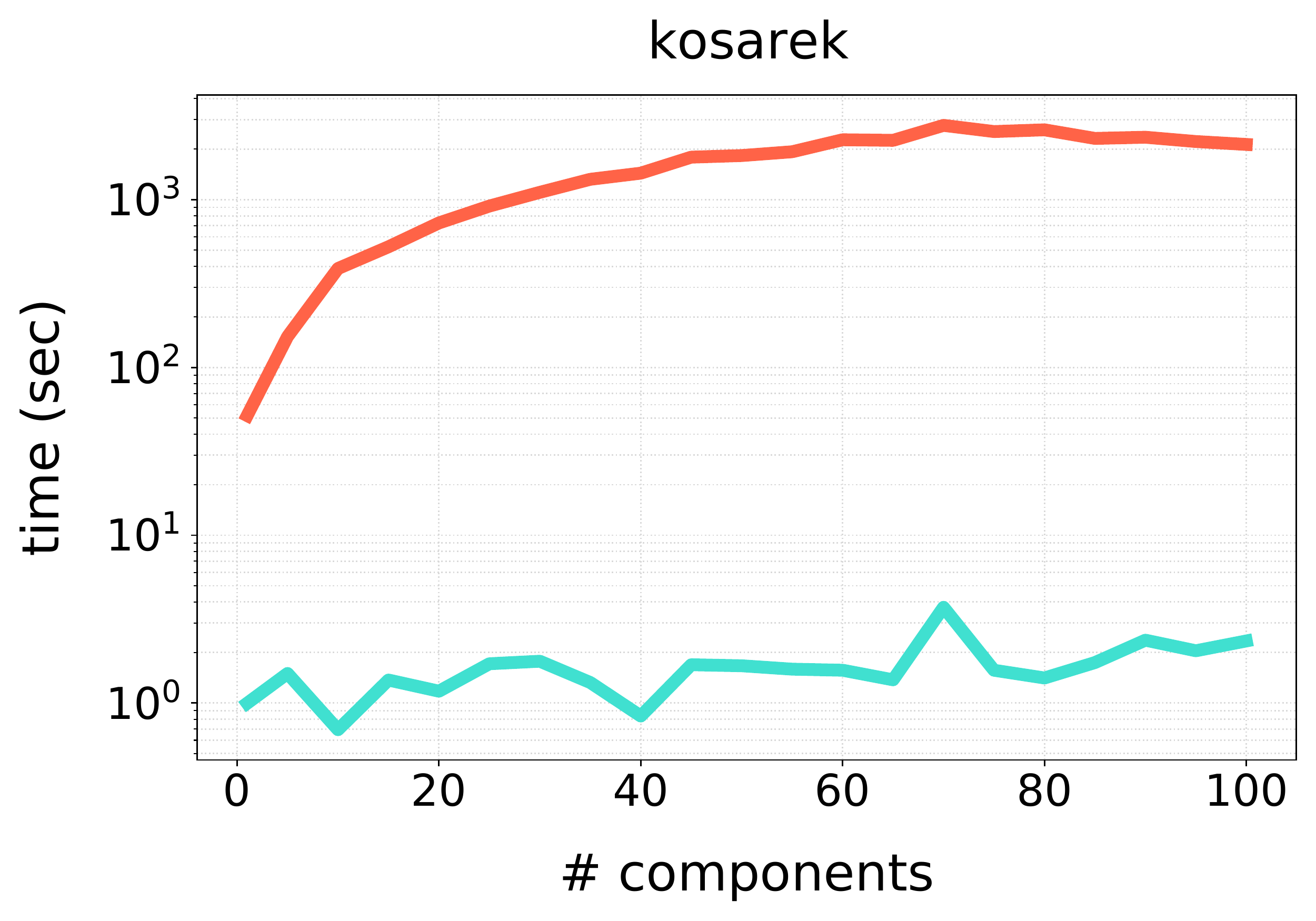}\\
    \includegraphics[width=0.23\textwidth]{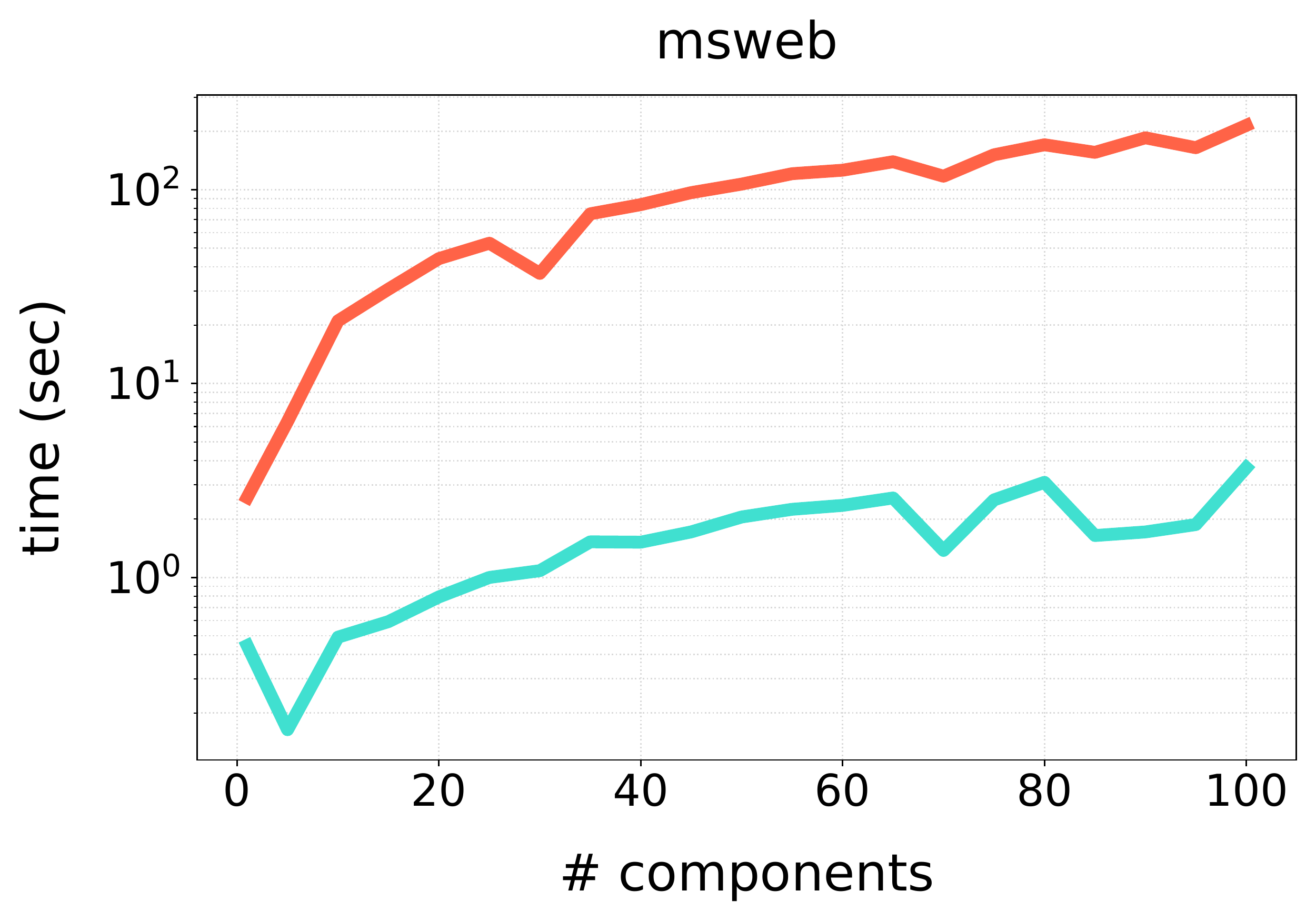}&
    \includegraphics[width=0.23\textwidth]{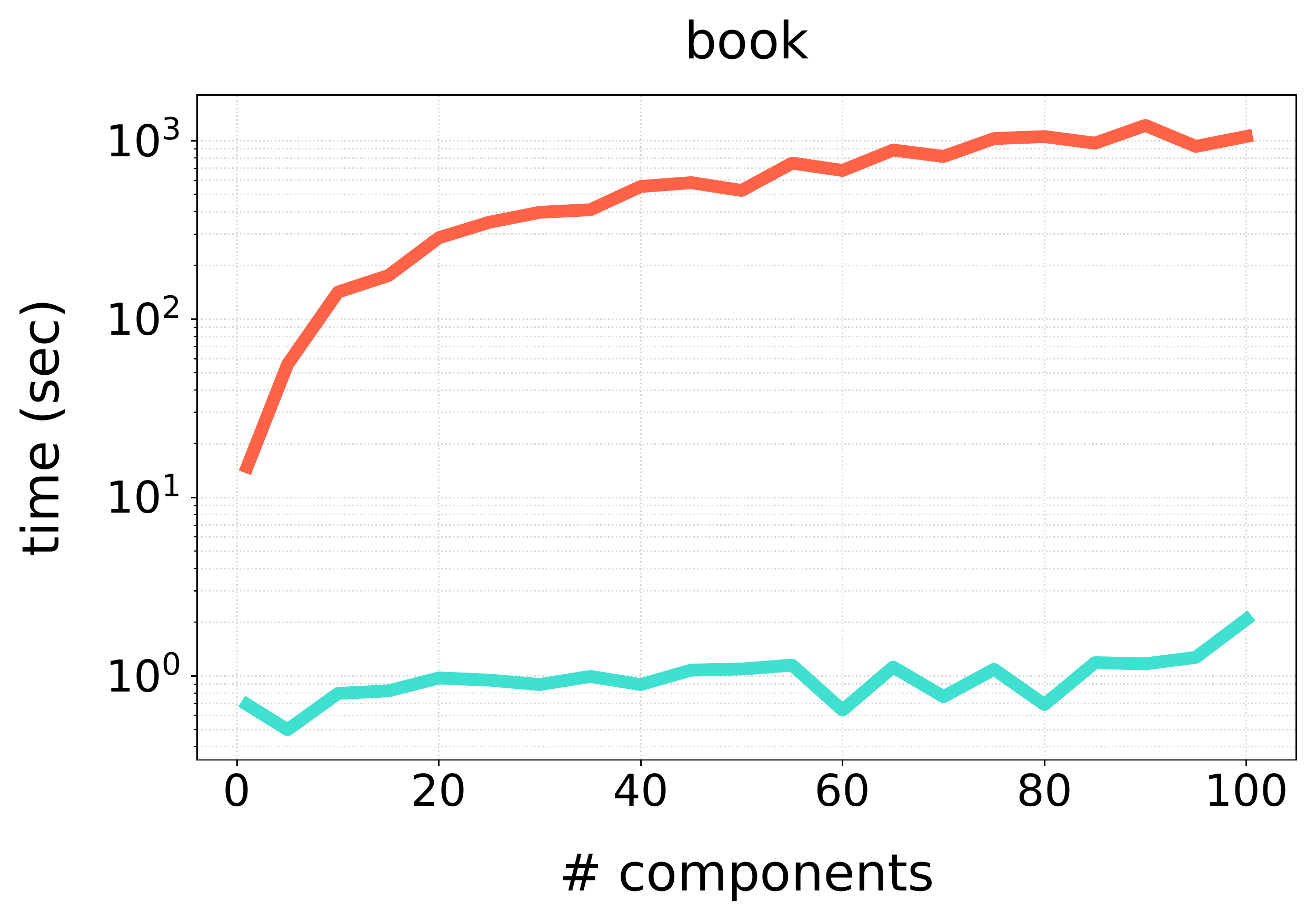}&
    \includegraphics[width=0.23\textwidth]{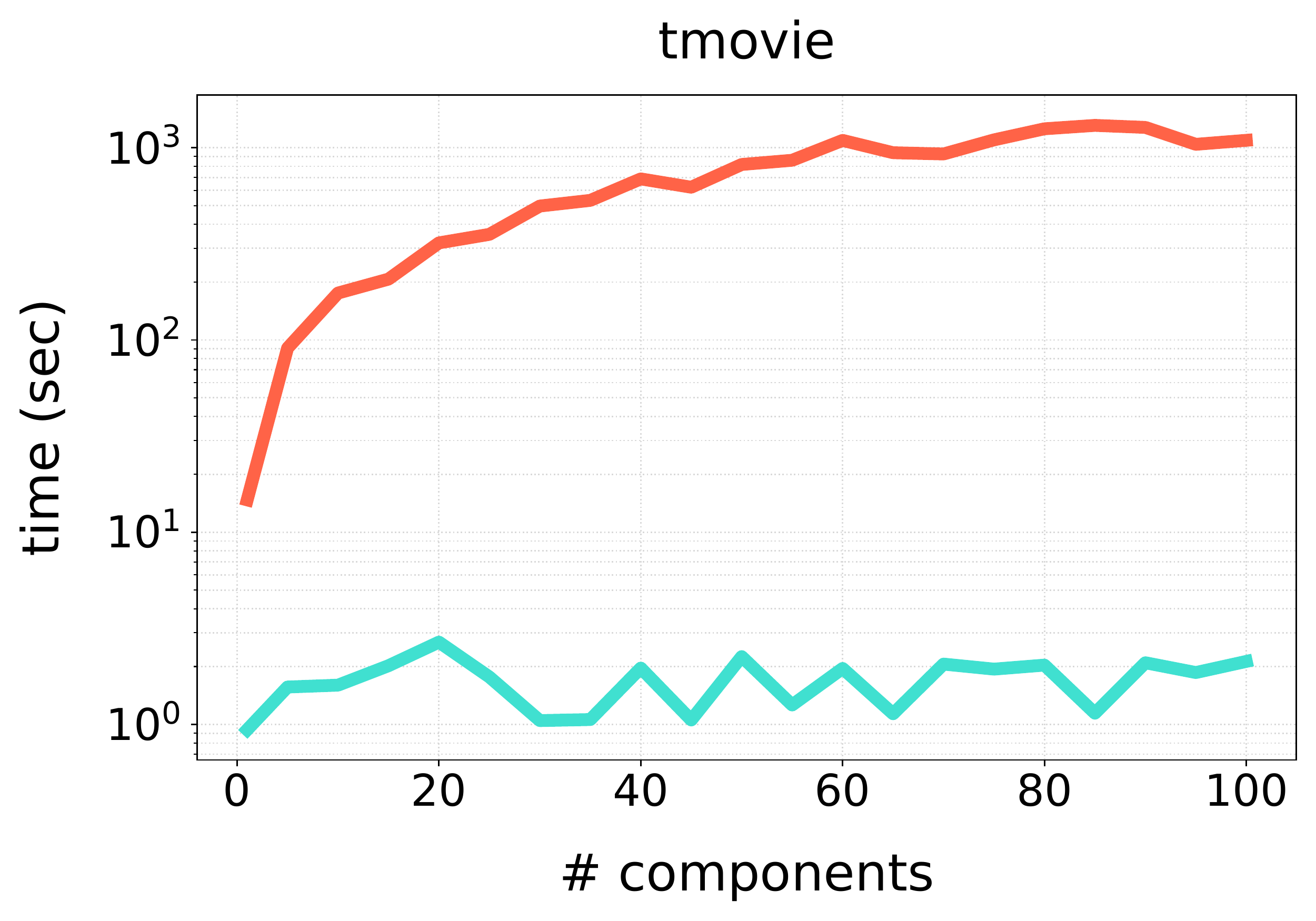}&
    \includegraphics[width=0.23\textwidth]{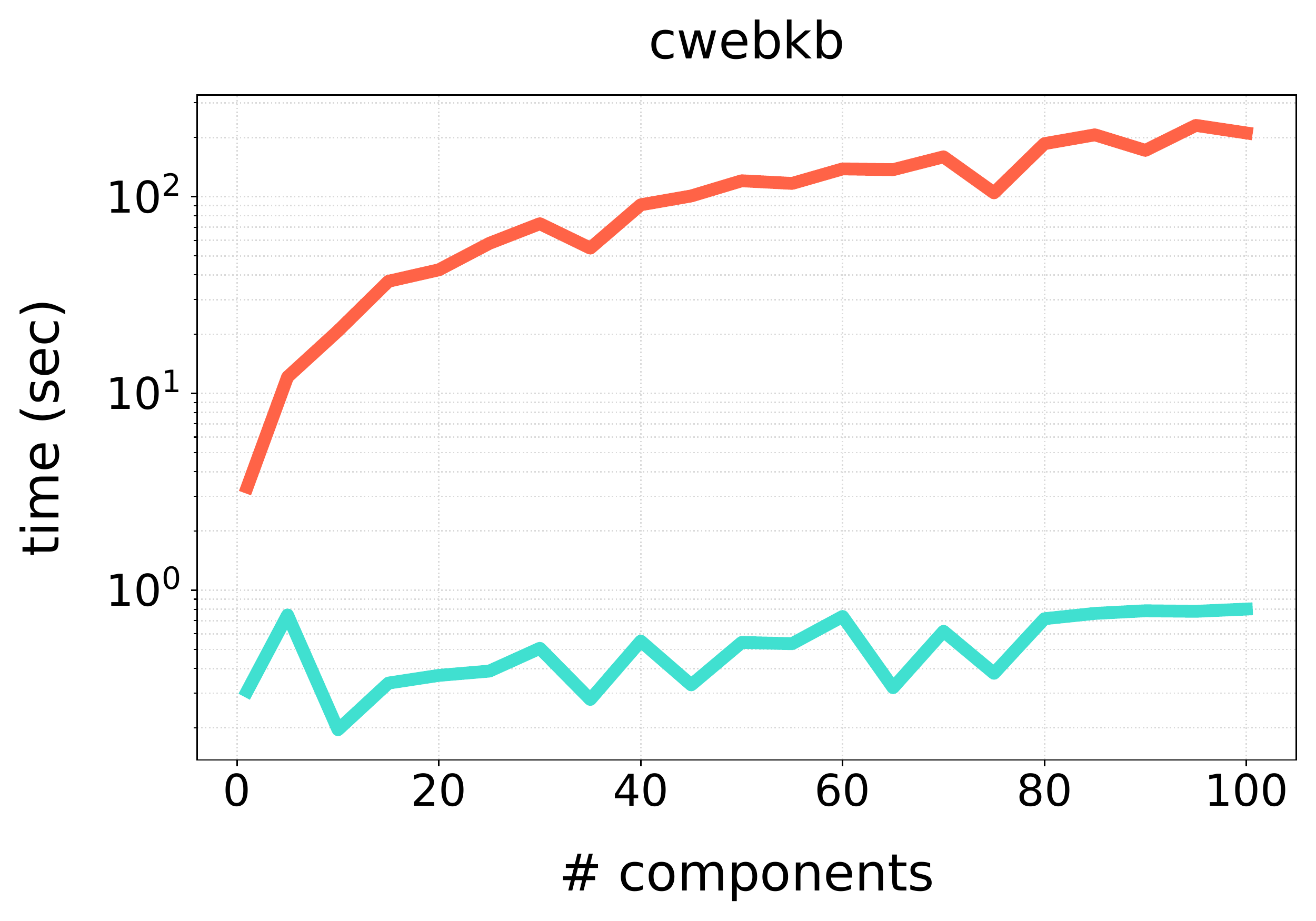}\\
    \includegraphics[width=0.23\textwidth]{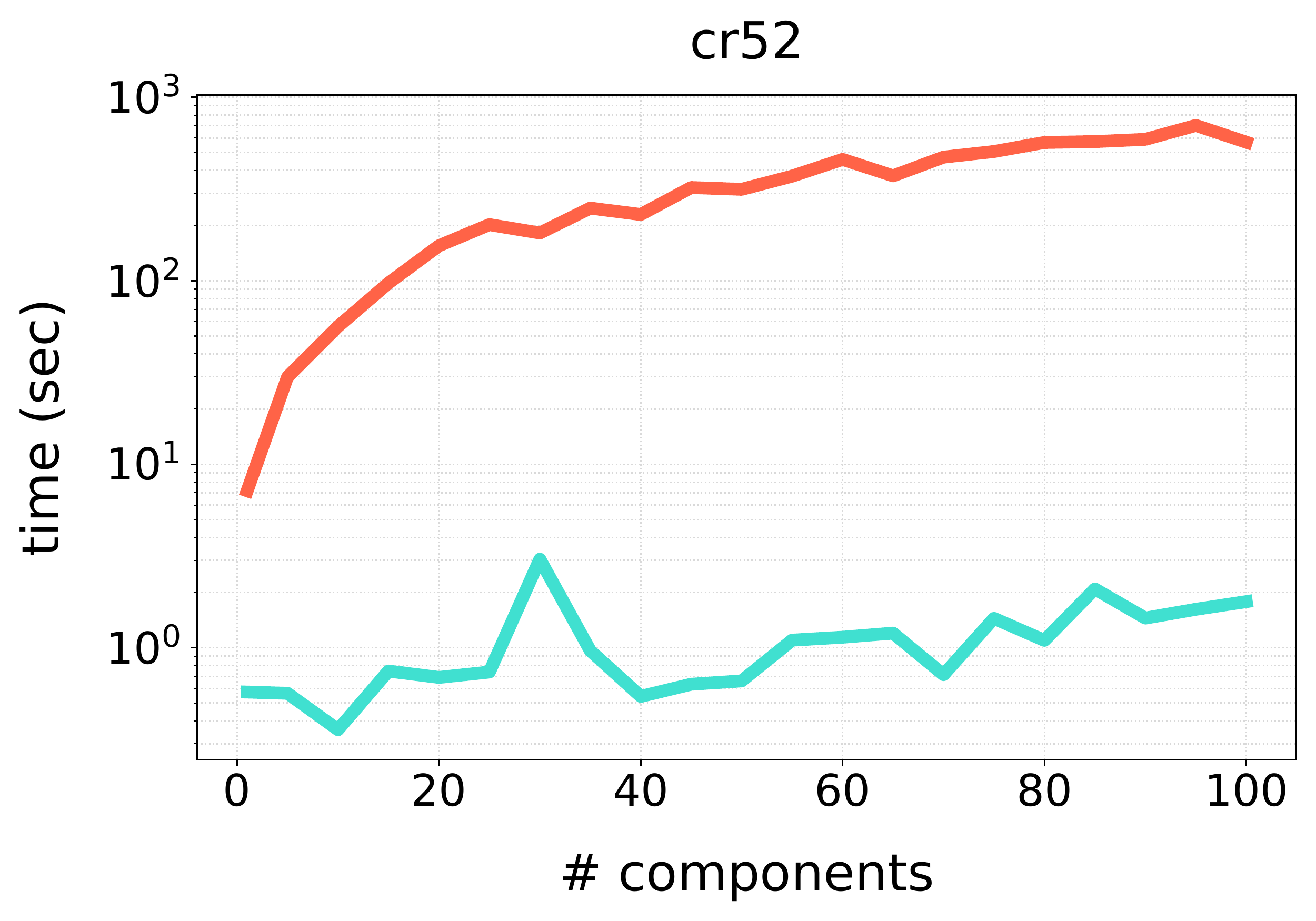}&
    \includegraphics[width=0.23\textwidth]{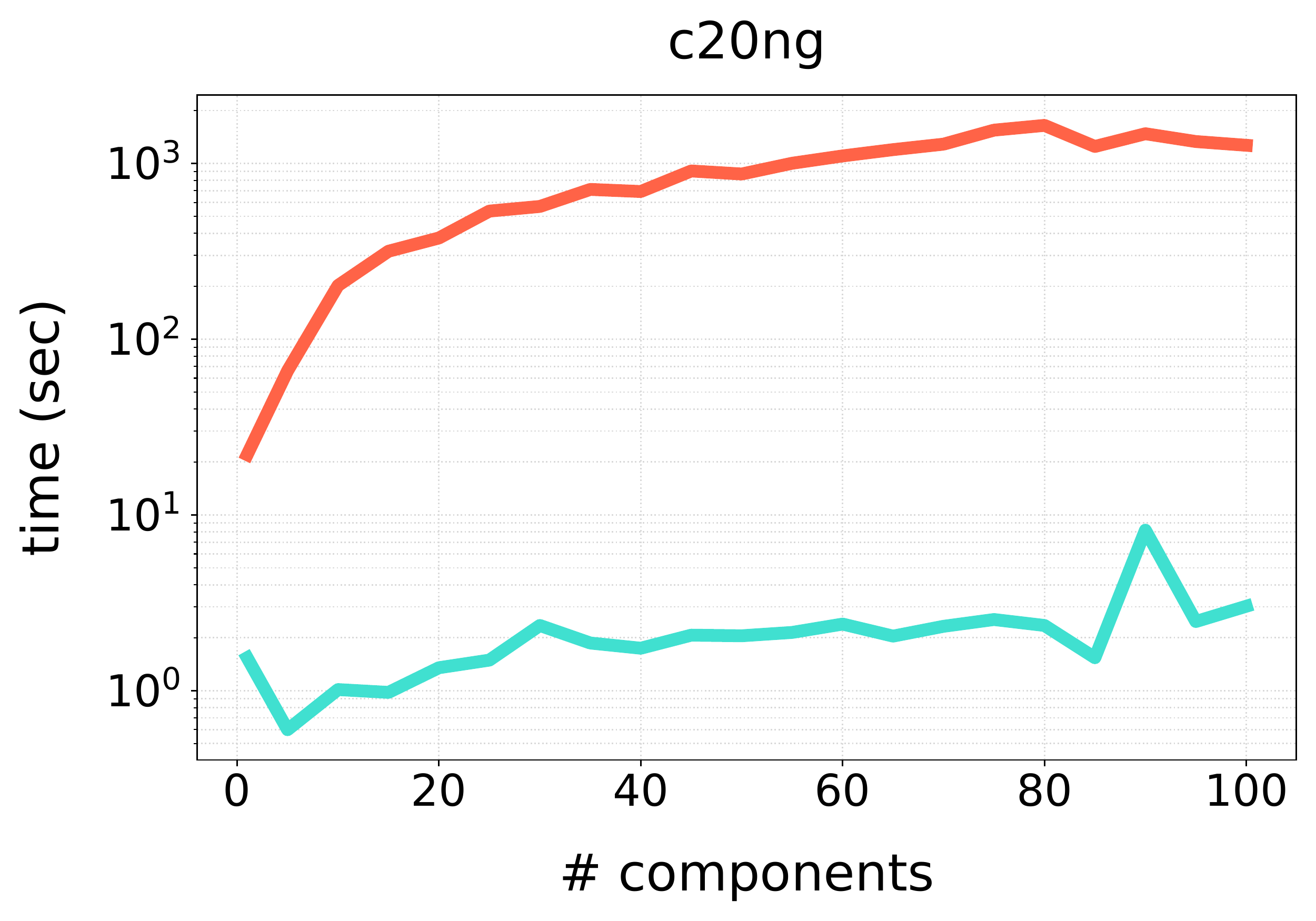}&
    \includegraphics[width=0.23\textwidth]{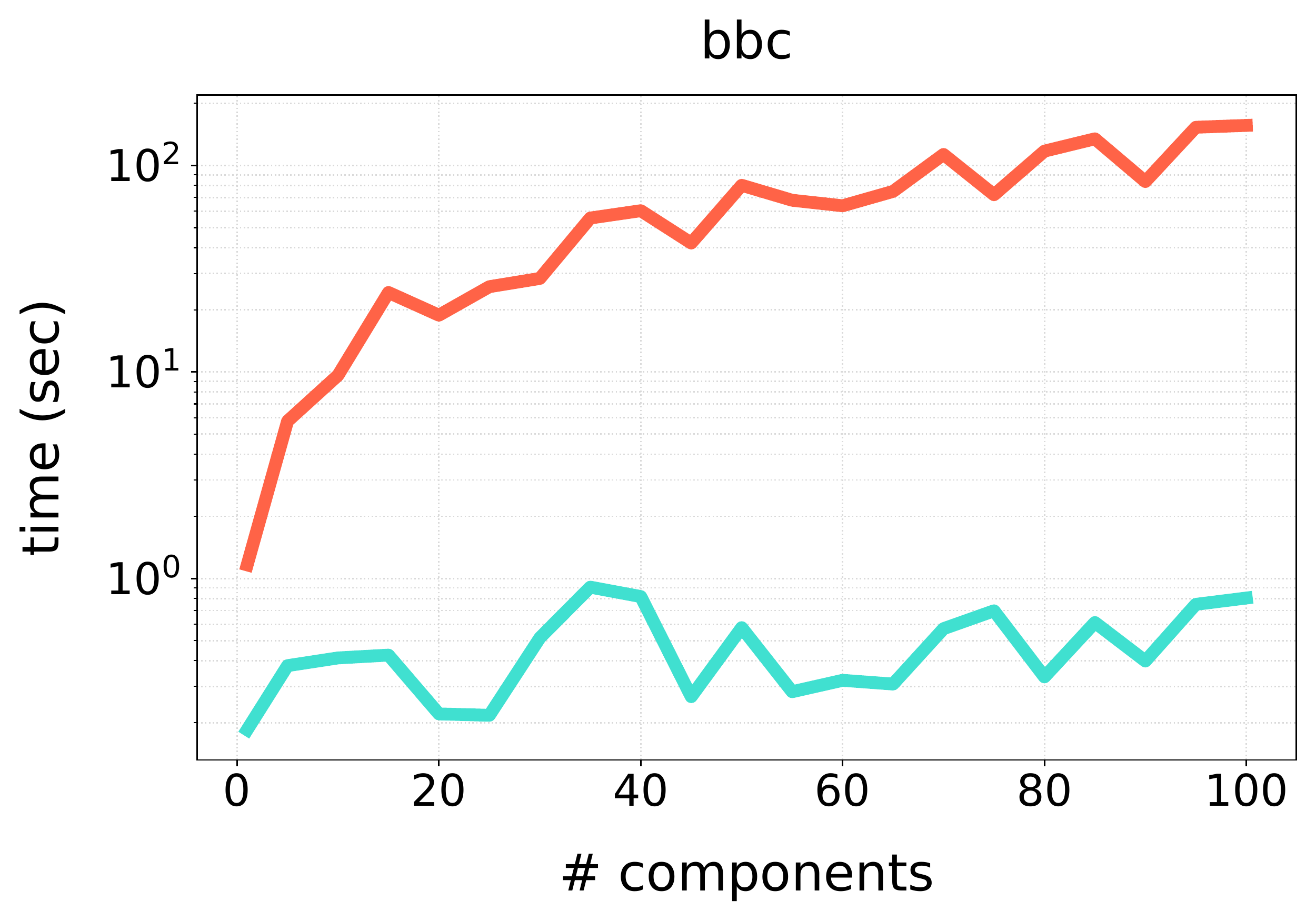}&
    \includegraphics[width=0.23\textwidth]{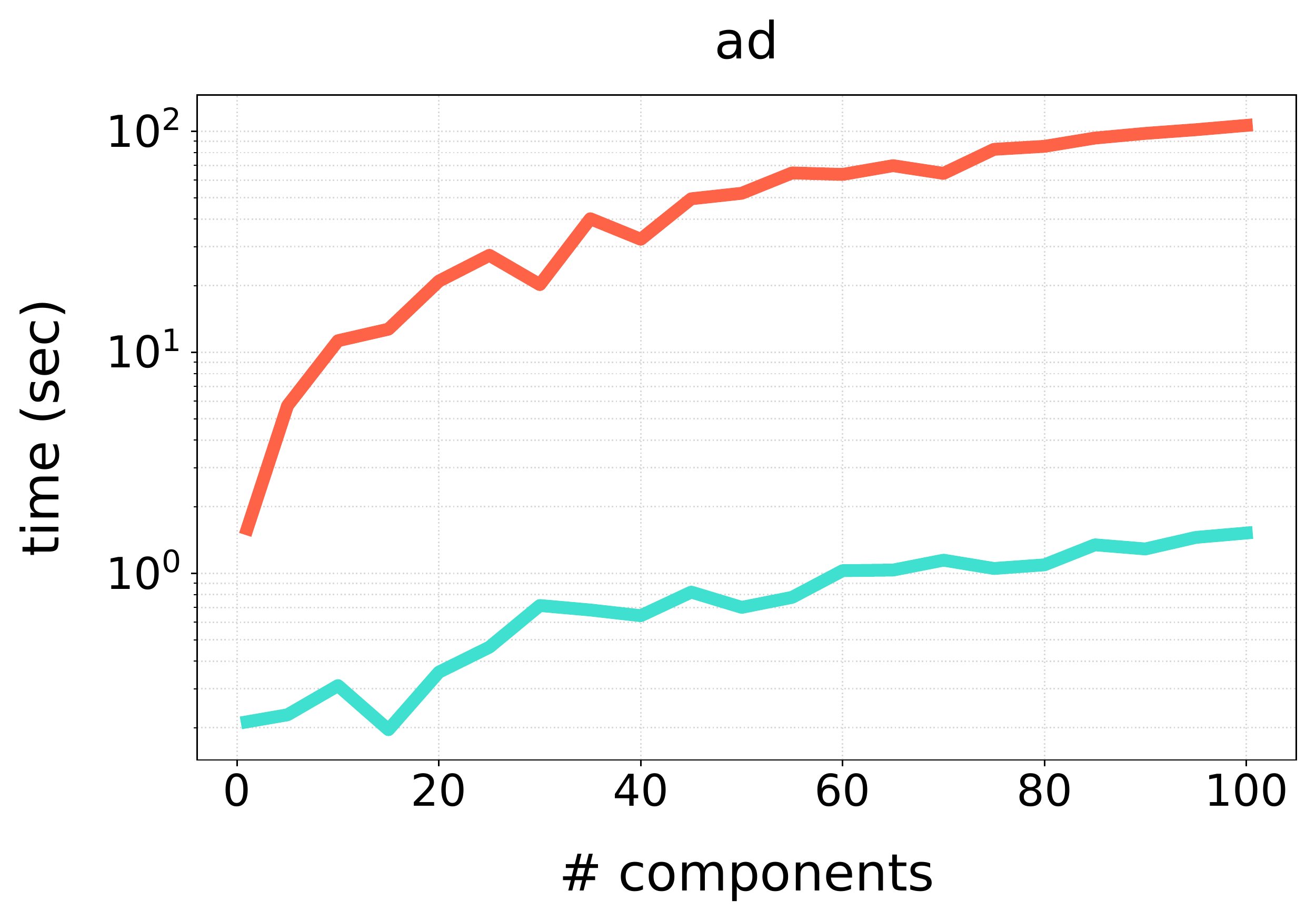}\\
    \caption{\textit{\textbf{Circuit flows for fast inference of ensembles for each dataset.}} We report the time (seconds) used evaluate the ensemble circuits on each dataset (y-axis) for different number of component in the ensemble (x-axis), comparing the circuit flows implementation (blue) and the classical algorithm of naively evaluating the circuits bottom-up (red).}
    \label{tab:fast-flows}
    
\end{longtable}

\newpage
\section{Single Models with \ourlearner}
\label{sec:app-single}
\subsection{Density Estimation Performance}
\label{sec:app-lls}
We compare \ourlearner\ and \learnpsdd\ as density estimators. The average test-set log-likelihood is reported in Table \ref{tab:single-model}.
\begin{table}[!ht]
    \centering
    \small
    \begin{sc}
    \begin{tabular}{r r r ||r r}
    \toprule
         dataset    & \learnpsdd   &\ourlearner  &selSPN   & seaSPN     \\
         \midrule
         nltcs      & -6.03               & \textbf{-6.06}                        & -6.03$\downarrow$& -6.07$\uparrow$\\
         msnbc      & -6.04               & -6.05              & -6.04$\downarrow$& -6.06$\uparrow$\\
         kdd        & -2.17          & -2.17                        & -2.16$\downarrow$& -2.16$\downarrow$\\
         plants     & -13.49             & -13.72            & -12.97$\downarrow$& -13.12$\downarrow$\\
         audio      & -41.51              & -42.26            & -41.23$\downarrow$& -40.13$\downarrow$\\
         jester     & -54.63              & -55.30            & -54.38$\downarrow$& -53.08$\downarrow$\\
         netflix    & -58.53              & -58.68            & -57.98$\downarrow$& -56.91$\downarrow$\\
         accidents  & -28.291              & -29.46            & -26.88$\downarrow$& -30.02$\downarrow$\\
         retail     & -10.92               & \textbf{-10.90}             & -10.88$\downarrow$& -10.97$\uparrow$\\
         pumsb-star & -25.40              & \textbf{-25.28}           & -22.66$\downarrow$& -28.69$\downarrow$\\
         dna        & -83.02               & -87.10           & -80.44$\downarrow$& -81.76$\downarrow$\\
         kosarek    & -10.99          & \textbf{-10.98}            & -10.85$\downarrow$& -11.00$\downarrow$\\
         msweb      & -9.93              & -10.19                    & -9.93$\downarrow$ & -10.25$\downarrow$\\
         book       & -36.06              & \textbf{-35.77}            & -36.01$\downarrow$ & -34.91$\downarrow$\\
         eachmovie  & -55.41              & -59.47                  & -55.73$\downarrow$ & -53.28$\downarrow$\\
         webkb      & -161.42             & \textbf{-160.50}           & -158.52$\downarrow$& -157.88$\downarrow$\\
         routers-52 & -93.30              & \textbf{-92.38}            & -88.48$\downarrow$& -86.38$\downarrow$\\
         20news-grp & -160.43         & \textbf{-160.77}           & -158.68$\downarrow$& -153.63$\downarrow$\\
         bbc        & -260.24             & \textbf{-258.96}           & -259.35$\uparrow$ & -253.13$\downarrow$\\
         ad         & -20.13           & \textbf{-16.52}          & -16.94$\uparrow$  & -16.77$\uparrow$\\
         \bottomrule
    \end{tabular}
    \end{sc}
    \caption{\textit{\textbf{Density estimation benchmarks: single models}}. Average test log-likelihood for \ourlearner\ and \learnpsdd\ models. The bold values indicate \ourlearner\ is better than or statistically equivalent with
    (cf.~Appendix~\ref{sec:app-tests}) \learnpsdd. 
    On the right other state-of-the-art structure learners, which are \textit{not} targeting structured-decomposable circuits (see text). $\uparrow$ (resp. $\downarrow$) indicates that \ourlearner\ is more accurate (resp. less accurate).}
    \label{tab:single-model}
\end{table}{}
\newpage

\subsection{Circuit Sizes}
\label{sec:app-circuit-sizes}
We compare the circuit size of models learned by \learnpsdd\ and \ourlearner\ in Table \ref{tab:sizes}. As expected, \ourlearner\ delivers larger circuits but almost always on the same order of magnitude, which makes sense since \ourlearner\ does not penalize on the circuit size. 

\begin{table}[!ht]
    \centering
    \setlength{\tabcolsep}{2pt}
    \small
    \begin{sc}
    \begin{tabular}{r r r }
    \toprule
         dataset    & \learnpsdd   &\ourlearner    \\
         \midrule
         nltcs      & 1304          & 4373                       \\
         msnbc      & 5465          & 20795            \\
         kdd        & 2915         & 6572            \\
         plants     & 11583         & 119194           \\
         audio      & 18208          & 55675           \\
         jester     & 11322          & 27713           \\
         netflix    & 10997          & 27173          \\
         accidents  & 8418          & 75363          \\
         retail     & 2989           & 3979            \\
         pumsb-star & 8298          & 108876          \\
         dna        & 3068          & 17507           \\
         kosarek    & 7173          & 37583           \\
         msweb      & 6581         & 2347          \\
         book       & 10978          & 54839          \\
         eachmovie  & 20648          & 123996      \\
         webkb      & 11033          & 25406      \\
         routers-52 & 10410          & 36343         \\
         20news-grp & 15793          & 58749      \\
         bbc        & 12335          & 29532       \\
         ad         & 12238           & 13152       \\
         \bottomrule
    \end{tabular}
    \end{sc}
    \caption{\textit{\textbf{Density estimation benchmarks: single models}}. The circuit sizes of best learned single models for \learnpsdd\ and \ourlearner. 
    As expected, \ourlearner\ learns larger PCs than \learnpsdd\ since the latter explicitly penalizes larger sizes via its likelihood score. 
    Nevertheless, the increase in size is contained with the \vMI\ heuristics (cf.~Figure~\ref{fig:heuristics}~(right)) and is a negligible price to pay for a much faster learner (cf.~Table~\ref{tab:single-seconds}).}
    \label{tab:sizes}
\end{table}{}

\subsection{Learning Times}
\label{sec:app-learn-times}
To compare the efficiency of \learnpsdd\ and \ourlearner, we reproduce the experimental setting of \learnpsdd\ and rerun the single model experiments with the latest version of the \learnpsdd\ code. Here we report the learning times -- both the seconds per iteration (\begin{sc}sec per iter\end{sc}) and total seconds during learning (\begin{sc}sec sum\end{sc}) -- in Table \ref{tab:single-seconds}, from which it is clear that \ourlearner\ is more efficient than~\learnpsdd.

\newpage
\begin{table}[!ht]
    \centering
    \setlength{\tabcolsep}{1.8pt}
    \small
    \begin{sc}
    \begin{tabular}{r r r | r r r r}
    \toprule
         dataset    & LearnPSDD             & \ourlearner   & LearnPSDD     & \ourlearner\\
                    & sec/iter          & sec/iter  & sec tot       &sec tot\\
         \midrule
         nltcs      &1.3                  & 0.1         &193.0        &42.1\\
         msnbc      &18.6                 &0.5          &12573.9      &1930.3\\
         kdd        &7.5                  &0.4          &2325.3       &131.4\\
         plants     &31.2                 &1.3          &27438.4      &1911.3\\
         audio      &59.0                 &0.8          &98457.1      &848.4\\
         jester     &23.0                 &0.4          &23573.1      &279.6\\
         netflix    &52.3                 &0.3          &51300.7      &436.4\\
         accidents  &54.0                 &0.8          &29664.2      &3393.9\\
         retail     &66.5                &0.1          &16018.3      &44.2\\
         pumsb-star &83.7                 &1.1          &40508.6      &4737.3\\
         dna        &3.0                &0.5          & 595.4       &86.6\\
         kosarek    &45.8                 &0.4          &30196.8      &398.9\\
         msweb      &176.6                &4.3          &93589.7      &12.9\\
         book       &28.2                 &1.1          &32567.2      &2316.8\\
         eachmovie  &31.1                 &0.7          &61792.3      &1427.2\\
         webkb      &5.6                  &0.4          &6333.0       &884.3\\
         routers-52 &93.2                 &0.5          &98796.0      &1561.0\\
         20news-grp &34.6                 &0.7          &53829.2      &2593.2\\
         bbc        &6.2                  &0.9          &8000.8       &2136.3\\
         ad         &3.7                   &1.1          &3748.6       &300.8\\          \bottomrule
    \end{tabular}
    \end{sc}
    \caption{Total times (in seconds) taken (\textsc{sec tot}) and averaged times per iteration (\textsc{sec/ iter}) to learn the best single models on each dataset for \learnpsdd\ and \ourlearner. \ourlearner\ requires a fraction of the time of \learnpsdd\ greatly speeding up learning.}
    \label{tab:single-seconds}
\end{table}{}

\begin{longtable}[!ht]{cccc}
    \centering
    \includegraphics[width=0.23\textwidth]{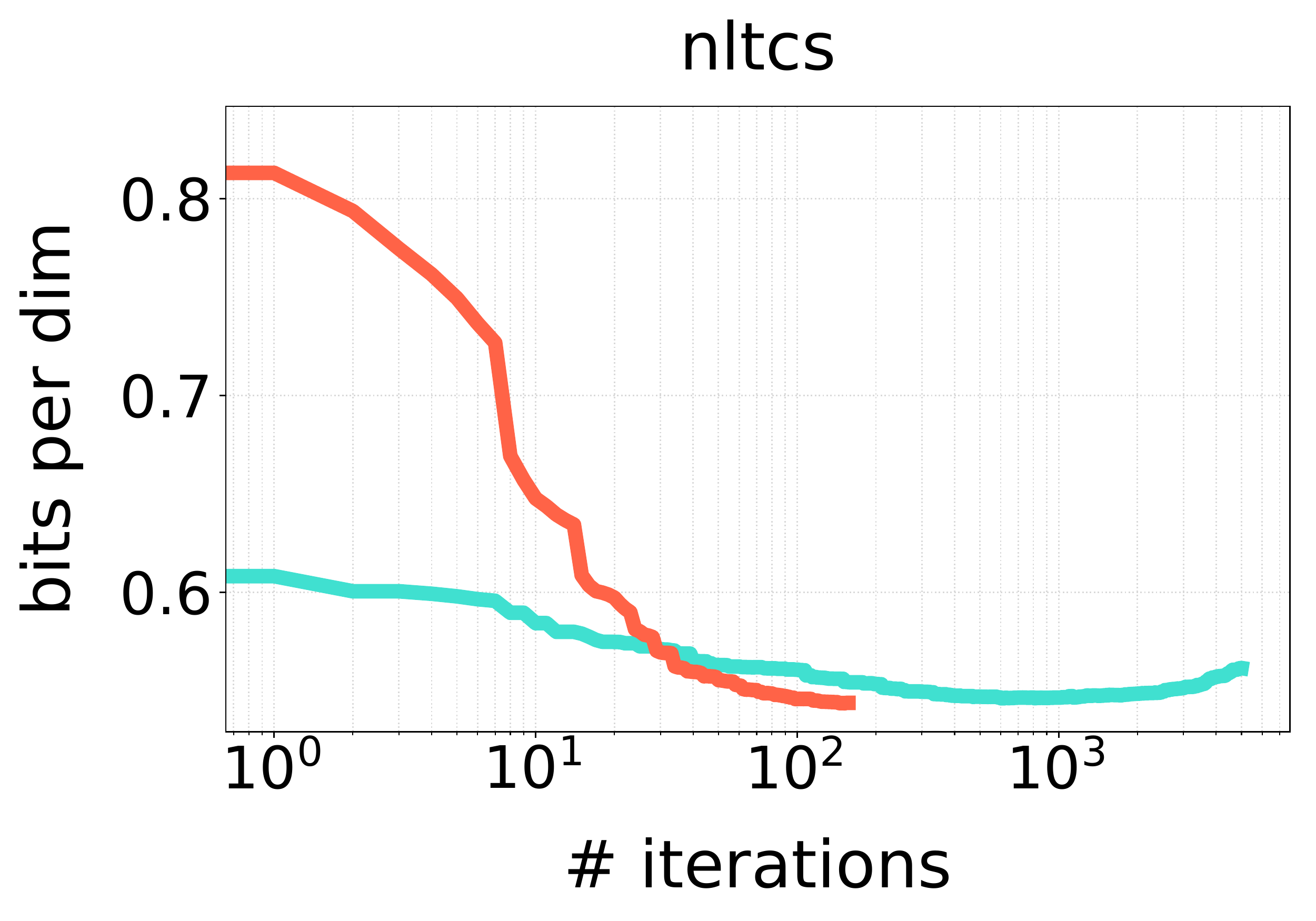}&
    \includegraphics[width=0.23\textwidth]{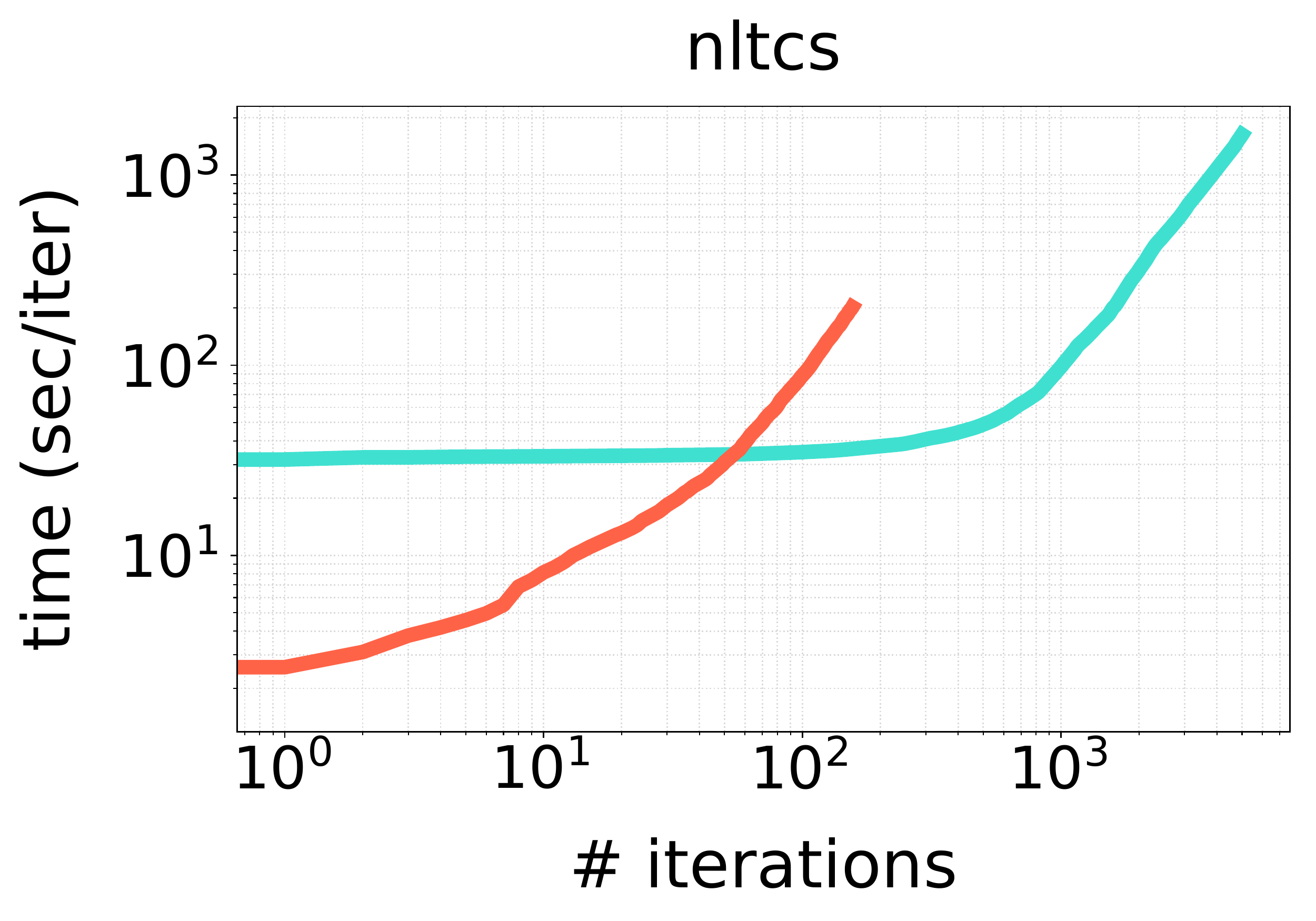}&
    \includegraphics[width=0.23\textwidth]{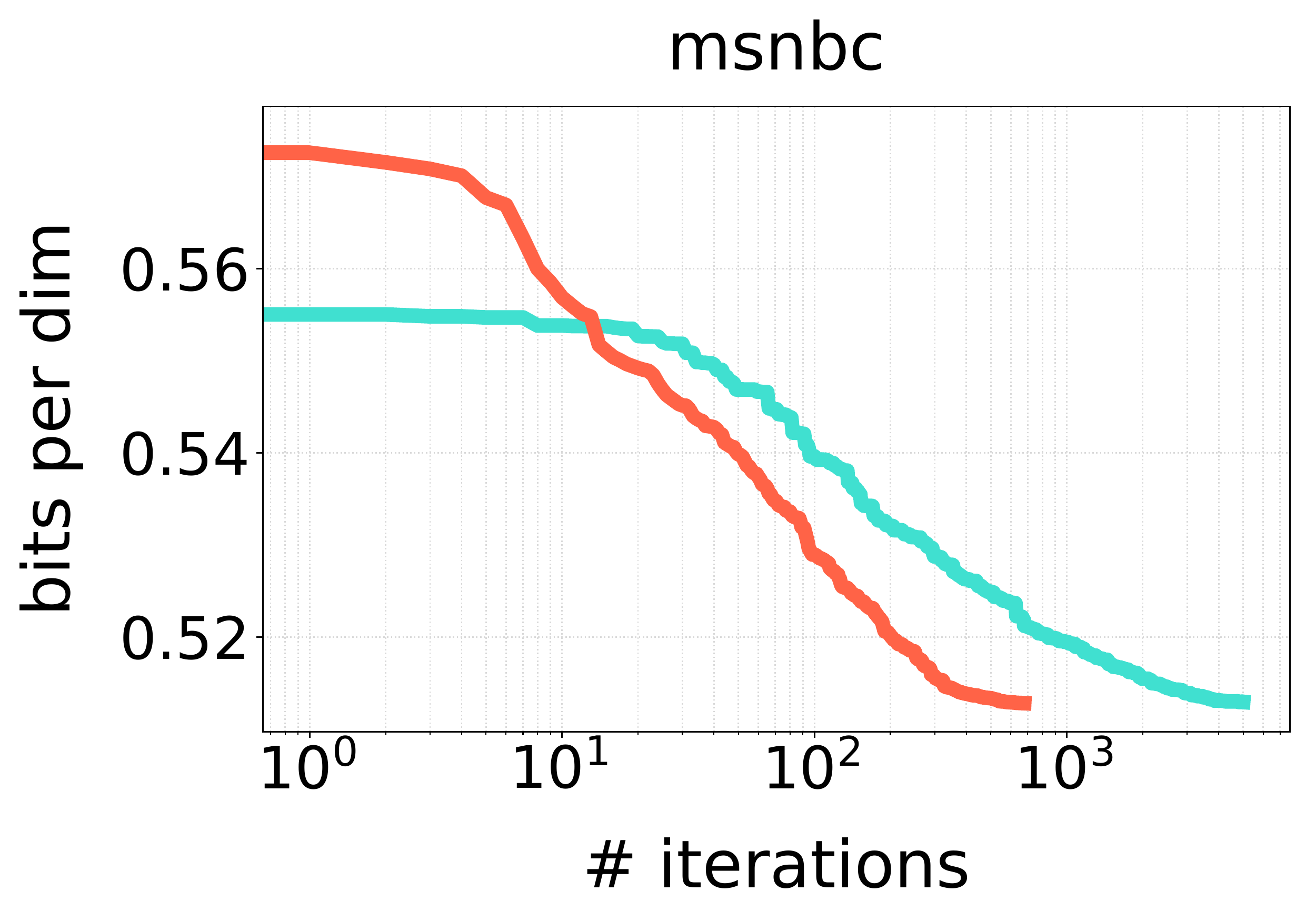}&
    \includegraphics[width=0.23\textwidth]{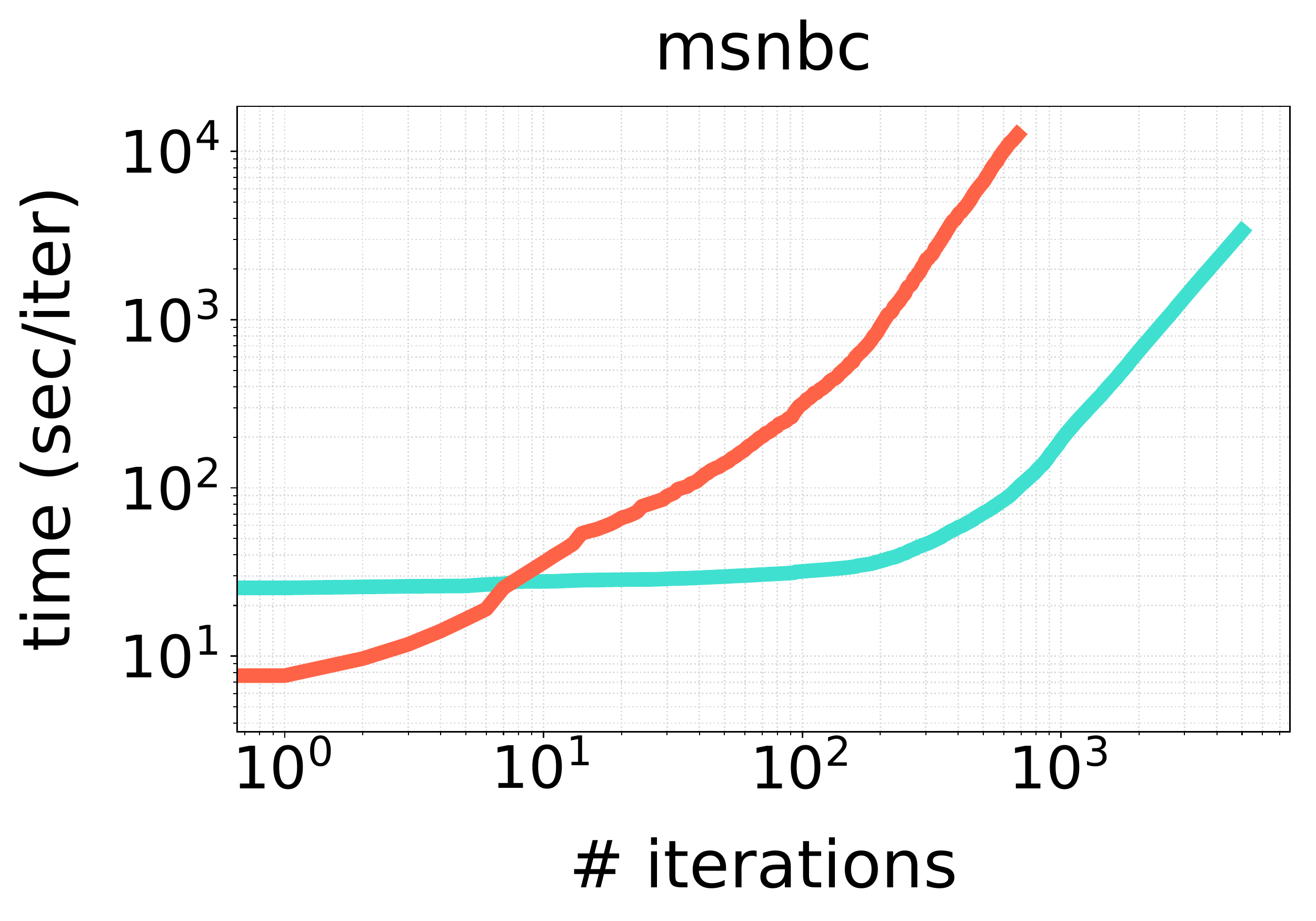}\\
    \includegraphics[width=0.23\textwidth]{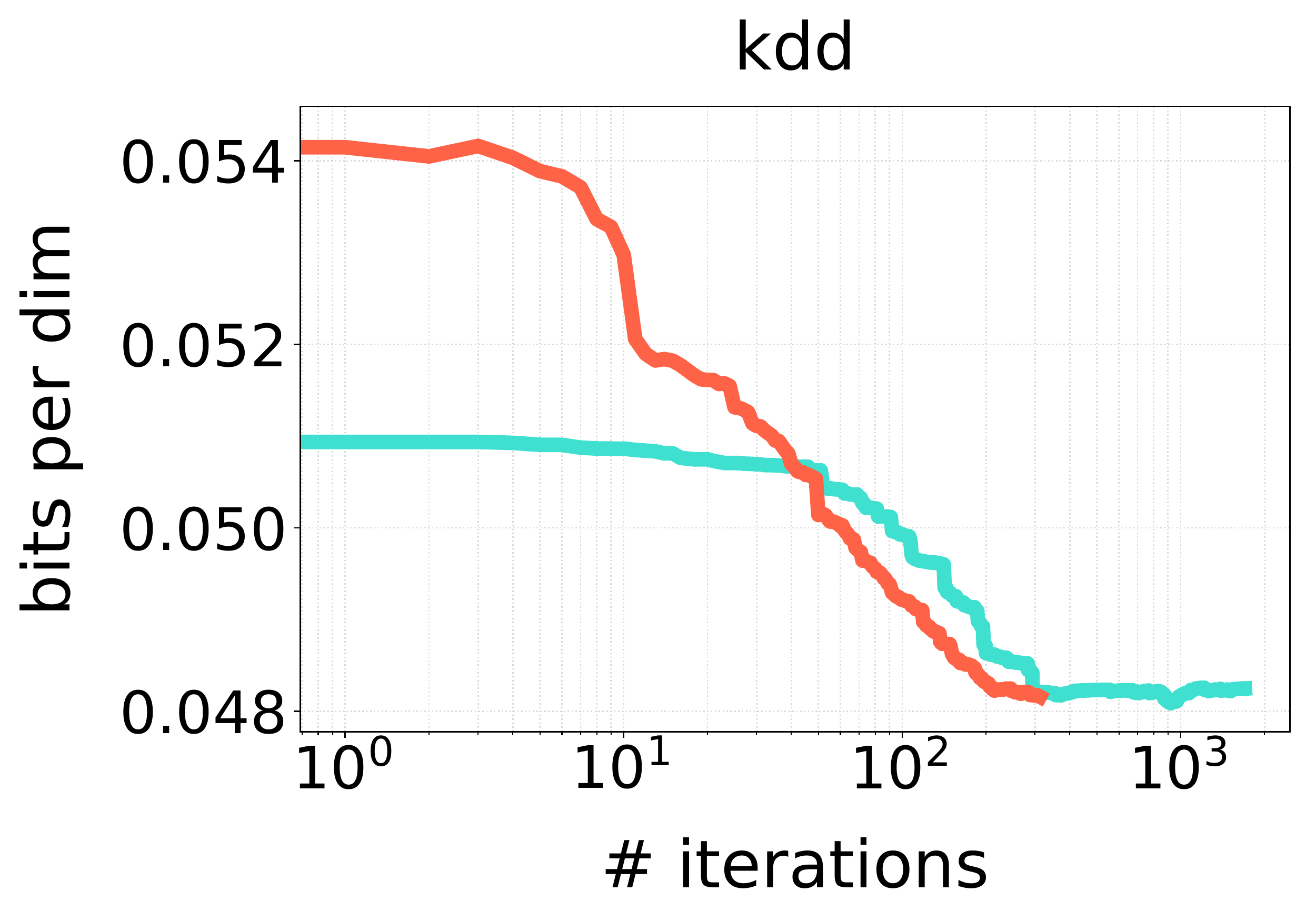}&
    \includegraphics[width=0.23\textwidth]{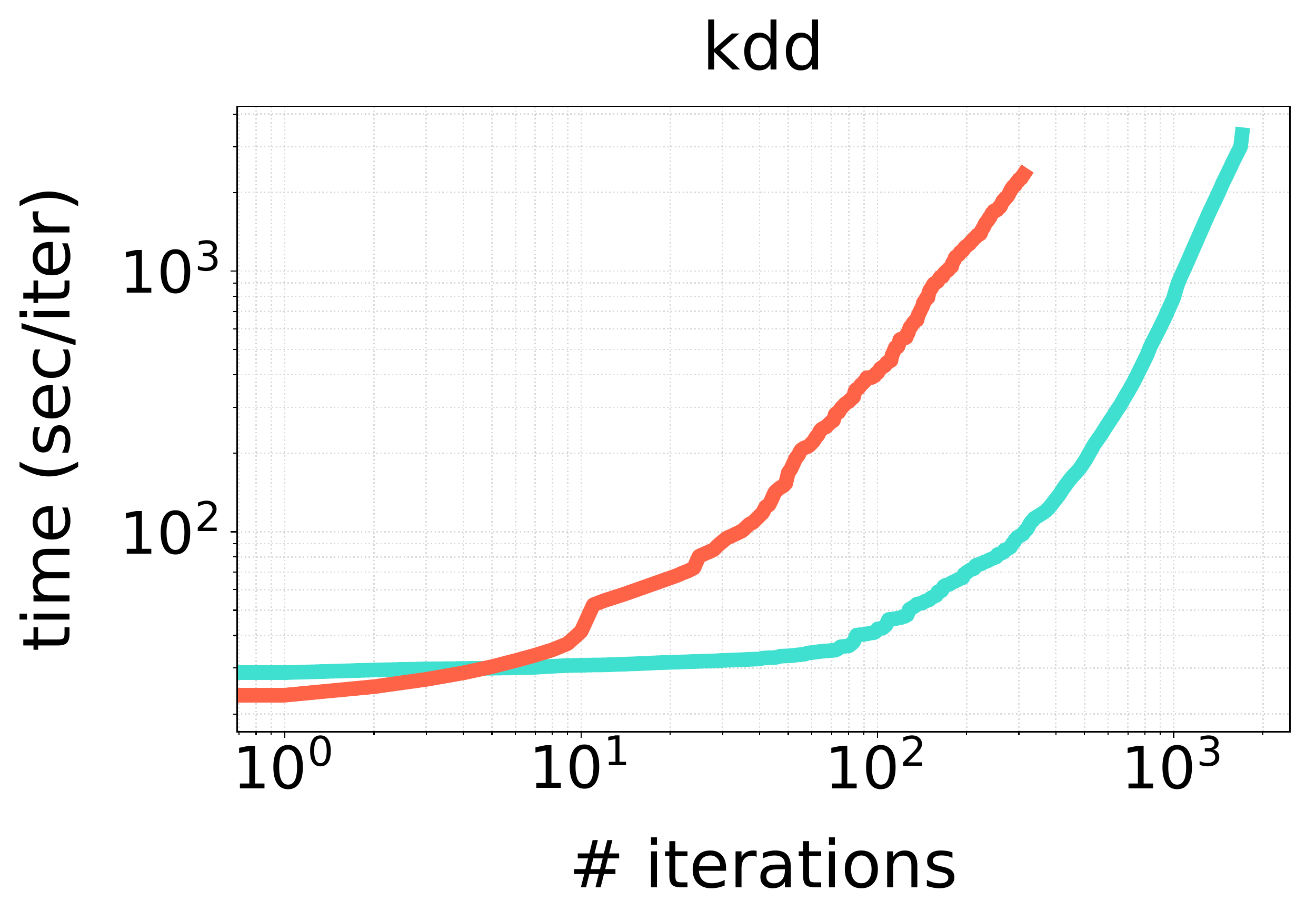}&
    \includegraphics[width=0.23\textwidth]{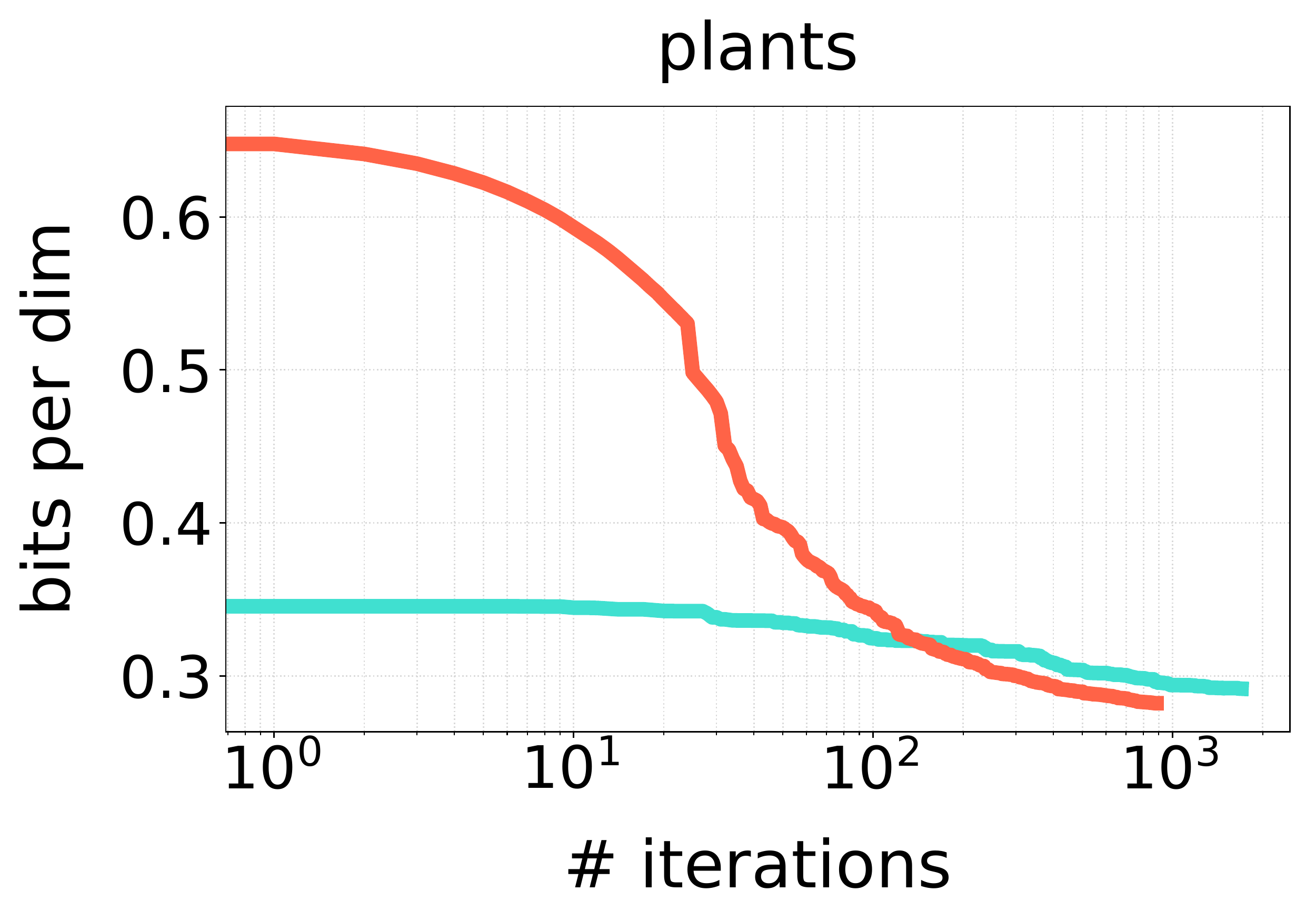}&
    \includegraphics[width=0.23\textwidth]{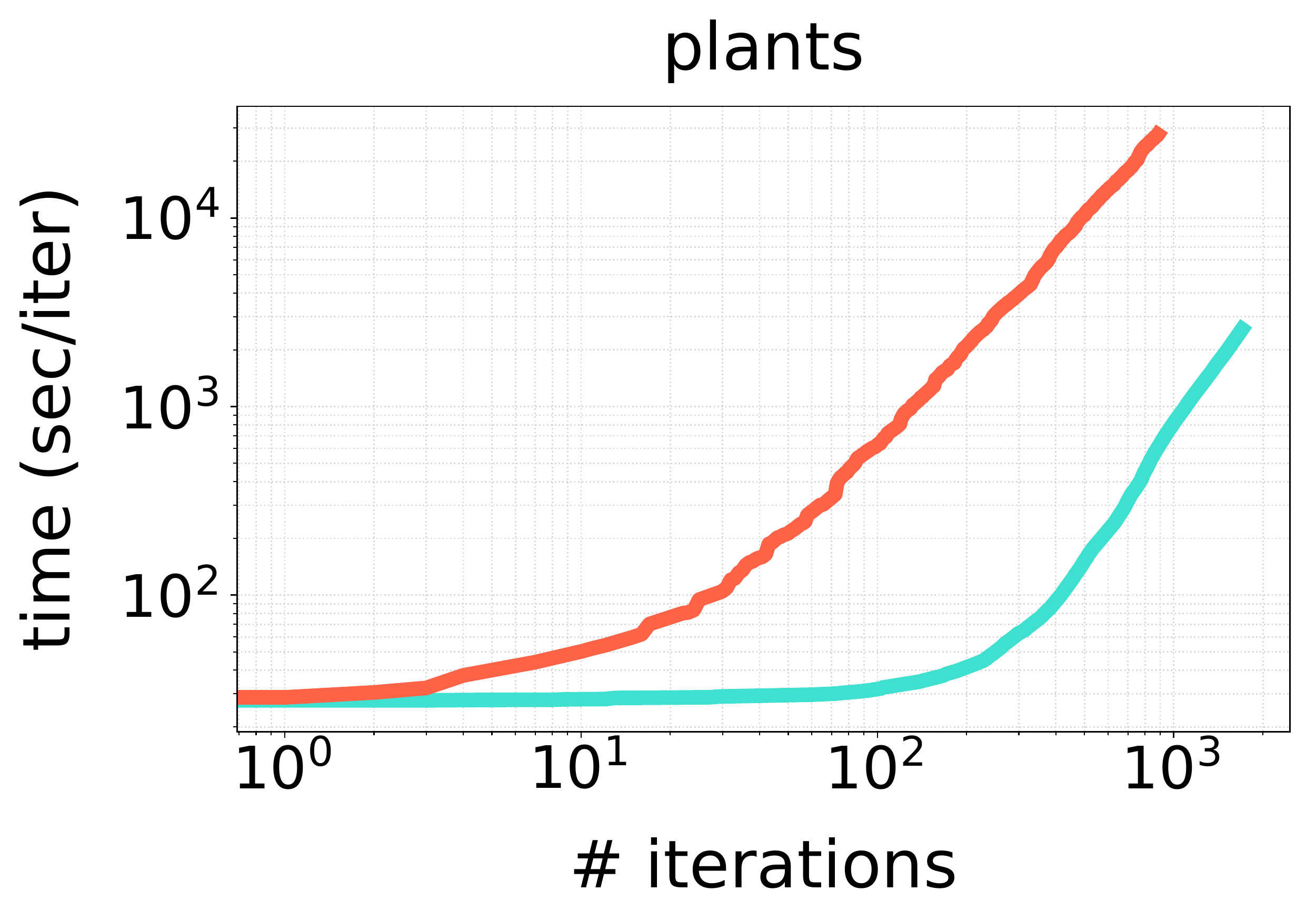}\\
    \includegraphics[width=0.23\textwidth]{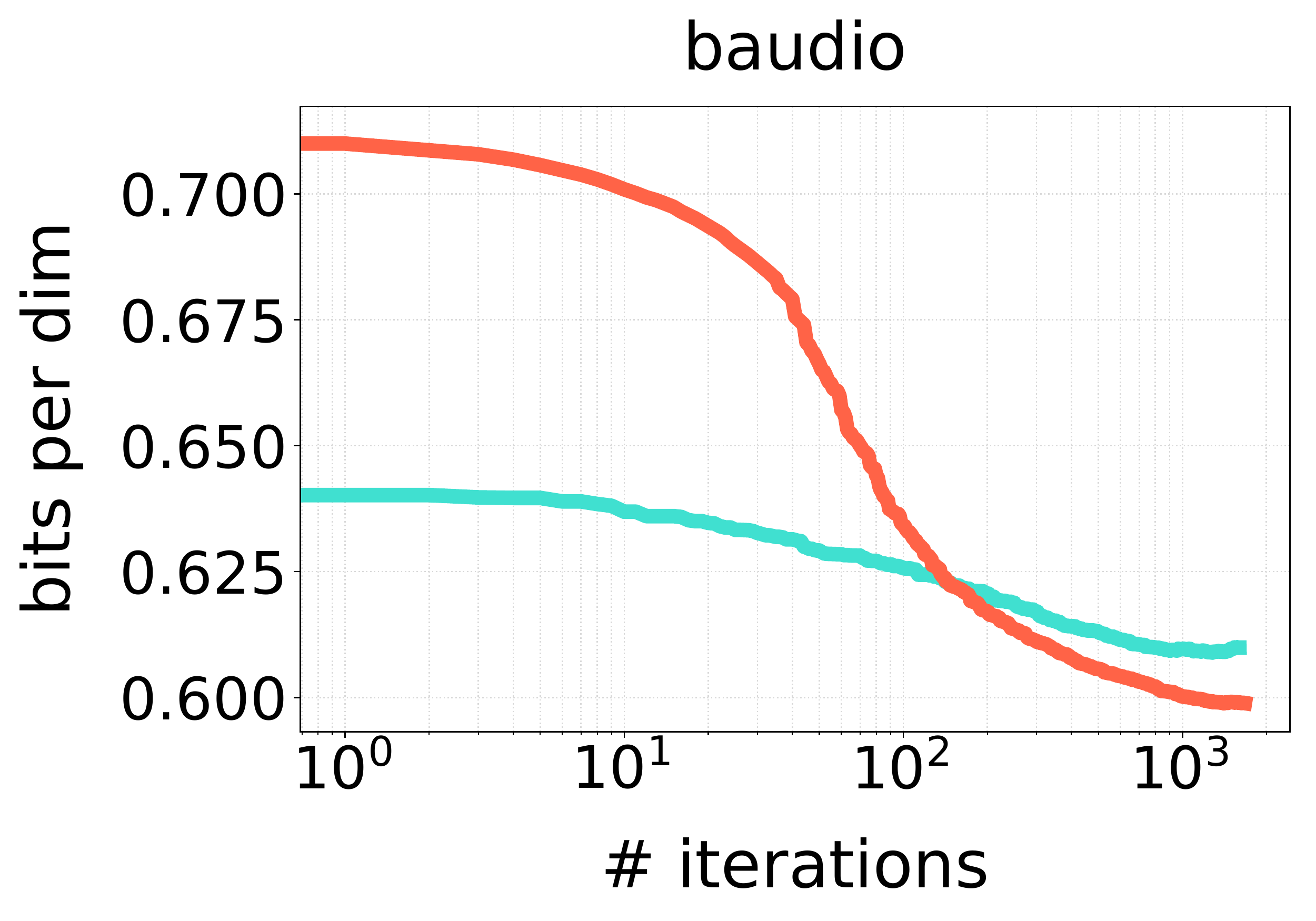}&
    \includegraphics[width=0.23\textwidth]{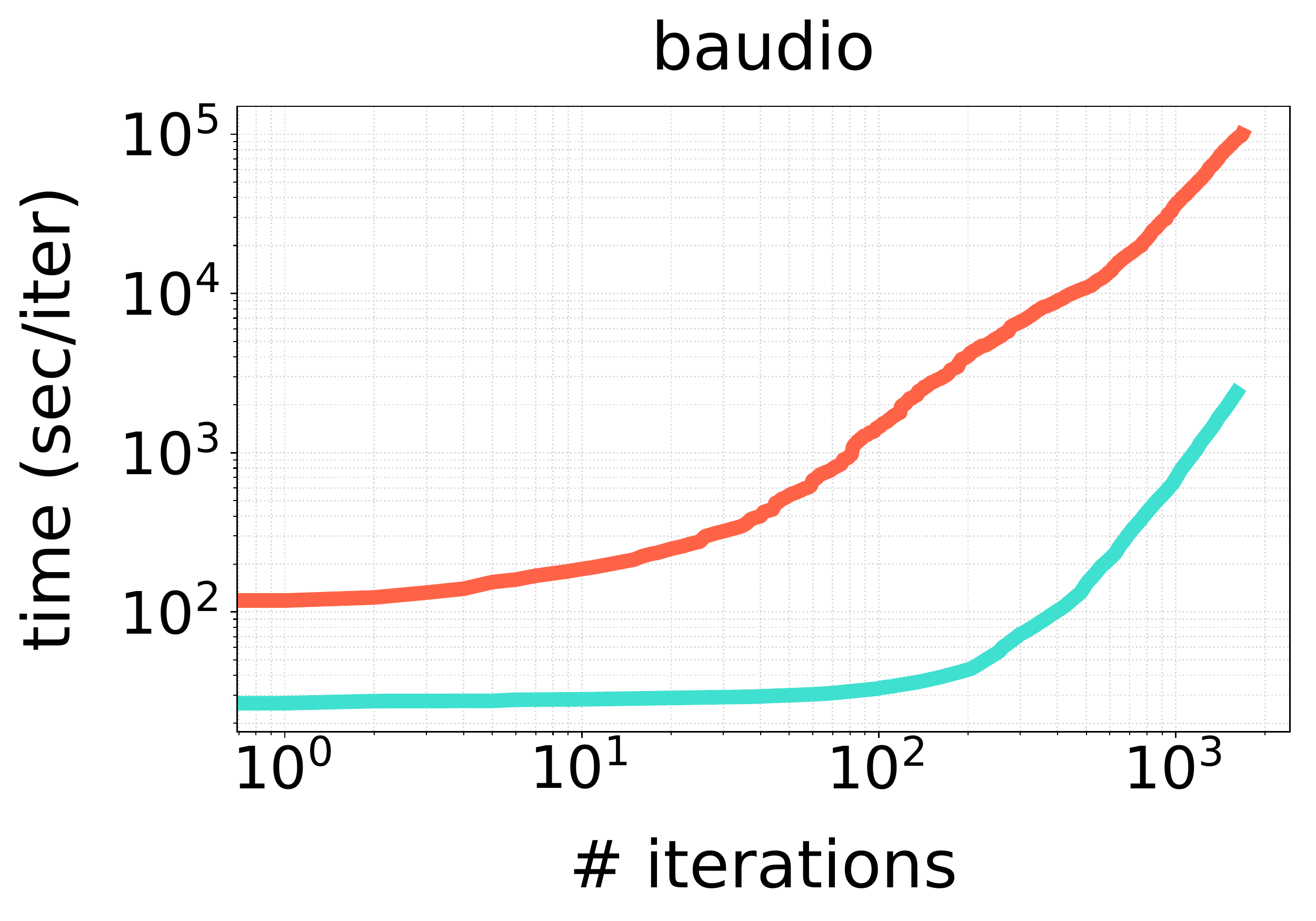}&
    \includegraphics[width=0.23\textwidth]{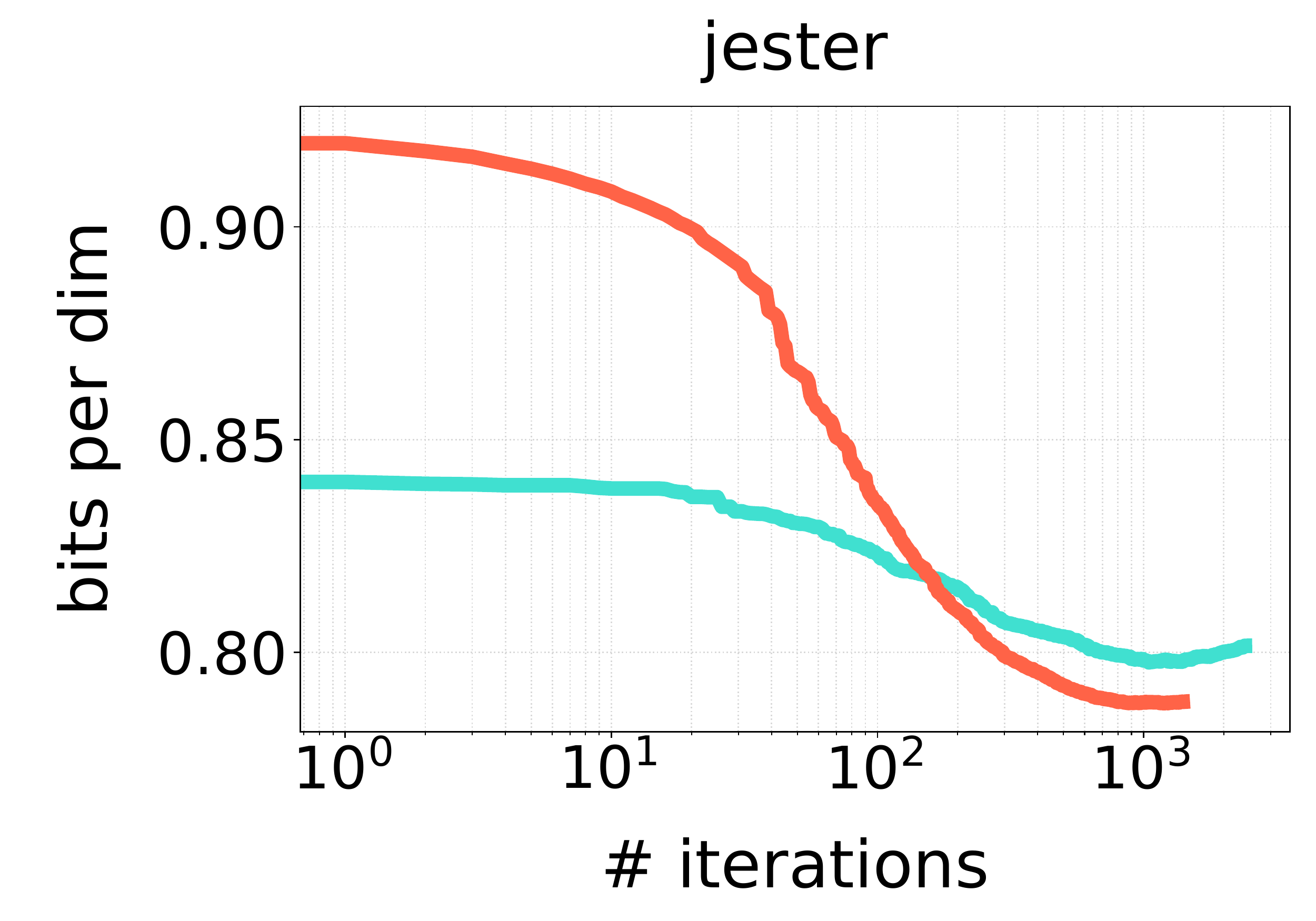}&
    \includegraphics[width=0.23\textwidth]{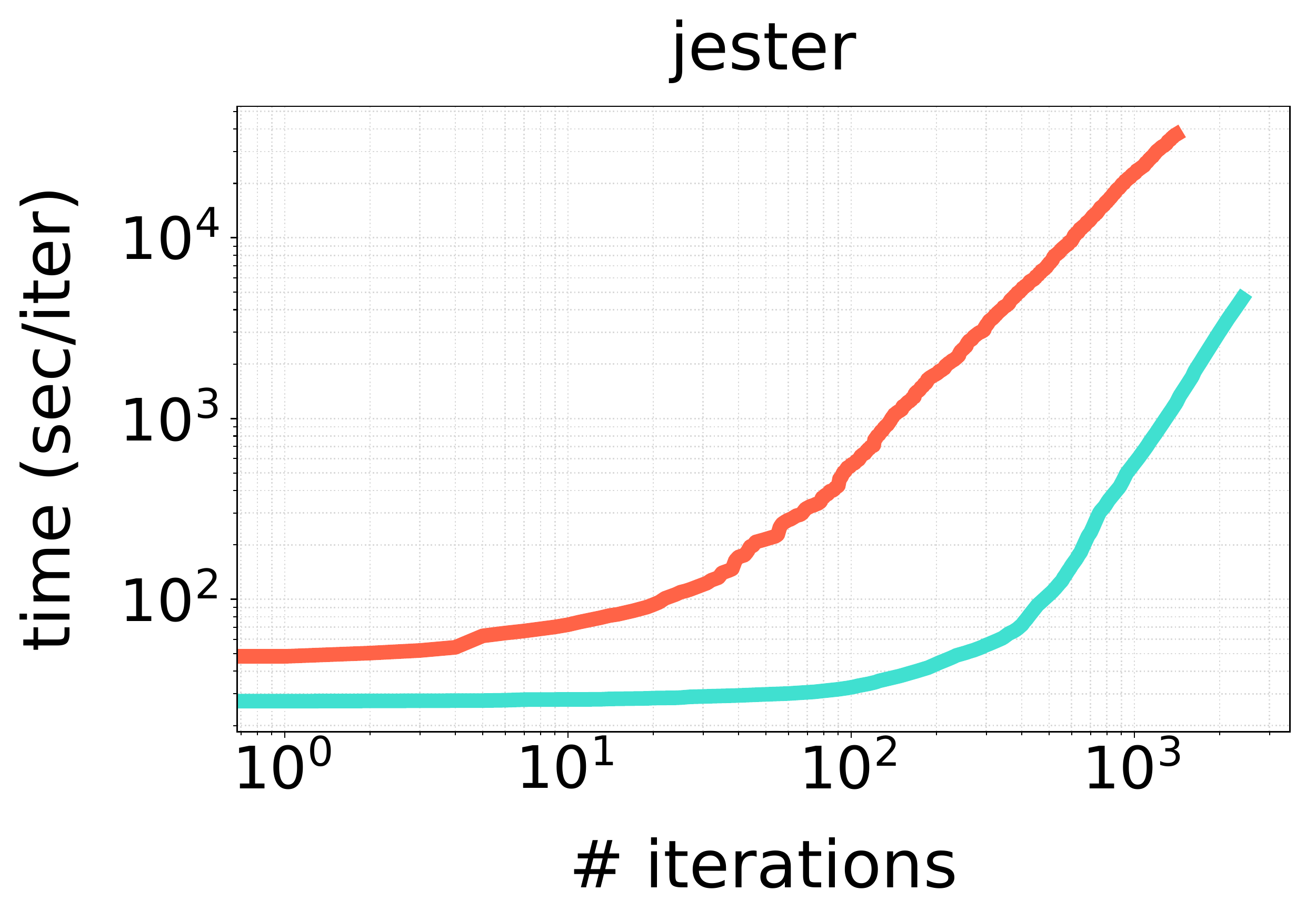}\\
    \includegraphics[width=0.23\textwidth]{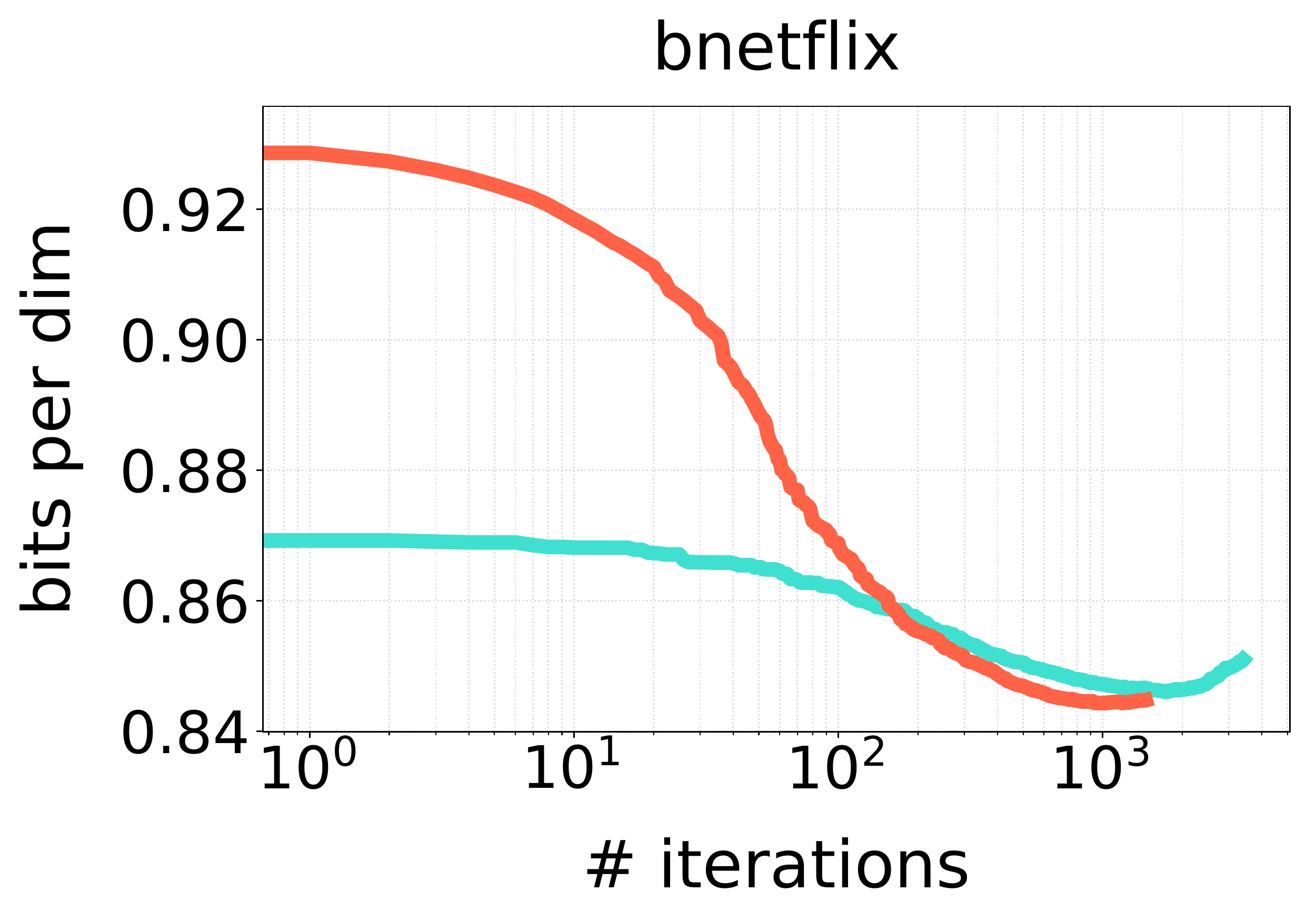}&
    \includegraphics[width=0.23\textwidth]{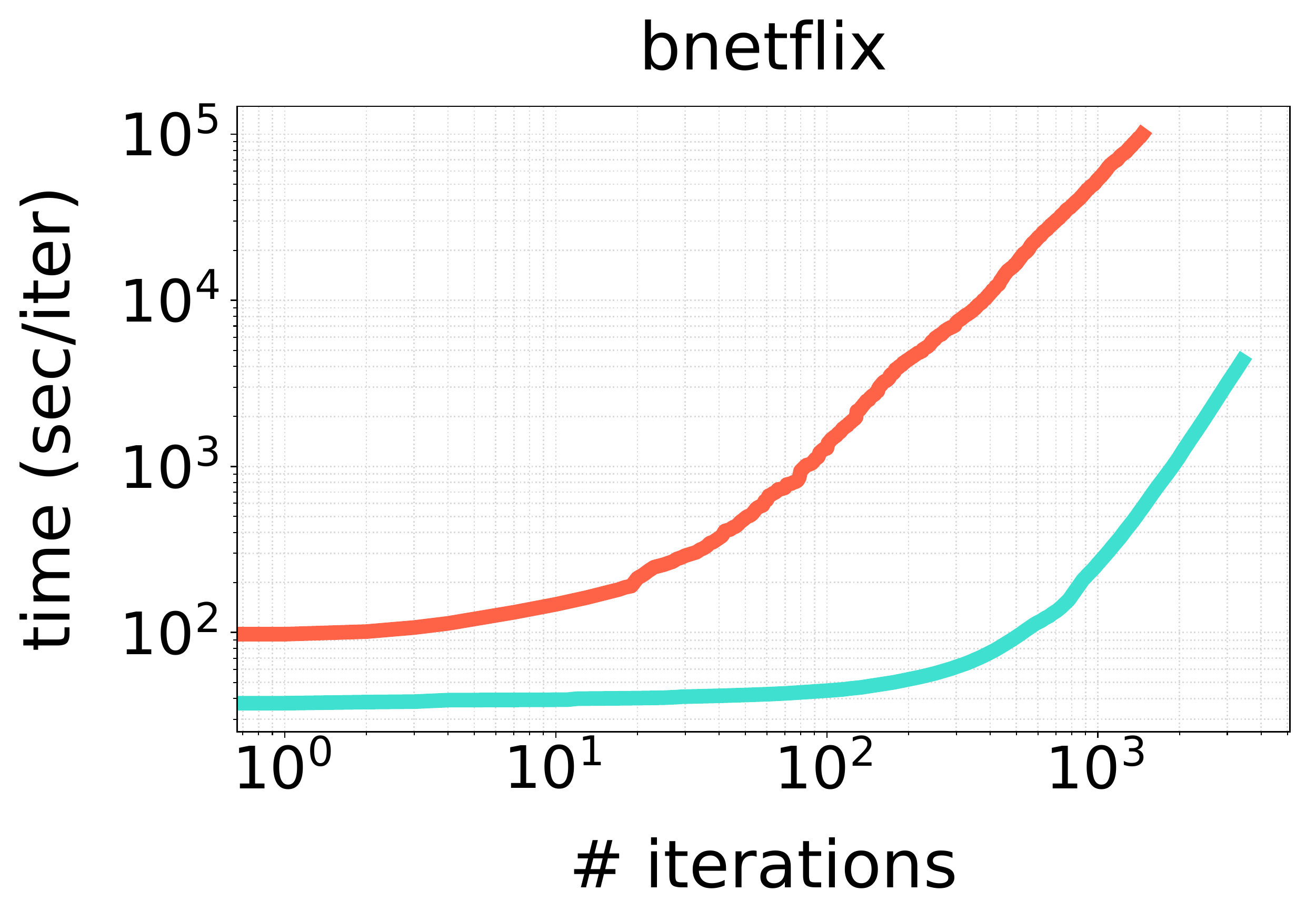}&
    \includegraphics[width=0.23\textwidth]{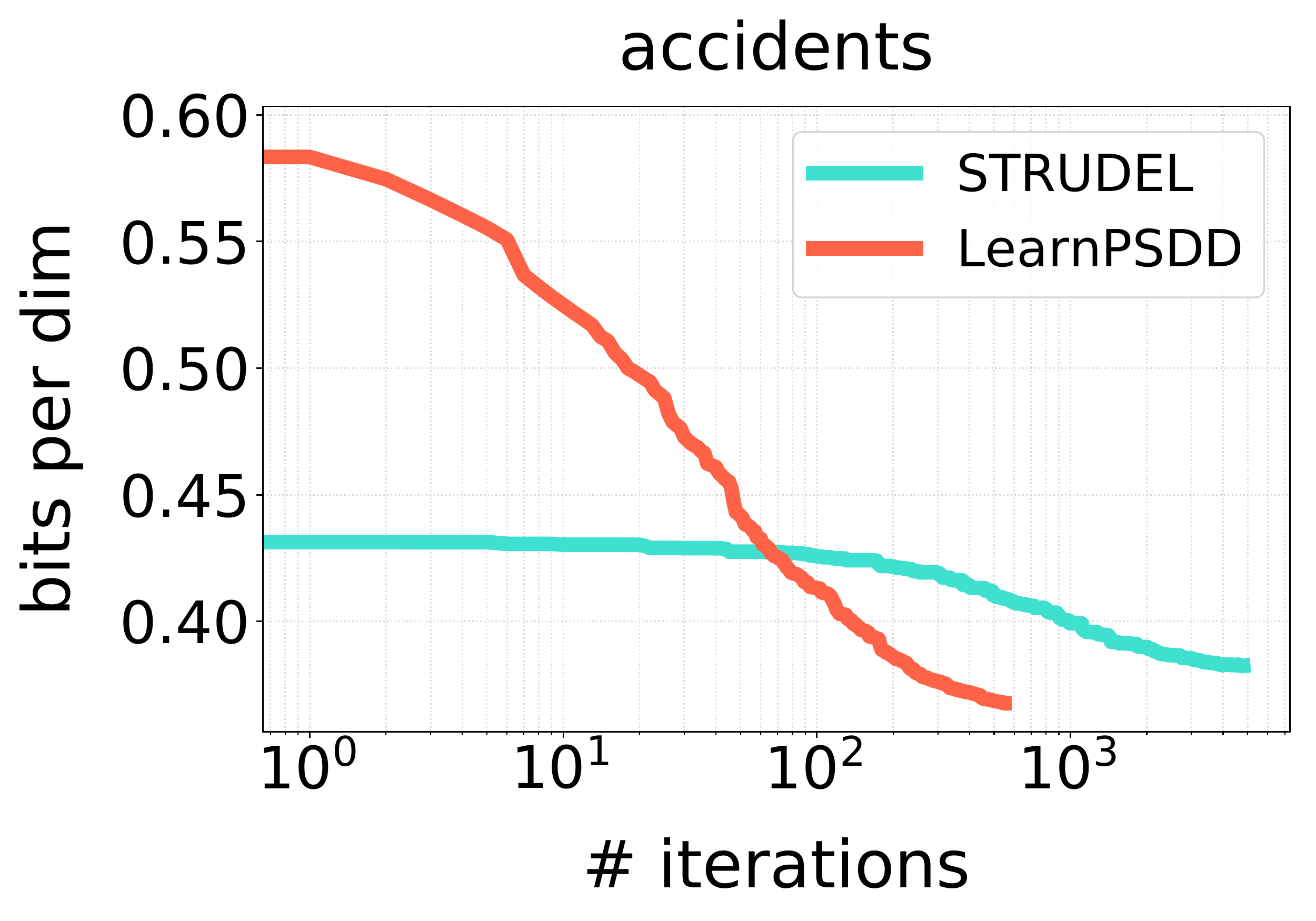}&
    \includegraphics[width=0.23\textwidth]{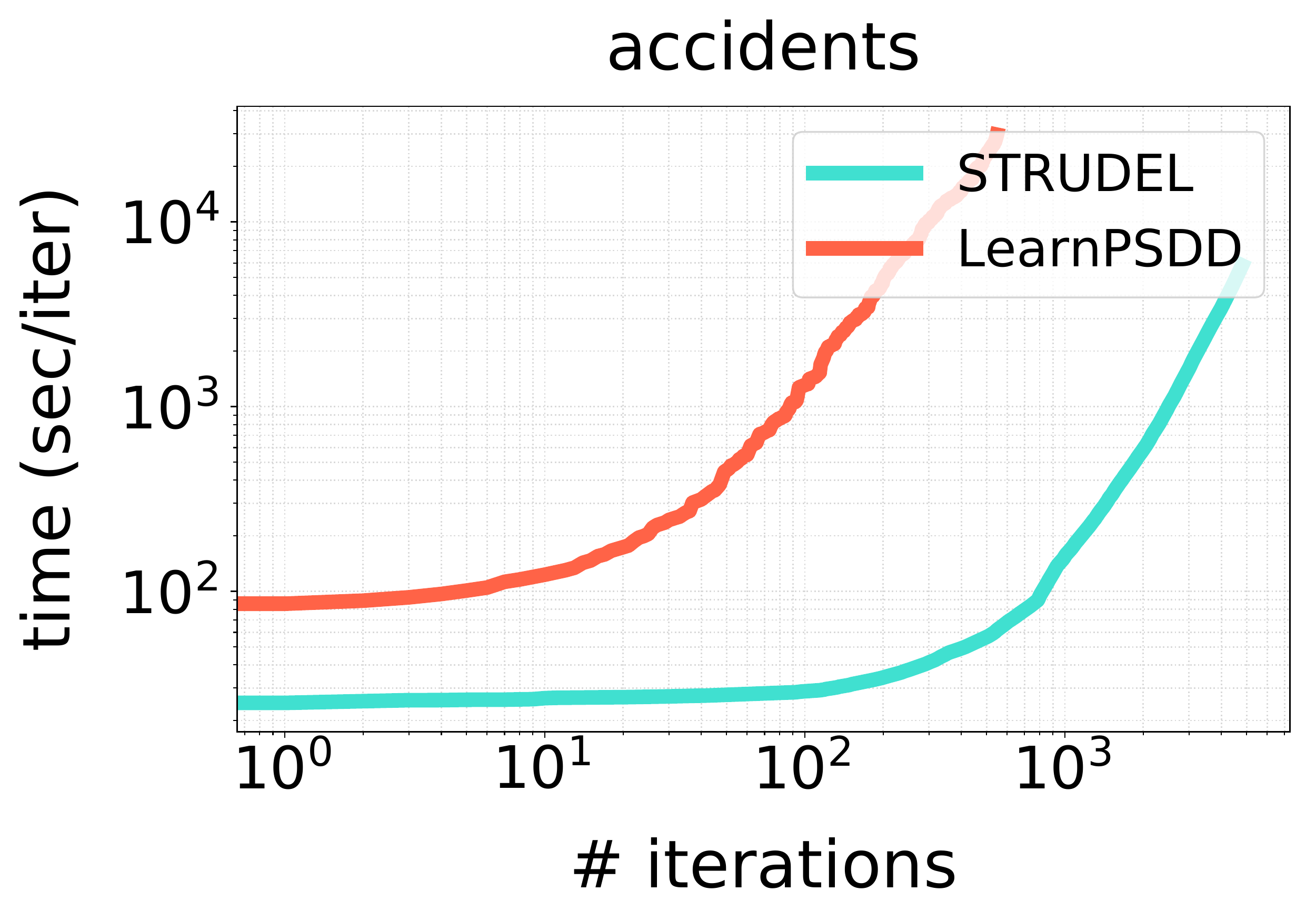}\\
    \includegraphics[width=0.23\textwidth]{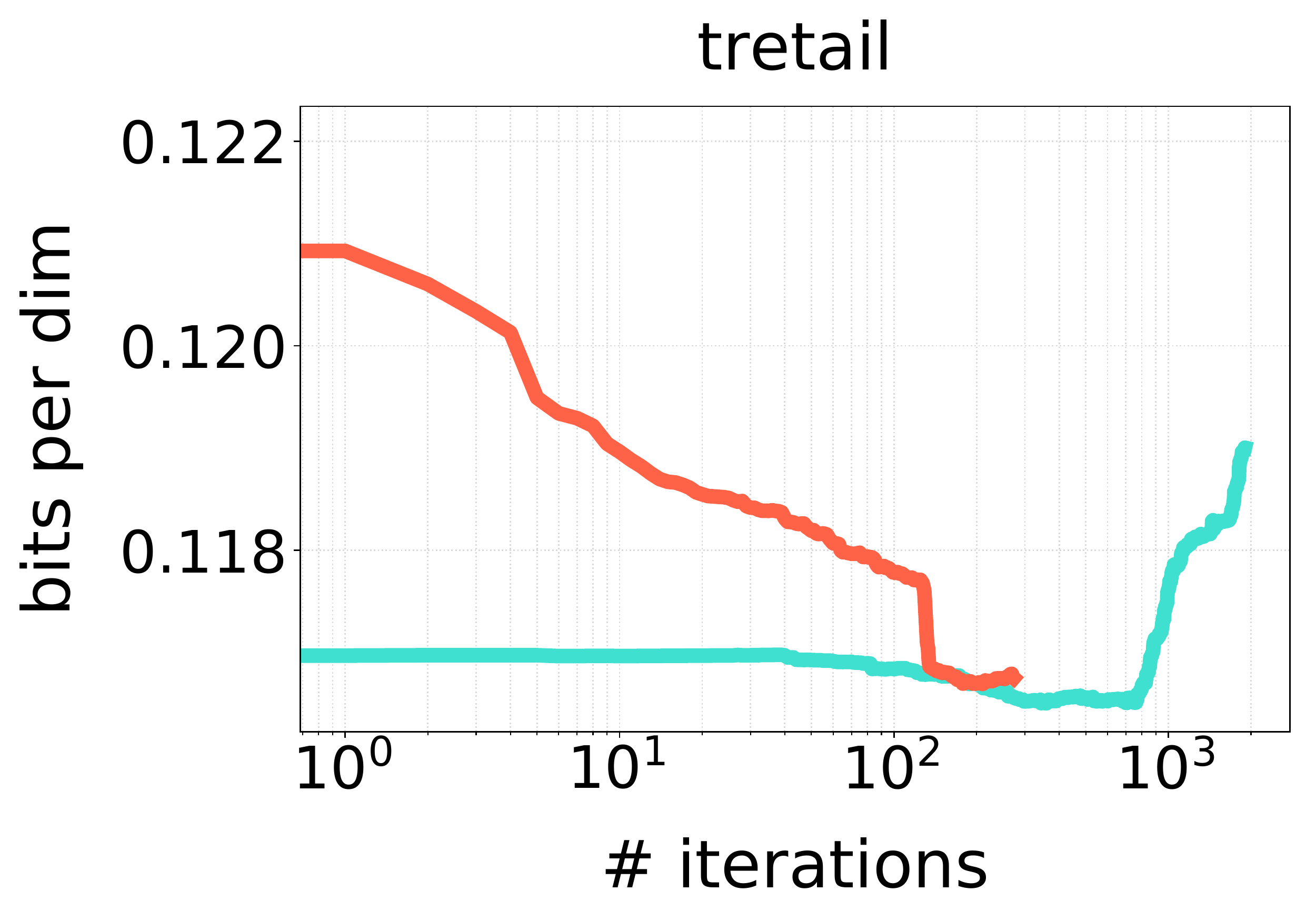}&
    \includegraphics[width=0.23\textwidth]{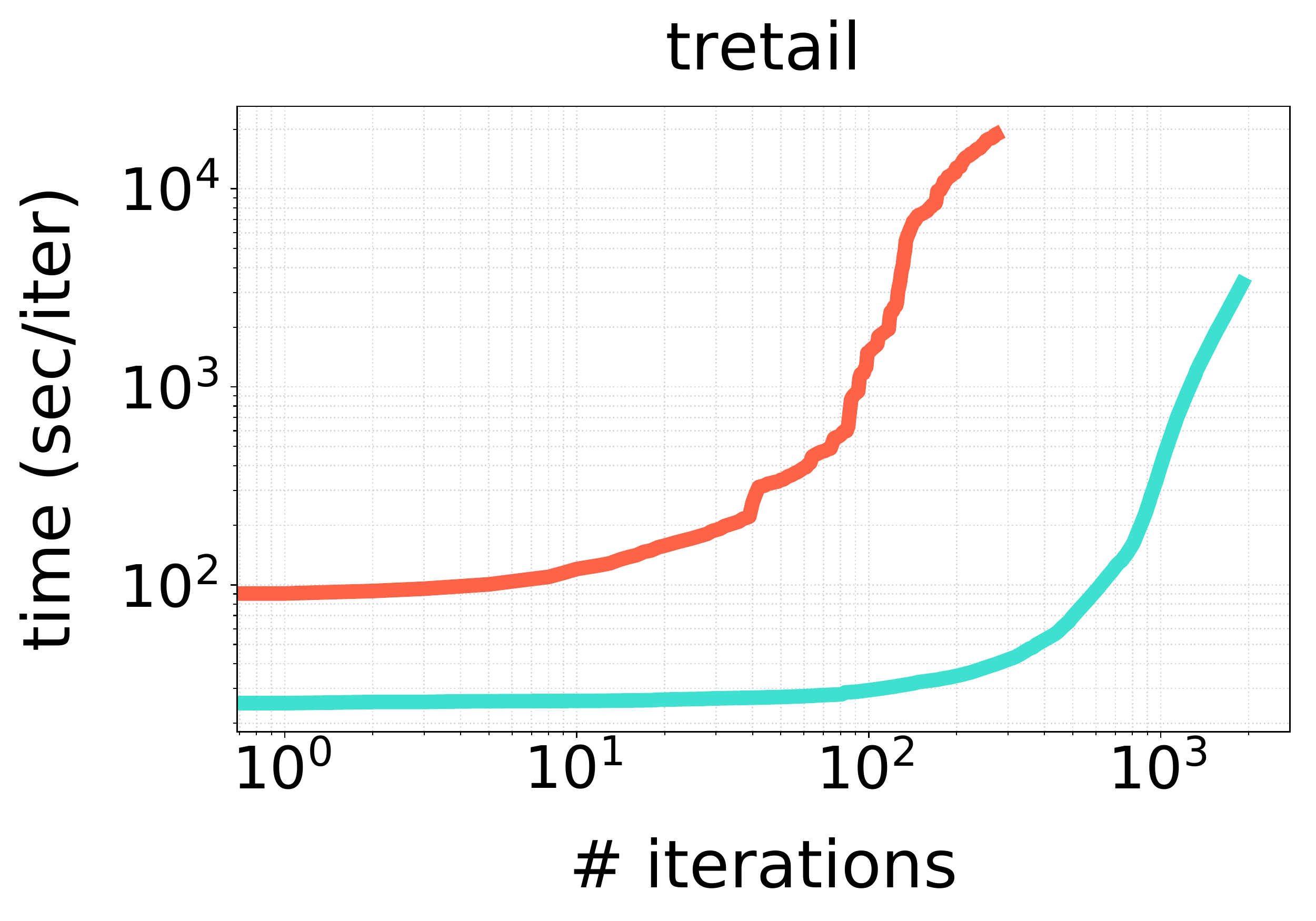}&
    \includegraphics[width=0.23\textwidth]{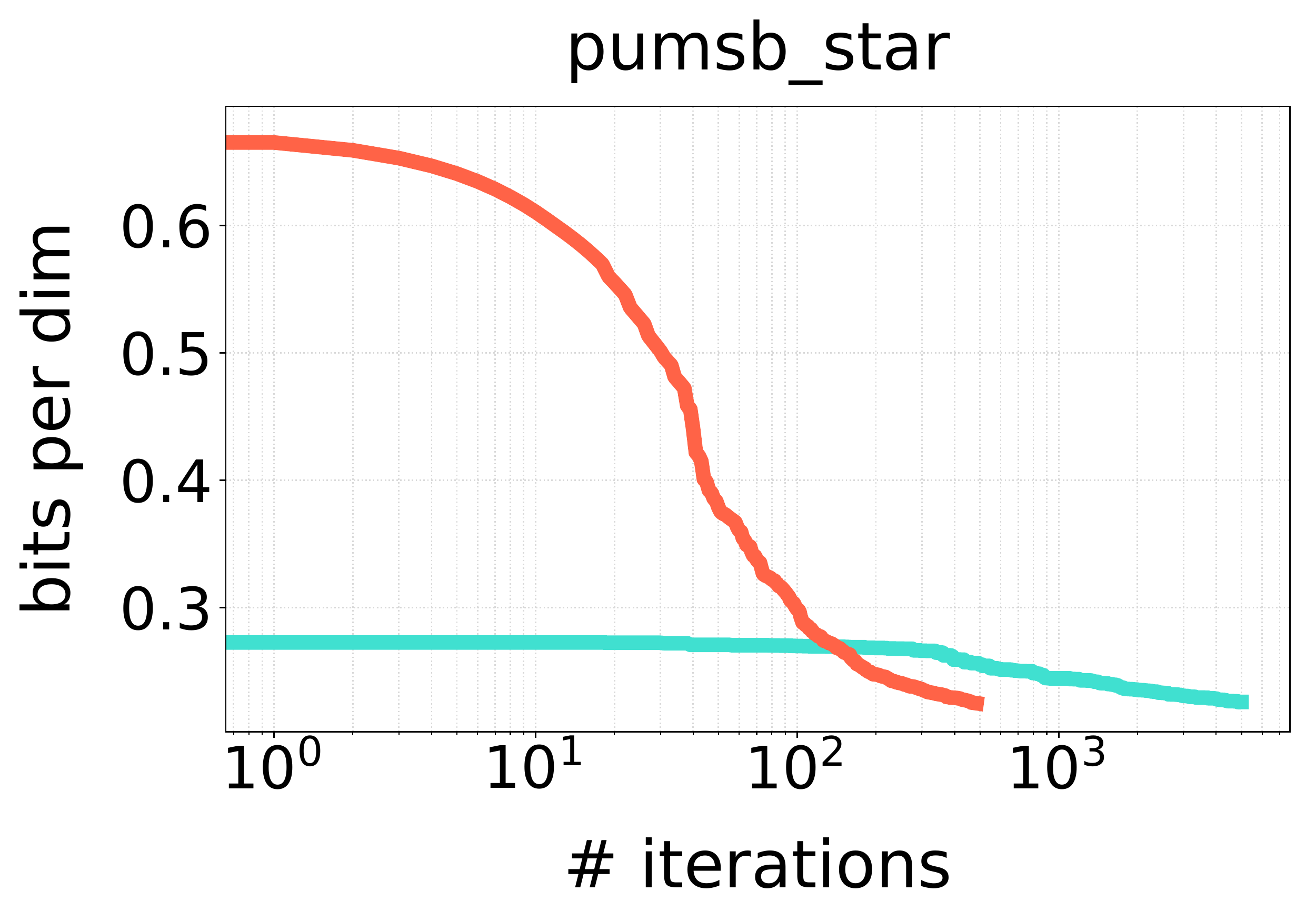}&
    \includegraphics[width=0.23\textwidth]{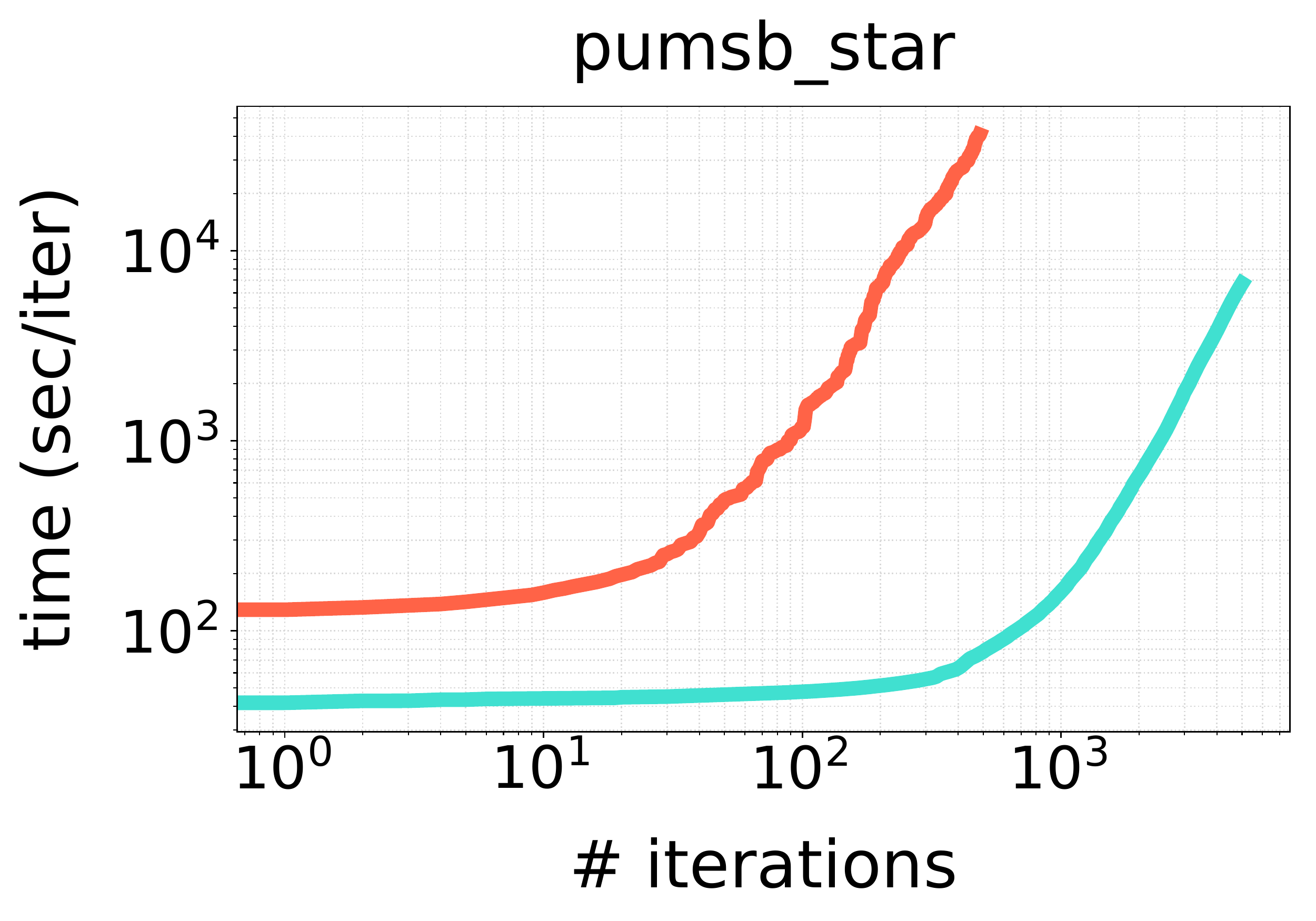}\\
    \includegraphics[width=0.23\textwidth]{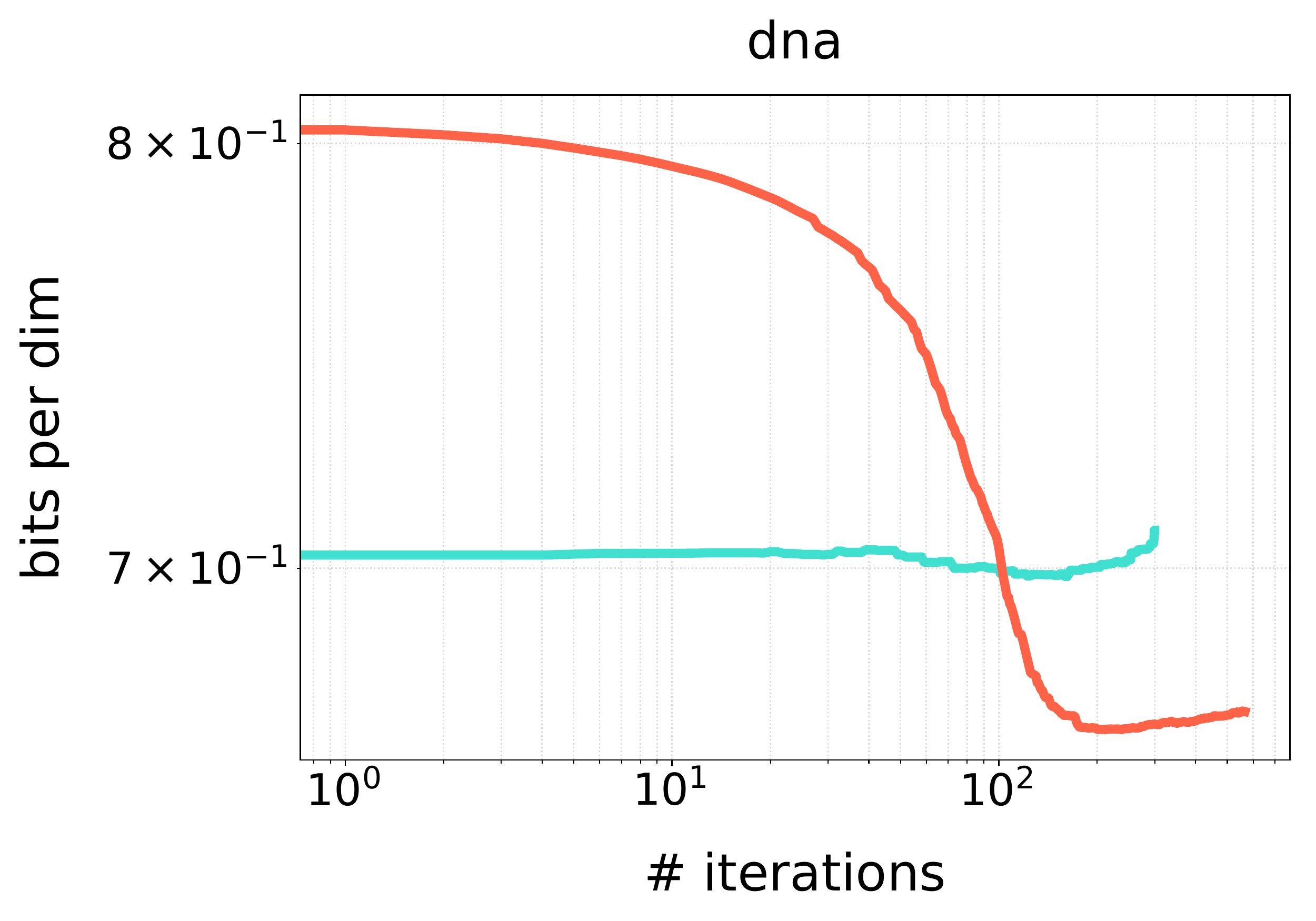}&
    \includegraphics[width=0.23\textwidth]{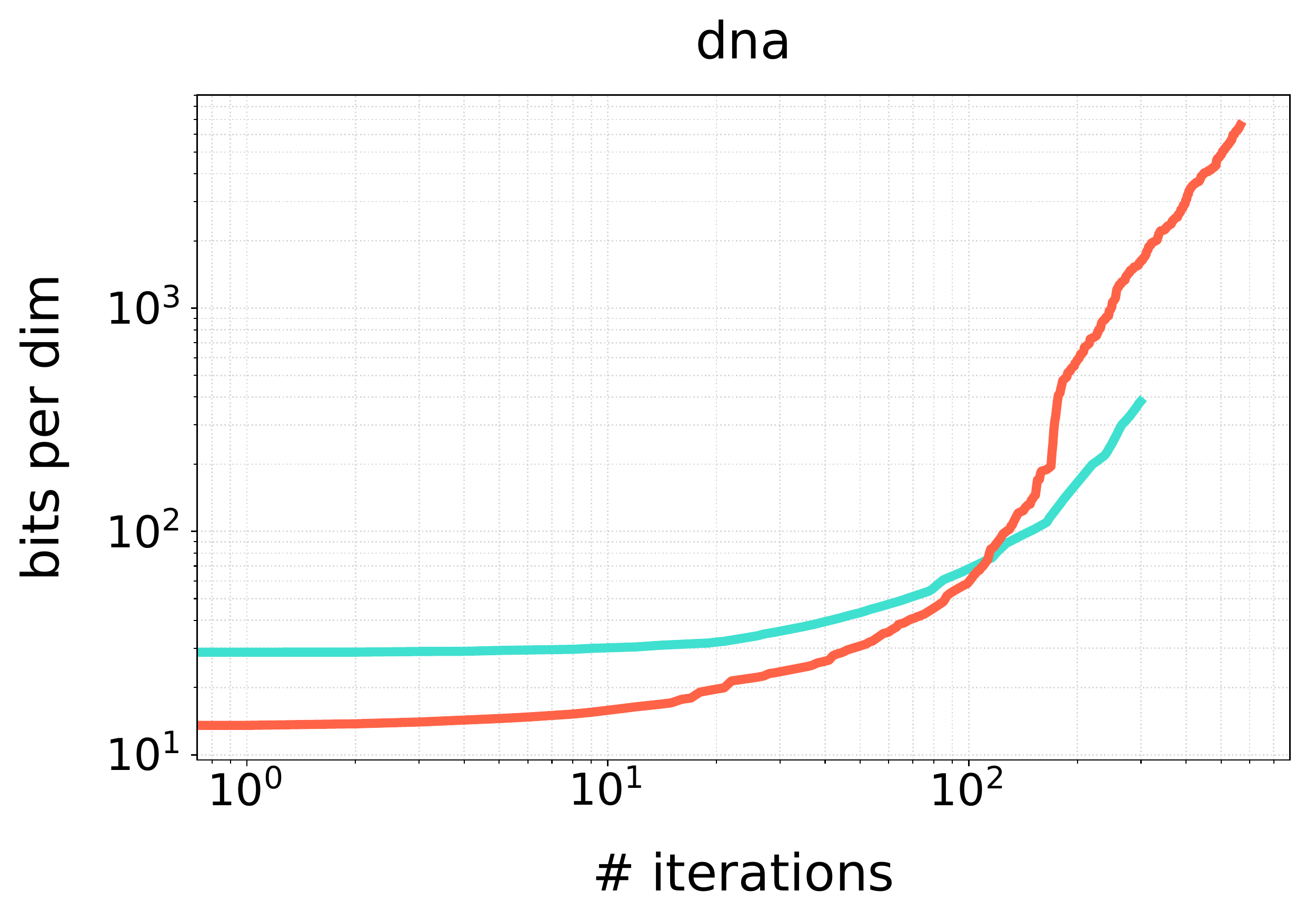}&
    \includegraphics[width=0.23\textwidth]{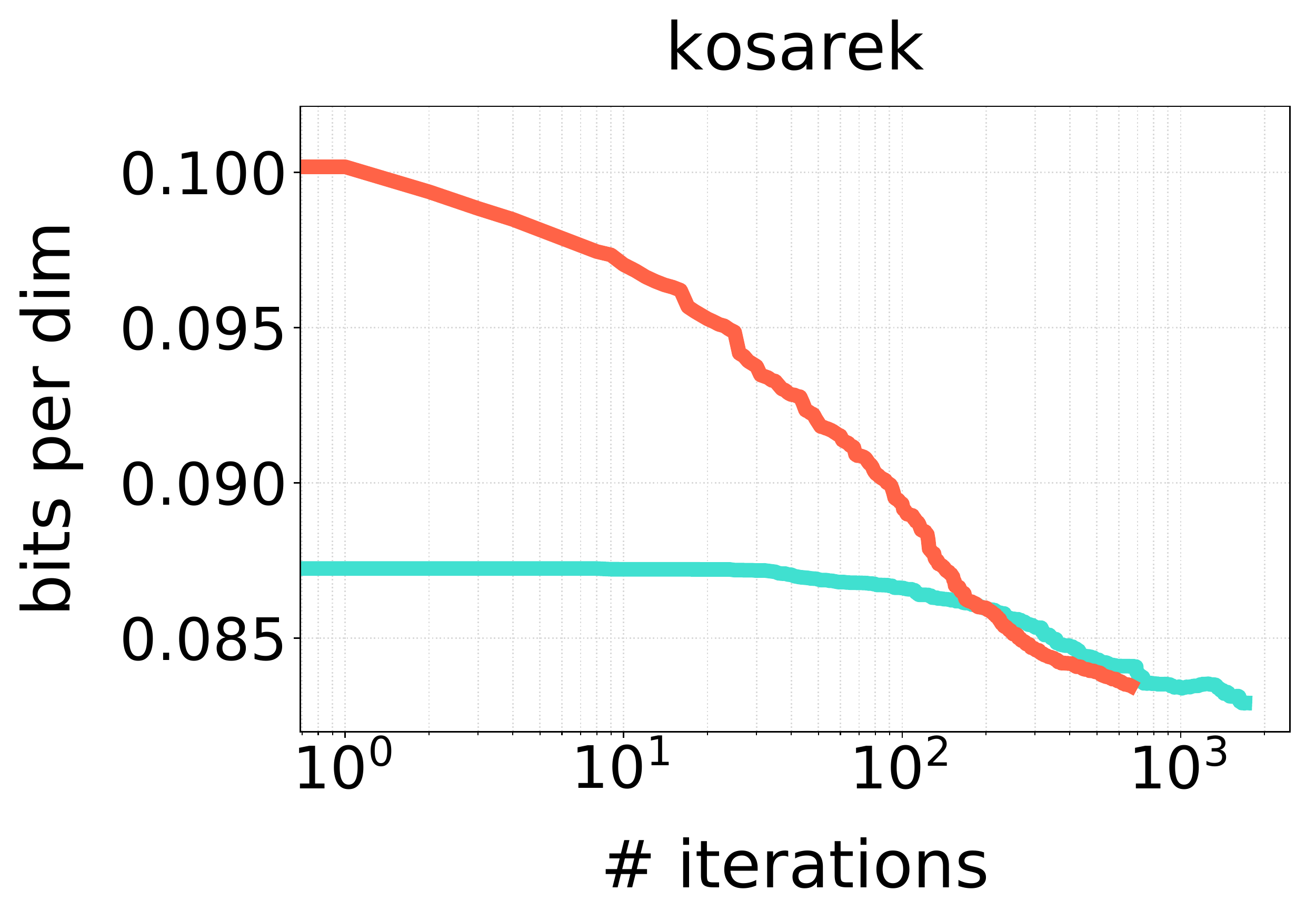}&
    \includegraphics[width=0.23\textwidth]{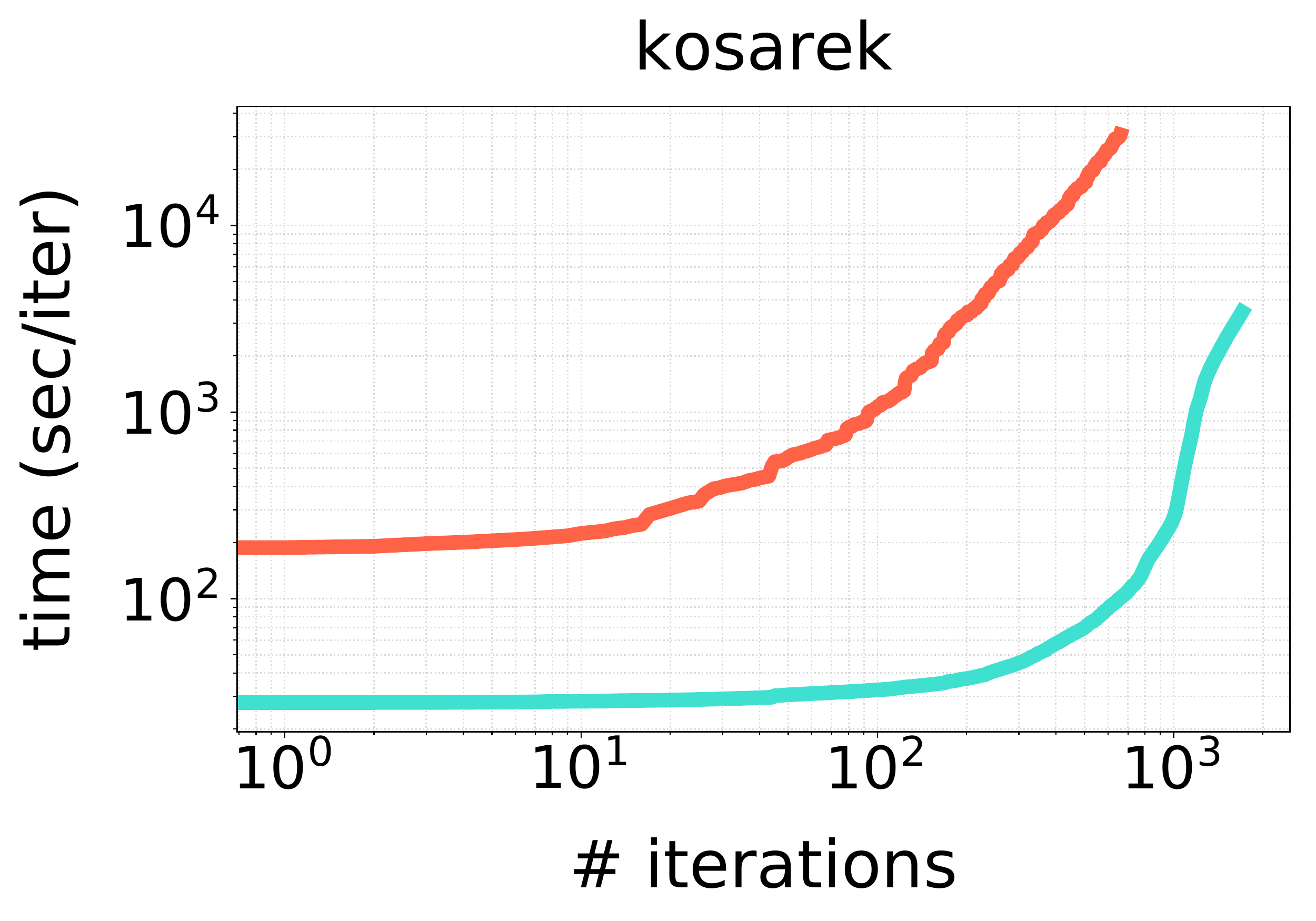}\\
    \includegraphics[width=0.23\textwidth]{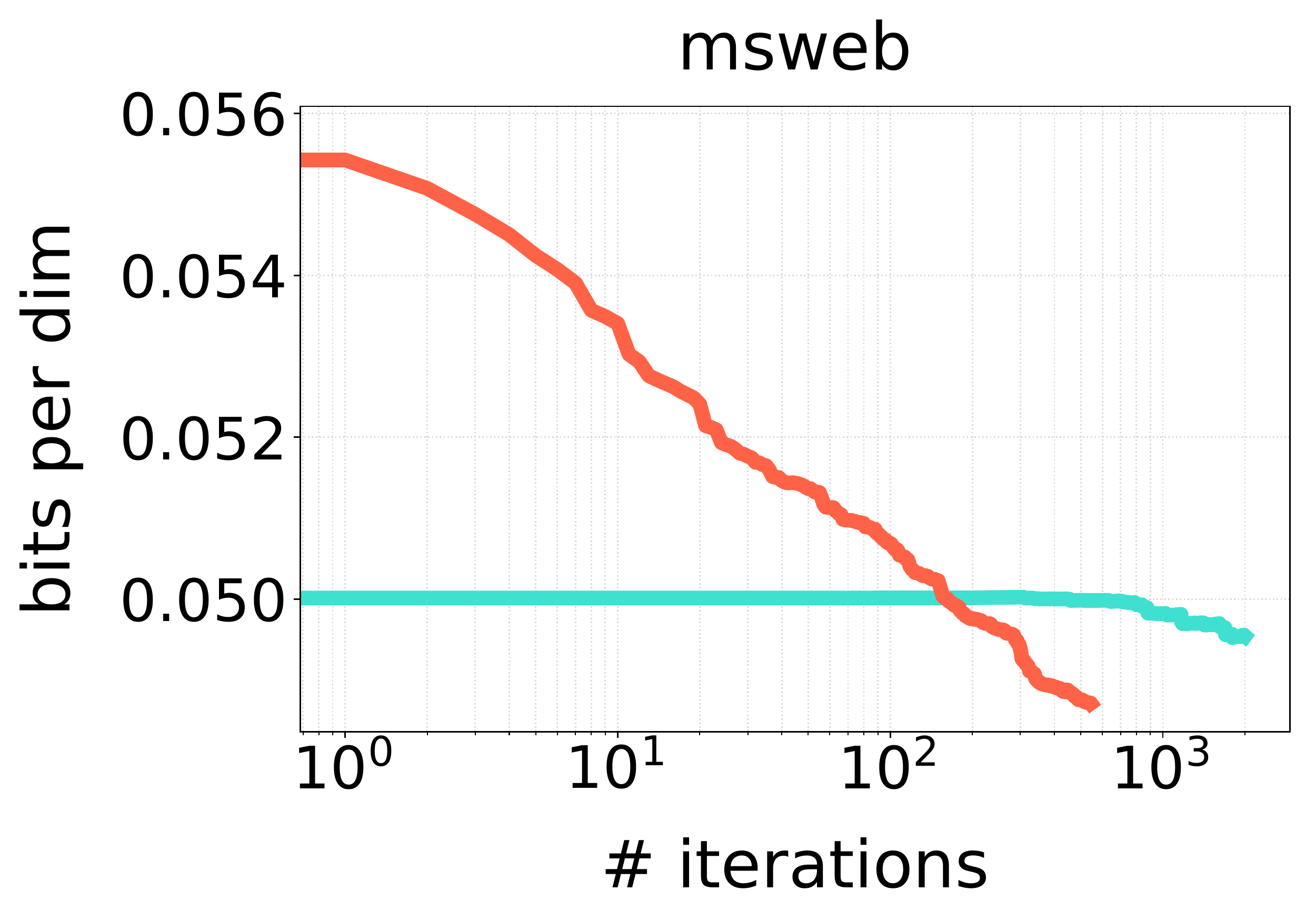}&
    \includegraphics[width=0.23\textwidth]{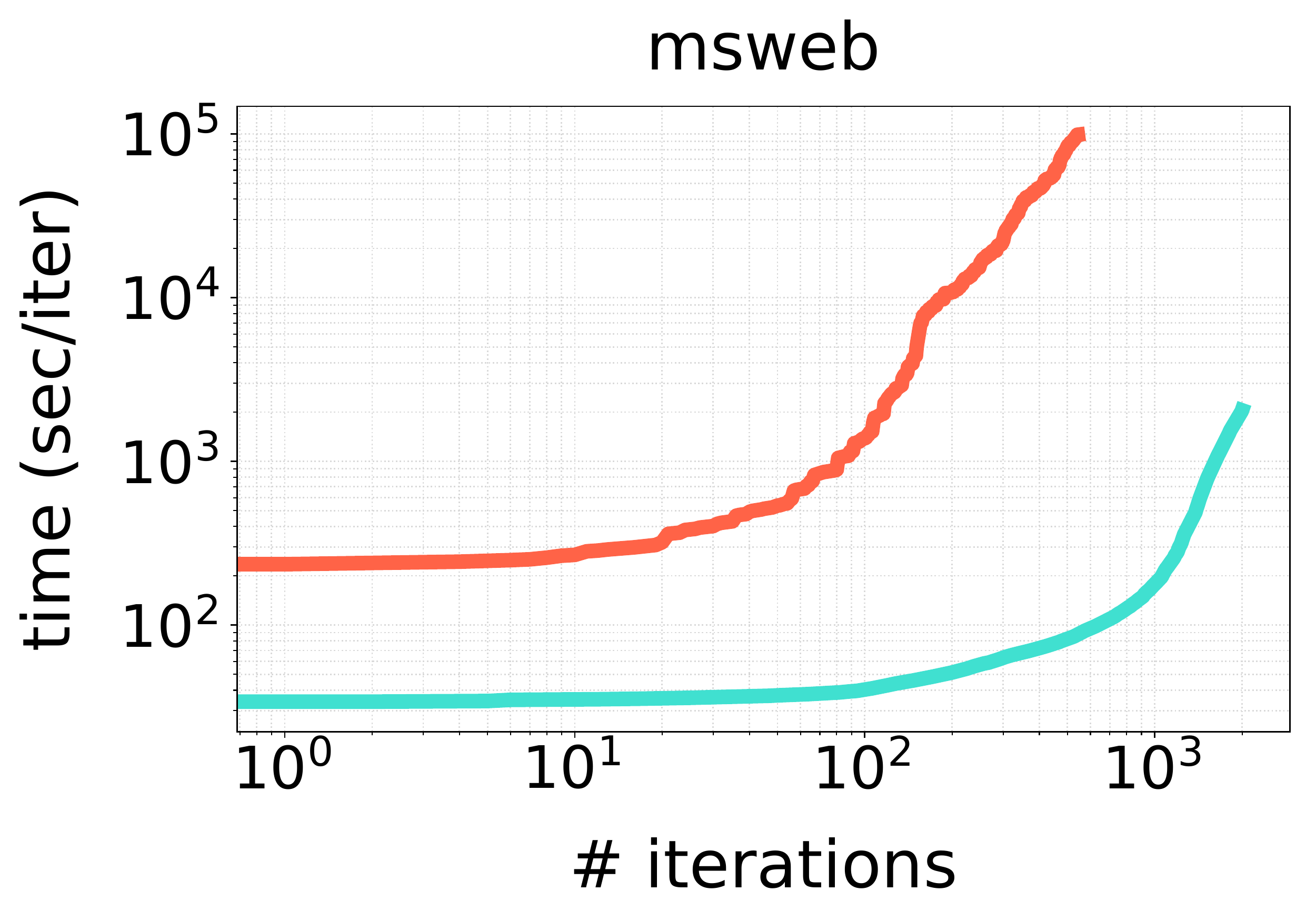}&
    \includegraphics[width=0.23\textwidth]{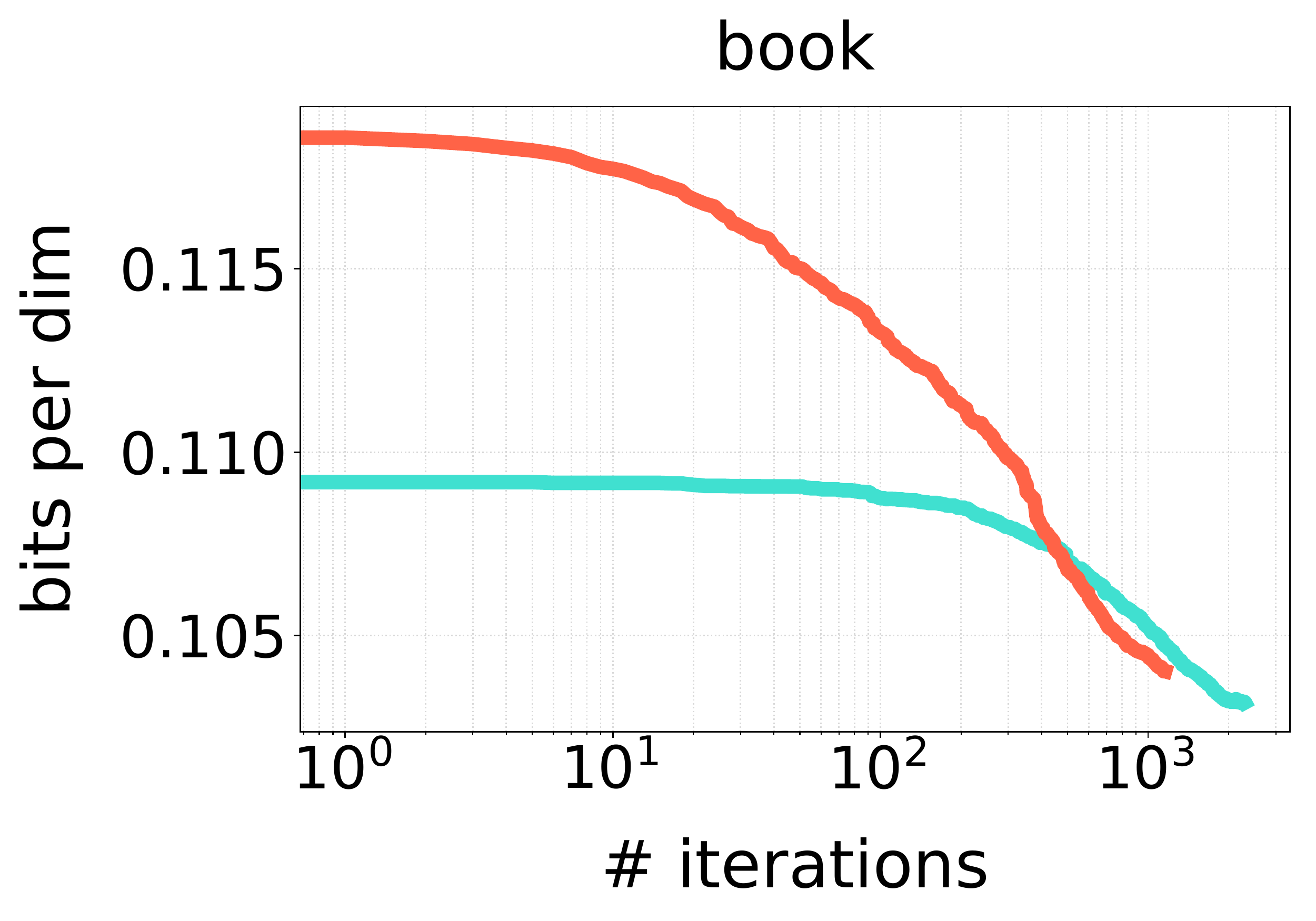}&
    \includegraphics[width=0.23\textwidth]{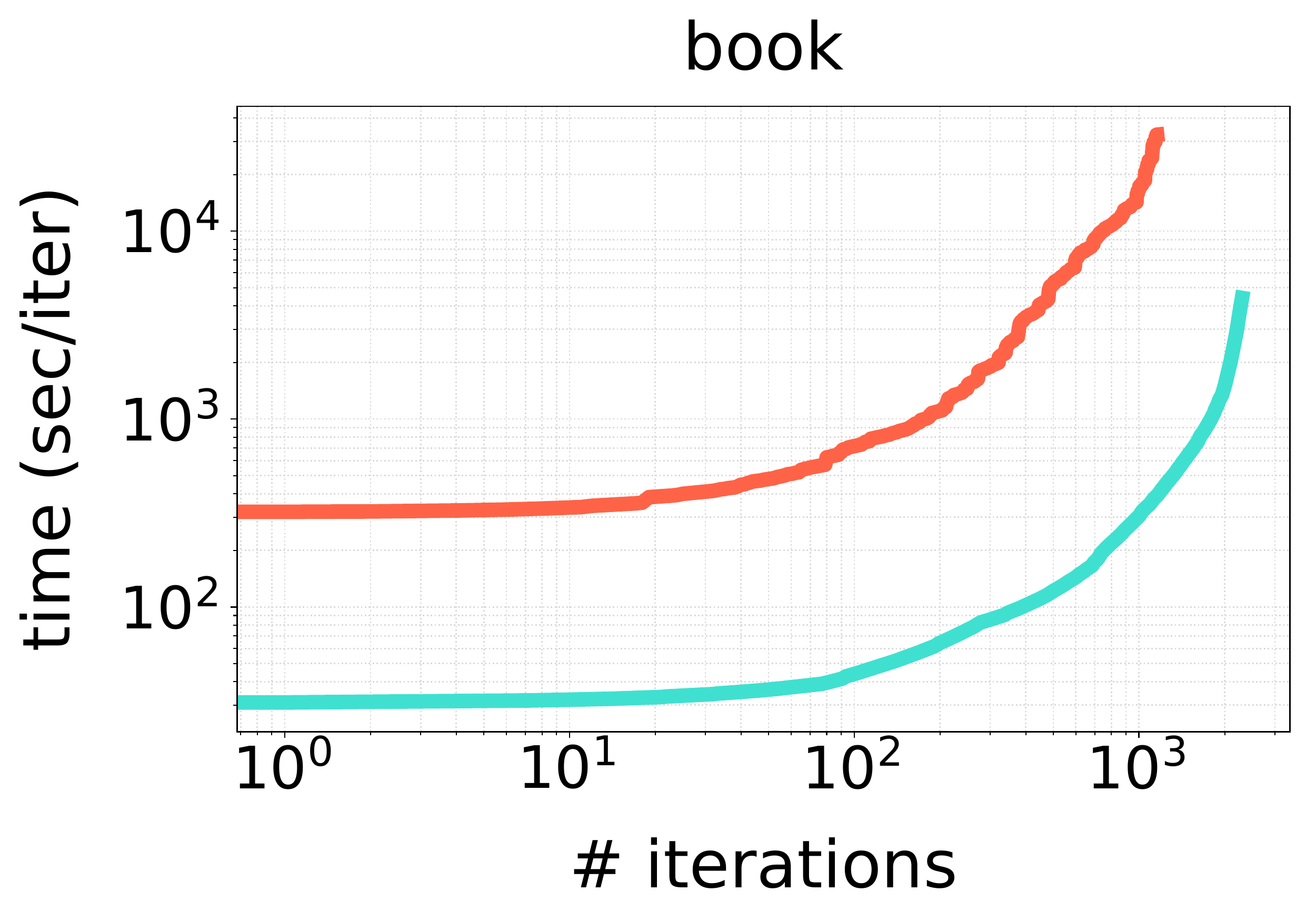}\\
    \includegraphics[width=0.23\textwidth]{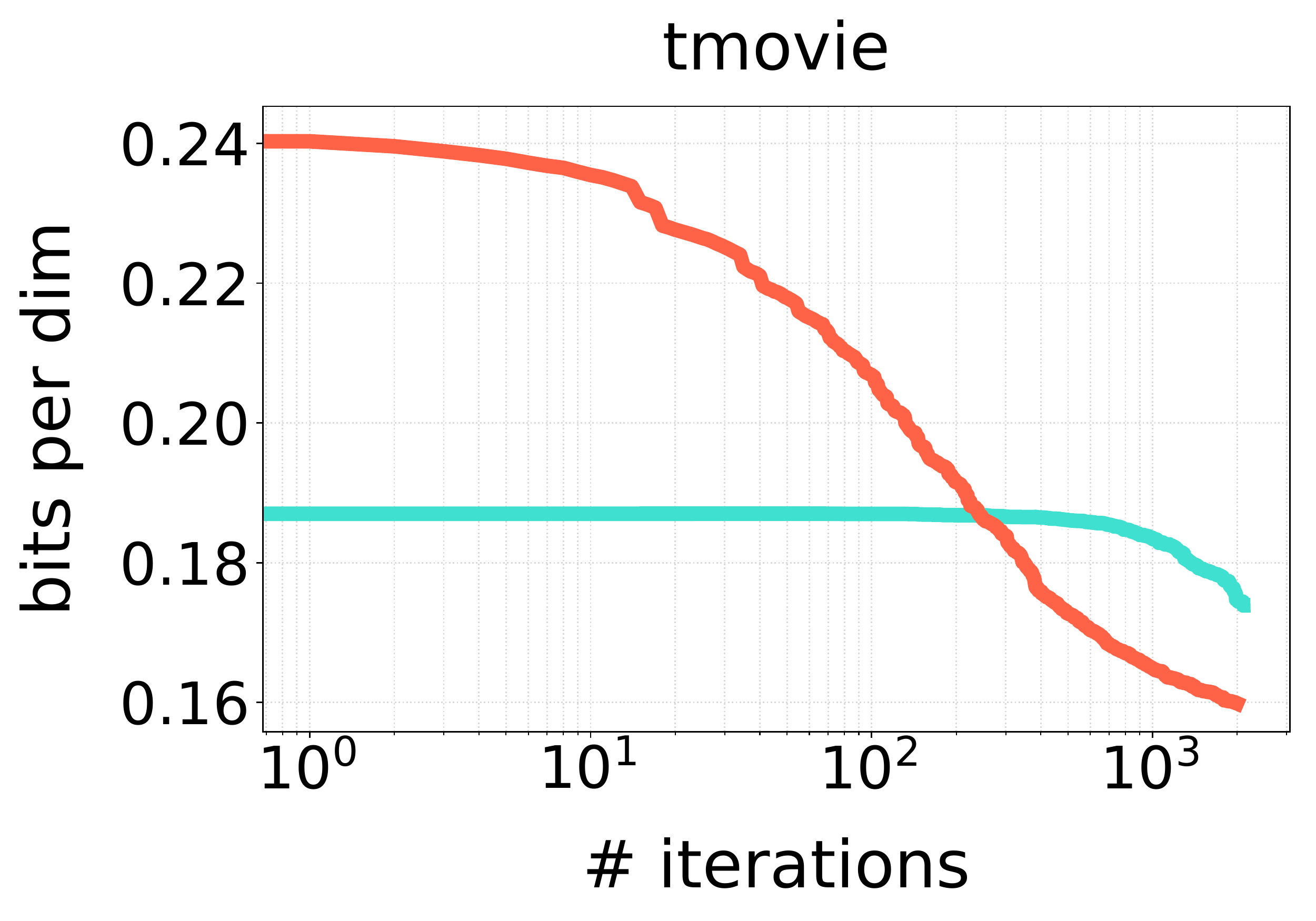}&
    \includegraphics[width=0.23\textwidth]{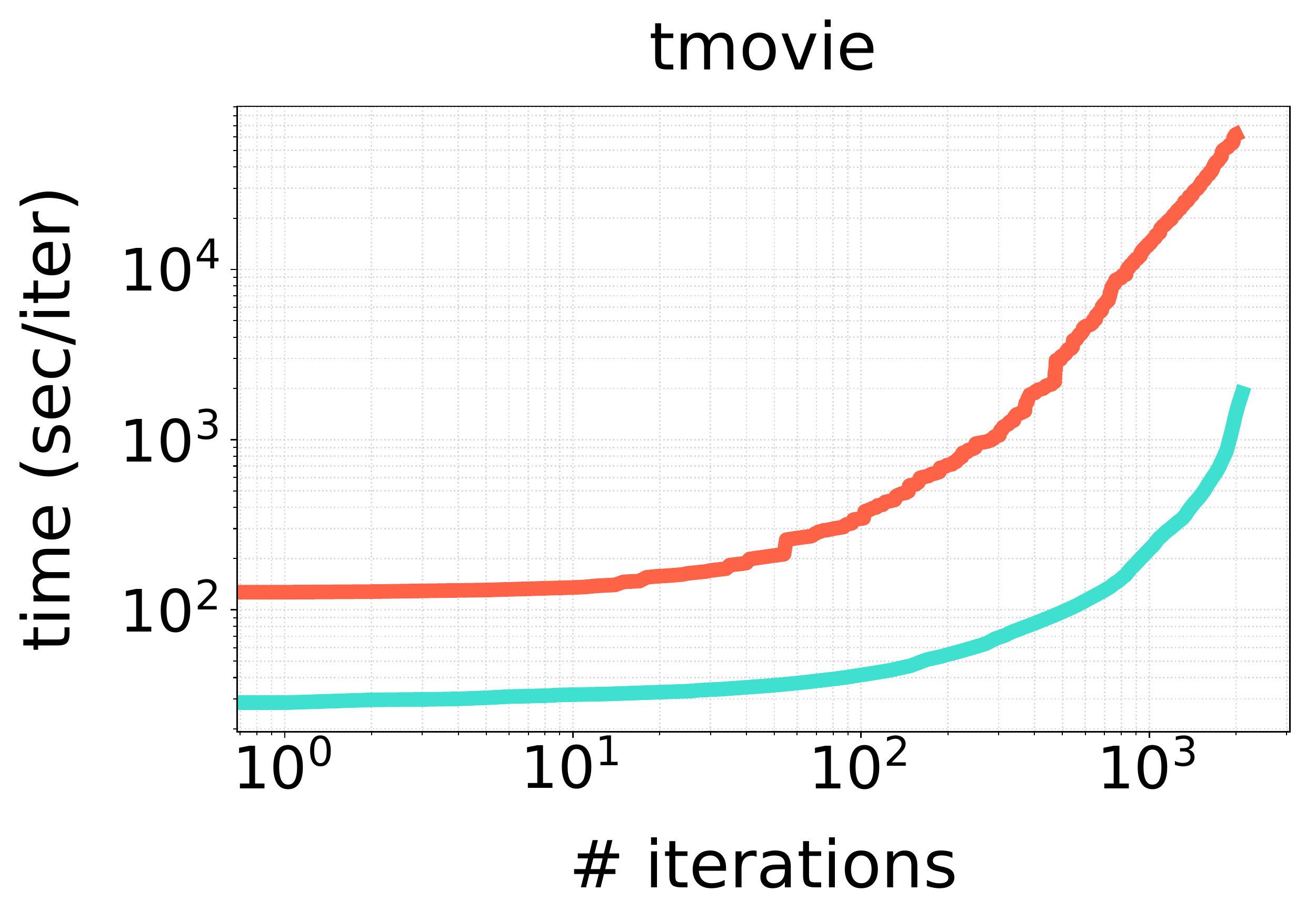}&
    \includegraphics[width=0.23\textwidth]{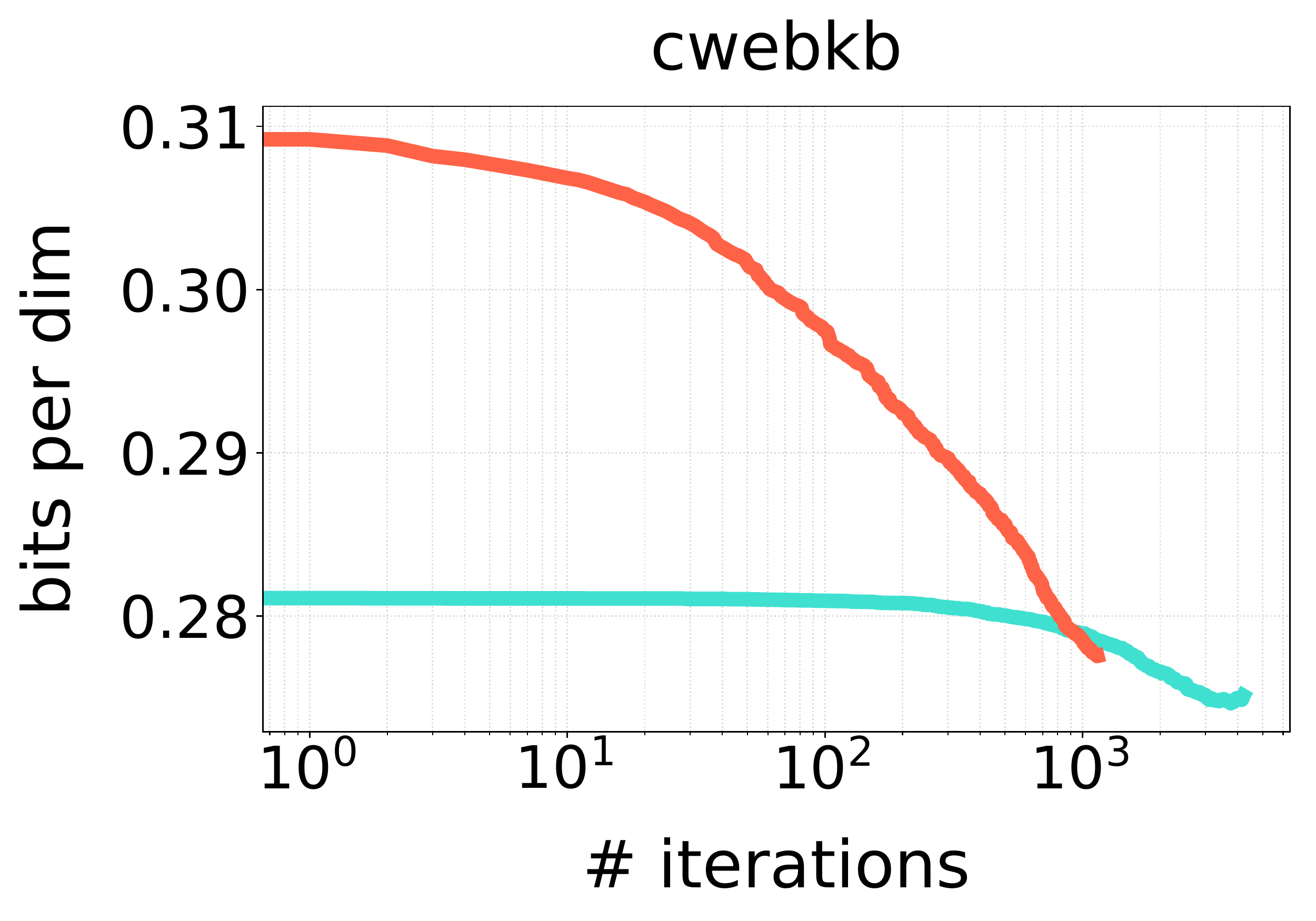}&
    \includegraphics[width=0.23\textwidth]{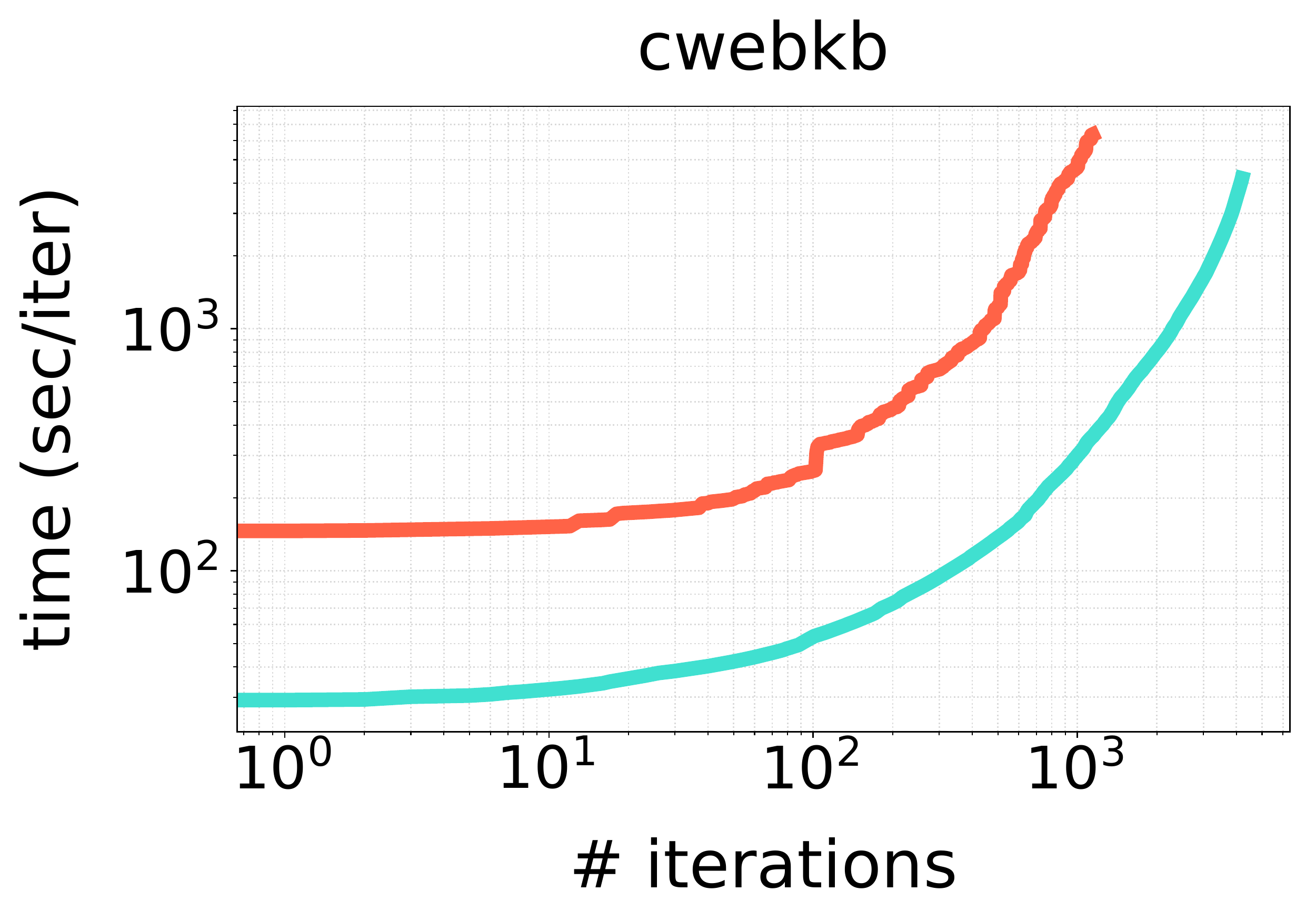}\\
    \includegraphics[width=0.23\textwidth]{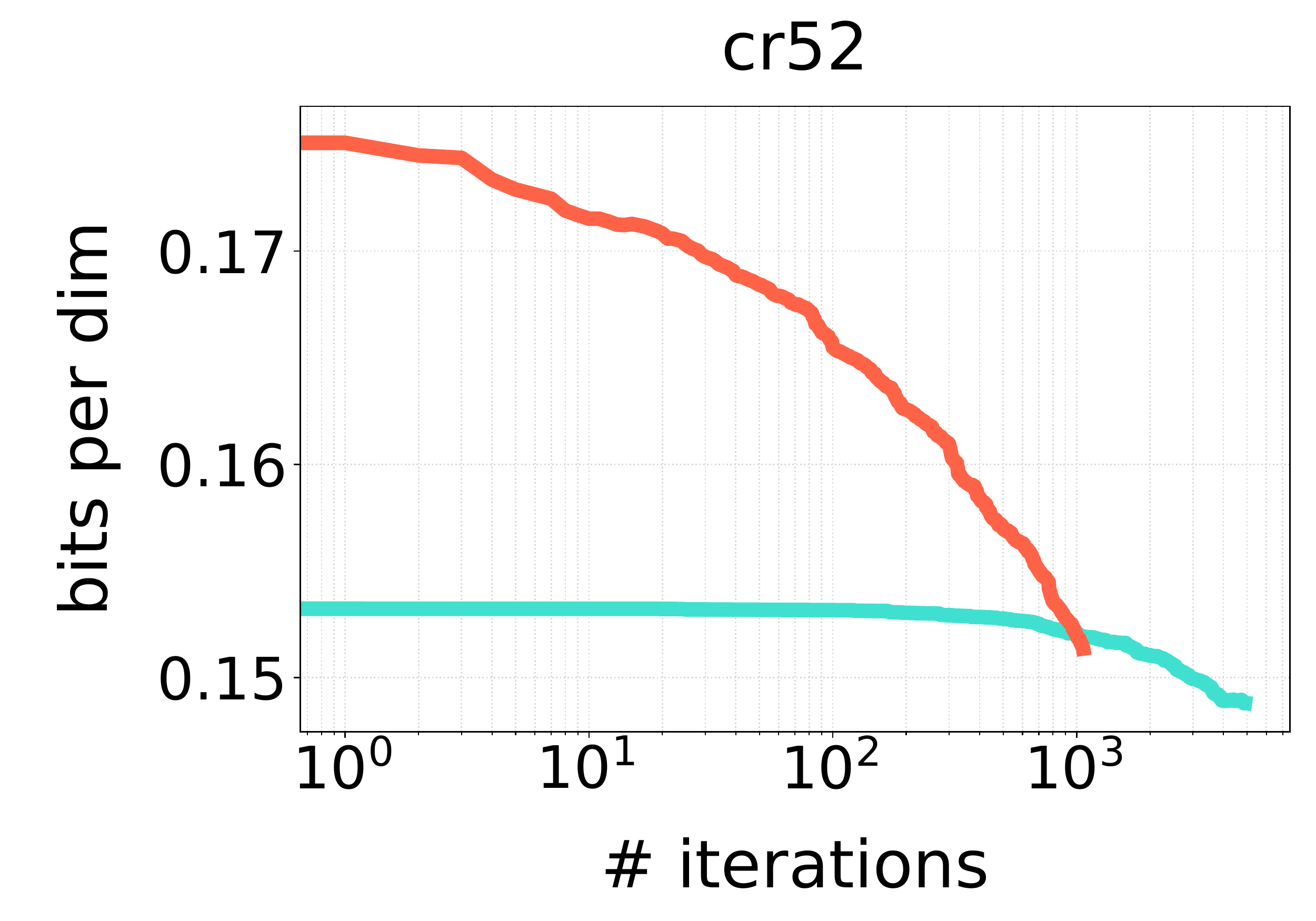}&
    \includegraphics[width=0.23\textwidth]{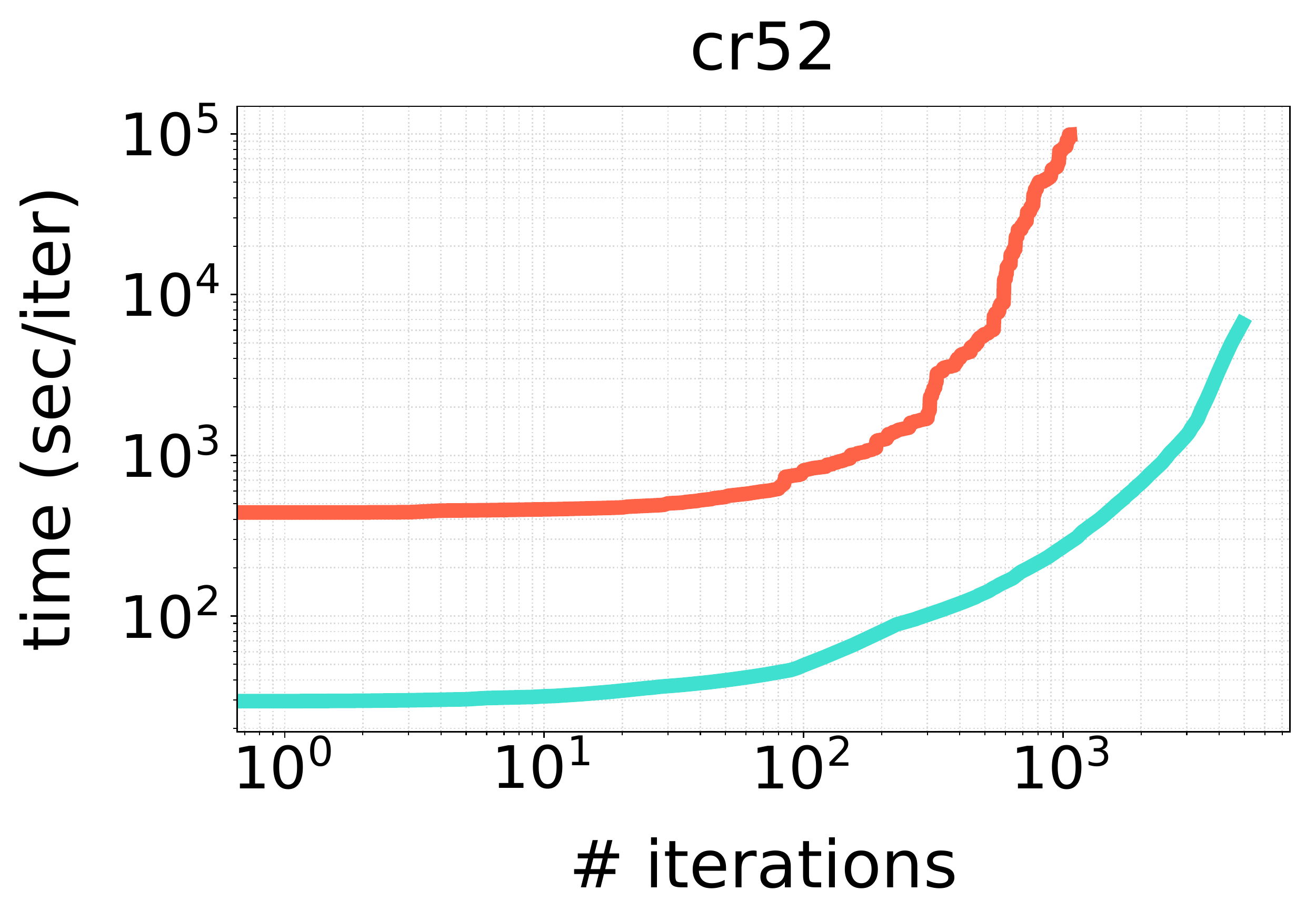}&
    \includegraphics[width=0.23\textwidth]{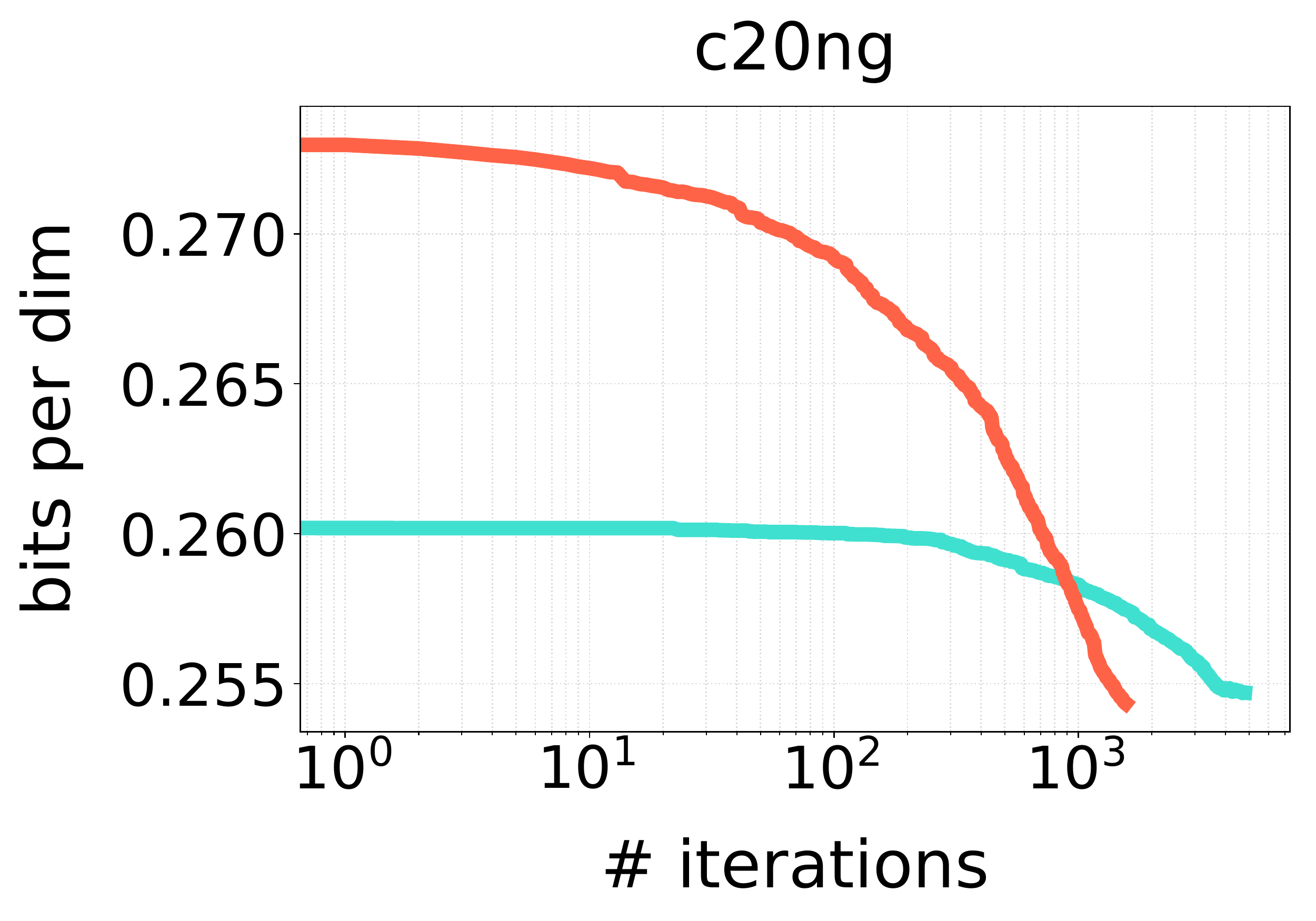}&
    \includegraphics[width=0.23\textwidth]{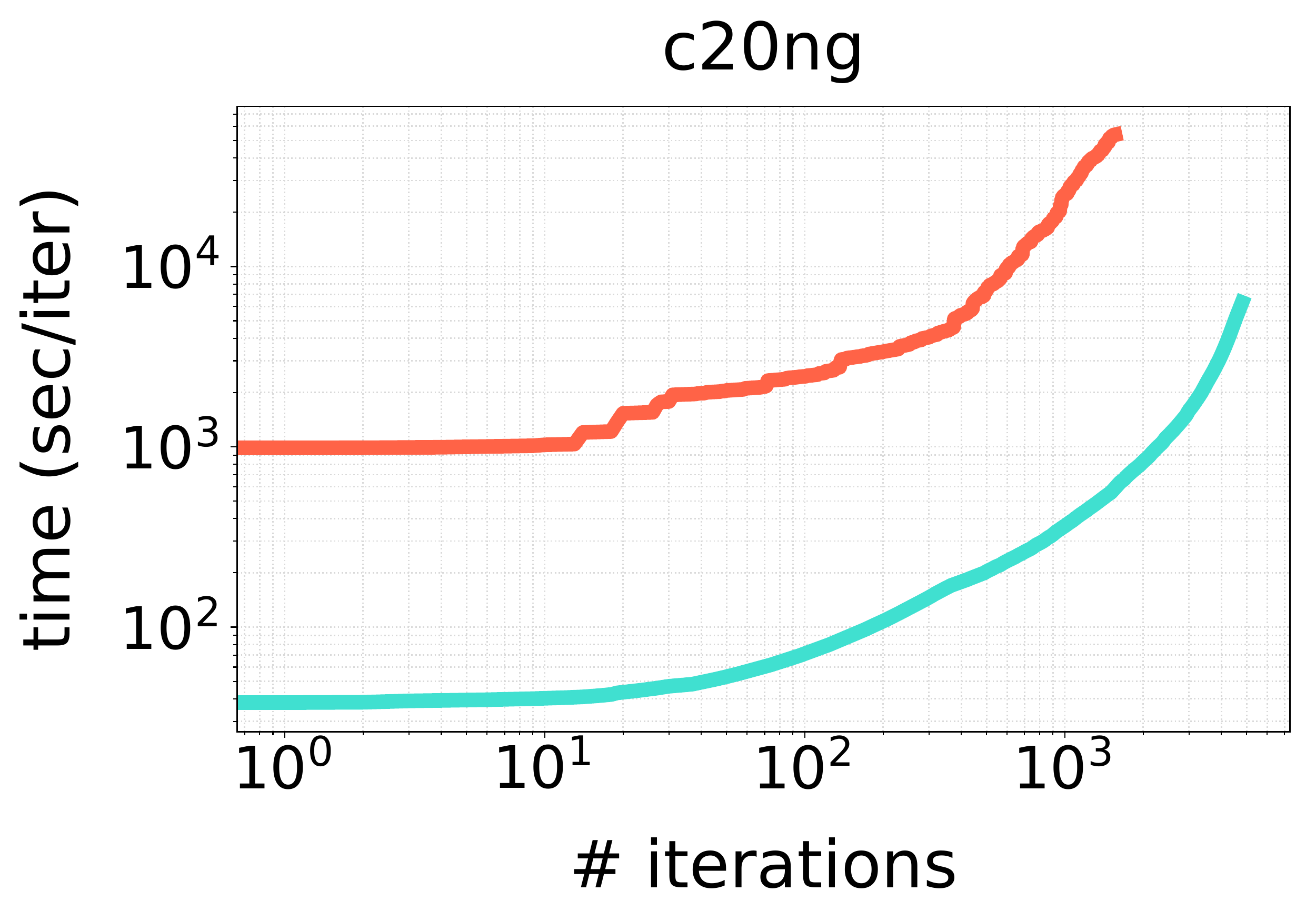}\\
    \includegraphics[width=0.23\textwidth]{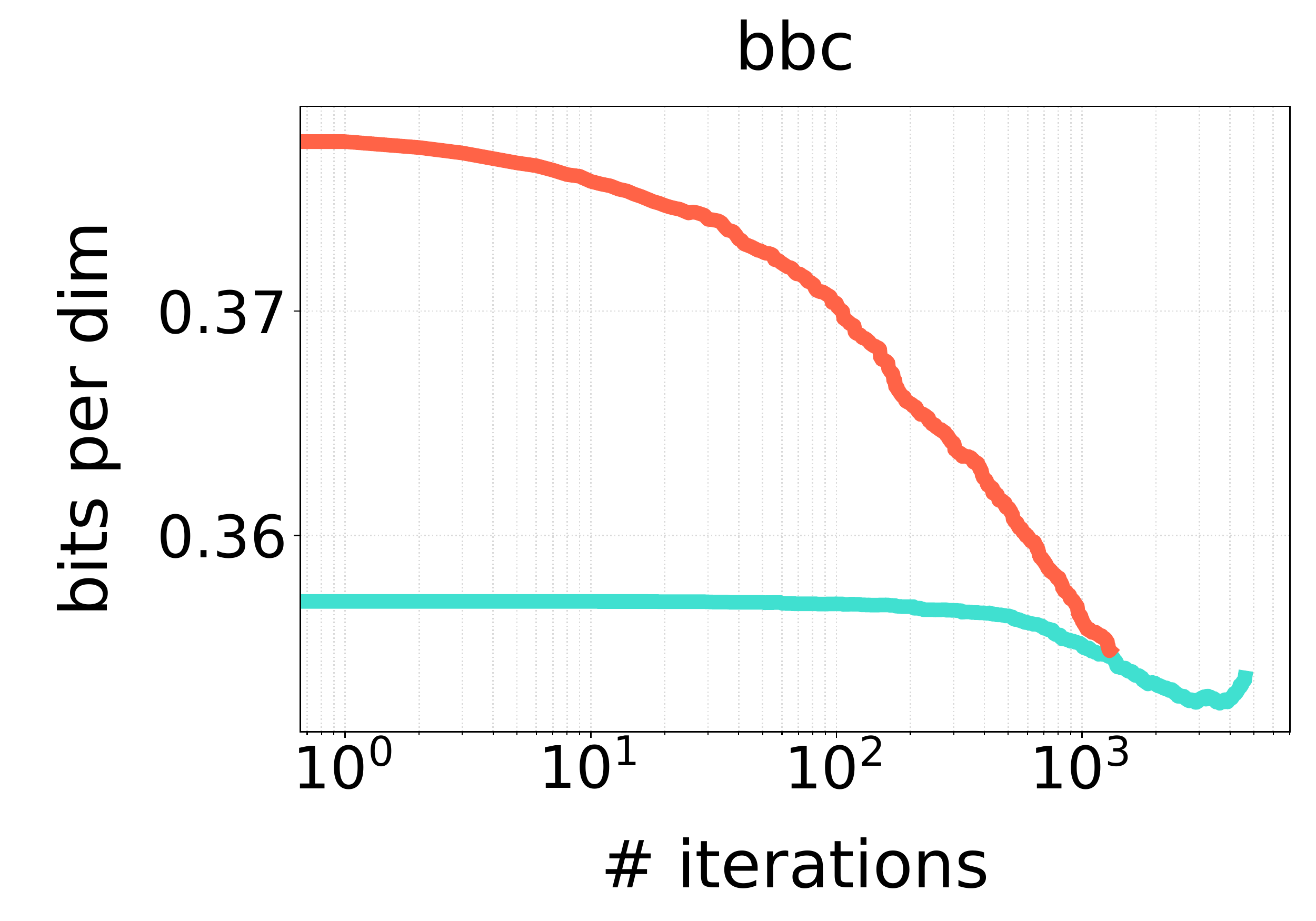}&
    \includegraphics[width=0.23\textwidth]{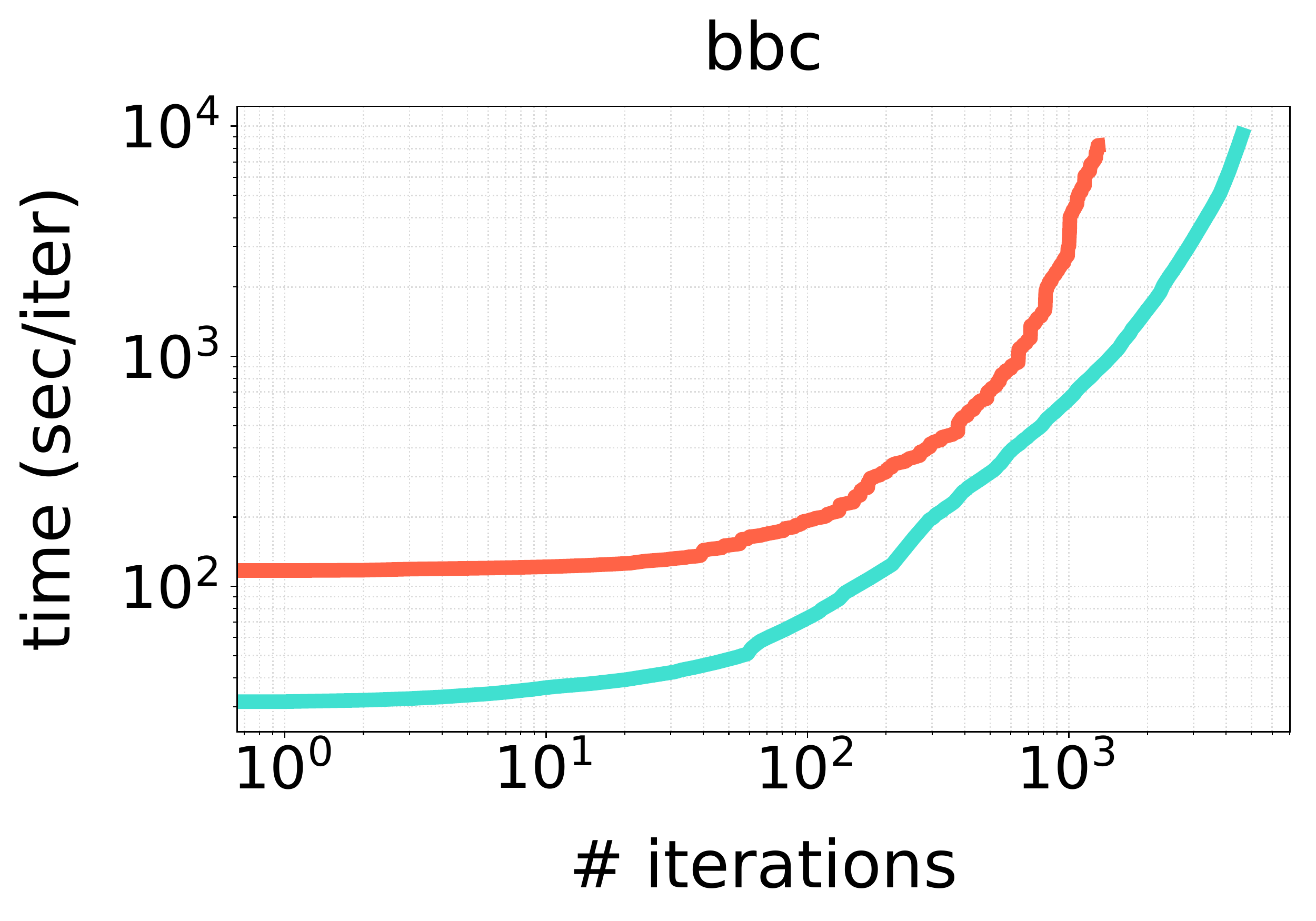}&
    \includegraphics[width=0.23\textwidth]{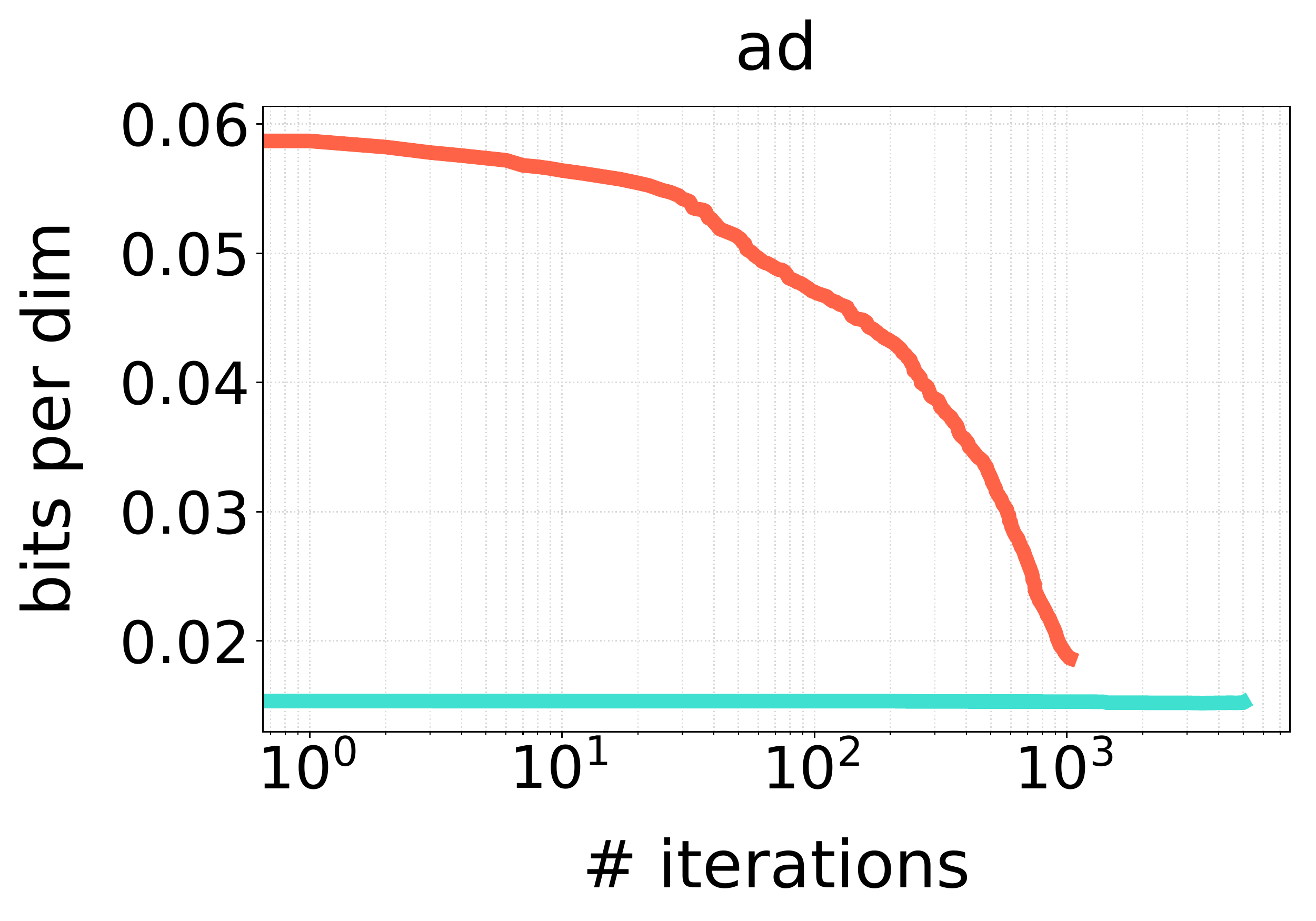}&
    \includegraphics[width=0.23\textwidth]{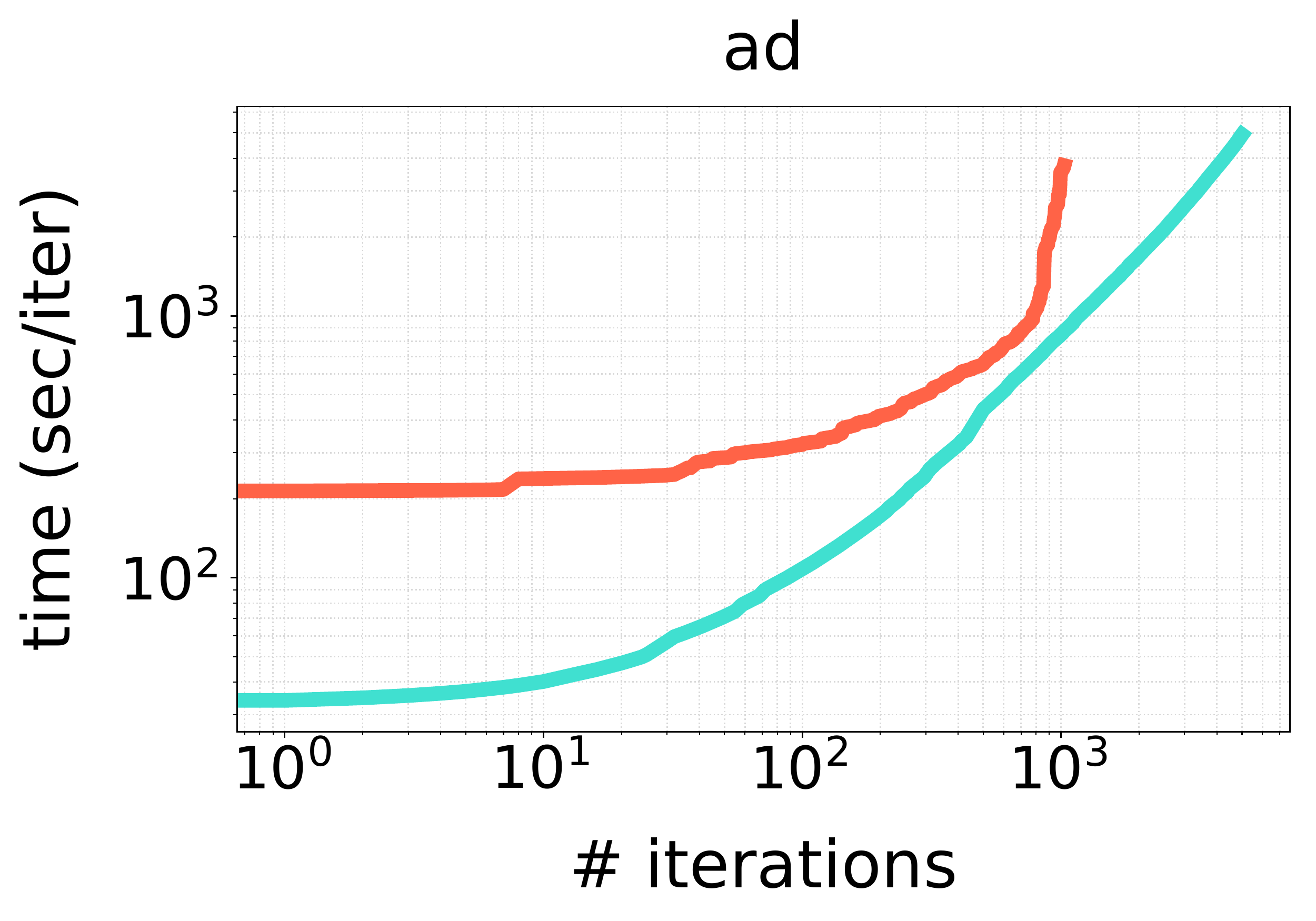}\\

    \caption{\textit{\textbf{Comparing \ourlearner\ and \learnpsdd\ single models performance and learning times.}} For each of the 20 benchmark datasets, we report 1) the test bits per dimensions (bpd) (y-axis) for each iteration (x-axis) on column 2 and column 4, and 2) seconds per iteration during learning (y-axis) for each iteration (x-axis) on column 2 and column 4 as scored by \ourlearner\ (blue) and \learnpsdd\ (red). The plots for the same dataset is aligned next to each other as comparison.}
    \label{tab:fig-vspsdd}
\end{longtable}
\newpage
\subsection{Statistical Tests}
\label{sec:app-tests}
Since we only have the log-likelihood per sample available for SelSPN, we run pairwise Wilcoxon signed rank tests to compare \ourlearner\ and SelSPN more rigorously. The p-values are reported in Table \ref{tab:p-value}, in 4 out of 20 datasets, the results are statistically equivalent

\begin{table}[!ht]
    \centering
    \setlength{\tabcolsep}{1.8pt}
    \small
    \begin{sc}
    \begin{tabular}{r r r}
    \toprule
         dataset    & \learnpsdd  & SelSPN          \\
         \midrule
         nltcs      & \textbf{5.34E-02} &\textbf{1.35e-02}\\
         msnbc      & 0.00E+00 &0.00e+00\\
         kdd        & 0.00E+00 &0.00e+00\\
         plants     & 6.58E-25 &3.46e-56\\
         audio      & 1.67E-30 &4.73e-38\\
         jester     & 2.37E-45 &2.05e-20\\
         netflix    & 5.29E-04 &1.38e-08\\
         accidents  & 1.46E-102 &3.08e-279\\
         retail     & \textbf{2.48E-01} &6.54e-05\\
         pumsb-star & 1.52E-12 &2.83e-269\\
         dna        & 8.39E-164 &2.95e-189\\
         kosarek    & 5.17E-03 &5.93e-21\\
         msweb      & 2.76E-100 &7.06e-114\\
         book       & \textbf{6.48E-02} &\textbf{7.60e-01}\\
         eachmovie  & 2.42E-53 &2.33e-45\\
         webkb      & \textbf{1.08E-01} &\textbf{4.07e-01}\\
         routers-52 &3.56E-04 & 2.51e-87\\
         20news-grp &\textbf{1.40E-02} & 2.37e-07\\
         bbc        &1.96E-04 & \textbf{4.33e-01}\\
         ad         &9.90E-65 &1.64e-09\\
         \bottomrule
    \end{tabular}
    \end{sc}
    \caption{Pairwise Wilcoxon signed rank test p-values for the comparisons of test log-likelihoods for \ourlearner\ against \learnpsdd\ SelSPN on all datasets. Bold values indicate the results are not statistically significant when picking confidence value 99\%.}
    \label{tab:p-value}
\end{table}{}

\newpage

\subsection{Initializations}
\label{sec:app-init}
Here we report the effect of different initialization methods (CLT and independent) on each dataset in Table~\ref{tab:fig-init-all}.
\begin{longtable}[!ht]{cccc}
    \centering
    \includegraphics[width=0.23\textwidth]{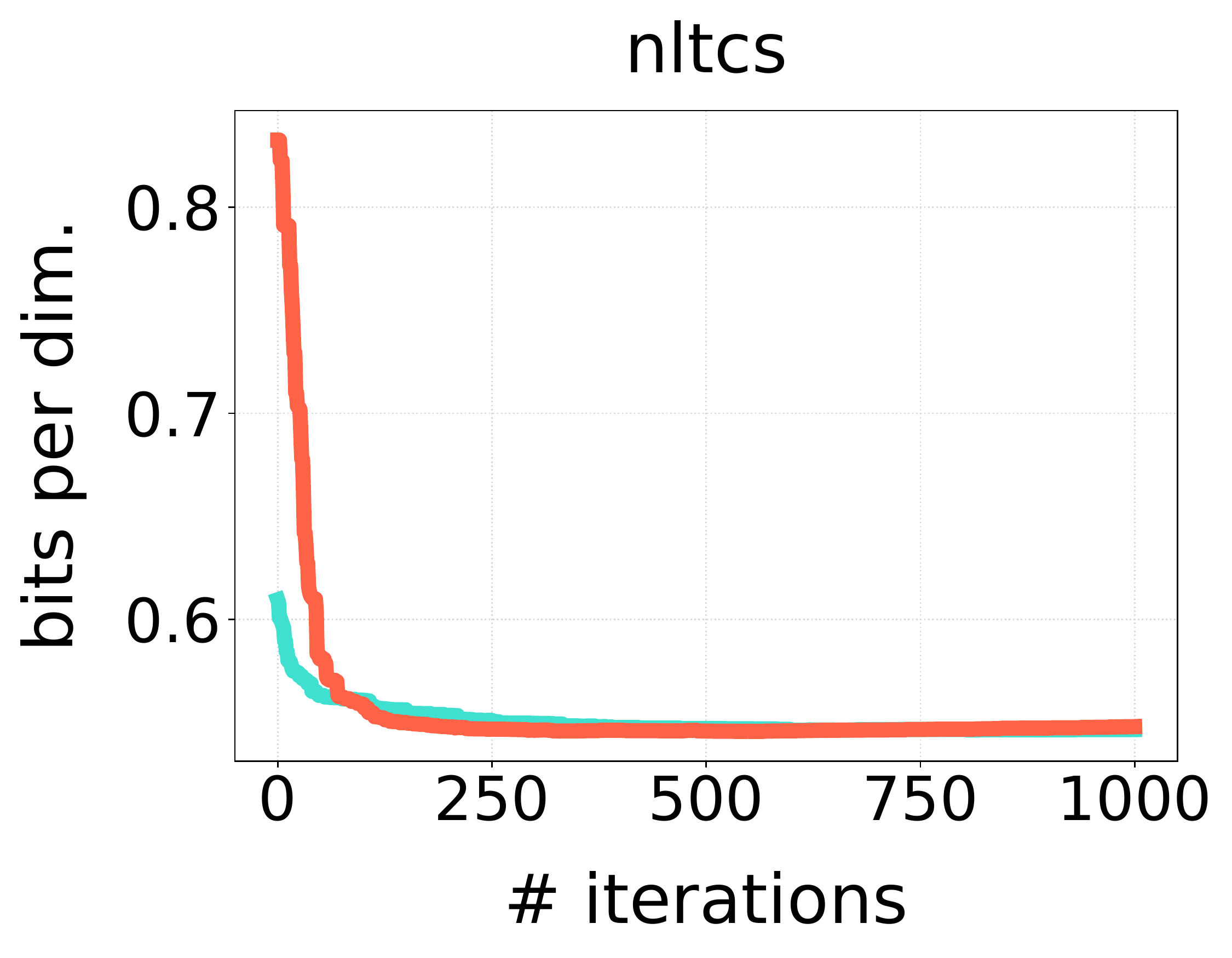}&
    \includegraphics[width=0.23\textwidth]{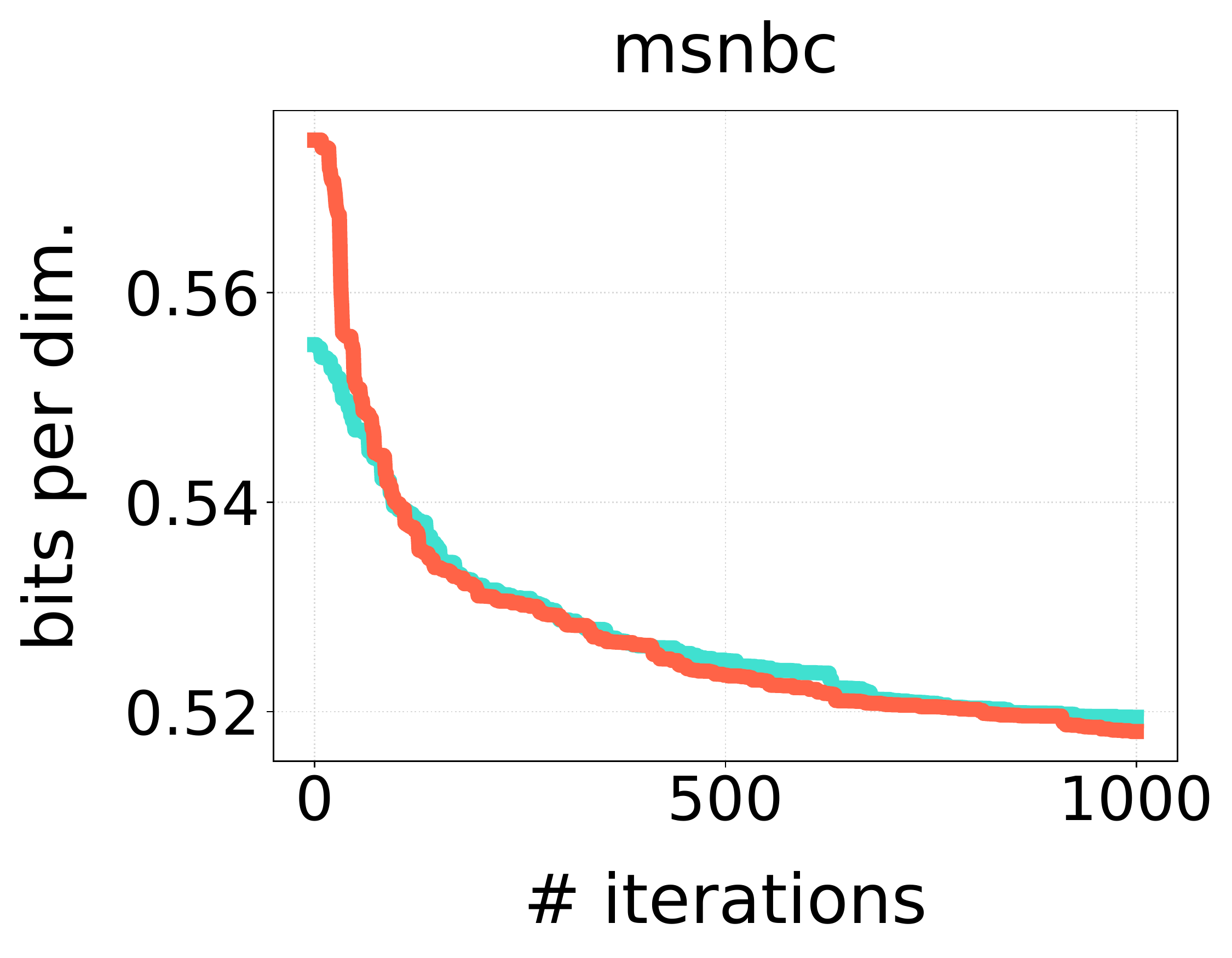}&
    \includegraphics[width=0.23\textwidth]{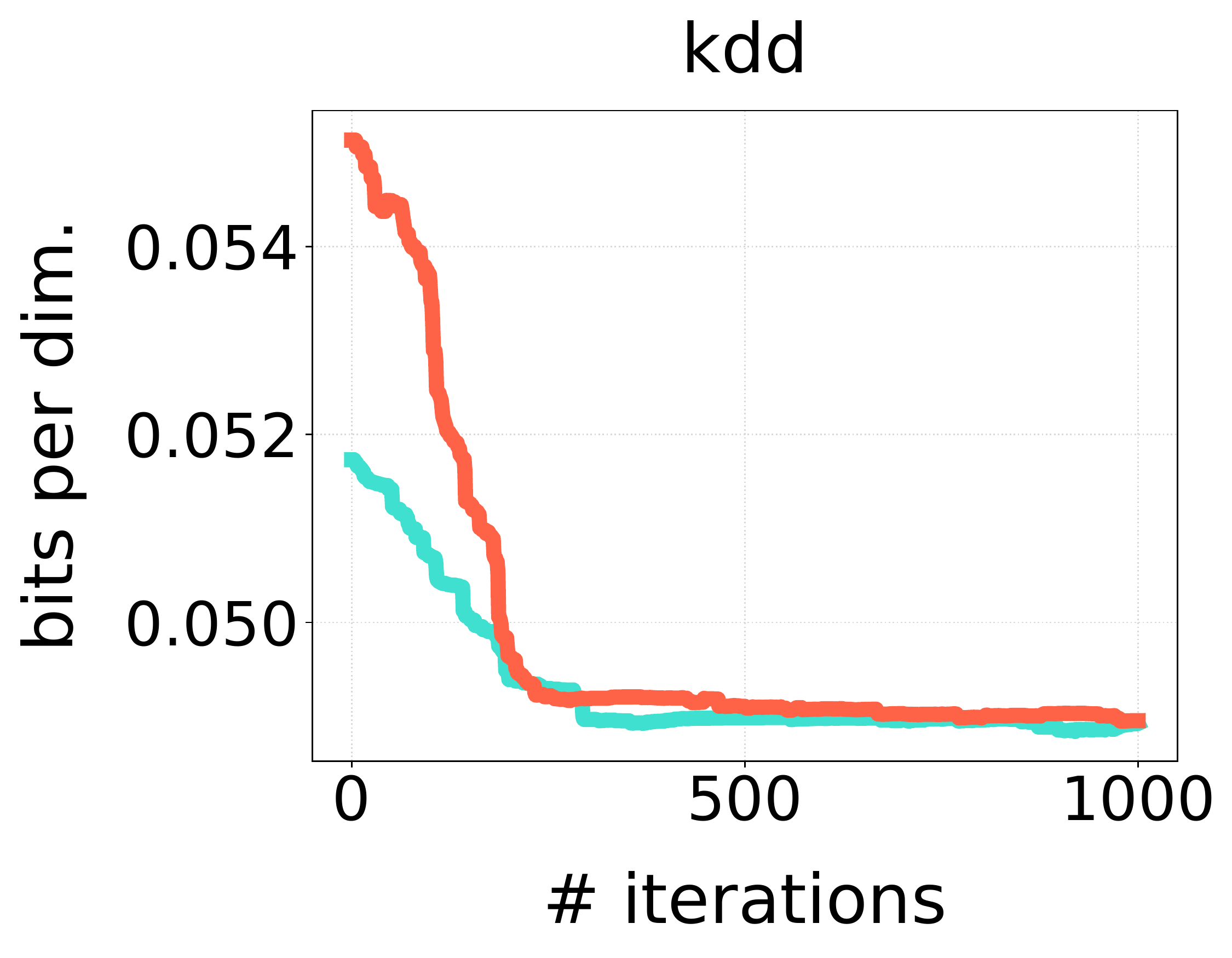}&
    \includegraphics[width=0.23\textwidth]{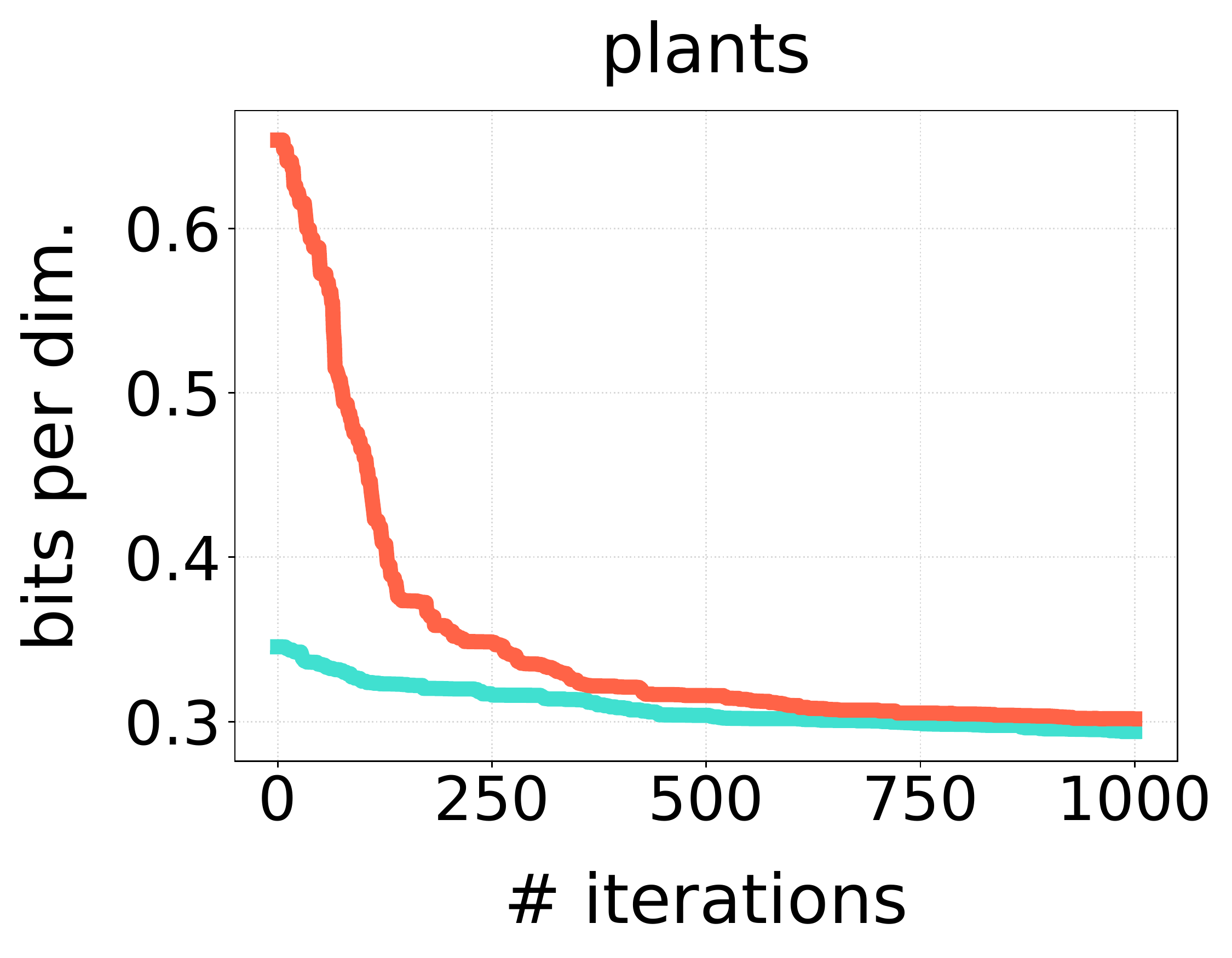}\\
    \includegraphics[width=0.23\textwidth]{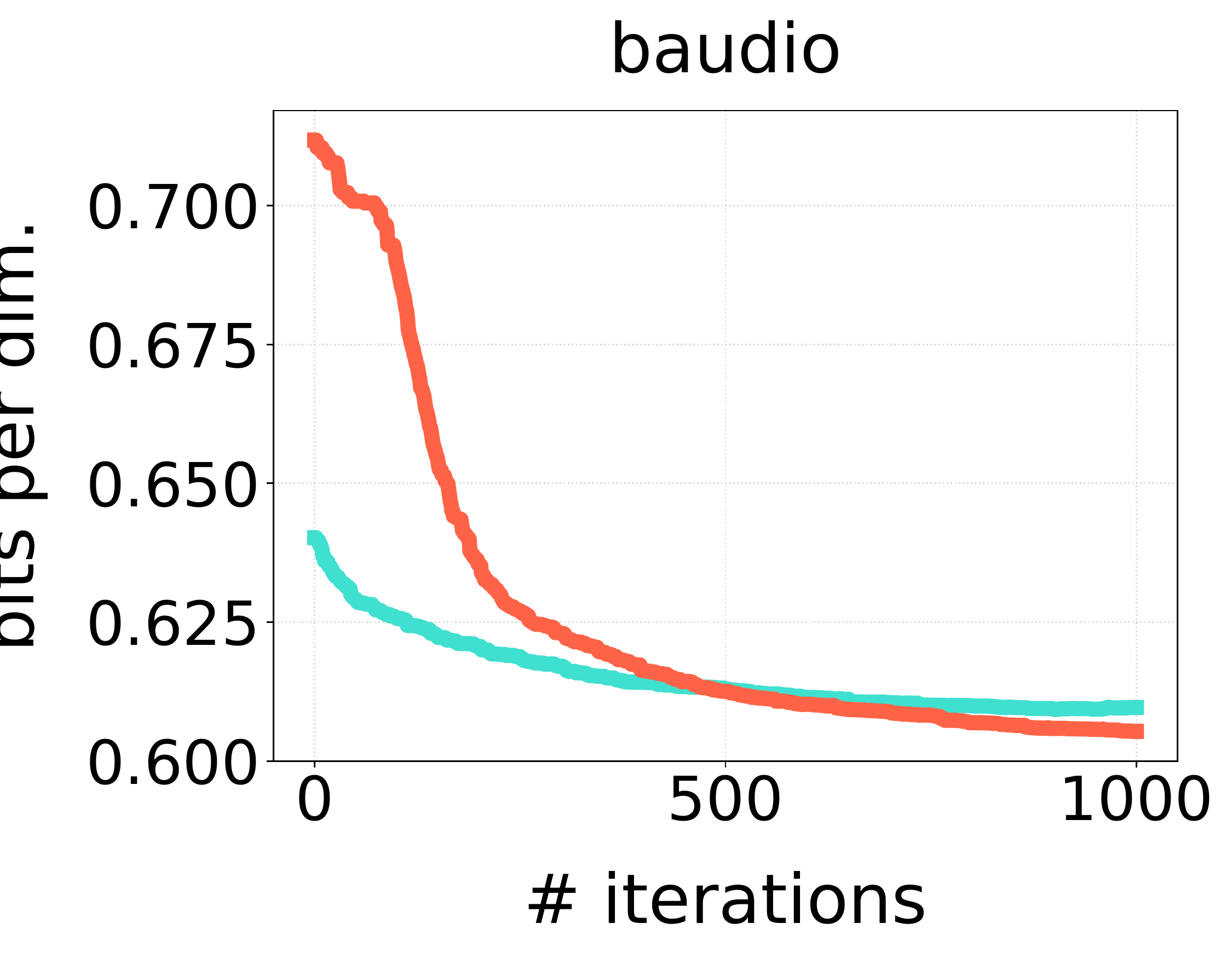}&
    \includegraphics[width=0.23\textwidth]{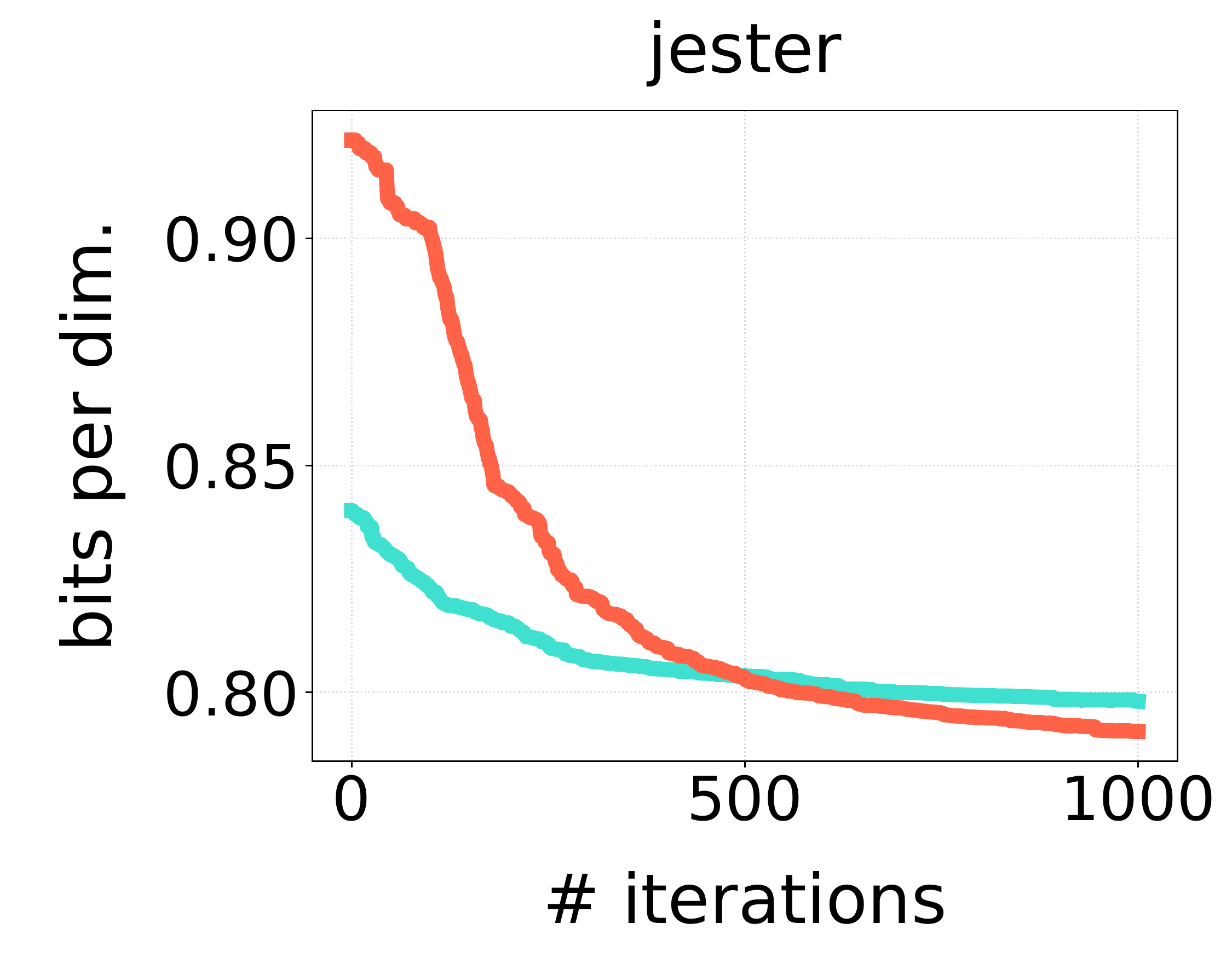}&
    \includegraphics[width=0.23\textwidth]{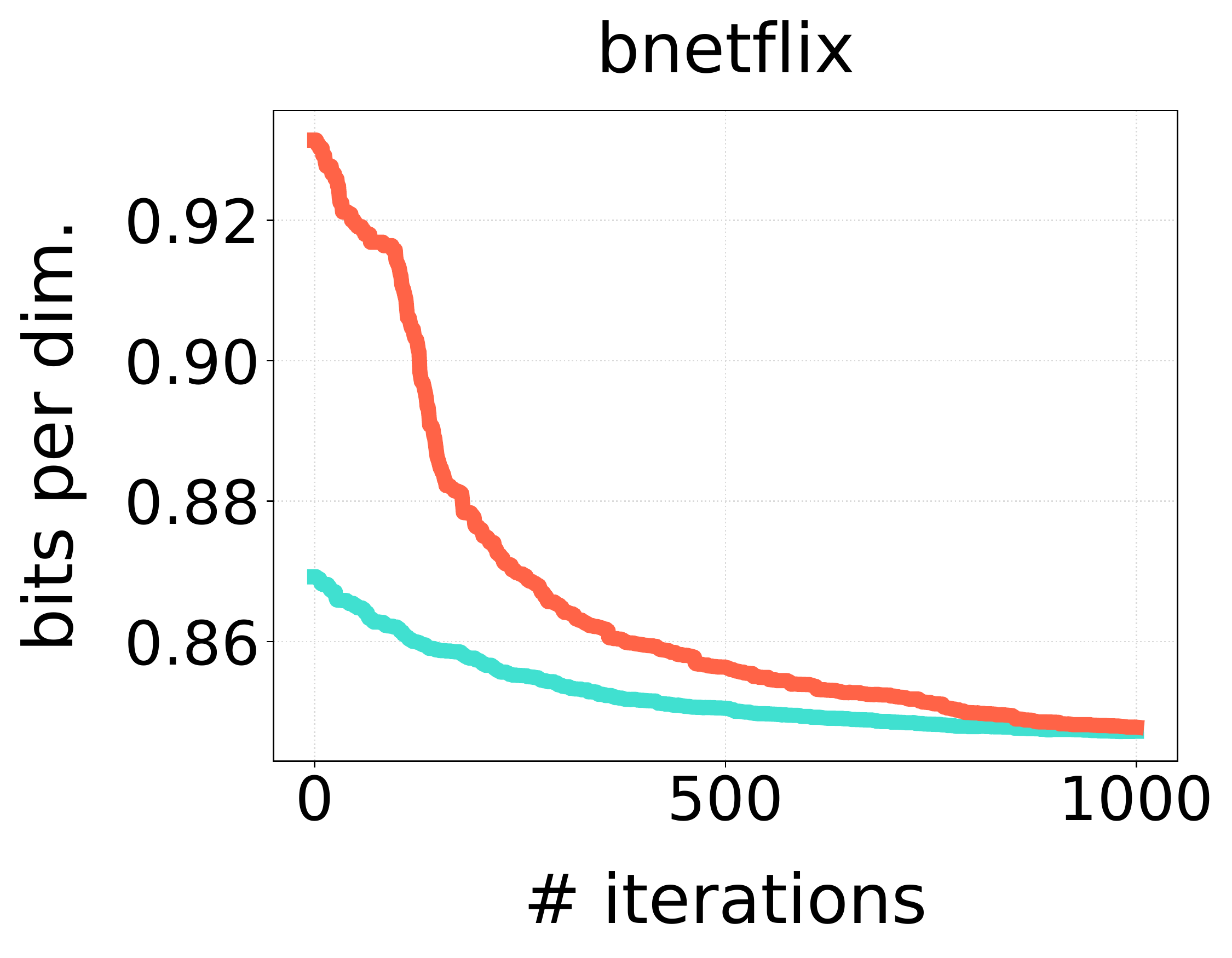}&
    \includegraphics[width=0.23\textwidth]{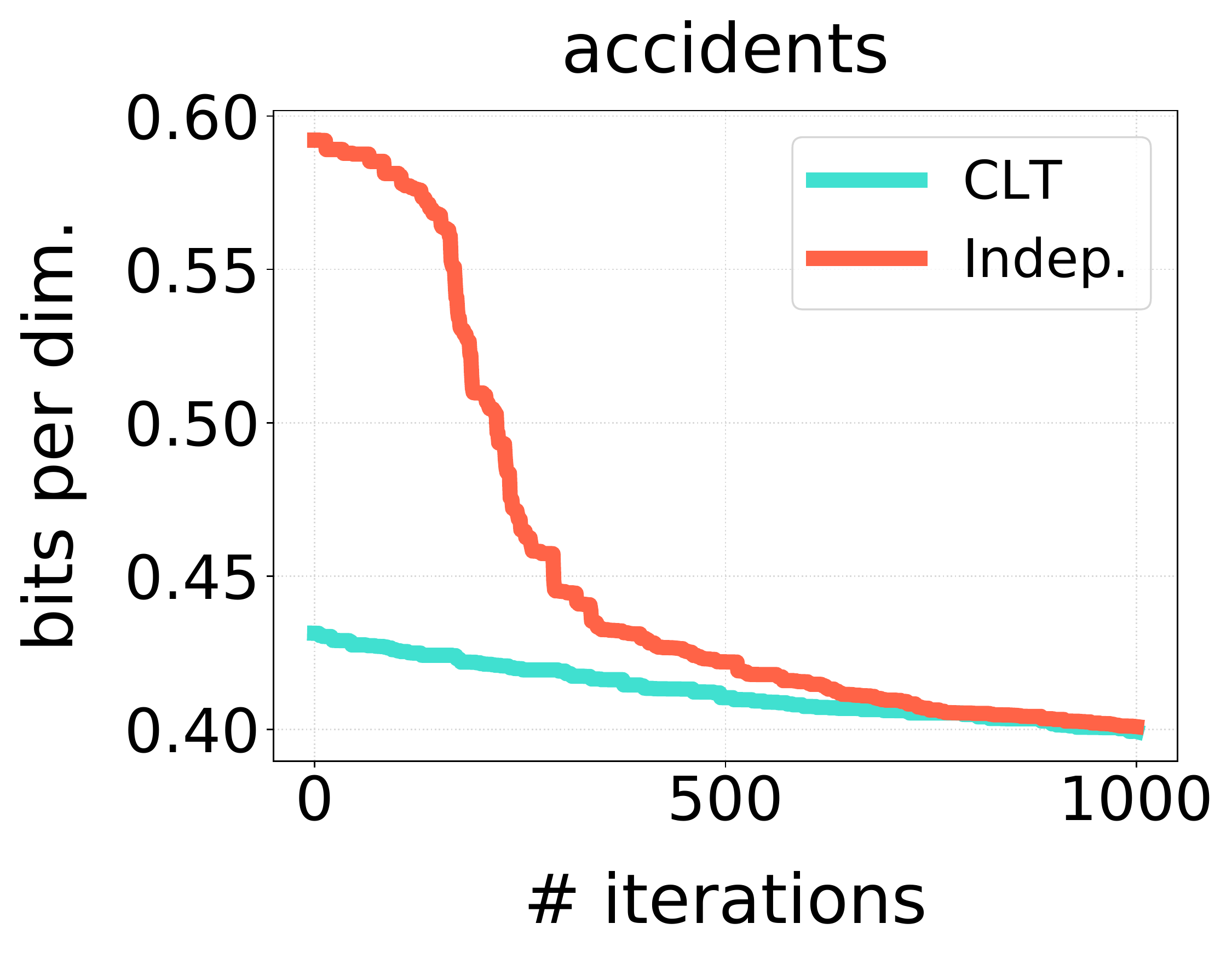}\\
    \includegraphics[width=0.23\textwidth]{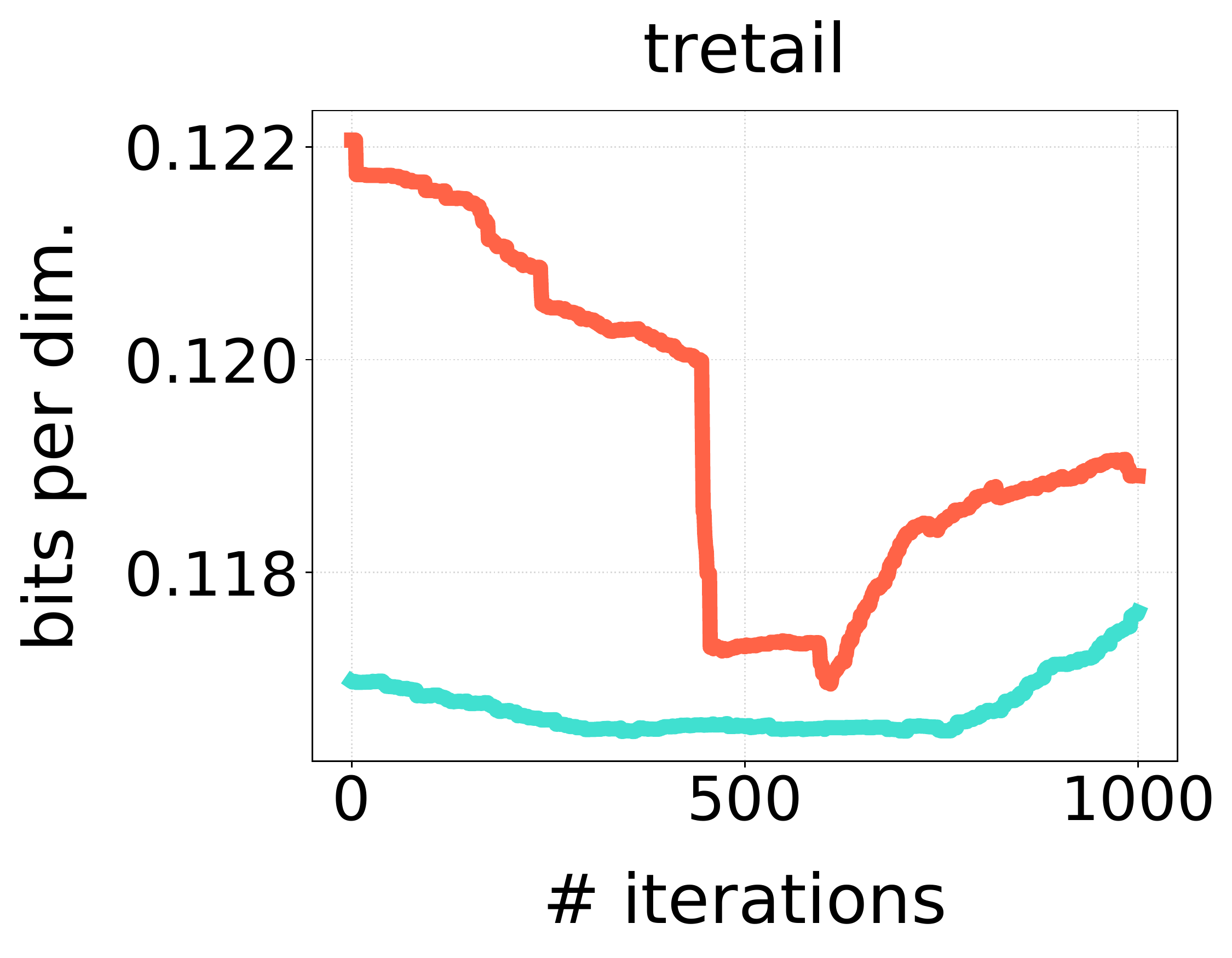}&
    \includegraphics[width=0.23\textwidth]{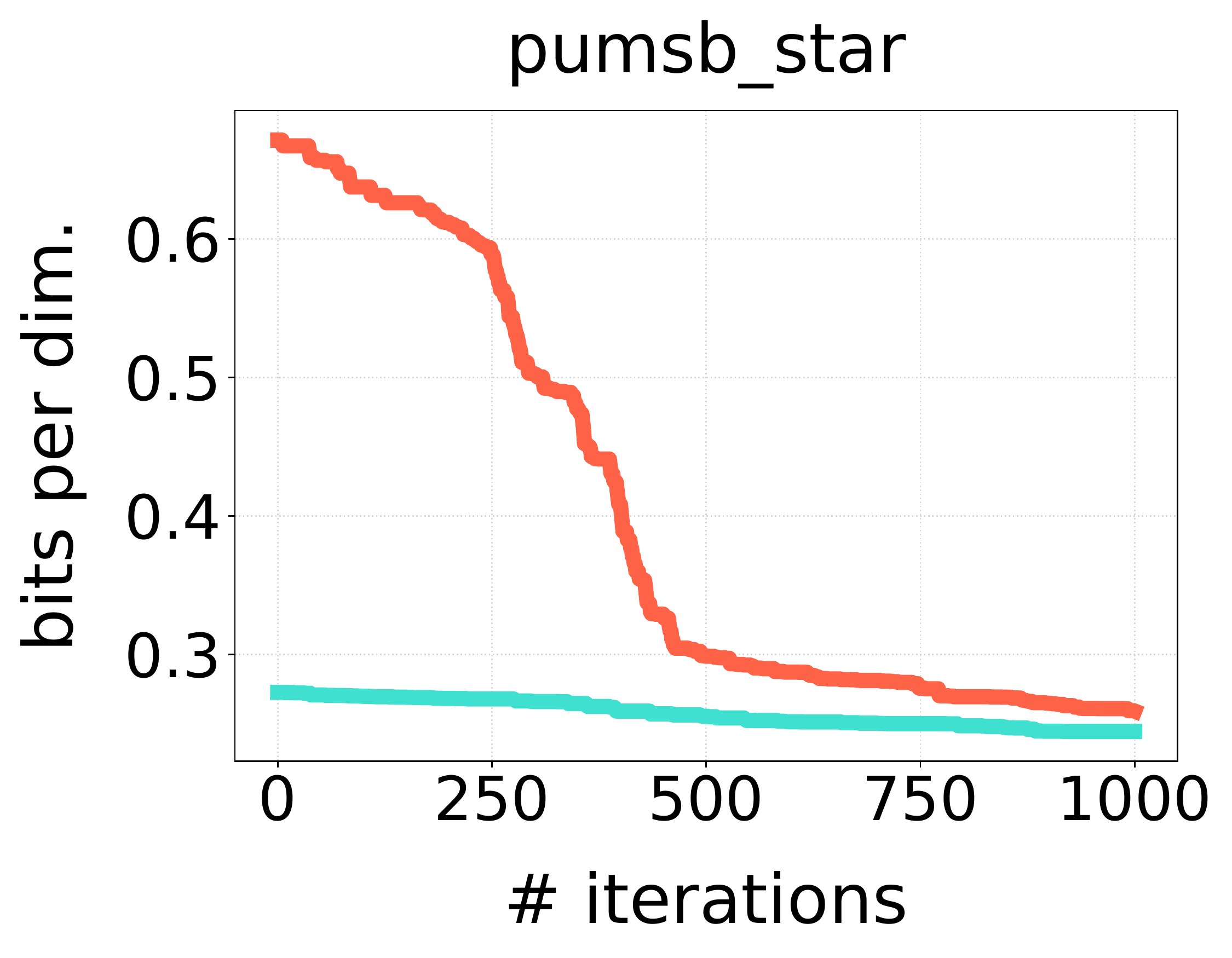}&
    \includegraphics[width=0.23\textwidth]{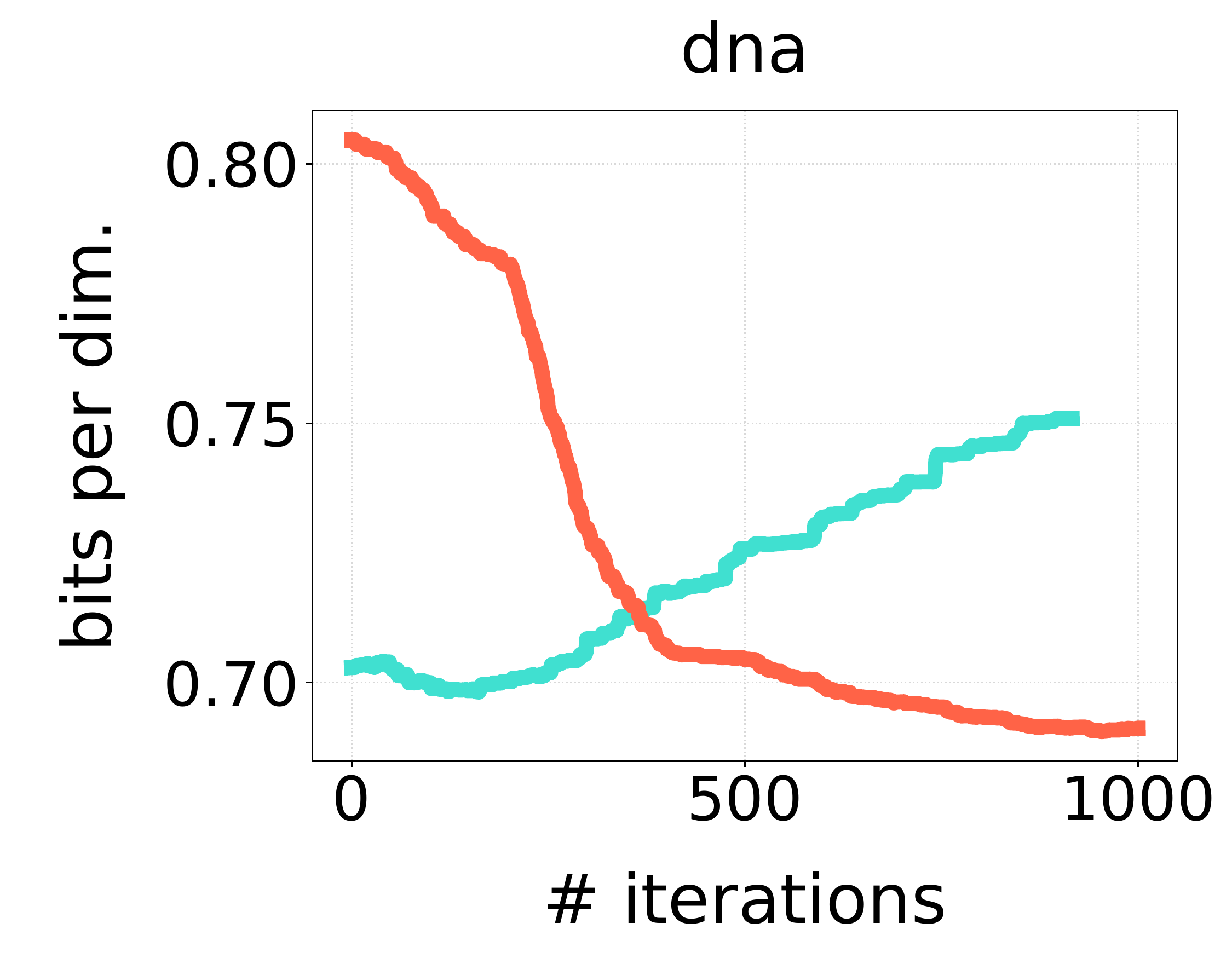}&
    \includegraphics[width=0.23\textwidth]{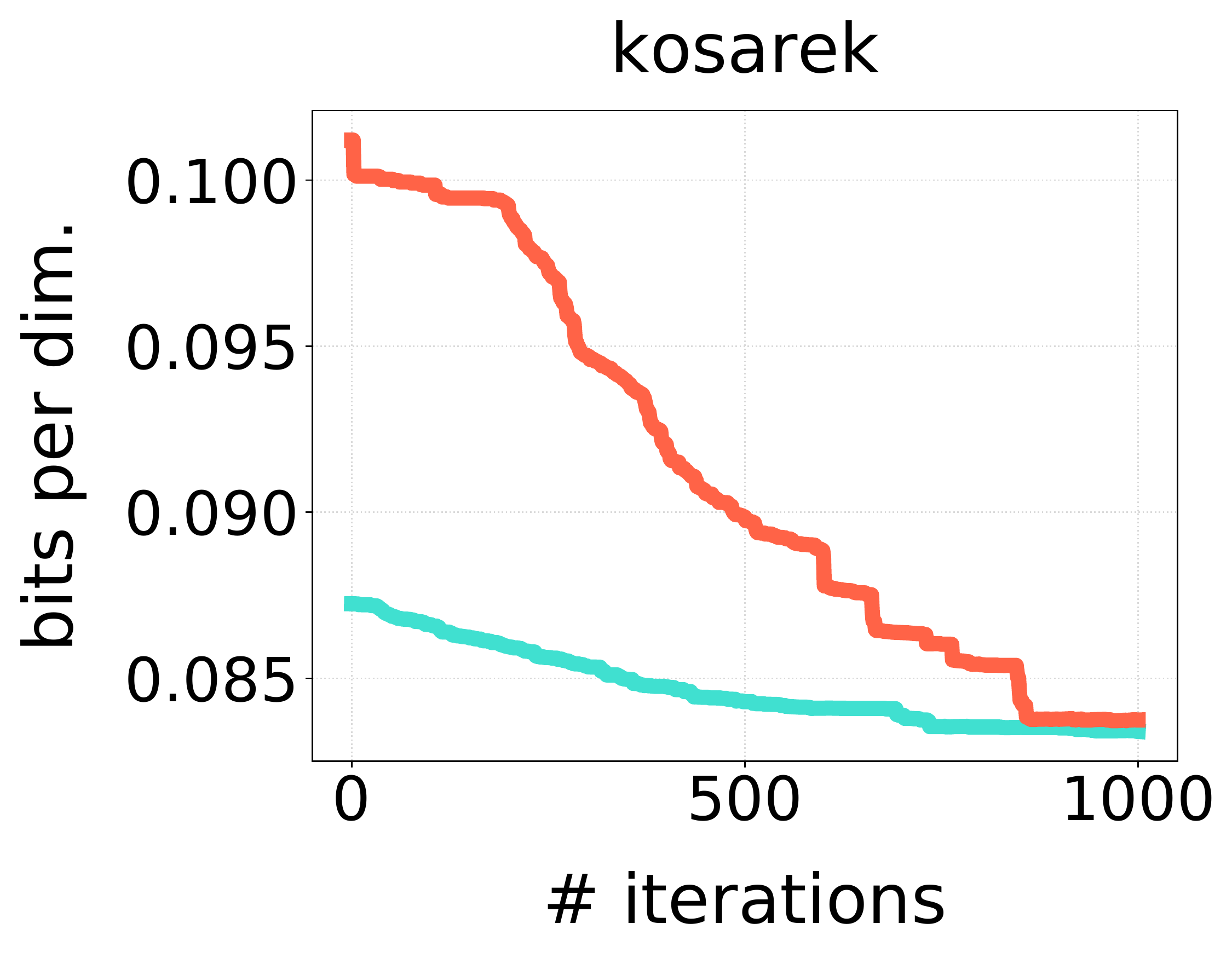}\\
    \includegraphics[width=0.23\textwidth]{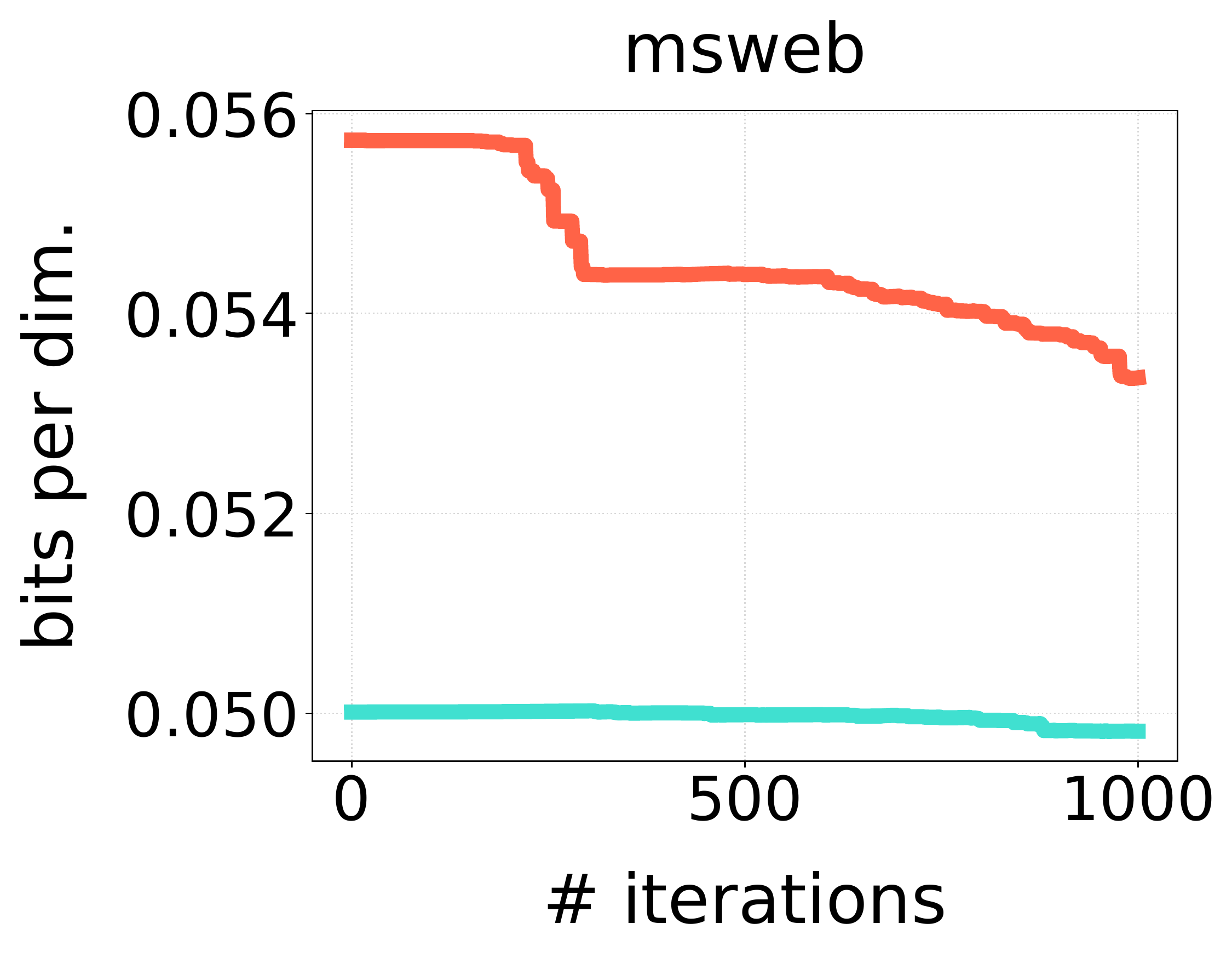}&
    \includegraphics[width=0.23\textwidth]{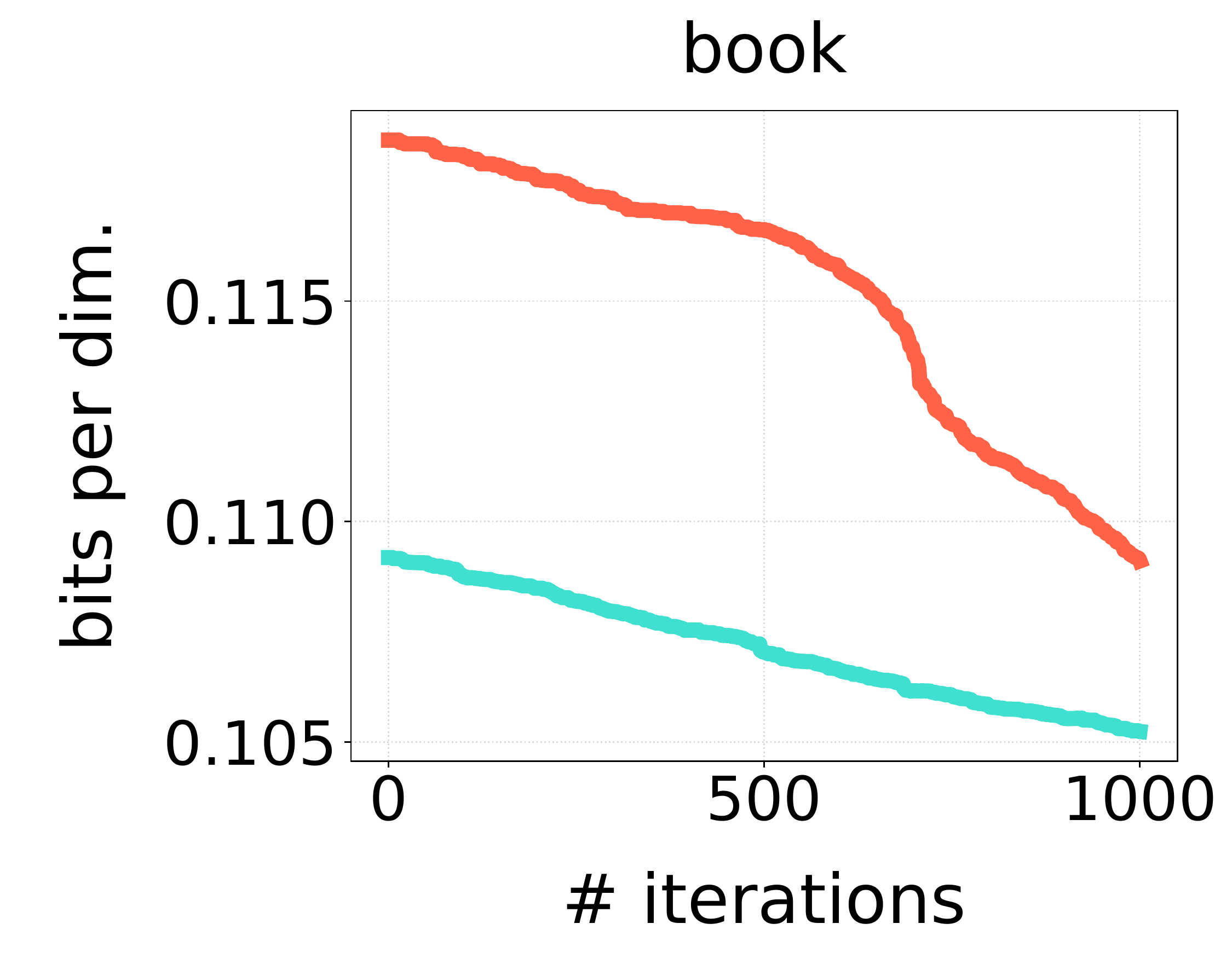}&
    \includegraphics[width=0.23\textwidth]{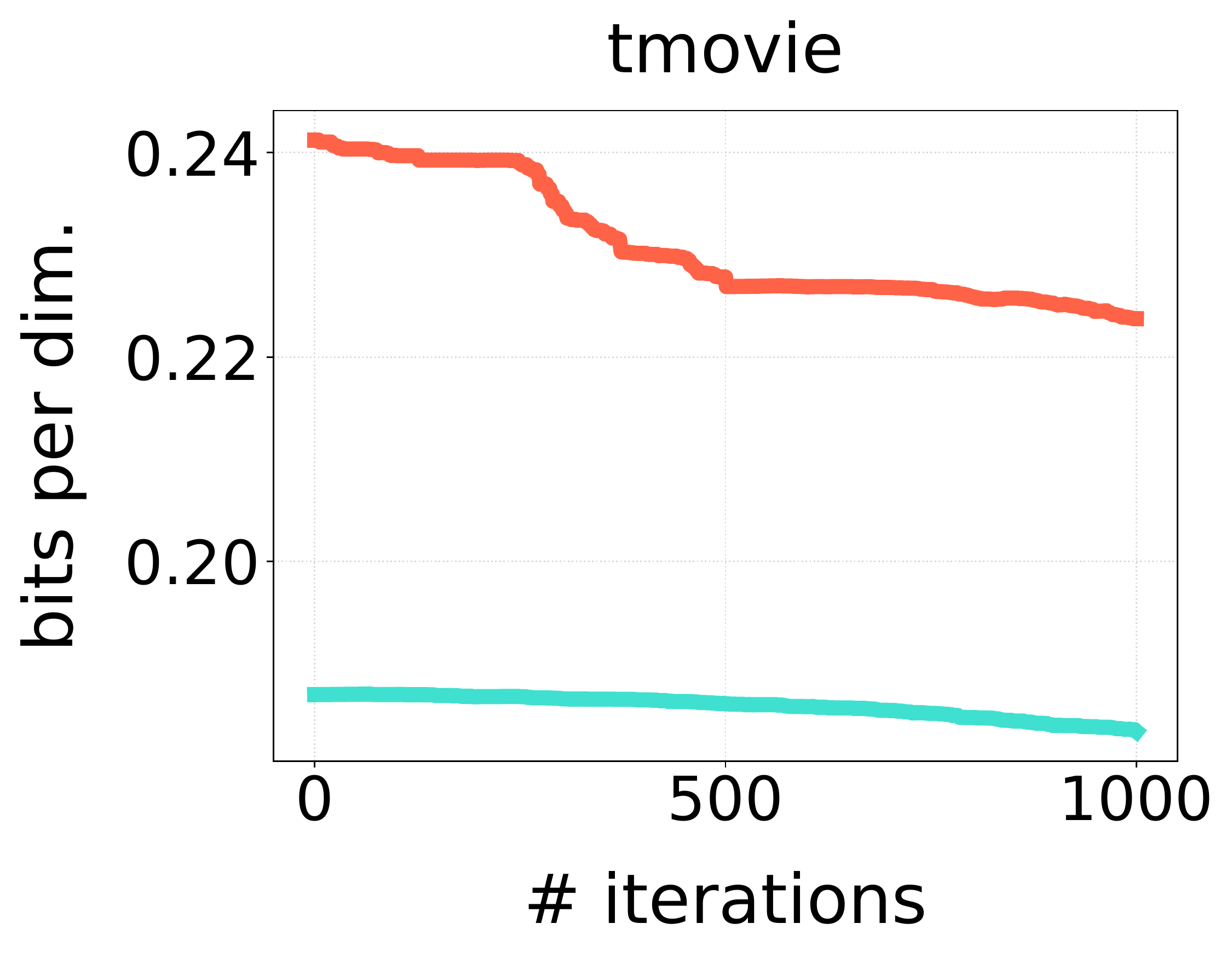}&
    \includegraphics[width=0.23\textwidth]{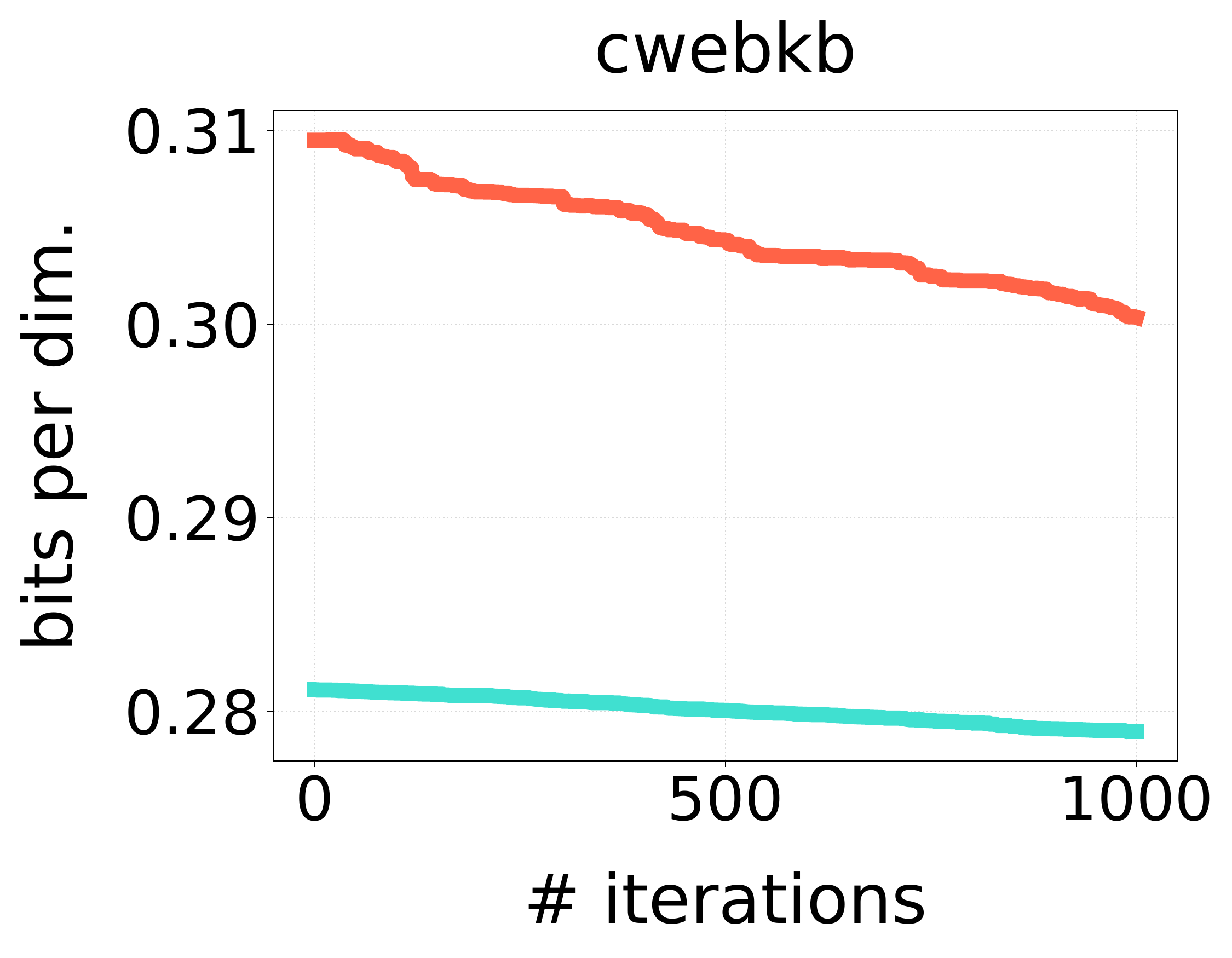}\\
    \includegraphics[width=0.23\textwidth]{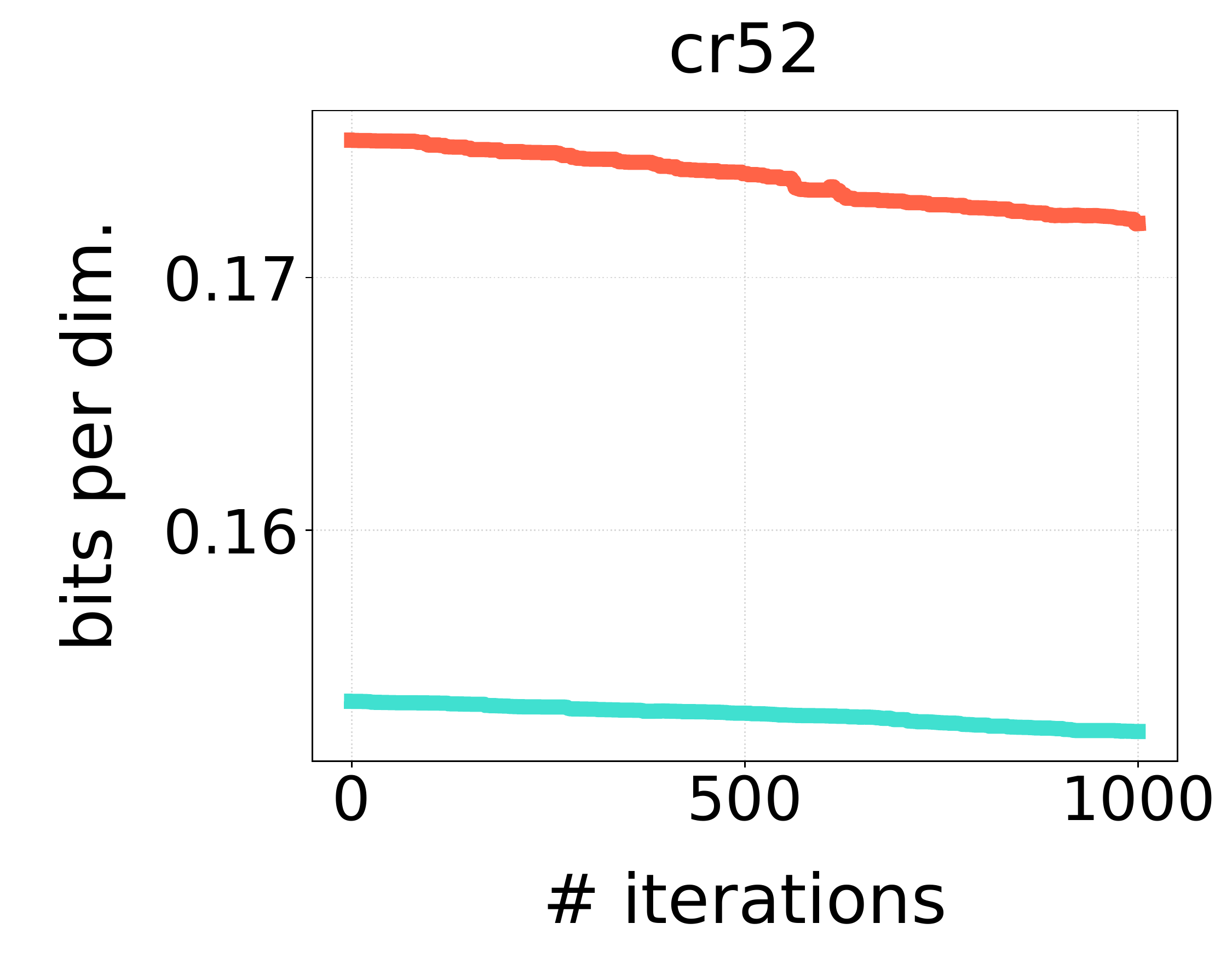}&
    \includegraphics[width=0.23\textwidth]{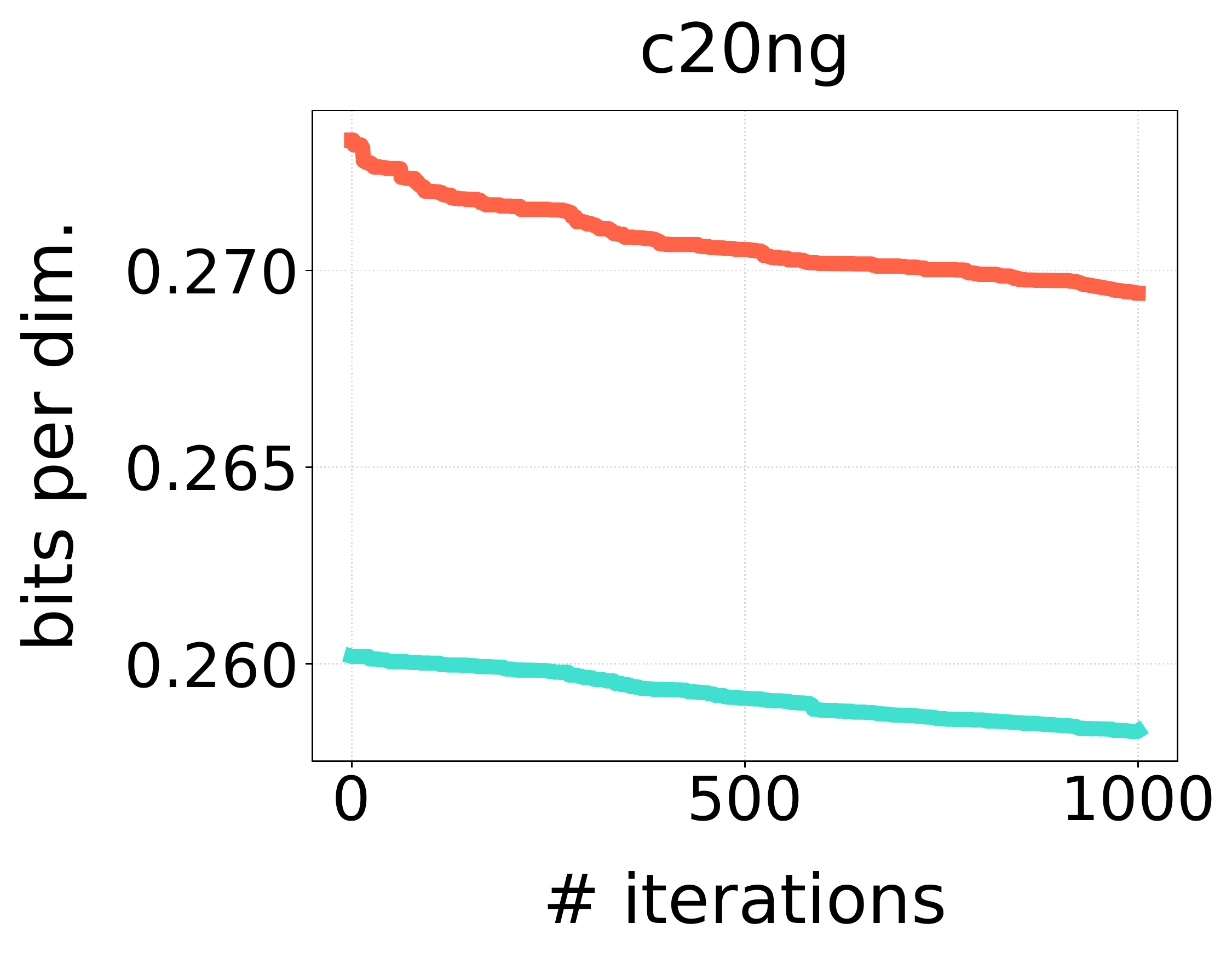}&
    \includegraphics[width=0.23\textwidth]{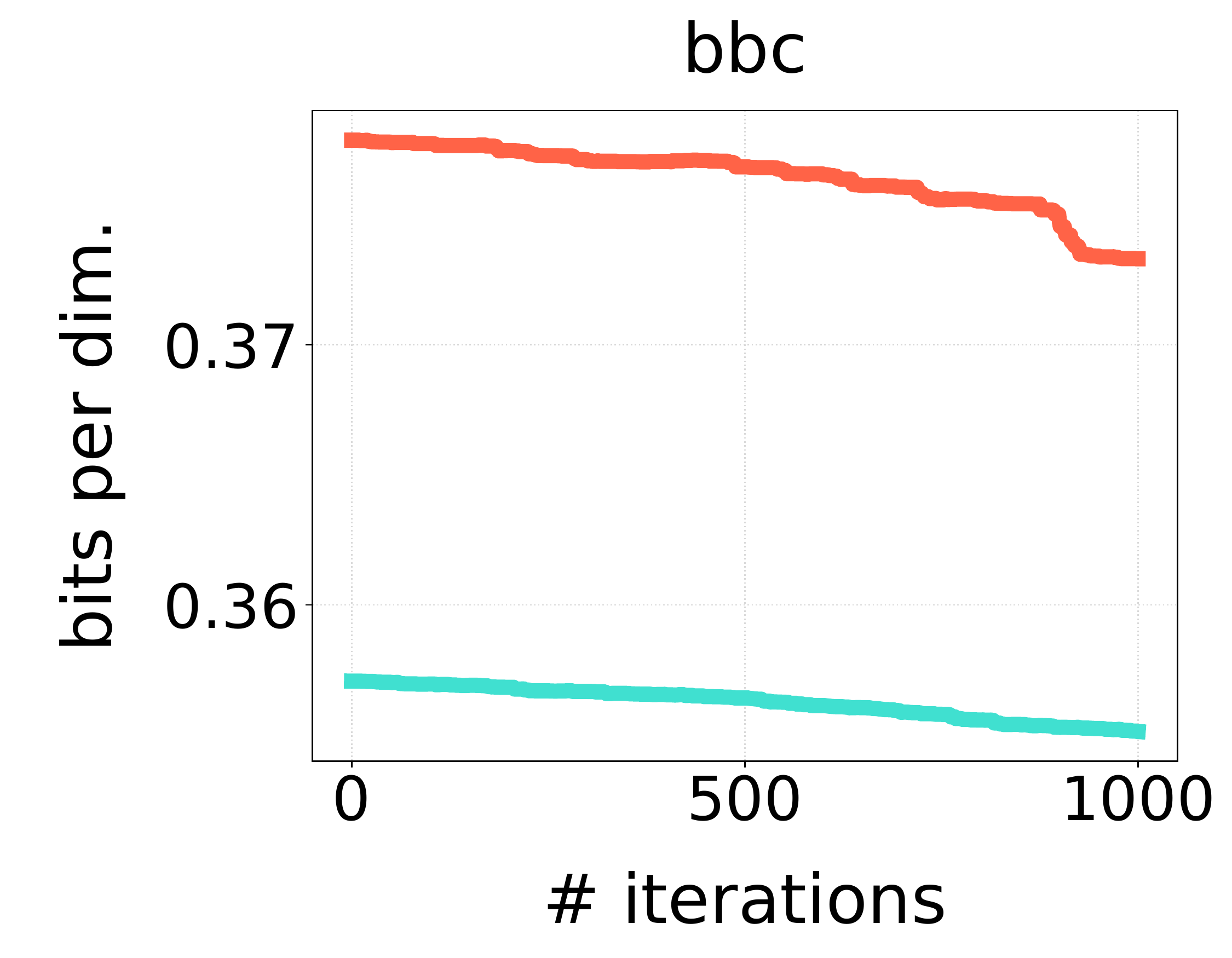}&
    \includegraphics[width=0.23\textwidth]{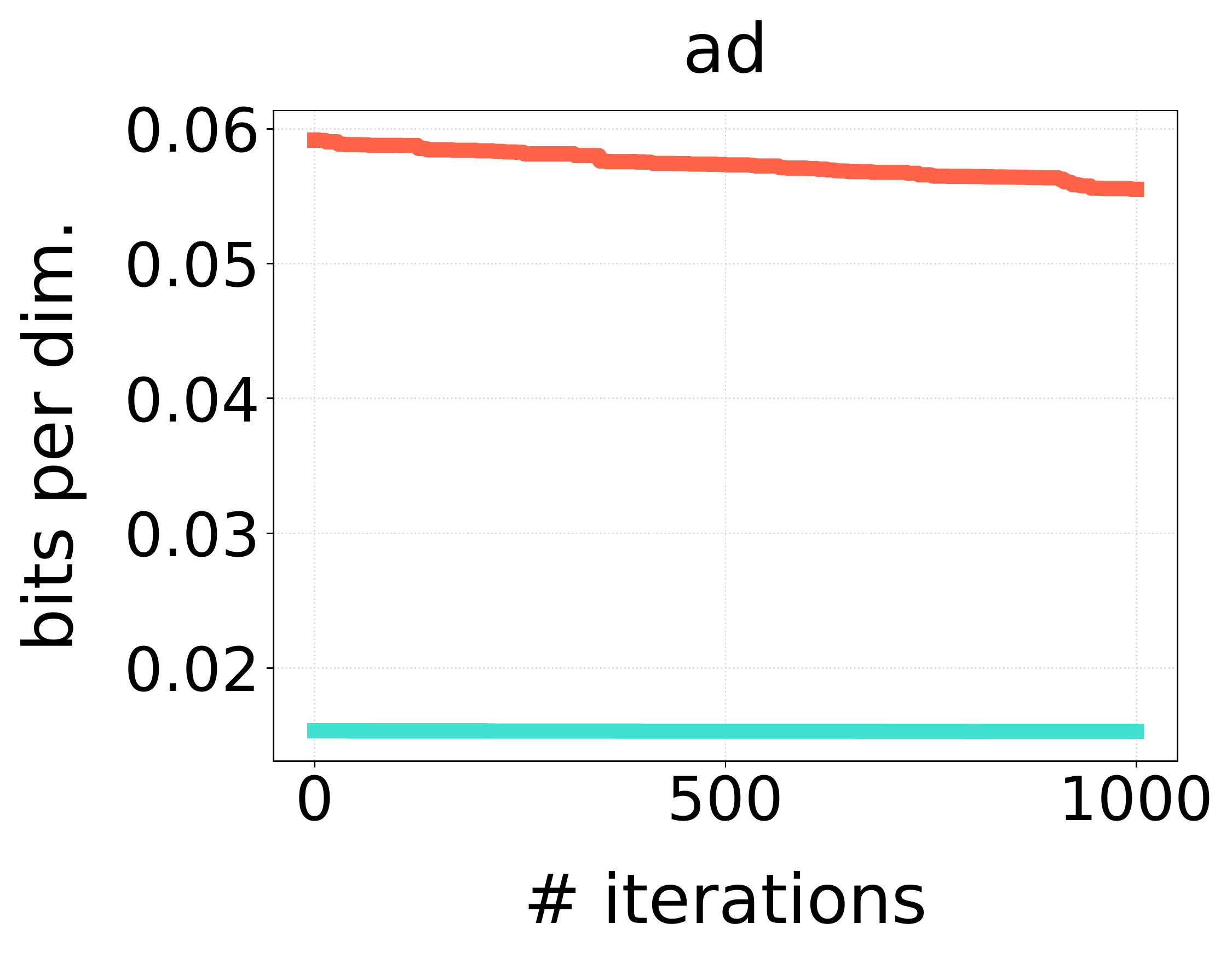}\\
    \caption{\textit{\textbf{Effect of different initializations in \ourlearner\ for each dataset.}} We report the mean test bits per dimensions (bpd) on each dataset (y-axis) for each iteration (x-axis) as scored by the different initializations: CLT (blue) and independent (red).}
    \label{tab:fig-init-all}
    
\end{longtable}

\newpage
\subsection{Heuristics}
\label{sec:app-heuristics}
Here we report the effect of different heuristics on each dataset in Table~\ref{tab:fig-heuristics-all}.
\begin{longtable}[!ht]{cccc}
    \centering
    \includegraphics[width=0.23\textwidth]{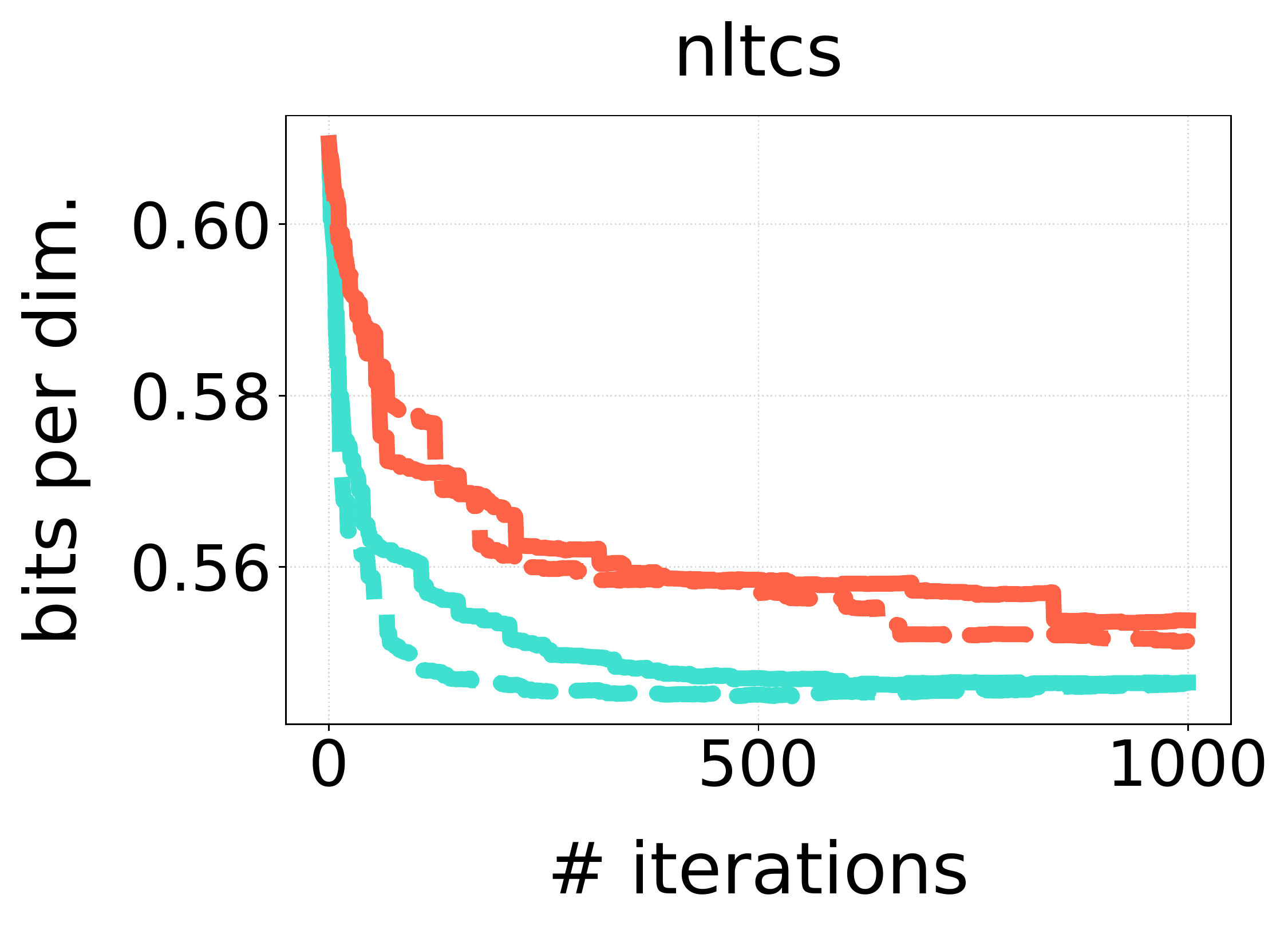}&
    \includegraphics[width=0.23\textwidth]{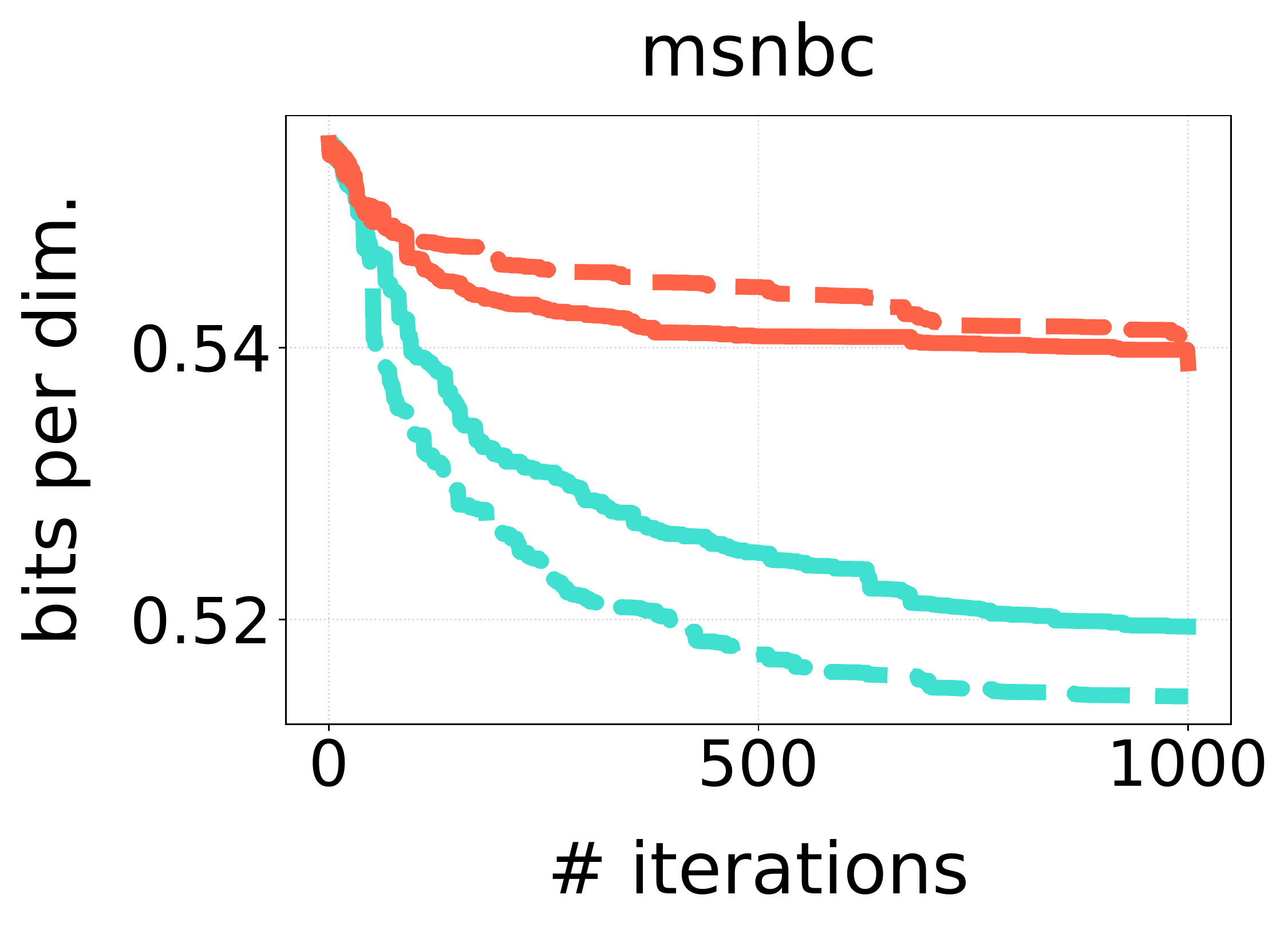}&
    \includegraphics[width=0.23\textwidth]{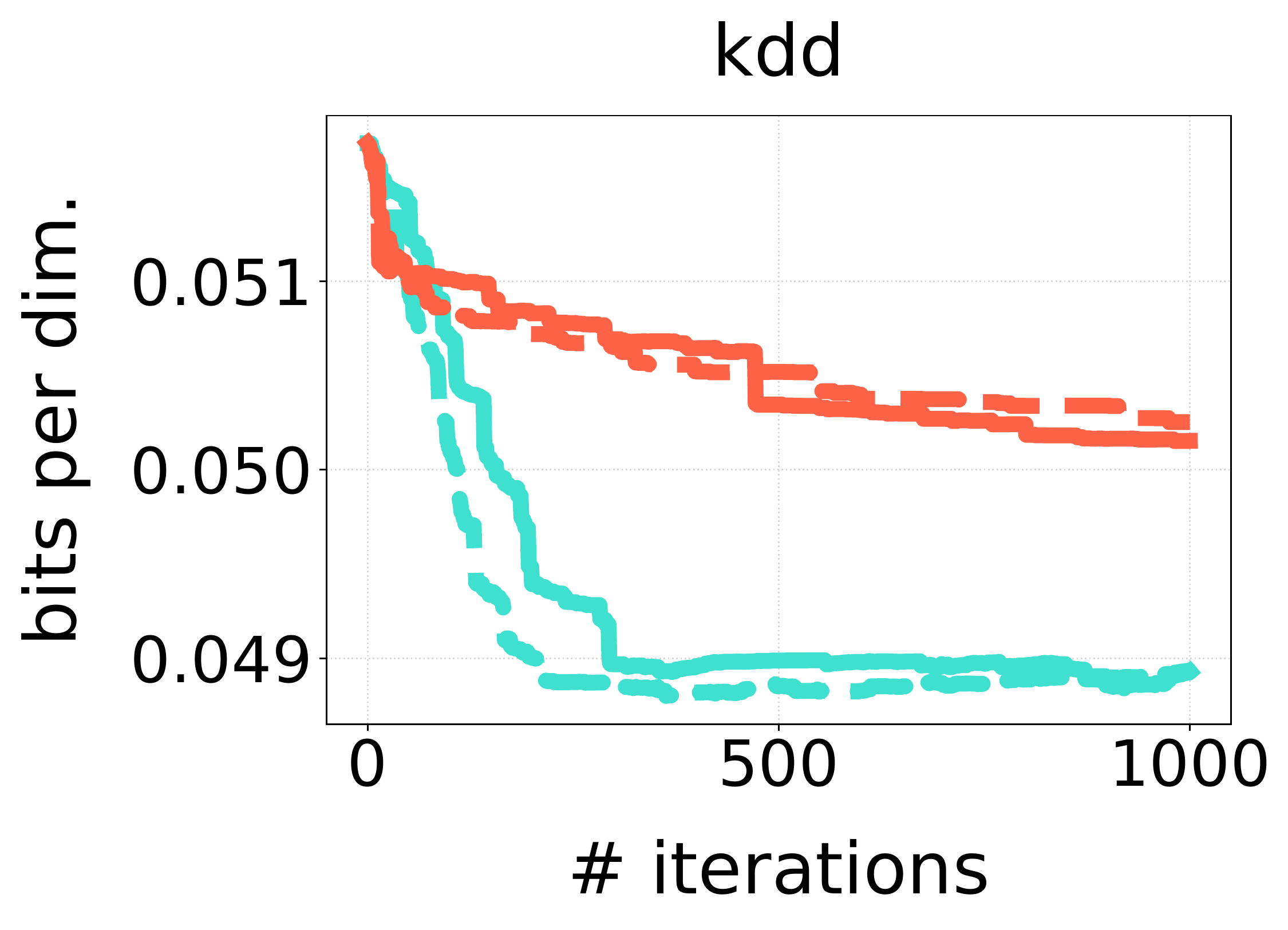}&
    \includegraphics[width=0.23\textwidth]{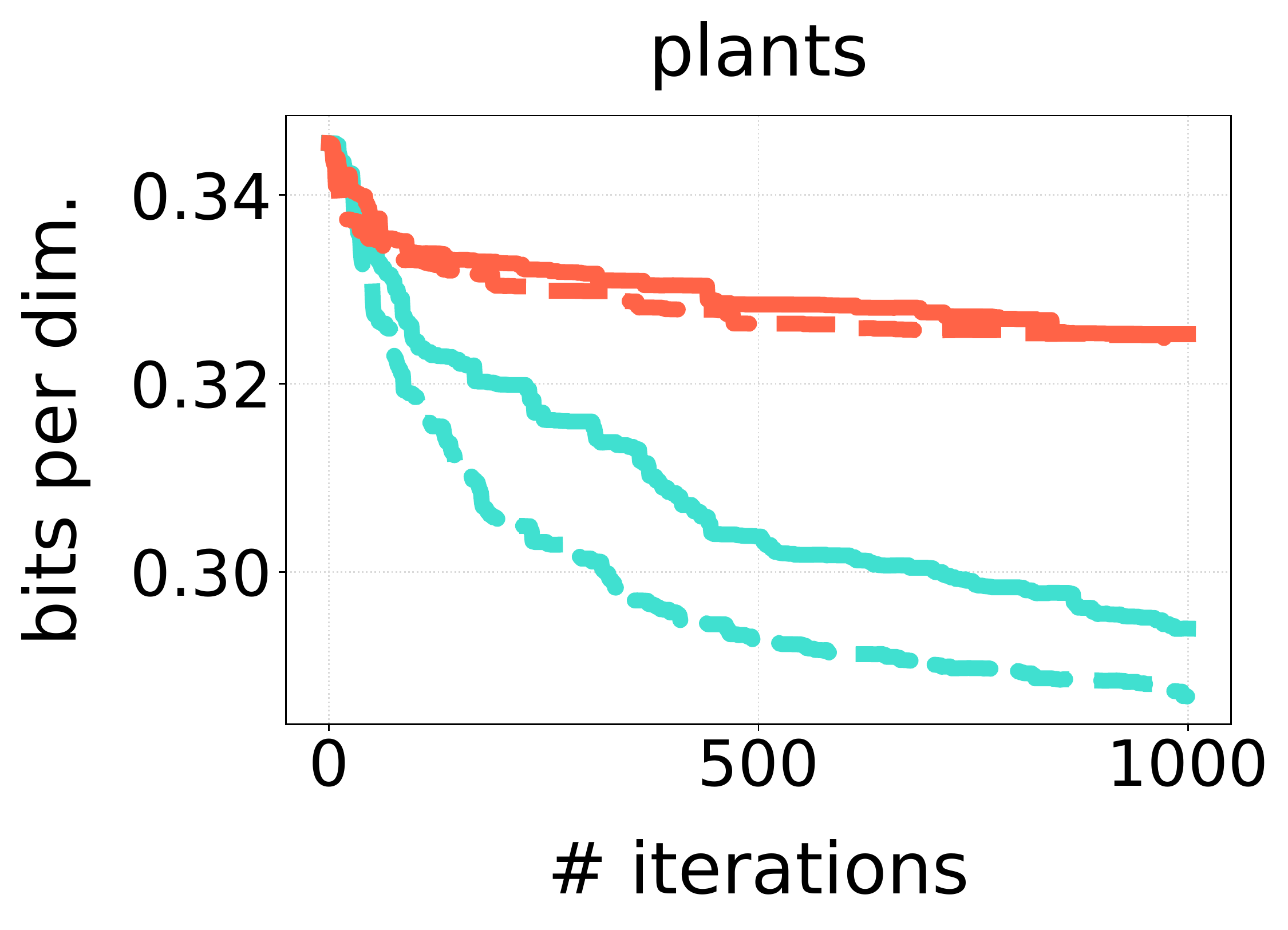}\\
    \includegraphics[width=0.23\textwidth]{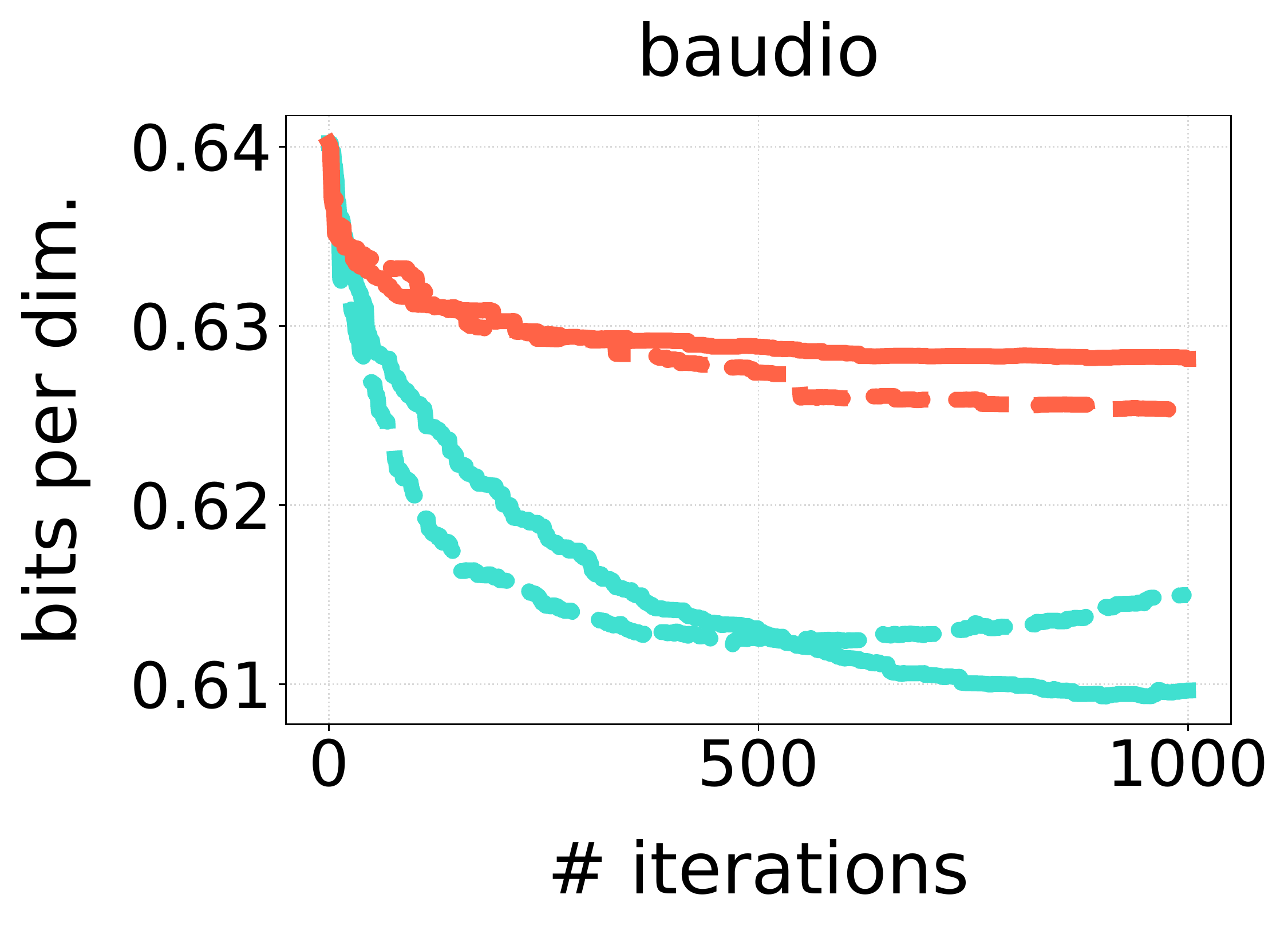}&
    \includegraphics[width=0.23\textwidth]{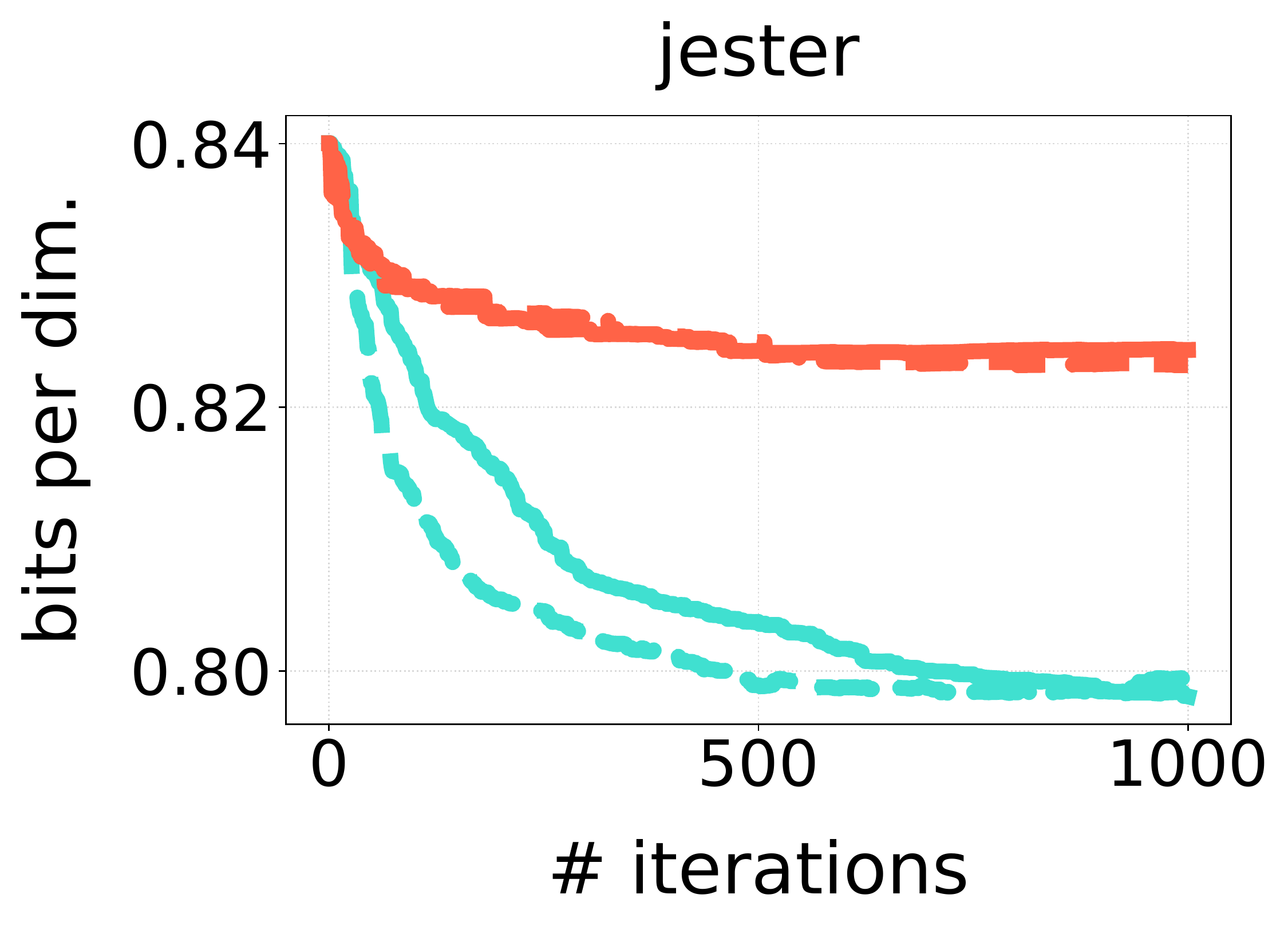}&
    \includegraphics[width=0.23\textwidth]{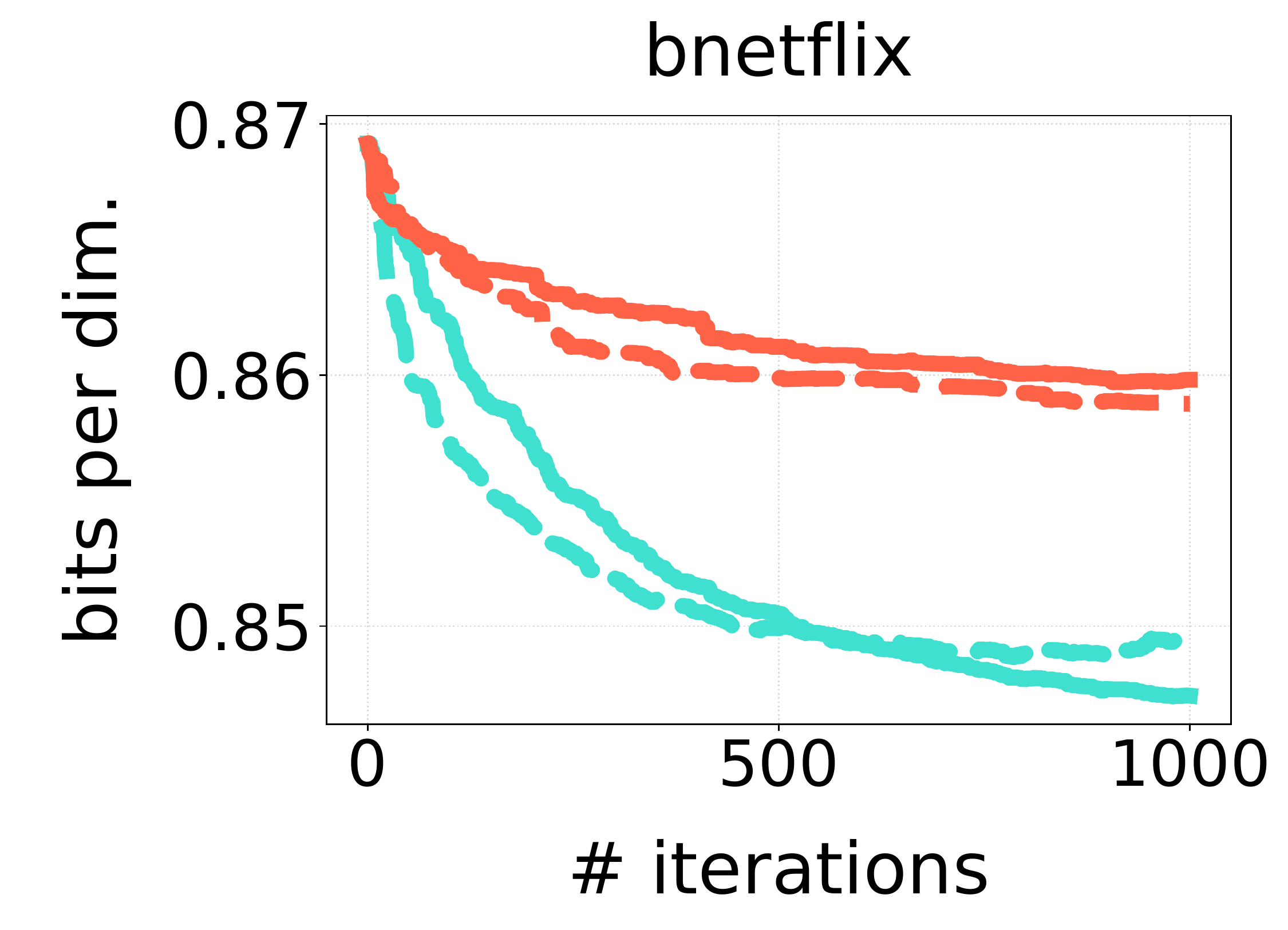}&
    \includegraphics[width=0.23\textwidth]{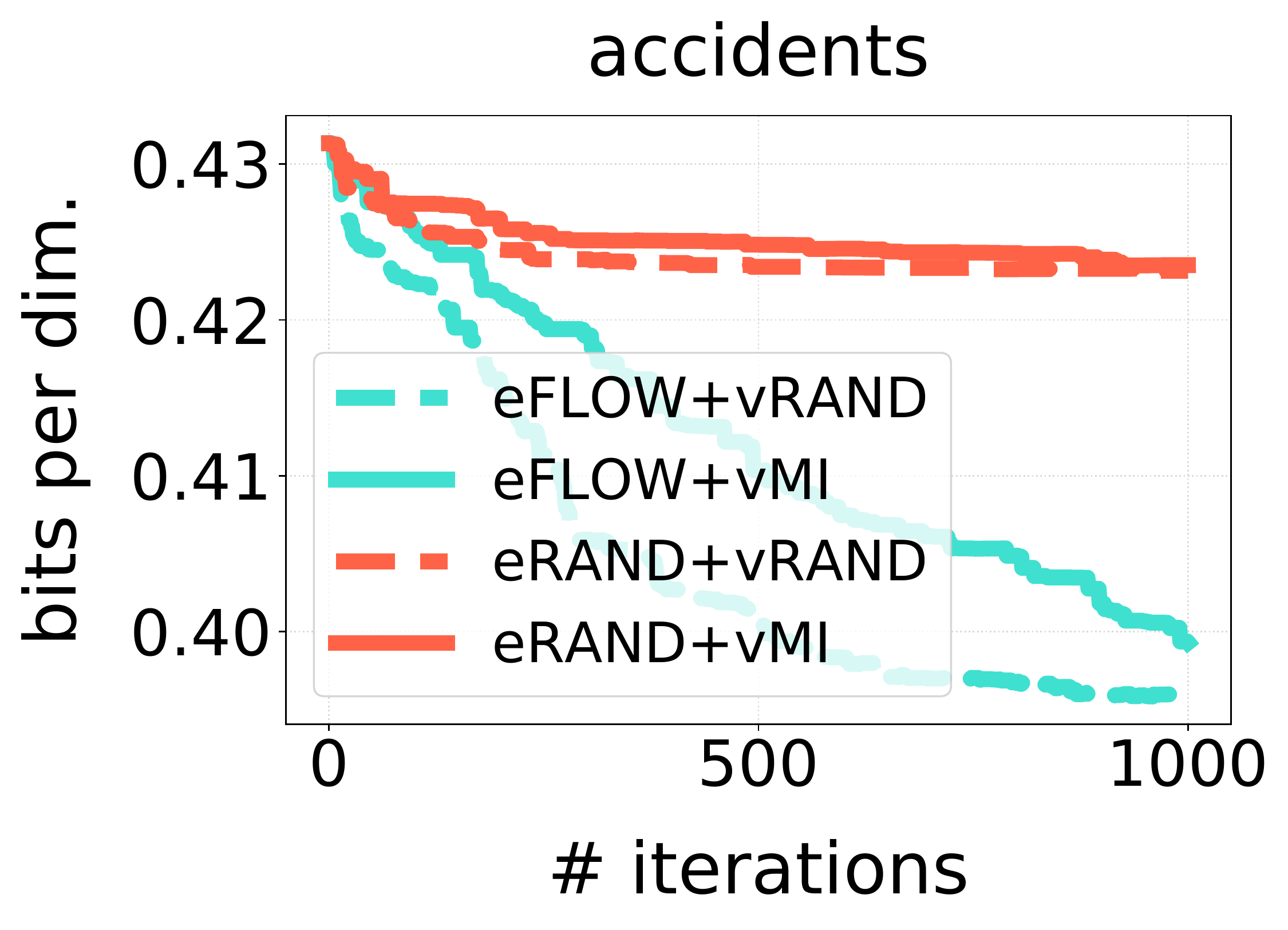}\\
    \includegraphics[width=0.23\textwidth]{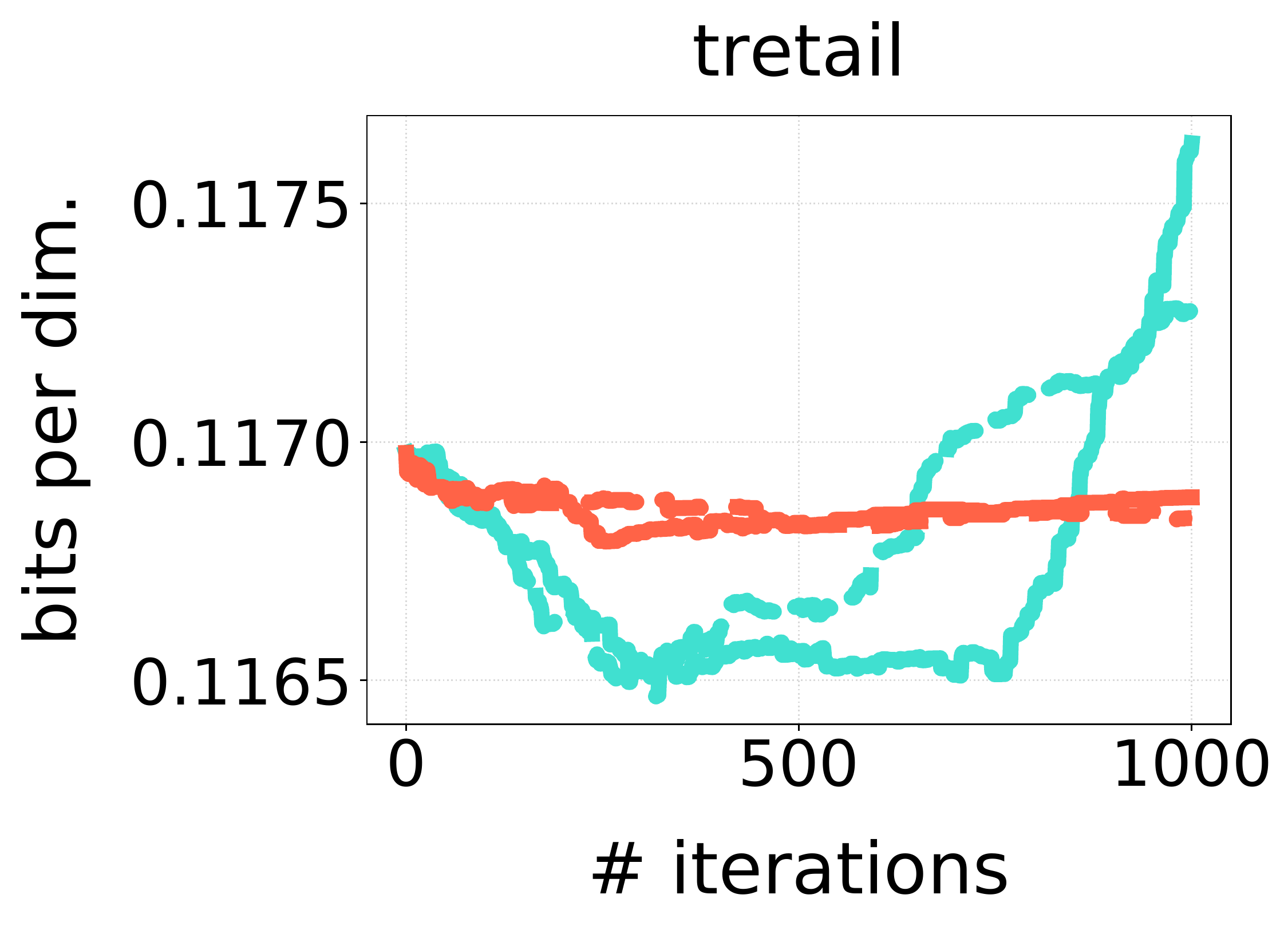}&
    \includegraphics[width=0.23\textwidth]{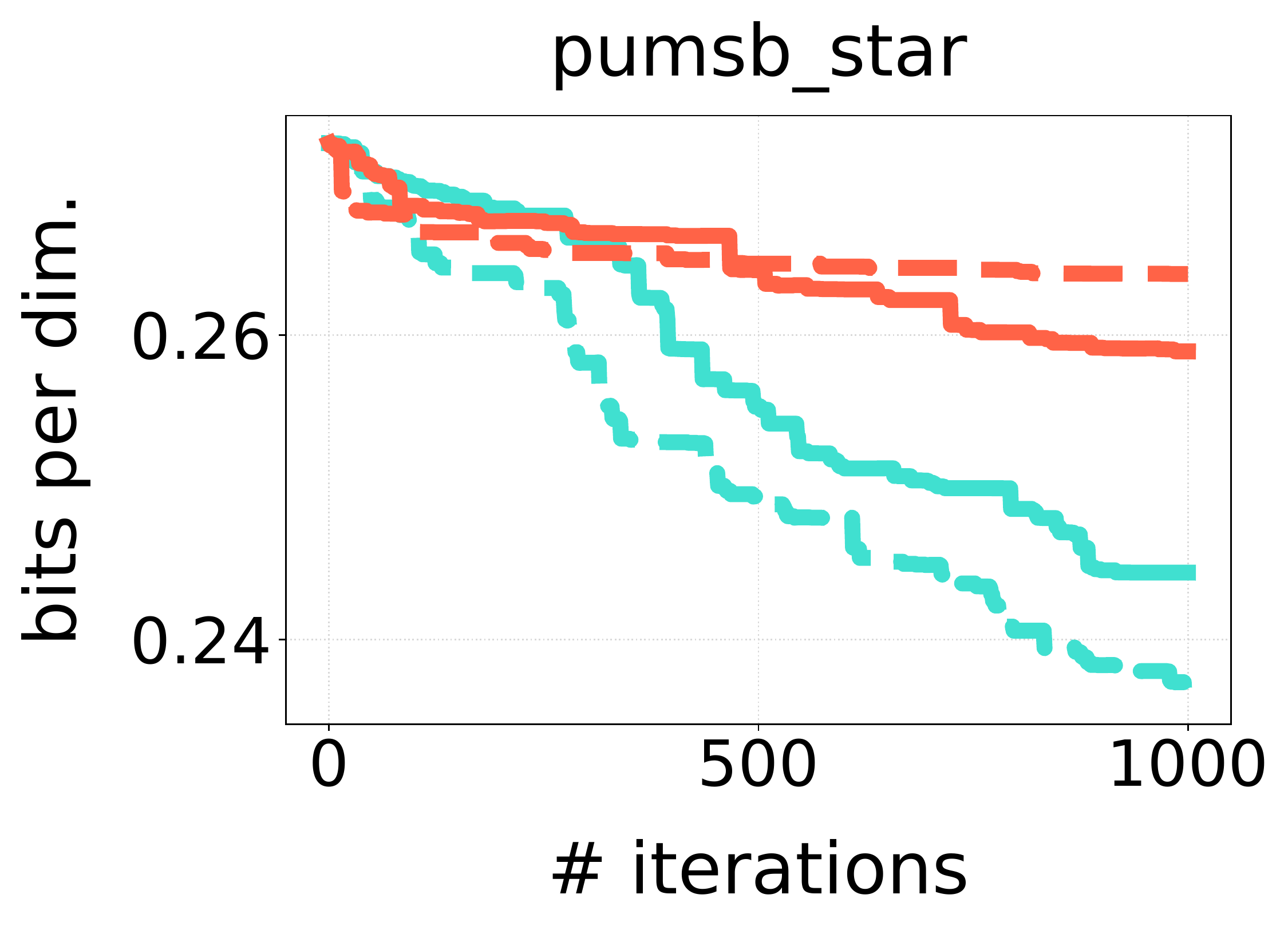}&
    \includegraphics[width=0.23\textwidth]{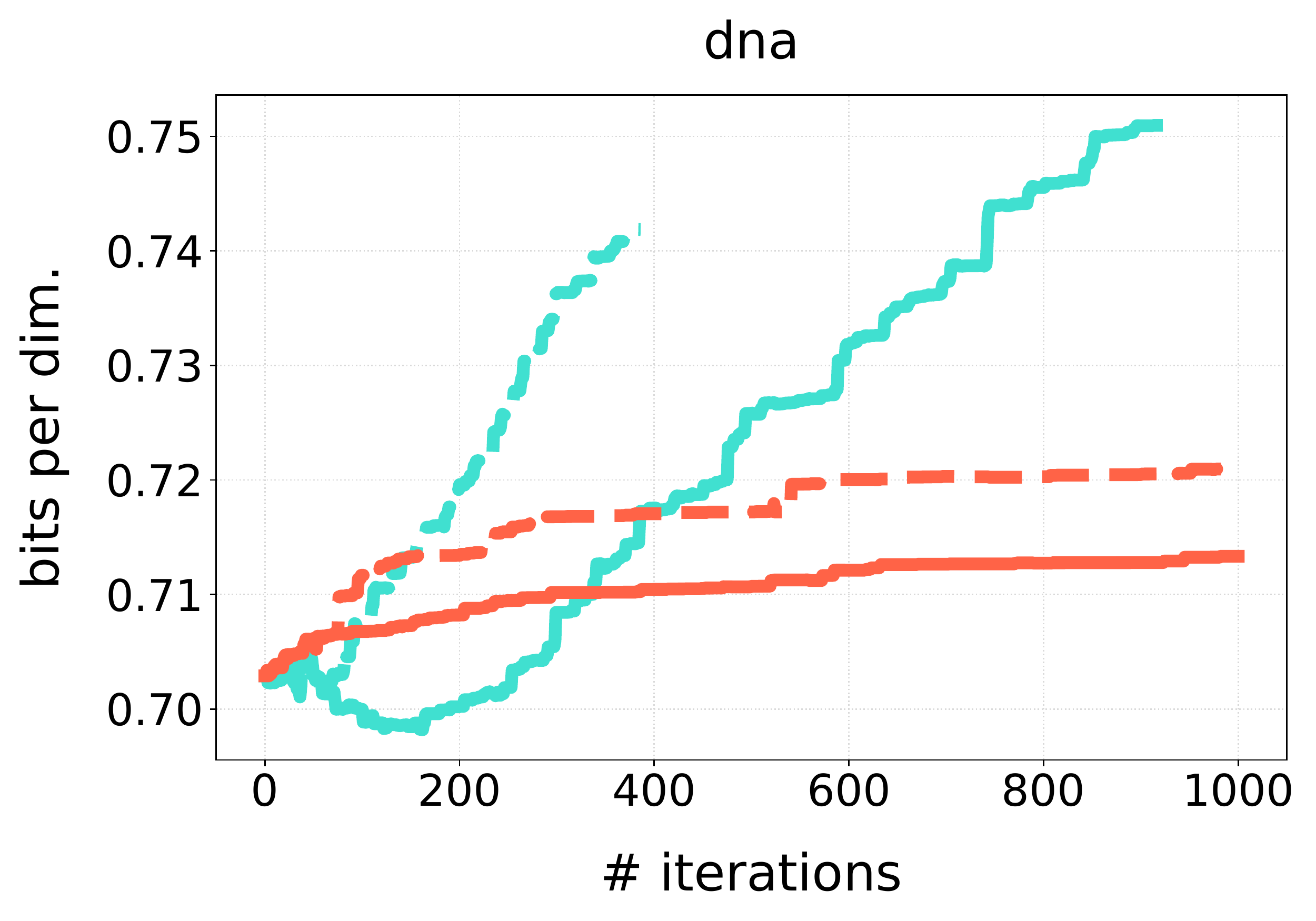}&
    \includegraphics[width=0.23\textwidth]{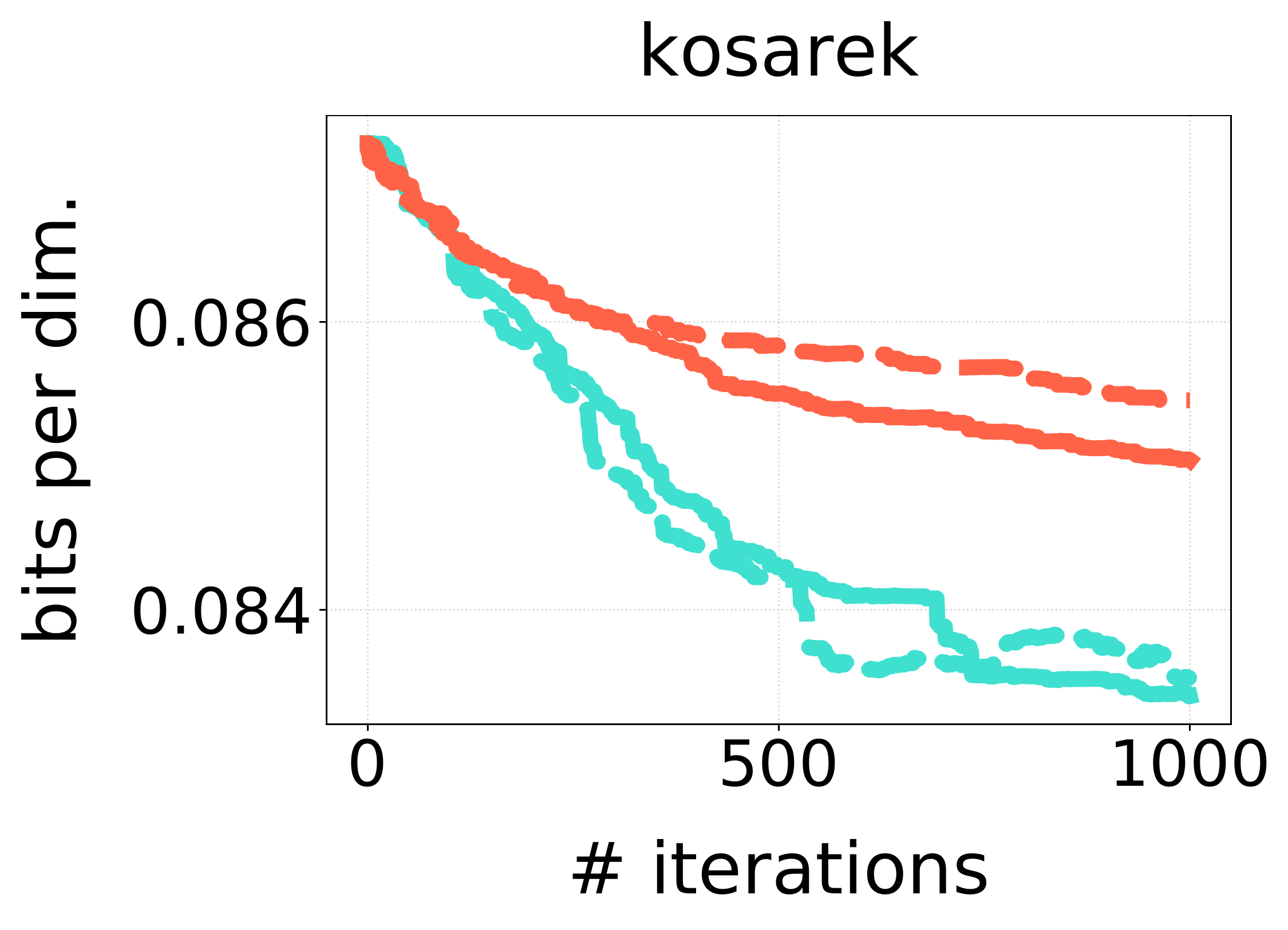}\\
    \includegraphics[width=0.23\textwidth]{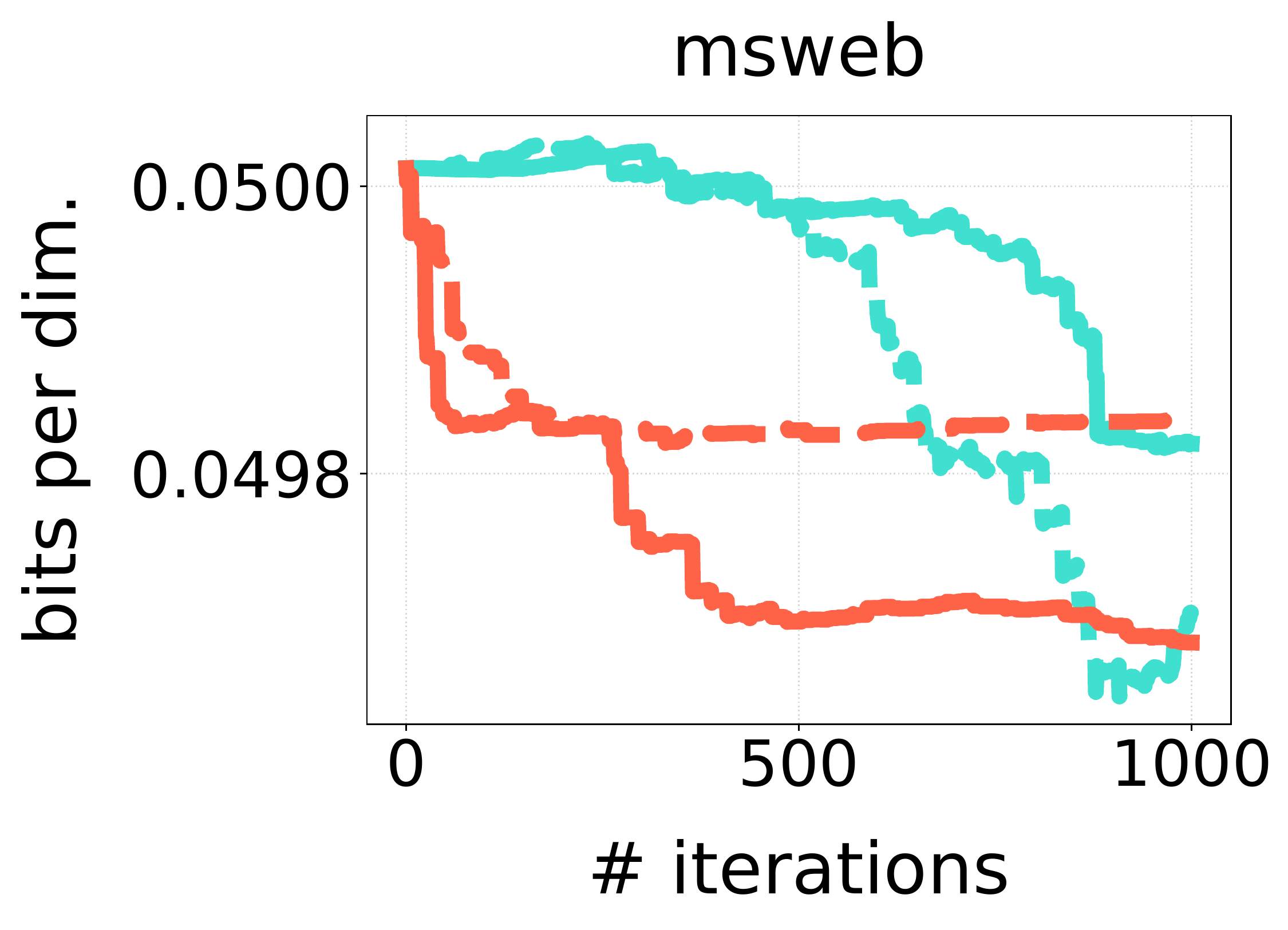}&
    \includegraphics[width=0.23\textwidth]{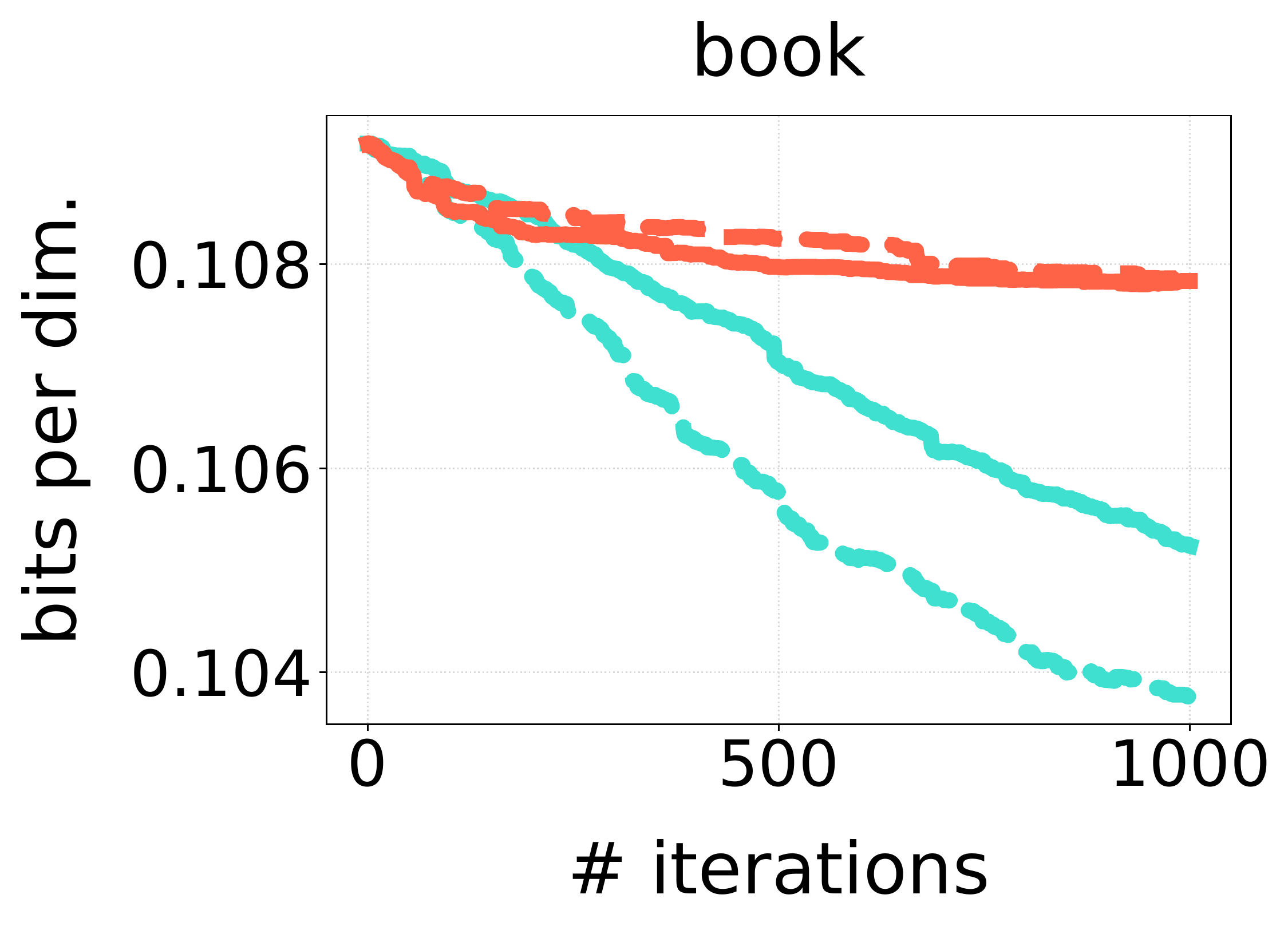}&
    \includegraphics[width=0.23\textwidth]{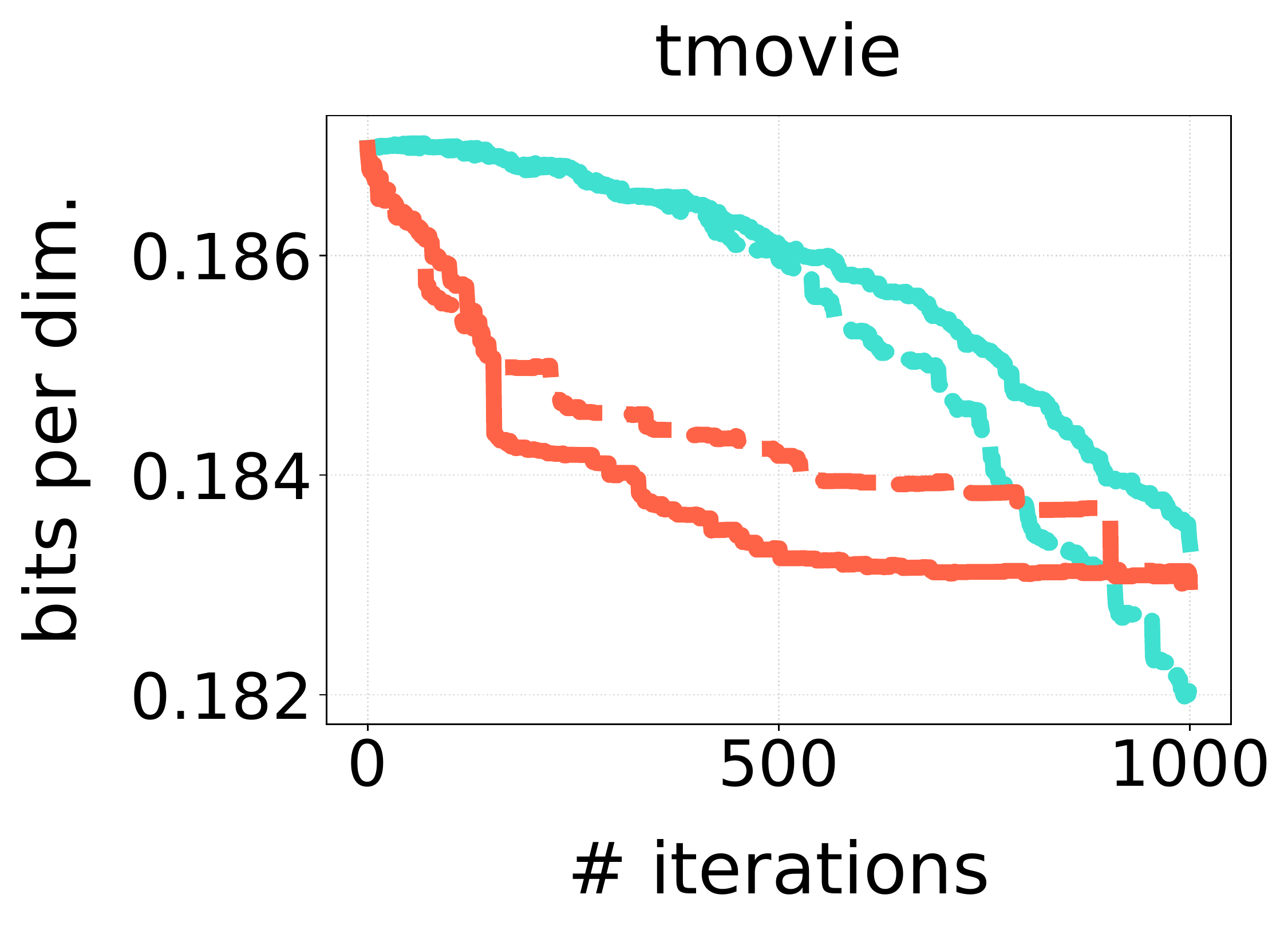}&
    \includegraphics[width=0.23\textwidth]{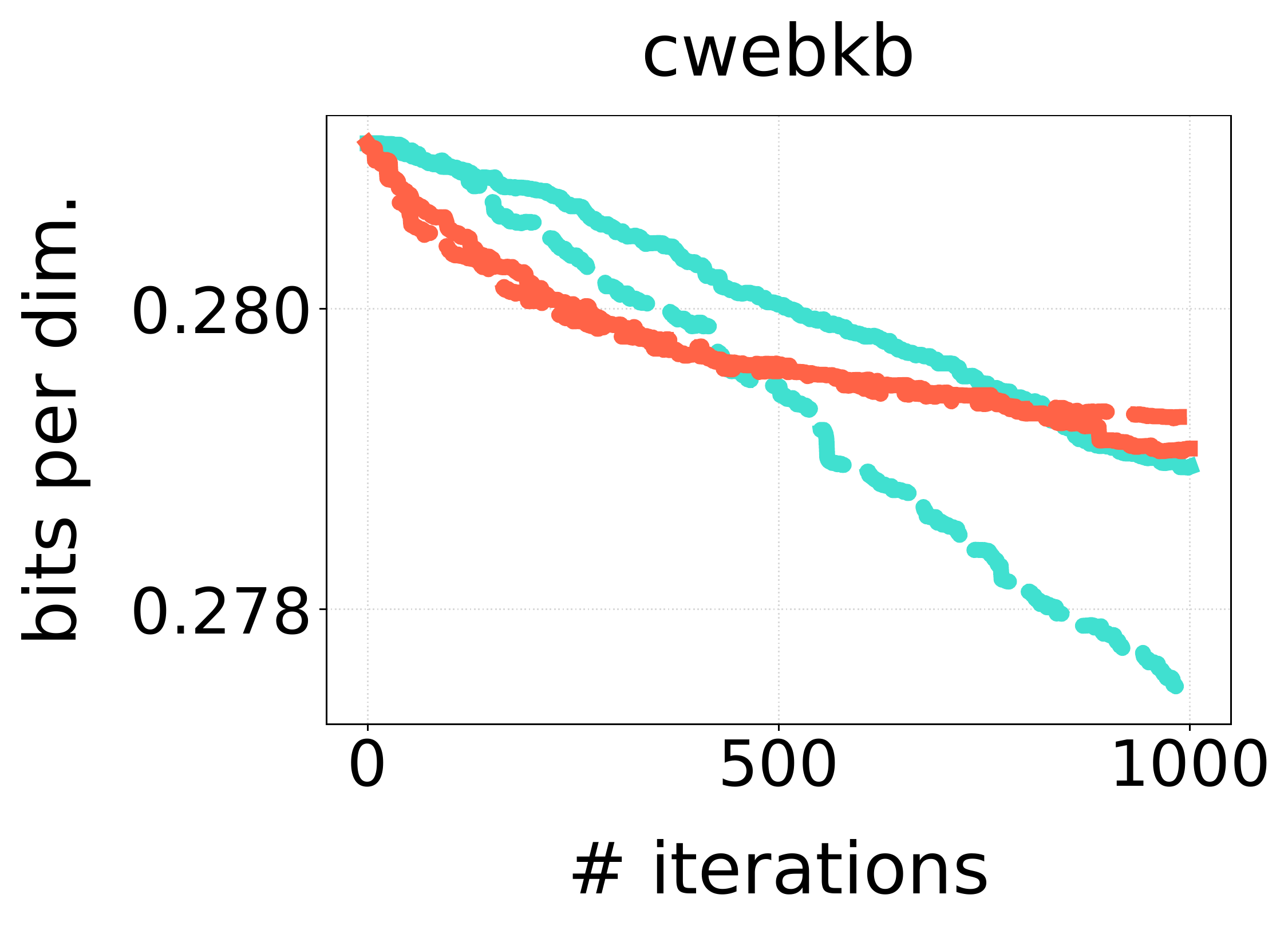}\\
    \includegraphics[width=0.23\textwidth]{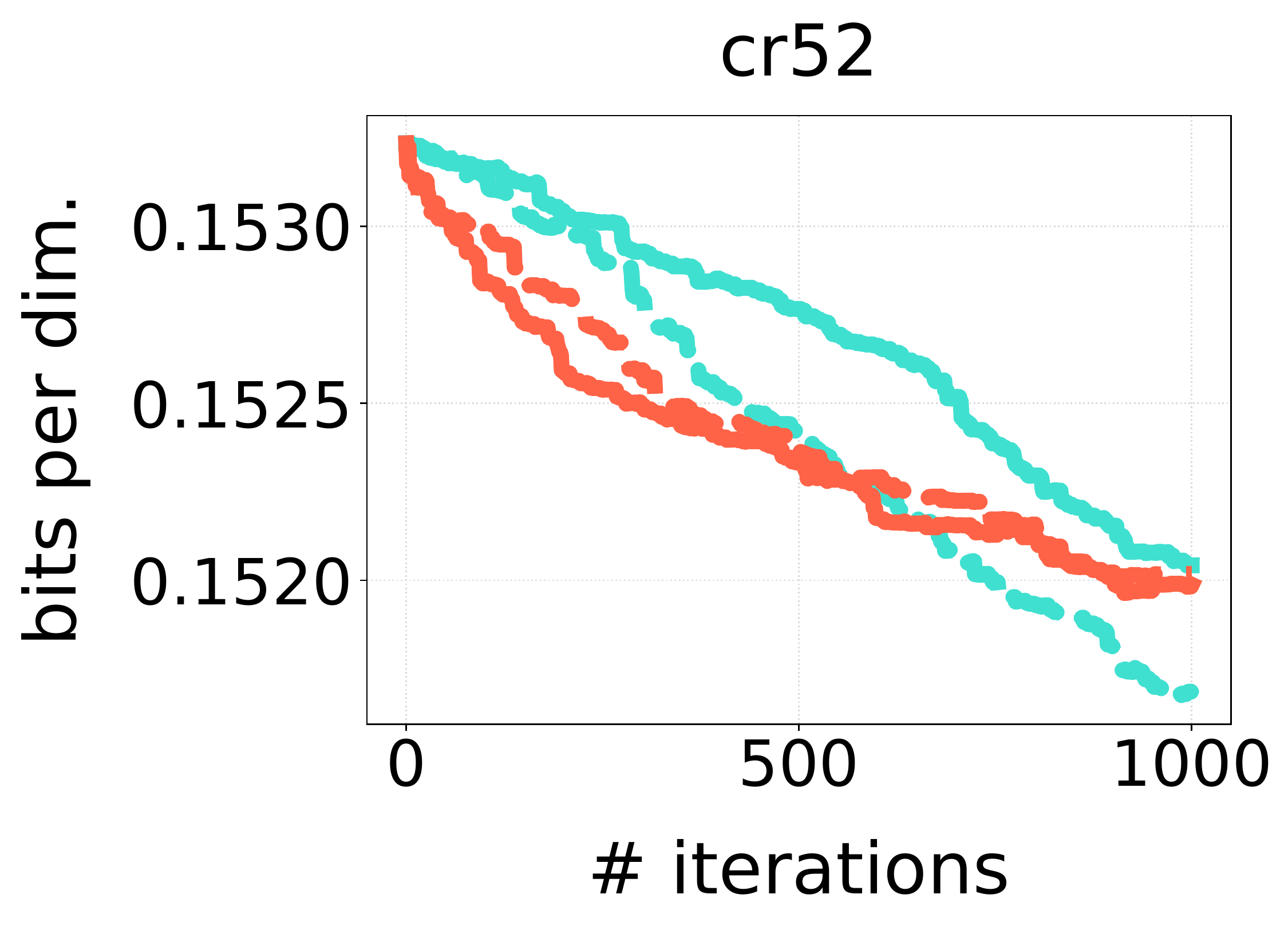}&
    \includegraphics[width=0.23\textwidth]{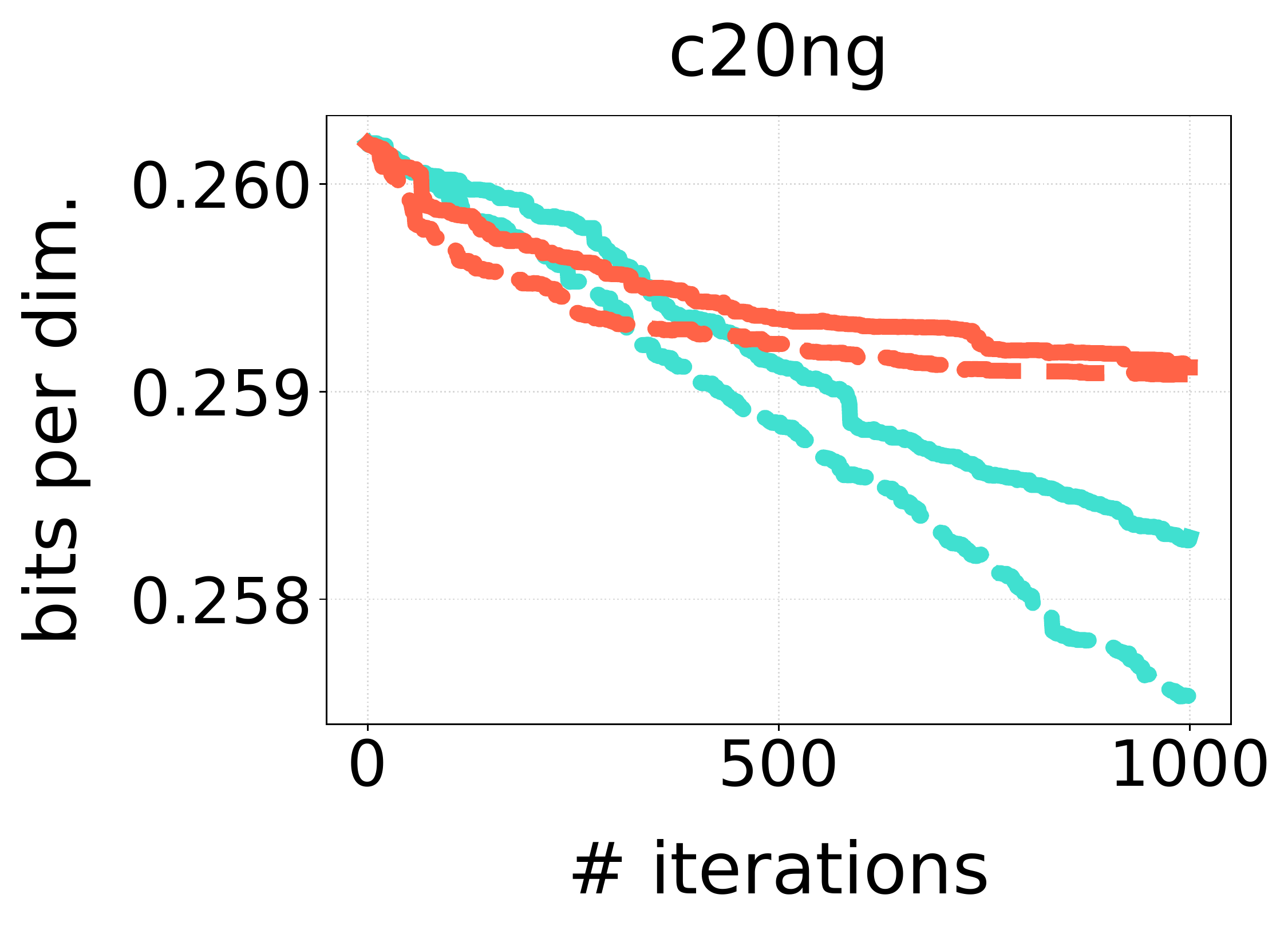}&
    \includegraphics[width=0.23\textwidth]{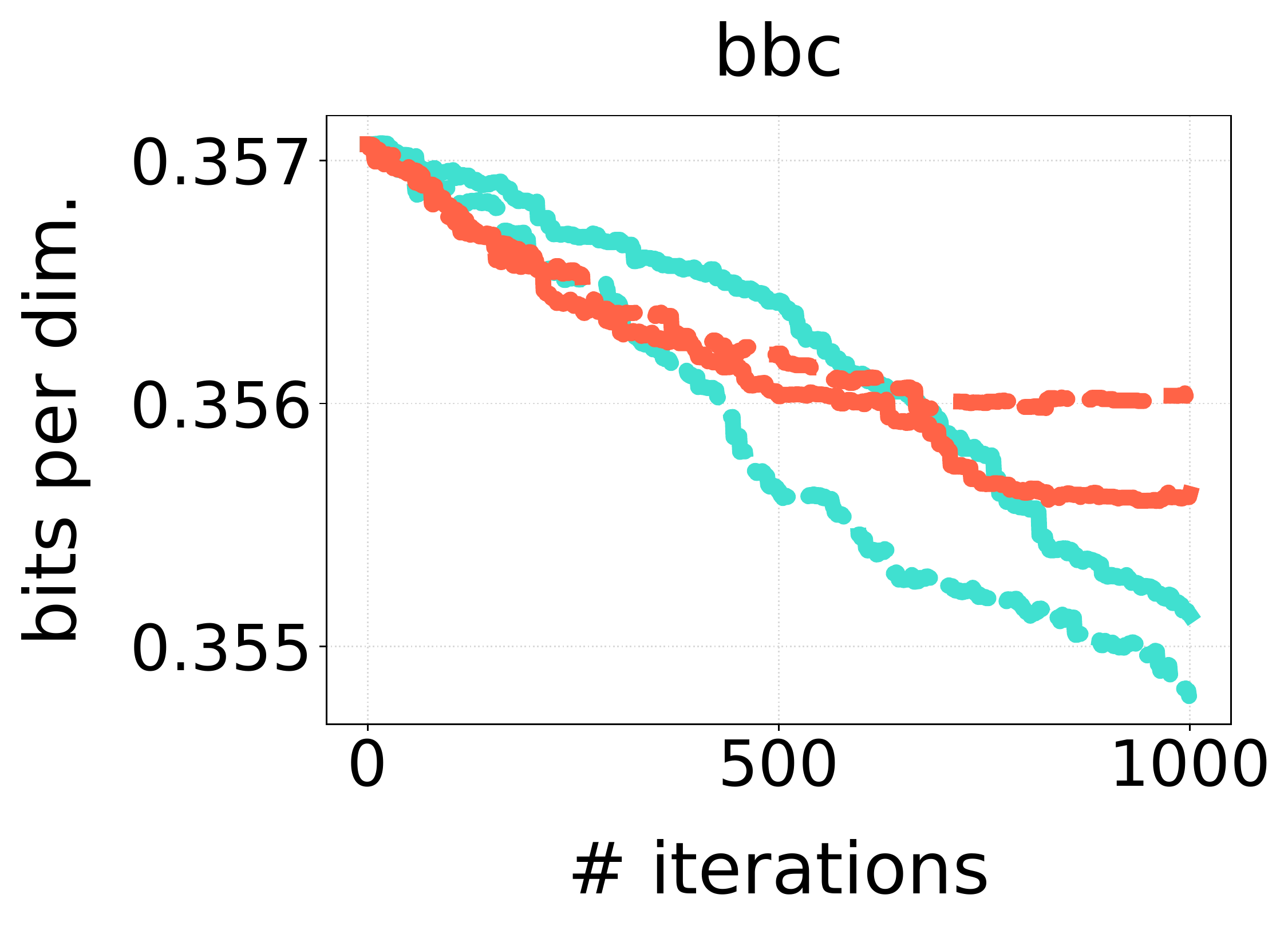}&
    \includegraphics[width=0.23\textwidth]{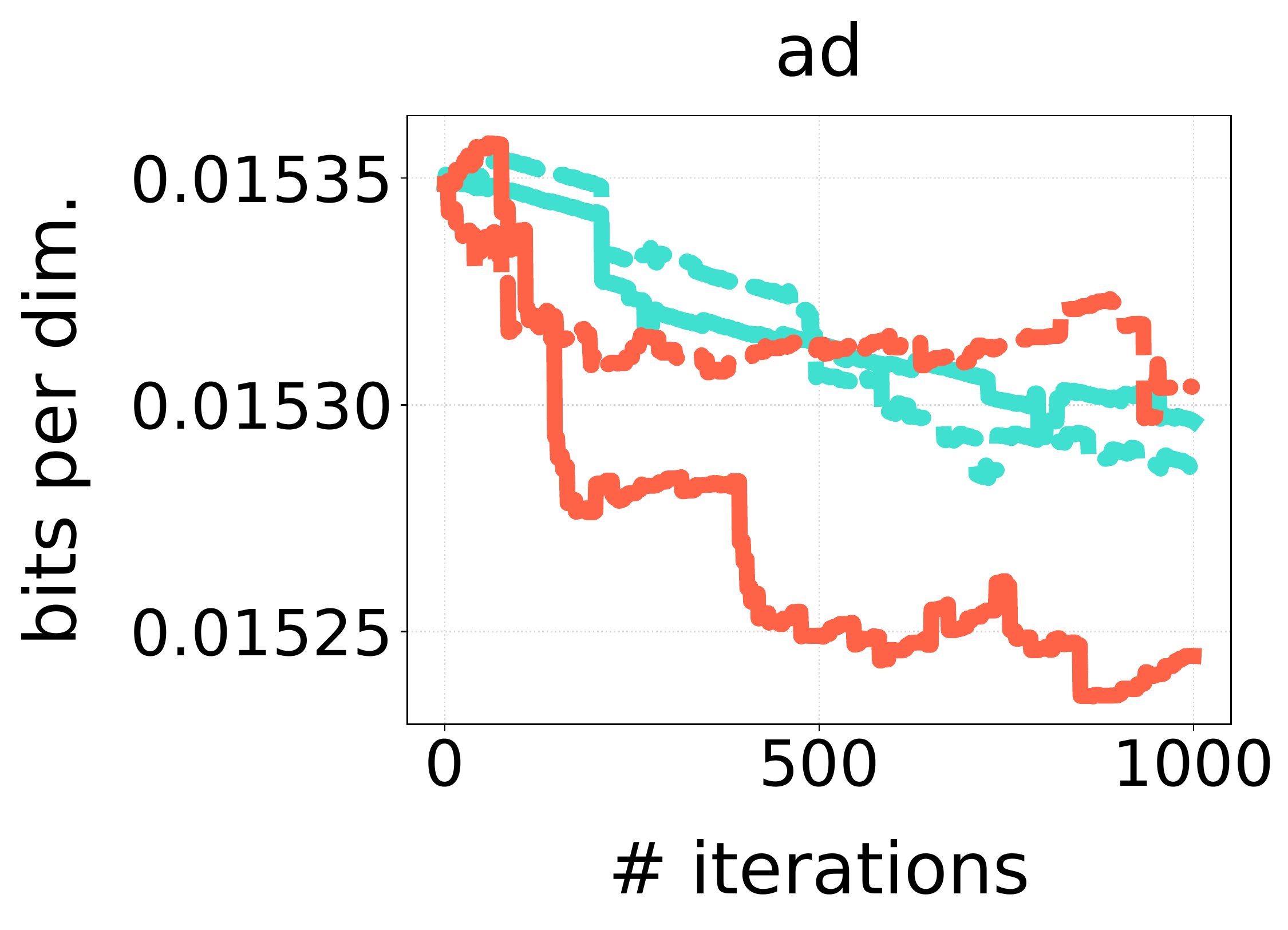}\\
    \caption{\textit{\textbf{Effect of different heuristics in \ourlearner\ for each dataset.}} We report the mean test bits per dimensions (bpd) on each dataset (y-axis) for each iteration (x-axis) as scored by the 4 combinations of different heuristics.}
    \label{tab:fig-heuristics-all}
    
\end{longtable}

\newpage
\section{Mixtures with \ourlearner}
We report the test set log-likelihood result in Table~\ref{tab:ensembles}  of mixtures with \ourlearner\ and mixtures with \learnpsdd.
\label{sec:app-ensemble}
\begin{table}[!ht]
    \centering
    \setlength{\tabcolsep}{1.8pt}
    \small
    \begin{sc}
    \begin{tabular}{r r r || r r r  r}
    \toprule
         dataset    & LearnPSDD     & \ourlearner   & LearnPSDD    & \ourlearner\\
                    & EM            & EM            & BEM      & BEM \\
         \midrule
         nltcs      & -6.03$\downarrow$         & -6.07     &\textbf{-5.99}$\downarrow$         &-6.06\\
         msnbc      & \textbf{-6.04}$\uparrow$         & \textbf{-6.04}     &\textbf{-6.04}$\uparrow$      &\textbf{-6.04}\\
         kdd        & -2.12$\downarrow$         & -2.14     &\textbf{-2.11}$\downarrow$         &-2.13\\
         plants     & -13.79$\uparrow$        & -13.22    & -13.02$\uparrow$    &\textbf{-12.98}\\
         audio      & -41.98$\uparrow$        & -41.20    & \textbf{-39.94}$\downarrow$     &-41.50\\
         jester     & -53.47$\downarrow$        & -54.24    &\textbf{-51.29}$\downarrow$    &-55.03\\
         netflix    & -58.41$\uparrow$        & -57.93    &\textbf{-55.71}$\downarrow$      &-58.69\\
         accidents  & -33.64$\uparrow$        & -29.05    & -30.16$\uparrow$   &\textbf{-28.73}\\
         retail     & -10.81$\downarrow$        & -10.83    &\textbf{-10.72}$\downarrow$      &-10.81\\
         pumsb-star & -33.67$\uparrow$        & -24.39    & -26.12$\uparrow$   &\textbf{-24.12}\\
         dna        & -92.67$\uparrow$        & -87.15    & -88.01$\uparrow$    &\textbf{-86.22}\\
         kosarek    & -10.81$\uparrow$        & -10.70    & \textbf{-10.52}$\downarrow$     &-10.68\\
         msweb      & -9.97$\uparrow$         & -9.74     & -9.89$\uparrow$     &\textbf{-9.71}\\
         book       & -34.97$\uparrow$        & \textbf{-34.49}    & -34.97$\downarrow$    &-34.99\\
         eachmovie  & -58.01$\uparrow$        & -53.72    & -58.01$\uparrow$    &\textbf{-53.67}\\
         webkb      & -161.09$\uparrow$       & \textbf{-154.83}   & -161.09$\uparrow$  &-155.33\\
         routers-52 & -89.61$\uparrow$        & -86.35    & -89.61$\uparrow$   &\textbf{-86.22}\\
         20news-grp & -160.09$\uparrow$       & \textbf{-153.87}   & -155.97$\uparrow$  &-154.47\\
         bbc        & -253.19$\downarrow$       & -256.53   & \textbf{-253.19}$\downarrow$    &-254.41\\
         ad         & -31.78$\uparrow$        & -16.52    & -31.78$\uparrow$   &\textbf{-16.38}\\          \bottomrule
    \end{tabular}
    \end{sc}
    \caption{\textit{\textbf{Density estimation benchmarks: ensembles}}. Average test log-likelihood for \ourlearner\ and \learnpsdd\ models. 
    Two versions of ensembles (EM vs. BEM) are compared separately, $\uparrow$ (resp.\ $\downarrow$) indicates that \ourlearner\ is more accurate (resp.\ less accurate). Bold values indicates the best on certain dataset over 4 methods.}
    \label{tab:ensembles}
\end{table}{}

\newpage
\section{Expected predictions}
\label{sec:app-exp}
\begin{figure*}[!h]
    
    \centering
    \begin{subfigure}[t]{0.22\textwidth}
    \includegraphics[width=0.99\textwidth]{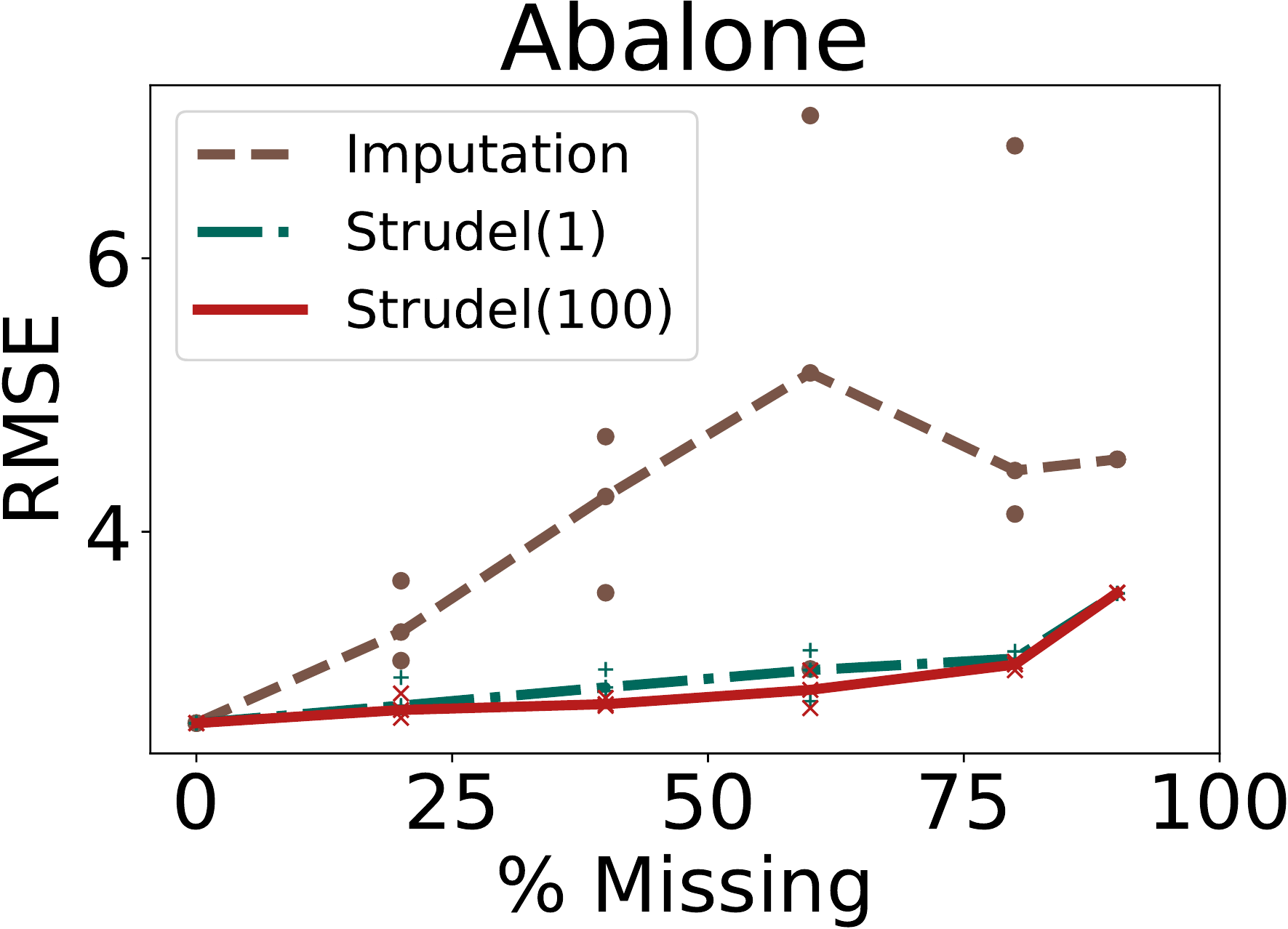}
    \caption{\label{fig:missing-abalone}}
    \end{subfigure}
    \begin{subfigure}[t]{0.22\textwidth}
    \includegraphics[width=0.99\textwidth]{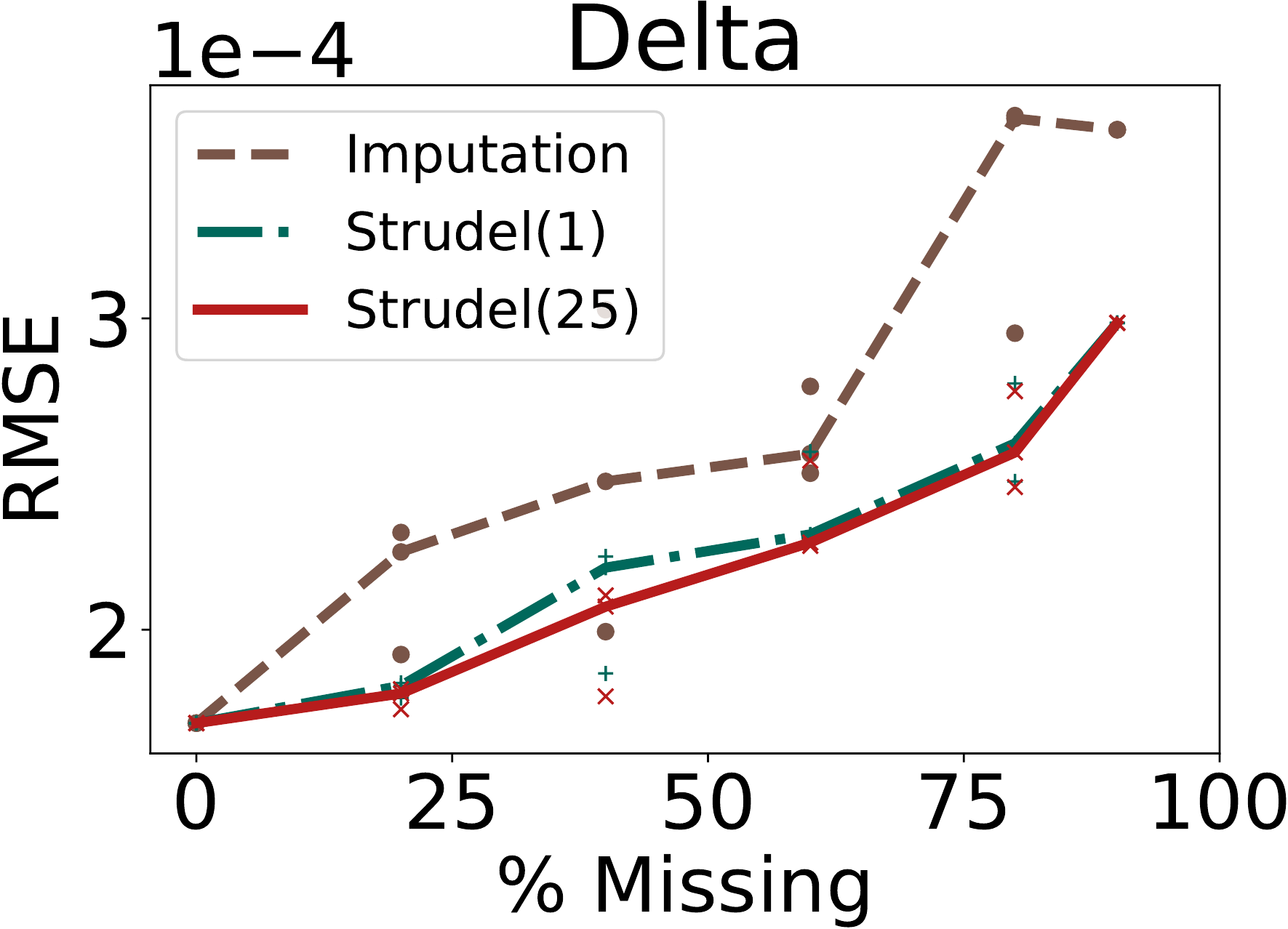}
    \caption{\label{fig:missing-delta}}
    \end{subfigure}
    \begin{subfigure}[t]{0.22\textwidth}
    \includegraphics[width=0.99\textwidth]{figs/exp-exp/Elevators-crop.pdf}
    \caption{\label{fig:missing-elevators}}
    \end{subfigure}
    \begin{subfigure}[t]{0.22\textwidth}
    \includegraphics[width=0.99\textwidth]{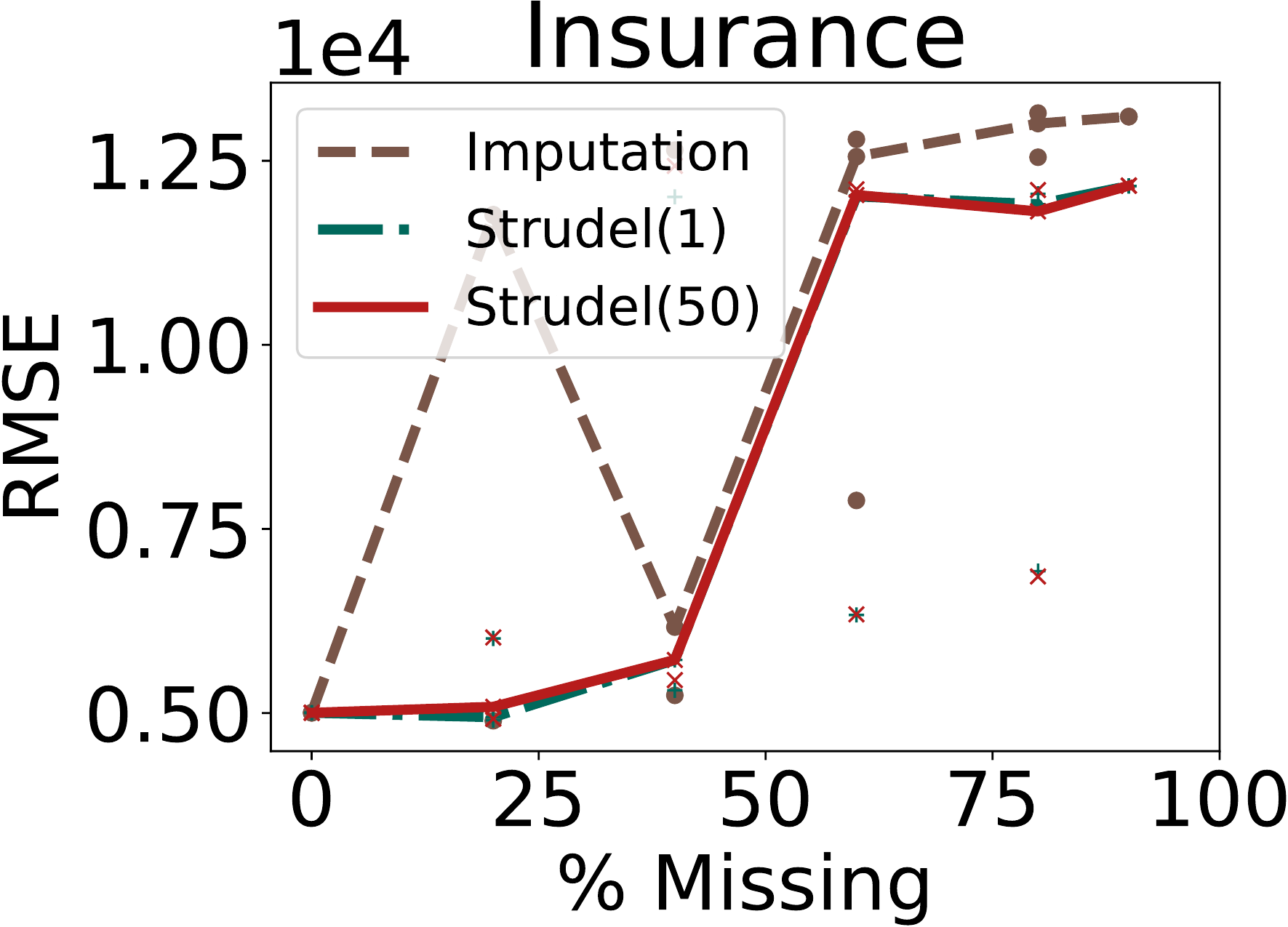}
    \caption{\label{fig:missing-insurance}}
    \end{subfigure}
    \caption{\textit{\textbf{Expected prediction benchmarks}}. Root Mean Square Error (RMSE) for median imputation, \ourlearner\ and \ourlearner-Ensemble models (number in parenthesis shows the number of components in the ensemble models).}
    \label{fig:missingvalue}
\end{figure*}

\end{document}